\pdfoutput=1
\documentclass[a4paper,12pt,numbered,print,index,  twoside]{Classes/PhDThesisPSnPDF}
\input{Preamble/preamble}

\title{Neural Networks Regularization Through Representation Learning
}

\author{Soufiane BELHARBI}

\dept{LITIS laboratory}

\university{INSA Rouen Normandie}
\crest{\includegraphics[scale=0.2]{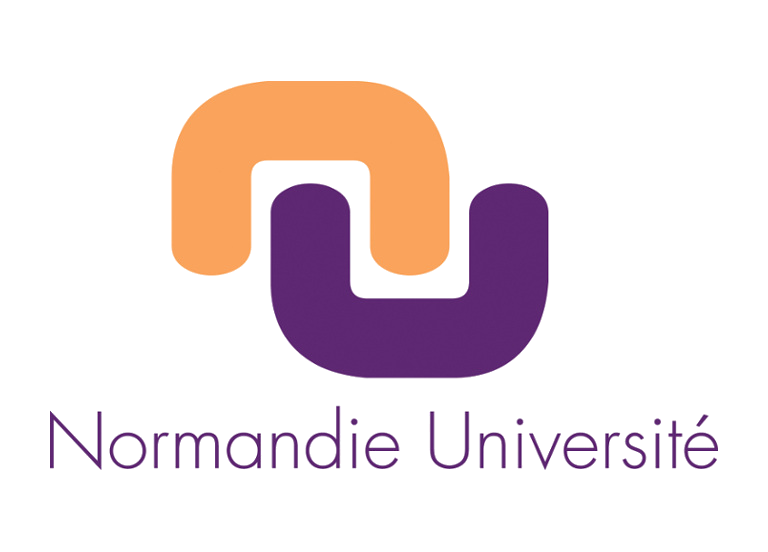}}

\collegeshield{\includegraphics[scale=0.6]{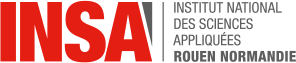} ~ \includegraphics[scale=1.7]{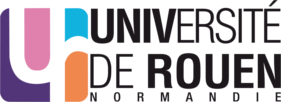} ~ \includegraphics[scale=0.17]{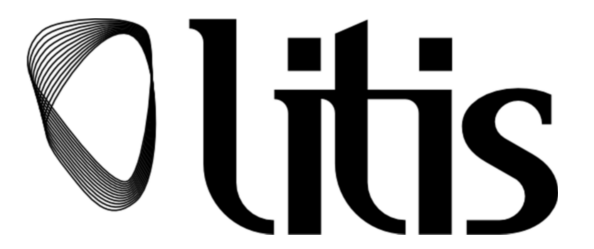}}

\supervisor{\textbf{Prof. Sébastien ADAM}}

\advisor{Assistant Prof. Clément CHATELAIN\newline
Assistant Prof. Romain HÉRAULT}
     
\degreetitle{Doctor of Philosophy}

\college{Normandie Université}

\subject{LaTeX} \keywords{{LaTeX} {PhD Thesis} {Engineering} {University of
Cambridge}}

\ifdefineAbstract
\fi

\ifdefineChapter
\fi

\begin{document}
\includepdf[pages={1-}]{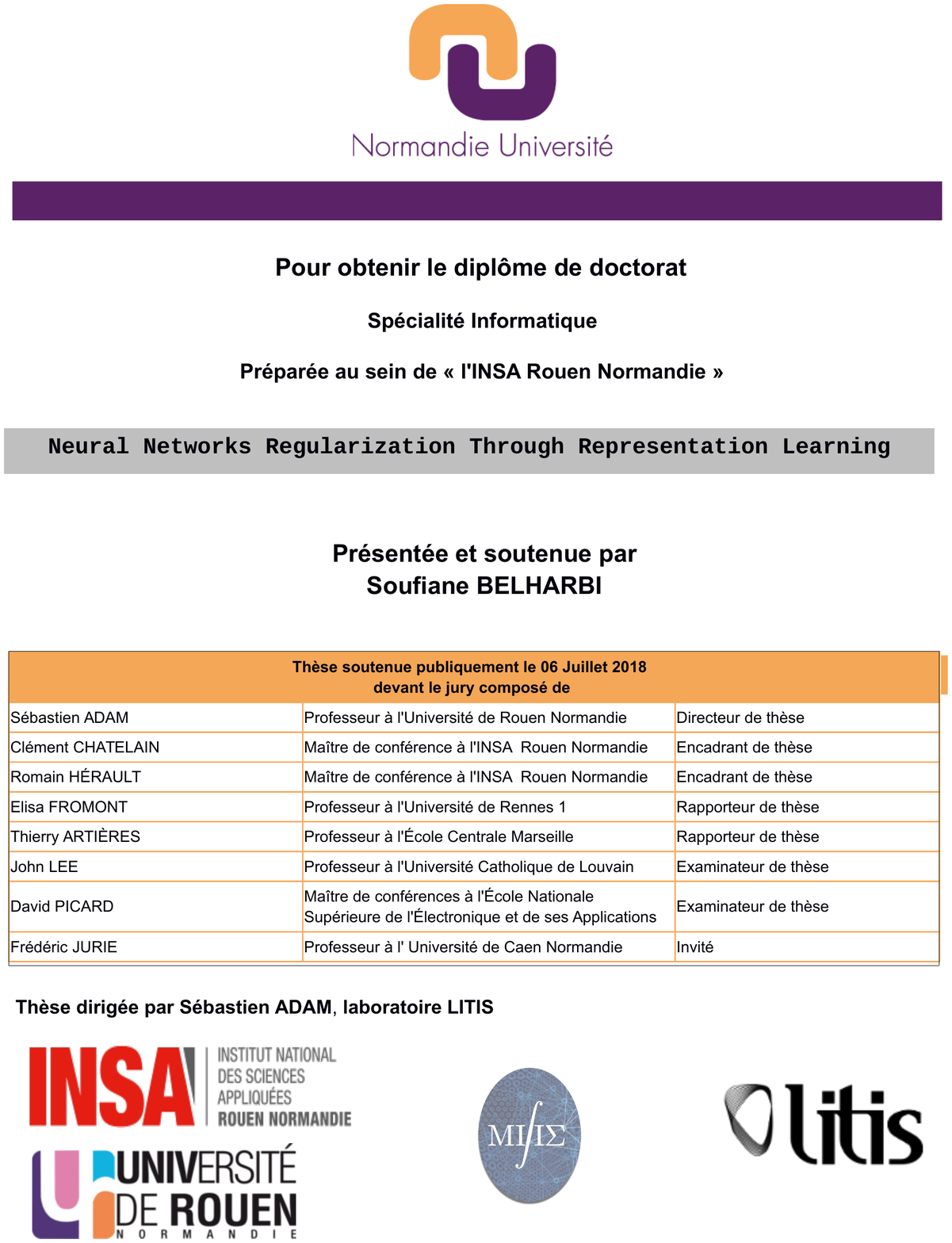}
\newpage

\frontmatter

\maketitle


\begin{dedication} 
I would like to dedicate this thesis to my parents and my grandparents.
\bigskip

\begin{center}
\begin{figure}[!htbp]
  \centering
  \includegraphics[scale=0.2]{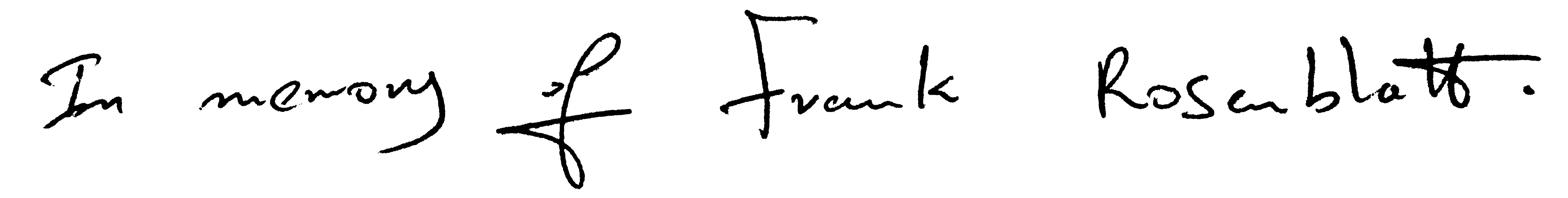}
\end{figure}
\end{center}

\begin{center}
\begin{figure}[!htbp]
  \centering
  \includegraphics[scale=0.5]{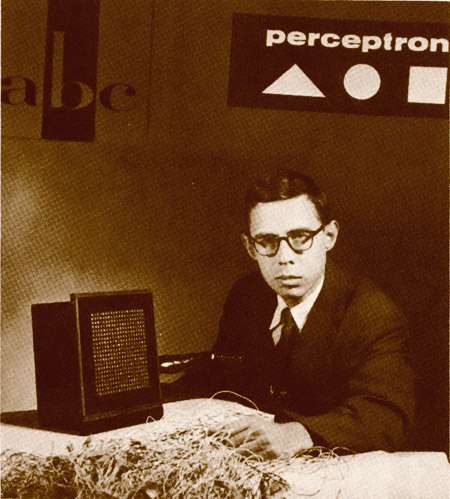}
  \caption[\bel{Frank Rosenblatt, with the image sensor of the Mark I Perceptron.}]{\bel{Frank Rosenblatt, with the image sensor of the Mark I Perceptron (Source: Arvin Calspan Advanced Technology Center; Hecht-Nielsen, R. Neurocomputing (Reading, Mass.: Addison-Wesley, 1990); Cornell Library)}}
\end{figure}
\end{center}

\end{dedication}


\begin{declaration}

I hereby declare that except where specific reference is made to the work of 
others, the contents of this dissertation are original and have not been 
submitted in whole or in part for consideration for any other degree or 
qualification in this, or any other university. This dissertation is my own 
work and contains nothing which is the outcome of work done in collaboration 
with others, except when specified in the text. 



\end{declaration}


\begin{acknowledgements}      
\markboth{Acknowledgements}{}

I would like to thank many people who helped me along my thesis.

I would like to thank my thesis supervisor Sébastien Adam, for taking me as his PhD student. I would like also to thank my advisors Clément Chatelain and Romain Hérault for all their advice and knowledge they have shared with me. I would like to thank as well Romain Modzelewski at Henri Becqurel Center at Rouen for his availability and help for a whole year. I would like to thank as well my collaborators at the same center Sébastien Thureau, and Mathieu Chastan.

Writing this manuscript was a separate challenge. I would like to thank, again, my supervisor Sébastien Adam, and my advisors Clément Chatelain, and Romain Hérault, for their patience, comments, constant criticism, and availability all along the 7 months of writing. It was through their help that this manuscript has reached such maturity.

Preparing the presentation of my PhD defense was the last challenge. Again, my supervisor Sébastien Adam, and my advisors, Clément Chatelain, and Romain Hérault were a major help. It was through their comments and criticism that I have successfully prepared my presentation in terms of content, pedagogy, and speech.

I would like to thank all the jury members of my PhD defense: John Lee, Elisa Fromont, Thierry Artières, David Picard, Frédéric Jurie, Clément Chatelain, Romain Hérault, and Sébastien Adam. I would like to thank all of them for their presence, constructive, detailed, and helpful questions and criticism that helped improving my manuscript and showed their interest to my work. I would like to thank them as well for their valuable time, and their availability to judge my work. It was an honor, and a pleasure defending my work in front of them.

I would like to thank Clément Chatelain for helping me managing my administrative procedure at INSA Rouen Normandie. 

I would like to thank Stéphan Canu for his constant insightful conversations and his many offered opportunities and help. He knocks on my office door every morning to say hi and to start the day with another interesting conversation. I thank him for his encouragement, guidance, constructive criticism, and inspiration all along these years.

I would like to thank Alain Rakotomamonjy for his interesting conversations when we have the chance to have one.

I would like to thank Gilles Gasso for all the interesting conversations that we had including science, and running.

I would like to thank Carole Le Guyader for her interesting conversations and her help.

I would like to thank Samia Ainouz for her constant encouragement and help.

I would like to thank Aziz Bensrhair for his encouragement and advice.

I would like to thank John Lee at the Université Catholique de Louvain who admitted me for two weeks in his laboratory. I would like to thank as well Frédéric Precioso at University of Nice-Sophia Antipolis for inviting me to give a talk in his deep learning summer school.

I would like to thank the staff at the computing centers for their constant support at INSA Rouen Normandie: Jean-François Brulard; Université de Rouen Normandie: Arnaud Citerin; and the CRIANN  computing center (\href{http://www.criann.fr/}{www.criann.fr}): Benoist Gaston, Patrick Bousquet-Melou, Beatrice Charton, and all the technical support team.

I would like to thank the secretariat staff for their help, and assistance including but not limited to: Brigitte Diara, Sandra Hague, Florence Aubry, Isabelle Poussard,  Fabienne Bocquet,  Marion Baudesson, Laure Paris, Leila Lahcen, and Rachel Maeght.


I would like to thank Alexis Lechervy at the Université de Caen Normandie, Kamal D S Oberoi, and Lorette Noiret for their time, and availability to revise many of my papers.

I would like to thank Alexis Lechervy for his help and advise since I met him when I was his student.

I would like to thank Gilles Gasso, Antoine Godichon, Alain Rakotomamonjy, Djamila Boukehil, and Yuan Liu for their time, availability, and insightful conversations that helped me improve one of my papers.

I would like to thank the ASI department at INSA Rouen Normandie for opening the door for me to integrate with them, including but not limited to:  Nicolas Malandain, Nicolas Delestre, Geraldine Del Mondo, Gilles Gasso, Benoit Gaüzère, Sébastien Bonnegent, Damien Guesdon, Elsa Planterose, Alexandre Pauchet, Michel Mainguenaud, and from the Université de Rouen Normandie: Pierrick Tranouez and Daniel Antelme and all the members of the \quotes{miam} group.

I would like to thank all the administration staff at INSA Rouen Normandie and at the Université de Rouen Normandie for their help and assistance.

I would like to thank my previous teachers at the Université de Rouen Normandie for their encouragement, including but not limited to: Thierry Paquet, Laurent Heutte, Stéphane Nicolas, Caroline Petitjean, Maxime Berar, Pierre Héroux, Su Ruan, and Christele Lecomte. I would like also to thank Simon Bernard. I would like to thank my previous teachers at the Université de Caen Normandie as well, including but not limited to: Gaël Dias, Frédéric Jurie, and Alexis Lechervy.

I would like to thank my office colleagues with whom I had insightful conversations and wonderful time inside and outside the office, including but not limited to: Yuan Liu, Linlin Jia, Imad Rida, Cyprien Ruffino, Djamila Boukehil, Denis Rousselle, Sokol Koço, and Kawtar Bomohamed.

I would like to thank Jean-Baptiste Louvet, and Mathieu Bourgais for all the fun conversations we had. Good luck to Jean-Baptiste in saving nature. I hope he can survive without meat.

I Would like to thank all the students, and PhD students that I have met within LITIS laboratory or outside, including but not limited to: Bruno Stuner, Wassim Swaileh, Rivière Marc-Aurèle, Sovann En, Meriem El Azami, Safaa Dafrallah, Imen Beji, Cloé Cabot, Manon Ansart, Laetitia Jeancolas, Tongxue Zhou, Noémie Debroux,  Fan Wang, Fadila Taleb, Imene khames, Romaric Pighetti, Rémi Cadène, Mélanie Ducoffe, Jean baptiste Charpentier, Nezha Bouhafs, Ennassiri Hamza, Harik Elhoussein Chouaib, Mohammed Ali Benhamida, Barange Mukesh, Tatiana Poletaeva, Christophe Tirel, Danut Pop, Sahba Zojaji, Alexander Dydychkin, Julien Lerouge, Grégoire Mesnil, Quentin Lerebours, Florian Leriche, Clara Gainon de Forsan.

I would like to thank my colleagues from my engineer school \href{https://www.esi.dz/}{www.esi.dz} for their help and encouragement, including but not limited to:  Arab Ghedamsi, Boualem Bouldja, Lyas Said Aissa , Sofiane Bellil, Moussa Ailane, and Yasmina Chikhi. 

I would like to thank Hanene Haouam for her encouragement.

I would like to thank Amine Frikha for his constant help during all my three years of studies at Rouen and Caen.

I would like to thank my colleagues and PhD representatives at the LITIS laboratory committee: Fabien Bonardi and Riadh Saada. I would like also to thank the members of ADDED association (\href{http://association-added.fr/}{association-added.fr}) who I had a wonderful time working with them for almost two years: Claire Debreux, Xavier Monnier, Steven Araujo, and Javier Anez Perdomo. I would like to thank as well all the PhD students who subscribed and participated into the animation of my club DOC-AC (\href{http://doctorants-actifs.fr/}{doctorants-actifs.fr}, \href{https://doctorants-actifs.github.io}{doctorants-actifs.github.io}).

I would like to thank everyone who helped in preparing the reception of my PhD defense, including but not limited to: Cyprien Ruffino, Linlin Jia, Ismaila Seck, Djamila Boukehil, Yuan Liu, Benoit Gaüzère, Pierrick Tranouez, Antoine Godichon, Sandra Hague, Alain Rakotomamonjy, and Simon Bernard. Particular thanks to Brigitte Diara who organized and prepared the whole defense.

I thank everyone that helped me in my PhD thesis, people that I know and do not know of. 

In this short acknowledgment, I am certain that I have forgot many other people. For that, I apologize.

This was a wonderful, and a rich experience.

\end{acknowledgements}

\begin{mycontext}
  The research that led to this PhD thesis was conducted at the \quotes{Institut National des Sciences Appliquées Rouen Normandie (INSA Rouen Normandie)}, at the \quotes{Laboratoire d’Informatique, de Traitement de l’Information et des Systèmes (LITIS)}\footnote{Avenue de l’Université, 76801 Saint Etienne du Rouvray Cedex}, in  over the course of 3 years, between 2014 and 2017. This work was done in close collaboration with my supervisor Prof. Sébastien ADAM at \quotes{l'Université de Rouen}, and with my advisors Clément CHATELAIN and Romain HÉRAULT at INSA de Rouen.
\end{mycontext}
\begin{abstract}
  \markboth{Summary}{}
  \addcontentsline{toc}{chapter}{Summary}
Neural network models and deep models are one of the leading and state of the art models in machine learning. \belx{They have been applied in many} different domains. Most successful deep neural models are the ones with many layers which highly increases their number of parameters. Training such models requires a large number of training samples which is not always available. One of the fundamental issues in neural networks is overfitting which is the issue tackled in this thesis. Such problem often occurs when the training of large models is performed using few training samples. Many approaches have been proposed to prevent the network from overfitting and improve its generalization performance such as data augmentation, early stopping, parameters sharing, unsupervised learning, dropout, batch normalization, etc.

In this thesis, we tackle the neural network overfitting issue from a representation learning perspective by considering the situation where few training samples are available which is the case of many real world applications. We propose three contributions. The first one presented in \autoref{chap:chapter4} is dedicated to dealing with structured output problems to perform multivariate regression when the output variable $\y$ contains structural dependencies between its components. Our proposal aims mainly at exploiting these dependencies by learning them in an unsupervised way. Validated on a facial landmark detection problem, learning the structure of the output data has shown to improve the network generalization and speedup its training. The second contribution described in \autoref{chap:chapter5} deals with the classification task where we propose to exploit prior knowledge about the internal representation of the hidden layers in neural networks. This prior is based on the idea that samples within the same class should have the same internal representation. We formulate this prior as a penalty that we add to the training cost to be minimized. Empirical experiments over MNIST and its variants showed an improvement of the network generalization when using only few training samples. Our last contribution presented in \autoref{chap:chapter6} showed the interest of transfer learning in applications where only few samples are available. The idea consists in re-using the filters of pre-trained convolutional networks that have been trained on large datasets such as ImageNet. Such pre-trained filters are plugged into a new convolutional network with new dense layers. Then, the whole network is trained over a new task. In this contribution, we provide an automatic system based on such learning scheme with an application to medical domain. In this application, the task consists in localizing the third lumbar vertebra in a 3D CT scan. A pre-processing of the 3D CT scan to obtain a 2D representation and a post-processing to refine the decision are included in the proposed system. This work has been done in collaboration with the clinic \quotes{\emph{Rouen Henri Becquerel Center}} who provided us with data.

\belx{\textbf{Keywords:} neural network, deep learning, regularization, overfitting, feedforawrd networks, convolutional networks, multi-task learning, unsupervised learning, representation learning, transfer learning, classification, univariate regression, multivariate regression, structured output prediction, prior knowledge.}

\end{abstract}


\begin{abstractfr}
  \markboth{Résumé}{}
  \addcontentsline{toc}{chapter}{Résumé}
  Les modèles de réseaux de neurones et en particulier les modèles profonds sont aujourd’hui l’un des modèles à l’état de l’art en apprentissage automatique et ses applications. Les réseaux de neurones profonds récents possèdent de nombreuses couches cachées ce qui augmente significativement
le nombre total de paramètres. L'apprentissage de ce genre de modèles nécessite
donc un grand nombre d'exemples étiquetés, qui ne sont pas toujours
disponibles en pratique. Le sur-apprentissage est un des problèmes fondamentaux des réseaux de neurones, qui se produit lorsque le modèle apprend par cœur les données d’apprentissage, menant à des difficultés à généraliser sur de nouvelles données. Le problème du sur-apprentissage des réseaux de neurones est le thème principal abordé dans cette thèse. Dans la littérature, plusieurs solutions ont été proposées pour remédier à ce problème, tels que l’augmentation de données, l’arrêt prématuré de l'apprentissage (\quotes{early stopping}), ou encore des techniques plus spécifiques aux réseaux de neurones comme le \quotes{dropout} ou la \quotes{batch normalization}.

Dans cette thèse, nous abordons le sur-apprentissage des réseaux de neurones profonds sous l'angle de l’apprentissage de représentations, en considérant l'apprentissage avec peu de données. Pour aboutir à cet objectif, nous avons proposé trois différentes contributions. La première contribution, présentée dans le chapitre \ref{chap:chapter4}, concerne les problèmes à sorties structurées dans lesquels les variables de sortie sont à grande dimension et sont généralement liées par des relations structurelles. Notre proposition vise à exploiter ces relations structurelles en les apprenant de manière non-supervisée avec des autoencodeurs. Nous avons validé notre approche sur un problème de régression multiple appliquée à la détection de points d’intérêt dans des images de visages. Notre approche a montré une accélération de l’apprentissage des réseaux et une amélioration de leur généralisation. La deuxième contribution, présentée dans le chapitre \ref{chap:chapter5}, exploite la connaissance \emph{a priori} sur les représentations à l'intérieur des couches cachées dans le cadre d'une tâche de classification. Cet \emph{a priori}
est basé sur la simple idée que les exemples d'une même classe doivent avoir la même représentation interne. Nous avons formalisé cet \emph{a priori} sous la forme d’une pénalité que nous avons rajoutée à la fonction de perte.
Des expérimentations empiriques sur la base MNIST et ses variantes ont montré des améliorations dans la généralisation des réseaux de neurones, particulièrement dans le cas où peu de données d’apprentissage sont utilisées. Notre troisième et dernière contribution, présentée dans le chapitre \ref{chap:chapter6}, montre l’intérêt du transfert d’apprentissage (\quotes{transfer learning}) dans des applications dans lesquelles peu de données d’apprentissage sont disponibles. L’idée principale consiste à pré-apprendre les filtres d’un réseau à convolution sur une tâche source avec une grande base de données (ImageNet par exemple), pour les insérer par la suite dans un nouveau réseau sur la tâche cible. Dans le cadre d'une  collaboration avec le centre de lutte contre le cancer \quotes{\emph{Henri Becquerel de Rouen}}, nous avons construit un système automatique basé sur ce type de transfert d’apprentissage pour une application médicale où l'on dispose d'un faible jeu de données étiquetées. Dans cette application, la tâche consiste à localiser la troisième vertèbre lombaire dans un examen de type scanner. L'utilisation du transfert d'apprentissage ainsi que de prétraitements et de post traitements adaptés a permis d'obtenir des bons résultats, autorisant la mise en \oe uvre du modèle en routine clinique.

\belx{\textbf{Mots clés:} réseaux de neurones, apprentissage profond, régularisation, sur-apprentissage, réseau de neurones à passe avant, réseaux de neurones convolutifs, apprentissage multi-tâches, apprentissage non supervisé, apprentissage des représentations, transfert d'apprentissage, classification, régression univariée, régression multiple, prédiction à sortie structurée, connaissances à priori.}
\end{abstractfr}

\begin{mypublications}
  \makeatletter
\newcommand\footnoteref[1]{\protected@xdef\@thefnmark{\ref{#1}}\@footnotemark}
\makeatother
  \textbf{Journals:}
  \begin{itemize}
      \item \textbf{Deep Neural Networks Regularization for Structured Output Prediction}. Soufiane Belharbi, Romain Hérault\footnote{\label{note2}Authors with equal contribution.}, Clément Chatelain\footnoteref{note2},  and Sébastien Adam. Neurocomputing Journal, 281C:169-177, 2018.
      \item \textbf{Spotting L3 Slice in CT Scans using Deep Convolutional Network and Transfer Learning}. Soufiane Belharbi, Clément Chatelain\footnoteref{note2}, Romain Hérault\footnoteref{note2}, Sébastien Adam, Sébastien Thureau, Mathieu Chastan, and Romain Modzelewski. \emph{Computers in Biology and Medicine, 87: 95-103, 2017}.
      \item \textbf{Neural Networks Regularization Through Class-wise Invariant Representation Learning}. Soufiane Belharbi, Clément Chatelain, Romain Hérault, and Sébastien Adam. \textbf{Under review}, \emph{2017}.
  \end{itemize}
  \bigskip
  
  \noindent \textbf{International conferences/workshops:}
  \begin{itemize}
      \item \textbf{Learning Structured Output Dependencies Using Deep Neural Networks}. Soufiane Belharbi, Clément Chatelain, Romain Hérault, and Sébastien Adam. Deep Learning Workshop in the $32^{nd}$ International Conference on Machine Learning (ICML), 2015.
      \item \textbf{Deep Multi-Task Learning with Evolving Weights}. Soufiane Belharbi, Romain Hérault, Clément Chatelain, and Sébastien Adam. European Symposium on Artificial Neural Networks (ESANN), 2016.
  \end{itemize}
  \bigskip
  
  \noindent \textbf{French conferences:}
  \begin{itemize}
      \item \textbf{A Unified Neural Based Model For Structured Output Problems}. Soufiane Belharbi, Clément Chatelain, Romain Hérault, and Sébastien Adam. Conférence Francophone sur l'Apprentissage Automatique (CAP), 2015.
      \item \textbf{Pondération Dynamique dans un Cadre Multi-Tâche pour Réseaux de Neurones Profonds}. Soufiane Belharbi, Romain Hérault, Clément Chatelain, and Sébastien Adam. Reconnaissance des Formes et l'Intelligence Artificielle (RFIA) (Session spéciale: Apprentissage et vision), 2016.
  \end{itemize}
\end{mypublications}


\tableofcontents

\listoffigures

\listoftables

\renewcommand{\nomname}{List of Symbols}
\markboth{\MakeUppercase\nomname}{\MakeUppercase\nomname}
\nomenclature[a-0]{$a$}{A scalar (integer or real)}
\nomenclature[a-1]{$\bm{a}$}{A vector}
\nomenclature[a-2]{$\bm{A}$}{A matrix}
\nomenclature[a-3]{$\bm{I}_n$}{Identity matrix with $n$ rows and $n$ columns}
\nomenclature[a-4]{$\bm{I}$}{Identity matrix with dimentionality implied by the context}
\nomenclature[a-5]{$\diag(\bm{a})$}{A square, diagonal matrix with diagonal entries given by $\bm{a}$}
\nomenclature[a-6]{a}{A scalar random variable}
\nomenclature[a-7]{$\mathbf{a}$}{A vector-valued random variable}
\nomenclature[a-8]{$\mathbf{A}$}{A matrix-valued random variable}
\nomenclature[a-9]{$\bm{0}$}{A vector or a matrix, depending on the context, full of $0$}

\nomenclature[b-0]{$\mathbb{A}$}{A set}
\nomenclature[b-1]{$a \in \mathbb{A}$}{Element $a$ in set $\mathbb{A}$}
\nomenclature[b-2]{$\mathbb{A} \subset \mathbb{B}$}{Set $\mathbb{A}$ is a subset of set $\mathbb{B}$}
\nomenclature[b-3]{$\abs{\mathbb{A}}$}{Number of elements in set $\mathbb{A}$}
\nomenclature[b-4]{$\R$}{Set of real numbers}
\nomenclature[b-5]{$\R_{+}$}{Set of non-negative real numbers}
\nomenclature[b-6]{$\R^n$}{Set of $n$-dimensional real-valued vectors}
\nomenclature[b-7]{$\R^{n \times m}$}{Set of $n\times m$-dimensional real-valued matrices}
\nomenclature[b-8]{$\N$}{Set of natural numbers, i.e., $\{0, 1, \cdots\}$}
\nomenclature[b-9]{$\{0, 1, \cdots, n\}$}{The set of all integers between $0$ and $n$}
\nomenclature[b-90]{$[a, b]$}{Closed interval between $a$ and $b$}
\nomenclature[b-91]{$\{a, b, c\}$}{Set containing elements $a$, $b$ and $c$}

\nomenclature[c-0]{$a_i$}{Element $i$ of vector $\bm{a}$}
\nomenclature[c-1]{$A_{i,j}$}{Element $i,j$ of matrix $\bm{A}$}
\nomenclature[c-2]{$\bm{A}_{i,:}$}{Row $i$ of matrix $\bm{A}$}
\nomenclature[c-3]{$\bm{A}_{:,i}$}{Column $i$ of matrix $\bm{A}$}

\nomenclature[d-0]{$\bm{A}^{\top}$}{Transpose of matrix $\bm{A}$}
\nomenclature[d-1]{$\bm{A} \odot \bm{B}$}{Element-wise (Hadamard) product of $\bm{A}$ and $\bm{B}$}
\nomenclature[d-2]{$\langle \bm{a}, \bm{b} \rangle$}{Inner product between vectors $\bm{a}$ and $\bm{b}$}

\nomenclature[e-1]{$\frac{\partial y}{\partial x}$}{Partial derivative of $y$ with respect to $x$}
\nomenclature[e-2]{$\nabla_{\xvec}y$}{Gradient of $y$ with respect to $\xvec$}
\nomenclature[e-3]{$\frac{\partial f}{\partial \xvec}$}{Jacobian matrix $\bm{J} \in \R^{m \times n}$ of $ f: \R^n \to R^m$}

\nomenclature[f-0]{$P(a)$}{A probability distribution over a discrete variable}
\nomenclature[f-1]{$p(a)$}{A probability distribution over a continuous variable, or over a variable whose type has not been specified}
\nomenclature[f-2]{$\text{a} \sim P$}{Random variable a has distribution $P$}
\nomenclature[f-3]{$\E\limits_{x \sim D}[\cdot]$}{Expectation over $x$ drawn from distribution $D$}
\nomenclature[f-4]{$\mathcal{N}(\xvec; \bm{\mu}, \bm{\Sigma})$}{Gaussian distribution over $\xvec$ with mean $\bm{\mu}$ and covariance $\bm{\Sigma}$}
\nomenclature[f-5]{$\mathcal{D}$}{Unspecified probability distribution}

\nomenclature[g-0]{$f: \X \to \Y$}{The function $f$ with domain $\X$ and range $\Y$}
\nomenclature[g-1]{$f(\bm{x}; \bm{\theta})$}{A function of $\bm{x}$ parametrized by $\bm{\theta}$. (Sometimeswe write $f(\bm{x})$ and omit the argument $\bm{\theta}$ to lighten the notation)}
\nomenclature[g-2]{$\log x$}{Natural logarithm of $x$}
\nomenclature[g-3]{$\log_a$}{Logarithm with base $a$}
\nomenclature[g-4]{$\norm{\xvec}$}{$L_2$ norm of $\xvec$}
\nomenclature[g-5]{$\norm{\xvec}_p$}{$L_p$ norm of $\xvec$}
\nomenclature[g-6]{$\ind_{\mathbb{A}}$}{Indicator function indicating membership in subset $\mathbb{A}$}
\nomenclature[g-7]{$R(\cdot)$}{Generalization error or risk}
\nomenclature[g-8]{$\hat{R}(\cdot)$}{Empirical error or risk}

\nomenclature[h-0]{$\X$}{Input space}
\nomenclature[h-1]{$\Y$}{Target space}
\nomenclature[h-2]{$p_{\text{data}}$}{The data generating distribution}
\nomenclature[h-3]{$\hat{p}_{\text{data}}$}{The empirical distribution defined by the training set}
\nomenclature[h-4]{$\mathbb{D}$}{A set of samples}
\nomenclature[h-5]{$\bm{x}^{(i)}$}{The $i$-th example (input) from a dataset}
\nomenclature[h-6]{${y}^{(i)}$ or $\bm{y}^{(i)}$}{The target associated with $\bm{x}^{(i)}$ for supervised learning}
\nomenclature[h-7]{$\bm{X}$}{The $m\times n$ matrix with input examples $\bm{x}^{(i)}$ in row $\bm{X}_{i,:}$}

\printnomenclature

\mainmatter




%

\begin{generalintroduction}
\markboth{General Introduction}{}
\addcontentsline{toc}{chapter}{General Introduction}

In the last years, \fromont{the neural network field} has seen large success for different applications that \bel{require large number of features} to solve complex tasks. This success has made neural networks one of the leading models in machine learning as well as the state of the art for different applications. Among \belx{the tasks that have been well modeled using neural networks}, one can mention image classification, image labeling, object detection, image description, speech recognition, speech synthesis, query answering, text generation, etc. However, in order to solve such complex tasks, neural network models \belx{rely on} a large number of parameters that may easily reach millions. Consequently, such  models require a large number of training data in order to avoid overfitting, a case where the model becomes too specific to the training data and looses its ability to generalize to unseen data. In practice, one usually deals with applications with few training samples. Therefore, one has the choice either to use neural network models with small capacity and loose much of their power, or stick with models with large capacity and employ what it is known as regularization techniques to save the generalization ability and to prevent the network from overfitting. We provide in \bel{ \autoref{chap:chapter0} a background on machine learning with a focus on regularization aspects}. The same chapter contains a brief introduction to neural network models and their regularization techniques. \belxx{The chapter is completed by Appendix \ref{chap:chapterapp4} that provides some basic definitions in machine learning, and more technical details on neural networks and regularization}.

In the literature of neural networks, we find different approaches of regularization that we describe in Sec.\ref{sec:neuralnetsregularization0}. Such regularization methods either aim at explicitly reducing the model complexity \bel{using for example} $L_p$ parameter norm (Sec.\ref{sub:explicityreg0}), or implicitly reducing its complexity \bel{using for instance} early stopping (Sec.\ref{subsub:earlystopping0}). Other methods do not alter the model complexity but tackle the issue of overfitting using different angles \bel{ where the aim is to improve the model's generalization. One can mention the following approaches:
\begin{itemize}
  \item parameter sharing (Sec.\ref{sub:parametersharing0}),
  \item data augmentation (Sec.\ref{sub:dataaugmentation0}),
  \item \belxx{batch normalization (Sec.\ref{subsub:batchnormalization0}),}
  \item and smart ensemble methods such as dropout (Sec.\ref{subsub:dropout0}).
\end{itemize}
}
\belxx{\cite{Goodfellowbook2016} covers most \fromont{of} the technical approaches used to regularize neural networks.}

In this thesis-by-article, we provide three different ways of regularizing neural network models \bel{when} trained using small dataset for the task of classification, univariate, and multivariate regression. \bel{The aim of such approaches is to directly improve the model's generalization without affecting its complexity. Our research direction to improve neural networks generalization in this thesis is throughout learning good internal representations within the network}.
\bigskip

Data representation within neural networks is a key component of their success. \bel{Such models are able to self-learn adequate internal representations to solve the task in hand. This ability is a major factor that separates such models from other type of models in machine learning. Although models with high capacity are able to build complex features that allow to solve complex tasks, they are more likely to fall into overfitting, particularly when they are trained with few samples. In this thesis, we provide three approaches to regularize neural networks. Two of them are based on theoretical frameworks which adapt learning internal representations to the task in hand. This includes structured output prediction  through unsupervised learning (\autoref{chap:chapter4}), and classification task  through prior knowledge (\autoref{chap:chapter5}). The last approach is based on transfer learning  with an applicative aspect in medical domain (\autoref{chap:chapter6}). Each of our contributions is presented in a chapter.
\begin{description}
  \item[$\bullet$ Structured output prediction] \hfill \\
  In \autoref{chap:chapter4}, we explore the use of unsupervised learning to build robust features in order to solve structured output problems, i.e., to perform multivariate regression where the output variable is high dimensional with possible structural relations between its components. The motivation behind this work is that feedforward networks lack, by their structural design, the ability to learn dependencies among the output variable. We recall that each output neuron in a feedforward network performs the prediction independently from the rest of the other output neurons. In order to learn the output structure, local or global,  the network needs a large \fromont{amount of} training data in order to learn the output variations. Otherwise, the network falls into overfitting or in the case of outputting the mean structure. In this work, we want the network to focus explicitly on learning the output structure. In order to do so, we propose a unified multi-task framework to regularize the training of the network when dealing with structured output data. The framework contains a supervised task that maps the input to the output and two unsupervised tasks where the first learns the input distribution while the second learns the output distribution. The later one, which is the key component of the framework, allows learning explicitly the output structure in an unsupervised way. This allows the use of labels only data to learn such structure. We validate our framework over a facial landmark detection problem where the goal is to predict a set of key points on face images. Our experiments show that the use of our framework speeds up the training of neural networks and improves their generalization.
  \item[$\bullet$ Prior knowledge and classification] \hfill \\
  In \autoref{chap:chapter5}, we explore another aspect of learning representations within neural network. We investigate the use of prior knowledge. Using prior knowledge can be considered as a regularization. It allows to promote generalization without the need to see large \fromont{amount of} training  data. For instance, in the case of a classification task, if someone provides us the information that \quotes{generally, a car has four wheels}, it could save us the need to see a large number of car images to understand such a key concept in order to figure out what a car is. In the context of classification, a general prior knowledge about the internal representation of a network is that samples within the same class should have the same internal representation. Based on this belief, we propose a new regularization penalty that constrains the hidden layers to build similar features for samples within the same class. Validated on MNIST dataset and its variants, incorporating such prior knowledge in the network training allows improving its generalization while using few training samples.
  \item[$\bullet$ Transfer learning and medical domain] \hfill \\
  Our last contribution, presented in \autoref{chap:chapter6}, consists in a different strategy to boost learning complex features using high capacity models while using few training samples. The hope is to maintain the model generalization. As it seems in contradiction with the generalization theory (Sec.\ref{sec:machinel0}) that states that to well fit a model with high capacity we need a large number of training data, empirical evaluation shows the possibility of such approach. In this work, we provide a real life application of a learning scheme that allows us to  train a model with high capacity using only few training samples. The learning scheme consists in training a model with high capacity over a task that has abundant training samples such as training large convolutional network over ImageNet dataset. Then, we re-use part of the trained model, responsible for features building, in the second task which has only few training samples. This training scheme, known as transfer learning, allows the second task which has a lack of data to benefit from another task with abundant data. We validate this learning scheme over a real life application based on medical image data. The task consists in localizing a specific vertebra, precisely the third lumbar vertebra, in a 3D CT scan of a patient's body. We provide adequate pre-processing and post-processing procedures. We report satisfying results. This work was done in collaboration with the clinic \quotes{Rouen Henri Becquerel Center} which provided us with the data.
\end{description}
}
\bigskip

{\begin{center} \Large \textbf{How to Read This Thesis} \end{center}}

This is a thesis by article. It is composed of 4 chapters. \autoref{chap:chapter0} is an introduction which is composed of four sections: Sec.\ref{sec:machinel0} presents machine learning backgrounds with more focus on regularization; Sec.\ref{sec:neuralnetsintro0} presents an introduction to feedforward neural networks; Sec.\ref{sec:neuralnetsregularization0} contains methods that are used to improve neural networks generalization; Sec.\ref{sec:conclusion0} contains the conclusion of the first chapter. If the reader is familiar with these subjects, we recommend skipping the first chapter. However, we recommend reading Sec.\ref{sub:modelselectionandregularization0} that presents the regularization concept and Sec.\ref{subsub:representationlearning0} which presents unsupervised and semi-supervised learning in neural networks as a regularization.

\autoref{chap:chapter4}, \autoref{chap:chapter5}, and \autoref{chap:chapter6} contain our three contributions.

\belxx{In order to keep this thesis short, straightforward, and self-contained, we provide one appendix that covers some basic definitions in machine learning and some technical details on neural networks and regularization (Appendix \ref{chap:chapterapp4}).}

\end{generalintroduction}


\chapter{Background}  
\label{chap:chapter0}

\ifpdf
    \graphicspath{{Chapter0/Figs/Raster/}{Chapter0/Figs/PDF/}{Chapter0/Figs/}}
\else
    \graphicspath{{Chapter0/Figs/Vector/}{Chapter0/Figs/}}
\fi

\makeatletter
\def\input@path{{Chapter0/}}
\newcommand\footnoteref[1]{\protected@xdef\@thefnmark{\ref{#1}}\@footnotemark}
\makeatother
We discuss in this first chapter three important subjects that \bel{constitute} the background of this thesis: machine learning, neural networks, and methods to improve neural networks generalization. In the first section (Sec.\ref{sec:machinel0}), we provide an introduction to machine learning problem with an emphasis on the generalization aspect and how one can improve it. As this thesis concerns specifically neural networks, we then provide an introduction to such models in the second section (Sec.\ref{sec:neuralnetsintro0}). Our main concern behind this thesis is to provide new techniques to improve the generalization performance of neural network, particularly when dealing with small training sets. Therefore, the last section (Sec.\ref{sec:neuralnetsregularization0}) contains a presentation of the most common methods used to improve the \bel{generalization} performance of neural networks.

This chapter is inspired from \bel{ machine learning and neural networks literature \cite{Mohri2012bookML, Shalev2014bookML, AbuMostafa2012MLBook, Kearns1994book, Goodfellowbook2016, Demyanovtehsis2015} including precise definitions and theorems}.

\section[Machine Learning]{Machine Learning}
\label{sec:machinel0}
This section contains an introduction to machine learning with a focus on the generalization aspect. Appendix \ref{chap:chapterapp1} contains basic definitions of machine learning if needed.
\subsection{Learning from Data}
\label{sub:introd0}
\bel{If a three-year-old kid is shown a picture and asked if there is a tree in it, it is more likely we get the right answer. If we ask the three-year-old kid what is a tree, we likely get an inconclusive answer. We, humans, did not learn what is a tree by studying the mathematical definition of trees. We learned it by looking at trees. In other words, we learned from \textit{data}, i.e., \textit{examples}.

The term \emph{machine learning} refers to the automated detection of meaningful patterns in data. In the past couple of decades, it has become a common tool in almost any task that requires information extraction from large datasets. Nowadays, we are surrounded by machine learning based technology almost anywhere: search engines learn how to bring us the best results, antispam software learns how to filter our email inbox, and credit card transactions are secured by a software that learns to detect frauds. Digital cameras learn to detect faces and intelligent personal assistance applications on smart-phones learn to recognize voice commands. Cars are equipped with accident prevention systems that are built using machine learning algorithms. Hospitals are equipped with programs that assist doctors in their diagnostics. Machine learning is also widely used in scientific applications such as bioinformatics, medicine, and astronomy.

A common feature of all these applications is that, in contrast to the traditional use of computers, in these cases, due to complexity of the patterns that need to be detected, a human programmer can not provide an explicit, fine-detailed specification of how such tasks should be executed. Taking example from intelligent beings, many of our skills are acquired or refined through \textit{learning} from our experience. Machine learning tools are concerned with endowing programs with the ability to \quotes{learn} and adapt.

Learning can be thought of as the process of converting experience into expertise or knowledge. Machine learning aims at incorporating such a concept into computers or any other computing device such as phones, tablets, etc. Now, why do we need machine learning tools? one of the reasons is to automate processes and eliminate humans routines to free them to do more intelligent and delicate tasks. Another important reason is to fill the gap in the human capabilities to perform tasks that go beyond their abilities. Such tasks require analyzing very large and complex data: astronomical data, turning medical archives into medical knowledge, weather prediction, analysis of genomic data, web search engines, electronic commerce, etc. With more and more available digitally recorded data, it becomes obvious that there is meaningful information buried in such data that are too large and too complex for human to process them or make sense of them. The last reason that we mention here is the limiting feature of traditional programmed tools which is their rigidity. Once installed, a programmed tool does not change. However, many tasks change over time or from a user to another. Machine learning based tools provide a solution to such issues. They are, by nature, adaptive to changes in the environment they interact with. A typical successful application of machine learning to such problems include programs that decode handwritten text, where a fixed program can adapt to variations between the handwriting of different users.

Machine learning covers a large spectrum of tasks to adapt to the human needs. Such tasks include classification, regression, ranking, clustering, dimensionality reduction, etc. Machine learning also provides different learning strategies as an adaptive way to the nature of available data and the environment including supervised/unsupervised/semi-supervised learning, transductive inference, online learning, reinforcement learning, active learning, etc. We assume the reader is familiar with basic definitions of machine learning. In the other case, Appendix \ref{chap:chapterapp1} contains such definitions.

Let us go back to our problem of tree detection in images. To solve this problem we decide to use a machine learning tool \belx{in a supervised context where we collect a set of images and annotate them to indicate if they contain a tree or not}. One may ask the following questions: is it possible to learn the concept of trees? which machine learning algorithm should we use? how many pictures of trees do we need for learning? is there any guarantee that our algorithm will succeed to detect a tree that was not seen during training? such questions are legitimate and they construct the foundation of machine learning. Computational learning theory \cite{Kearns1994book}, which is a subfield of artificial intelligence devoted to studying the design and analysis of machine learning algorithms, provides us with well studied and formal frameworks to answer the aforementioned questions and more. For the sake of simplicity, and in order to address such questions, we decide to use in this thesis the \textit{probably approximately correct} learning framework, also known as PAC learning, proposed by Leslie Valiant \cite{Valiant1984} in the $80^{\prime}$s.

A fundamental pillar in the learning concept is the ability of the learner to perform well for unseen situations. This aspect in learning is known as \textit{generalization}. It is likely in practice that the learner fails to acquire such capability and falls into what is known as \textit{overfitting}, a situation where the learner performs well on the training data but fails to generalize to unseen data. \textit{Regularization} is a well known approach in machine learning that we use to deal with such issue. For the sake of simplicity and coherence with our contributions in this thesis, we will present in the next section only some selected concepts in the PAC learning framework in order to answer the aforementioned questions while building our way and justifications toward the concept of regularization.}

\subsection{Statistical Learning and Generalization: PAC Learning}
\label{sub:learningframework0}
\bel{In this section, we present a simplified version of PAC learning framework in order to define the learnability of a concept, the generalization aspect, and the number of samples needed to learn. For illustration, we consider PAC learning framework for learning a binary classification task. Extension to other tasks is possible as well \cite{Mohri2012bookML, Shalev2014bookML, AbuMostafa2012MLBook}.}

\bel{We denote by $\X$ the set of all possible examples or instances. $\X$ is also referred to as the input space. The set of all possible labels or target values is denoted by $\Y$ which is referred to as the output space. $\Y$ is limited to two labels, $\Y=\{0, 1\}$, where $1$ refers to the class \texttt{tree} while $0$ refers to the class \texttt{not-tree}}.

\bel{A concept $c: \X \to \Y$ is a mapping function. A concept class is a set of concepts we may wish to learn one of them and is denoted by $\mathbb{C}$. We note by $h$ a hypothesis which is also a mapping function: $h: \X \to \Y$. $\mathbb{H}$ is the hypothesis set. Concepts and hypotheses have the same nature, both of them are mapping functions that take an element from $\X$ and map it into $\Y$. Therefore, the terms hypothesis and concept can be interchangeable. In this context, the main difference between a concept and a hypothesis is that the learner does not have access to the concept $c$ while it does know the hypothesis $h$. Also, this distinction makes it possible that the set of possible hypotheses may not contain $c$, the true hypothesis that associates a label to a random input. In this context, the aim of the learning algorithm is to pick a hypothesis $h$ that approximates $c$. We note that in the case of classification, a hypothesis $h$ and a concept $c$ can be called a \emph{classifier}.}

\bel{Let us assume that examples are independently and identically distributed (i.i.d.) according to some fixed but unknown distribution $\mathcal{D}$. The learning problem is then formulated as follows. The learning algorithm considers a fixed set of possible concepts $\mathbb{H}$, which may not coincide with $\mathbb{C}$. It receives $N$ samples ${\mathbb{S}=(\bm{x}^{(1)}, \cdots, \bm{x}^{(N)})}$ drawn i.i.d. according to $\mathcal{D}$ as well as the labels ${(c(\bm{x}^{(1)}),\cdots, c(\bm{x}^{(N)}))}$, which are based on a specific target concept $c \in \mathbb{C}$ to learn. The set $\{(\bm{x}^{(i)}, c(\bm{x}^{(i)}))\}_{i=1}^N$ is known as the training set.
In order to measure the success of a selected hypothesis $h$, we need an error measure. In the case of classification, we can use a 0-1 loss function defined as an indicator function that indicates whether $h$ makes a mistake or not over a sample $\bm{x}$,
\begin{equation}
  \label{eq:eq0-00}
  \ind_{h(\bm{x}) \neq c(\bm{x})} = 
  \begin{cases}
    1       & \quad \text{if } h(\bm{x}) \neq c(\bm{x})\\
    0  & \quad \text{if } h(\bm{x}) = c(\bm{x}) \; .
  \end{cases}
\end{equation}
The loss function, defined to measure the error committed by $h$, depends on the task in hand. For instance, in a regression task, a possible loss is the square loss $(h(x) - c(x))^2$. The error of a hypothesis over all examples that can be sampled from $\mathcal{D}$ is referred to as \emph{generalization error}.

 The generalization error of a hypothesis $h \in \mathbb{H}$, also referred to as the true error, or just error of $h$, is denoted by $R(h)$ and defined as follows \cite{Mohri2012bookML, Shalev2014bookML, AbuMostafa2012MLBook}
\begin{mydef} 
  \label{def:def0-0}
\textbf{Generalization error}\\
Given a hypothesis $h \in \mathbb{H}$, a target concept $c \in \mathbb{C}$, and an underlying distribution $\mathcal{D}$, the generalization error or risk of $h$ is defined by
\begin{equation}
    \label{eq:eq0-0}
    R(h) = \Prob_{\mathbf{x} \sim \mathcal{D}}[h(\bm{x}) \neq c(\bm{x})] = \E_{\mathbf{x} \sim \mathcal{D}}[\ind_{h(\bm{x}) \neq c(\bm{x})}] \; .
\end{equation}
\end{mydef}
The learning algorithm task is to use the labeled samples $\mathbb{S}$ to select a hypothesis $h_{\mathbb{S}} \in \mathbb{H}$ that has a small generalization error with respect to the concept $c$.

The generalization error of a hypothesis is not directly accessible to the learning algorithm since both distribution $\mathcal{D}$ and the target concept $c$ are unknown. Therefore, another method is required in order to measure how well does a hypothesis $h$. A useful notion of error that can be calculated by the learning algorithm is the \textit{training error}, which is the error of the classifier $h$ over the training set. The term \textit{empirical error} or \textit{empirical risk} are also used for this error. It is defined as follows \cite{Mohri2012bookML, Shalev2014bookML, AbuMostafa2012MLBook}
\begin{mydef}
\label{de:def0-1}
\textbf{Empirical error} \\
Given a hypothesis $h \in \mathbb{H}$, a target concept $c \in \mathbb{C}$, and a set of samples $\mathbb{S}=(\bm{x}^{(1)}, \cdots, \bm{x}^{(N)})$, the empirical error or the empirical risk of $h$ is defined by
\begin{equation}
    \label{eq:eq0-1}
    \hat{R}(h) = \frac{1}{N} \sum^N_{i=1} \ind_{h(\bm{x}^{(i)}) \neq c(\bm{x}^{(i)})} \; .
\end{equation}
\end{mydef}
Thus, the empirical error of $h \in \mathbb{H}$ is the average error over the samples $\mathbb{S}$, while the generalization error is the expected error based on the distribution $\mathcal{D}$. Since the training samples $\mathbb{S}$ is a snapshot of $\mathcal{D}$ that is available to the learning algorithm, it makes sense to search for a solution that works well on $\mathbb{S}$. This learning paradigm that outputs a hypothesis $h$ that minimizes $\hat{R}(h)$ is called \textit{Empirical Risk Minimization} or ERM for short.

The following introduces the PAC learning framework. We denote by $O(n)$ an upper bound on the cost of the computational representation of any $\bm{x} \in \X$ and by size$(c)$ the maximal cost of the computational representation of $c \in C$. For example, $\bm{x}$ may be a vector in $\R^n$, for which the cost of an array-based representation would be $O(n)$.
\begin{mydef}
  \textbf{PAC-learning} \\
  \label{de:def0-10paclearning}
  A concept class $\mathbb{C}$ is said to be \textit{PAC-learnable} if there exists an algorithm $\mathcal{A}$ and a polynomial function $poly(\cdot, \cdot, \cdot, \cdot)$ such that for any $\epsilon > 0$ and $\delta > 0$, for all distributions $\mathcal{D}$ on $\X$ and for any target concept $c \in \mathbb{C}$, the following holds for any number of samples $N \geq poly(1/\epsilon, 1/\delta, n, \text{size}(c))$:
  \begin{equation}
    \label{eq:eq0-10pac}
    \Prob_{\mathbf{x} \sim \mathcal{D}^N} [R(h_{\mathbb{S}}) \leq \epsilon] \geq 1 - \delta \; .
  \end{equation}
\end{mydef}
A concept class $\mathbb{C}$ is thus PAC-learnable if the hypothesis returned by the learning algorithm $\mathcal{A}$ after observing a number of points is \textit{approximately correct} (with error at most $\epsilon$) with high \textit{probability} (at least $1-\delta$), which justifies the PAC-learning terminology. The definition of PAC-learning contains two approximation parameters which are predefined beforehand. The accuracy parameter $\epsilon$ determines how far the output hypothesis (classifier) is from the optimal one, and a confidence parameter $\delta$ which indicates how likely the classifier to meet that accuracy requirement.

Several key points of the PAC-learning definition are worth mentioning. First, the PAC-learning framework is distribution-free model, no particular assumption is made about the distribution from which samples are drawn. Second, the training data and the test data are drawn from the same distribution $\mathcal{D}$. This is a necessary assumption for the generalization to be possible in most cases. Finally, the PAC-learning framework deals with the learnability of a concept class $\mathbb{C}$ not a particular concept $c$.

Up to now, we have provided only the definition of a learnable concept class. The next paragraph provides a condition on the sample complexity \cite{Mohri2012bookML, Shalev2014bookML}, i.e., the minimal number of samples needed in order to guarantee a probably approximately correct solution. An upper bound of the generalization error is as well provided. First, let us consider the case of a consistent hypothesis \belx{$h_{\mathbb{S}}$  which is a hypothesis that admits no error on the training samples $\mathbb{S}$: $\hat{R}(h_{\mathbb{S}}) = 0$. Whereas, $h_{\mathbb{S}}$ is said to be inconsistent when it has errors on the training samples: $\hat{R}(h_{\mathbb{S}}) > 0$.}

\begin{mytheor}
\label{theo:theo0-0-1}
\textbf{Learning bounds: finite $\mathbb{H}$, consistent case} \\
Let $\mathbb{H}$ be a finite set of functions mapping from $\X$ to $\Y$. Let $\mathcal{A}$ be an algorithm that for any target concept $c \in \mathbb{H}$ and any i.i.d. samples set $\mathbb{S}$ returns a consistent hypothesis ${h_{\mathbb{S}}: \hat{R}(h_{\mathbb{S}}) = 0}$. Then, for any $\epsilon, \delta > 0$, the inequality ${\Prob\limits_{\mathbb{S} \sim D^N} [R(h_{\mathbb{S}}) \leq \epsilon] \geq 1 - \delta}$ holds if 

\begin{equation}
    \label{eq:eq0-2-0-pac}
    N \geq \frac{1}{\epsilon} \left(\log\abs{\mathbb{H}} + \log\frac{1}{\delta}\right) \; .
\end{equation}
\noindent This sample complexity result admits the following equivalent statement as a generalization bound: for any ${\epsilon, \delta > 0}$, with probability at least $1-\delta$,
\begin{equation}
    \label{eq:eq0-2-1-pac}
    R(h_{\mathbb{S}}) \leq \frac{1}{N} \left(\log\abs{\mathbb{H}} + \log\frac{1}{\delta}\right) \; .
\end{equation}
\end{mytheor}
Theorem \ref{theo:theo0-0-1} shows that when the hypothesis set $\mathbb{H}$ is finite, $\mathcal{A}$, that returns a consistent hypothesis, is a PAC-learning algorithm under the condition of the availability of enough training samples. As shown by Eq.\ref{eq:eq0-2-1-pac}, the generalization error of consistent hypotheses is upper bounded by a term that decreases as a function of the number of training samples $N$. This is a general fact, as expected, learning algorithms benefit from large labeled training samples. The decrease rate of $O(1/N)$ guaranteed by this theorem, however, is particularly favorable. The price to pay for coming up with a consistent hypothesis is the use of a larger hypothesis set $\mathbb{H}$ in order to increase the chance to find target concepts. As shown in Eq.\ref{eq:eq0-2-1-pac}, the upper bound increases with the cardinality $\abs{\mathbb{H}}$. However, that dependency is only logarithmic. We note that the term $\log\abs{\mathbb{H}}$ can be interpreted as the number of bits needed to represent $\mathbb{H}$. Therefore, the generalization guarantee of the theorem is controlled by the ratio of this number of bits $\log\abs{\mathbb{H}}$ and the number of training samples $N$.

In most general case, there may be no hypothesis in $\mathbb{H}$ consistent with the labeled training samples. This, in fact, is the typical case in practice where the learning problem may be somewhat difficult or the concept class may be more complex than the hypothesis set used by the learning algorithm. The learning guarantees in this more general case can be derived under an inequality form that relates the generalization error and empirical error for all the hypotheses $h \in \mathbb{H}$ \cite{Mohri2012bookML, Shalev2014bookML}.

\begin{mytheor}
\label{theo:theo0-0}
\textbf{Learning bound: finite $\mathbb{H}$, inconsistent case} \\
Let $\mathbb{H}$ be a finite hypothesis set. Then, for any $\delta > 0$, with probability at least $1-\delta$, the following inequality holds:
\begin{equation}
    \label{eq:eq0-2}
    \forall h \in \mathbb{H}, \quad R(h) \leq \hat{R}(h) + \sqrt{\frac{\log(\abs{\mathbb{H}}) + \log(\frac{2}{\delta})}{2N}} \; .
\end{equation}
\end{mytheor}
Thus, for a finite hypothesis set $\mathbb{H}$,
\begin{equation}
    \label{eq:eq0-3}
    R(h) \leq \hat{R}(h) + O\left(\sqrt{\frac{\log_2(\abs{\mathbb{H}})}{N}}\right).
\end{equation}
\noindent The sample complexity required in this case is,
\begin{equation}
    \label{eq:eq0-2-1-pac-incon}
    N \geq \frac{1}{2\epsilon^2} \left(\log\abs{\mathbb{H}} + \log\frac{1}{\delta}\right) \; .
\end{equation}
Several remarks similar to those made on the generalization bound in the consistent case can be made here: a larger number of training samples $N$ guarantees better generalization, and the bound increase with the size $\abs{\mathbb{H}}$, but only logarithmically. But, here, the bound is a less favorable function of $\frac{\log_2\abs{\mathbb{H}}}{N}$; it varies as the square root of this term. This is not a minor price to pay: for a fixed $\abs{\mathbb{H}}$, to attain the same guarantees as in the consistent case, a quadratically larger labeled samples is needed.

We note that the bound suggests seeking a trade-off between reducing the empirical error versus controlling the size of the hypothesis set: larger hypothesis set is penalized by the second term but could help reducing the empirical error, that is the first term. But, for a similar empirical error, it suggests using a small hypothesis set. This can be viewed as an instance of the so-called \emph{Occam's razor} principle\footnote{\bel{\emph{Occam's razor} principle, after William of Ockham, a 14th-century English Logician, is a problem-solving approach that, when presented with competing explanations, a short explanation (that is, a hypothesis with short length) tends to be more valid than a long explanation. In computational learning theory context, Theorem \ref{theo:theo0-0-1} and Theorem \ref{theo:theo0-0} provide a justification of such principle \cite{Mohri2012bookML, Shalev2014bookML}.}} which states that: all other things being equal, a simpler (smaller) hypothesis set is better \cite{Mohri2012bookML}.

This concludes our introduction of the PAC-learning framework. We have seen during this short review the definition of  learnability which indicates if an algorithm has learned to do some task. We investigated as well the number of training samples required for a learning algorithm to achieve some predefined accuracy. We have seen also the relation between the generalization error and the empirical error as a function of the size of the hypothesis space $\mathbb{H}$ and the number of training samples $N$. This relation suggests that in order to obtain better generalization error, it is better to choose small hypothesis set and use large number of training samples.

As we mentioned in the beginning of this section, PAC-learning framework is a theoretical learning framework that provides learning guarantees for finite hypothesis sets. However, in practice, we mostly deal with infinite hypothesis set such as the set of all hyperplanes. While this framework remains theoretical, it provides us with general guidelines to build good learning algorithms. In the case of infinite $\mathbb{H}$, other learning frameworks can be required to provide learning guarantees. We can mention Vapnick and Chervonenkis learning framework that introduces the concept of Vapnik–Chervonenkis dimension \cite{vapnik1971}, also known as VC-dimension which is covered in Appendix \ref{sub:generalizationtheoryapp1}.

As we saw earlier, the generalization of a set of hypothesis $\mathbb{H}$ depends on two main aspects: the number of training samples and the size of the hypothesis set. In practice, we usually have a fixed number of training samples. The only variable that we can control is $\abs{\mathbb{H}}$. Now, given this fixed number of samples, how one can choose the right size of a hypothesis set $\abs{\mathbb{H}}$? In the following, we address such issue by introducing the concept of \textit{regularization}.
}

\subsection{Empirical Risk Minimization with Inductive Bias}
\label{sub:modelselectionandregularization0}
We discuss here some model selection and algorithmic ideas based on the theoretical results presented in previous section. Let us assume an i.i.d.  labeled training \bel{set $\mathbb{S}$ with $N$ samples} and denote the empirical  error of a hypothesis $h$ on $\mathbb{S}$ by $\hat{R}_{\mathbb{S}}(h)$ to explicitly indicate its dependency on $\mathbb{S}$.

\bel{In the following, we show that an Empirical Risk Minimization algorithm can result in the hypothesis that best fits the training data but lacks the generalization aspect. As a possible solution, we present \emph{hypothesis selection} approaches that, while seeking a hypothesis that minimizes the ERM, it prevents overfitting the training data and promotes generalization for unseen situations \cite{Mohri2012bookML, Shalev2014bookML}. Such hypothesis selection methods are known under the name of \emph{inductive biases} \cite{Mohri2012bookML, Shalev2014bookML, Mitchell80indbias, Gordon1995}.}

\subsubsection{\bel{Inductive Bias}}
\label{subsub:inductivebias0}

\bel{\textit{Empirical Risk Minimization}} algorithm (ERM), which only seeks to minimize the error on the training samples \cite{Mohri2012bookML}
\begin{equation}
    \label{eq:eq0-4}
    h_{\mathbb{S}}^{ERM} = \argmin_{h \in \mathbb{H}}\hat{R}_{\mathbb{S}}(h) \; ,
\end{equation}
\noindent might not be successful, since it disregards the complexity term of $\mathbb{H}$ \cite{Mohri2012bookML}. In practice, the performance of the ERM algorithm is typically poor, particularly when using limited training data, \bel{since the learning algorithm may fail to find a hypothesis that is able to generalize well for data outside the training set \cite{Mohri2012bookML, Shalev2014bookML, Mitchell80indbias,Gordon1995}. This situation is known as \textit{overfitting}}. Additionally, in many cases, determining the ERM solution is computationally intractable. For example, finding a linear hypothesis with smallest error on the training samples is NP-hard (as a function of the dimension of the space) \cite{Mohri2012bookML}.

\bel{A common solution to the overfitting issue of ERM algorithm is to apply the ERM learning rule over a restricted search space \cite{Shalev2014bookML}. Formally, a set of predictors $\mathbb{H}$ is chosen in advance before seeing the data. Such prior restrictions are often called an \textit{inductive bias}. Since the choice of such restrictions is determined before the learning algorithm sees the training data, it should ideally be based on some prior knowledge about the problem in hand. The hope is that the learning algorithm will search for a hypothesis such that when the ERM predictor has good performance with respect to the training data, it is more likely to perform well over the underlying data distribution.

\emph{Inductive bias} of a learning algorithm, also known as \emph{learning bias}, can be defined as the set of assumptions that the learner uses to predict correct outputs given inputs that have not been seen before \cite{Mitchell80indbias}. Therefore, the aim of inductive bias is to promote the generalization. A \textit{bias} refers to any basis for choosing a generalization hypothesis over another, other than strict consistency with the observed training instances \cite{Mitchell80indbias, Utgoff1986}. A fundamental question is raised here: what kind of assumptions and over which hypothesis classes an ERM learning algorithm will not result in an overfitting?

In machine learning, different inductive biases can be found \cite{mitchell1997, Mitchell80indbias, Utgoff1986} including:
\begin{itemize}
\item Factual knowledge about the domain and the training data.
\item Maximum margin when attempting to separate two classes. The assumption behind is that distinct classes tend to be separated by wide boundaries. Such assumption is behind support vector machines \cite{Cortes1995svm}.
\item Minimum features: unless there is a good justification that a feature is useful, it should be deleted. This is the assumption behind feature selection approach \cite{Hastie2015book, James2014book, bermingham2015application}.
\item Manifold assumption and nearest neighbors: assumes that most of the samples in a small neighborhood in representation space belong to the same class \cite{chapelle06MIT}. Given a sample with an unknown class label, guessing that it belongs to the same class as the majority in its immediate neighborhood is a direct application to such assumption. $k$-nearest neighbors algorithm \cite{James2014book, Mohri2012bookML, Shalev2014bookML} is based on such assumption.
\item Minimum description length and the bias toward simplicity and generality: the minimum description length principal (MDL) is a formalization of \emph{Occam's razor} principle in which the best hypothesis for a given set of data is the one that leads to the best compression of data. This paradigm was introduced in \cite{Rissanen1978} and it is an important concept in information theory and computational learning theory \cite{Grunwald2007}. Considering that any set of data can be represented by a string of symbols from a finite alphabet, the MDL principle is based on the intuition that any regularity in a given set of data can be used to compress it, i.e., such data can be described using fewer symbols than needed to describe the data literally \cite{Grunwald05atutorial}. In inductive and statistical inference theory, such concept promotes the idea that all statistical learning is about finding regularities in data, and the best hypothesis to describe the regularities in data is also the one that is able to compresses the data the most. Theorem \ref{theo:theo0-0-1} and Theorem \ref{theo:theo0-0} are based on such concept where the notion $\log_2(\abs{\mathbb{H}})$ is used as a description language to measure the length of a hypothesis set. Such theorems suggest that having two hypotheses sharing the same empirical risk, the true error of the one that has shorter description can be bounded by a lower error value. Seeking a hypothesis that best fits the data while keeping its complexity low is known as \emph{regularization}. This aspect is described more in details in the next section.
\end{itemize}
A bias needs to be justified in order to be used to constrain the hypothesis search \cite{Mitchell80indbias}. While inductive biases can help preventing overfitting the training data, strong biases can prevent learning from data \cite{Shalev2014bookML}. Therefore, a tradeoff is required.

In the following, we present one of the well known and well studied inductive biases which based on preferring hypothesis with low complexity \cite{Hastie2015book, Shalev2014bookML, Mohri2012bookML}. Such bias is justified theoretically as well (Sec.\ref{sub:learningframework0}).
}

\subsubsection{Example of Inductive Bias: Preference of Hypothesis Sets with Low Complexity (Regularization)}
\label{subsub:regularizationexampleofindbias0}

While guarantees of Theorem \ref{theo:theo0-0} \bel{and Theorem \ref{theo:theo0-0-1} hold} for finite hypothesis sets, they already provide some useful insights for the design of \bel{learning algorithms}.  Similar guarantees hold in the case of infinite hypothesis sets \cite{Mohri2012bookML}. Such results invite us to consider \bel{in a learning context} two terms: the empirical error and a complexity term, which here are a function of $\abs{\mathbb{H}}$ and the sample size $N$.

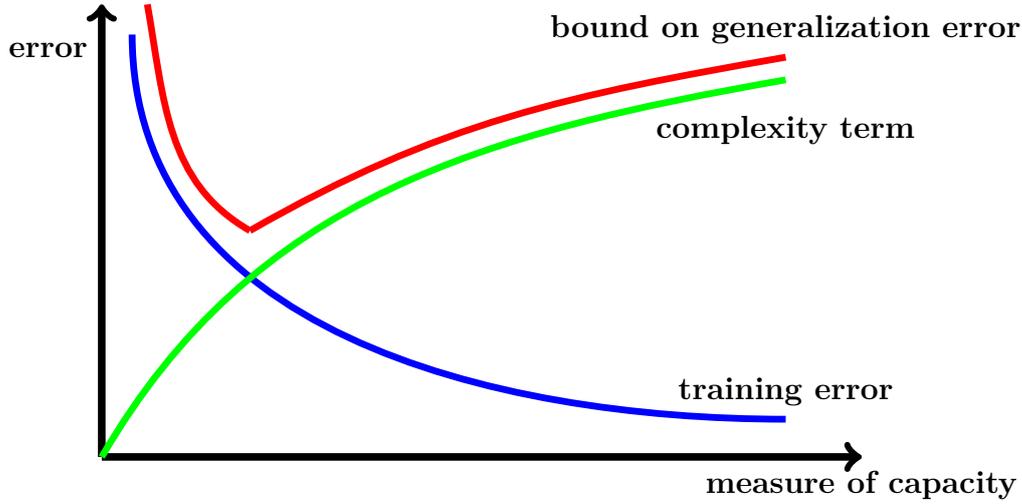
\begin{figure}[!htbp]
  \begin{center}
		\begin{tikzpicture}
  \coordinate (center) at (0, 0);
  \coordinate (x) at (10, 0);
  \coordinate (y) at (0, 6);
  \draw[->, line width=1.mm] (center) -- (x) node[align=right, below] (xlabel) {\textbf{measure of capacity}};
  \draw[->, line width=1.mm] (center) -- (y) node[left, pos=0.9] (ylabel) {\textbf{error}};
  \coordinate (trs) at (0.4, 5.6);
  \coordinate (tre) at (9, 0.5);
  \coordinate (cpxe) at (9, 5);
  \coordinate (bounds) at (0.6, 6);
  \coordinate (boundm) at (1.95, 3);
  \coordinate (bounde) at (9, 5.3);
    
  \draw[line width=0.9mm, color=blue] (trs) to[out=-90,in=180] (tre) node[align=right, above, color=black] (trainingerror) {\textbf{training error}} ; 
  
  \draw[line width=0.9mm, color=green] (center) to[out=60,in=190] (cpxe) node[below=0.3cm, color=black] (complexityterm) {\textbf{complexity term}}; 
  
  \draw[line width=0.9mm, color=red] (bounds) to[out=-80,in=150] (boundm);
  
  \draw[line width=0.9mm, color=red] (boundm) to[out=30,in=190] (bounde) node[above, color=black] (generalization) {\textbf{bound on generalization error}};
\end{tikzpicture}
	\end{center}
  \caption[Illustration of structural risk minimization (SRM).]{Illustration of structural risk minimization (SRM). The plots of three errors are shown as a function of a measure of capacity. As the size or capacity of the hypothesis set increases, the training error decreases, while the complexity term increases. SRM (shown in red) selects the hypothesis minimizing a bound on the generalization error, which is a sum of the empirical error, and the complexity term. (Reference: \cite{Mohri2012bookML})}
  \label{fig:fig0-1}
\end{figure}

\bel{\textit{Structural Risk Minimization} learning paradigm (SRM) comes to alleviate the overfitting issue of the ERM learning paradigm. While the ERM considers only the empirical error, the SRM considers the empirical error and the hypothesis set complexity which are both involved in bounding the generalization error. SRM consists in considering an infinite sequence of hypothesis sets with increasing sizes} \cite{Mohri2012bookML}
\begin{equation}
    \label{eq:eq0-5}
    \mathbb{H}_0 \subset \mathbb{H}_1 \subset \cdots \mathbb{H}_t \cdots  \; ,
\end{equation}
\noindent and find the ERM solution $h_t^{ERM}$ for each $\mathbb{H}_t$. The selected hypothesis is the one among the $h_t^{ERM}$ solutions with the smallest sum of the empirical error \bel{\textit{and}} a complexity term \bel{ $complexity(\mathbb{H}_t, N)$} that depends on the size (or more generally the capacity, that is, another measure of the richness of $\mathbb{H}$) of $\mathbb{H}_t$, and the sample size $N$ \cite{Mohri2012bookML}
\begin{equation}
    \label{eq:eq0-6}
    h_{\mathbb{S}}^{SRM} = \argmin_{\substack{h \in \mathbb{H} \\ t \in \N}} \hat{R}_{\mathbb{S}}(h) + complexity(\mathbb{H}_t, N) \; .
\end{equation}
\noindent Fig.\ref{fig:fig0-1} illustrates the SRM. While SRM benefits from strong theoretical guarantees, it is typically computationally expensive, since it requires \bel{to determine} the solution of multiple ERM problems \cite{Mohri2012bookML}.

\bel{Instead of performing an exhaustive search among different hypothesis sets $\mathbb{H}_t$ with increasing sizes $\abs{\mathbb{H}_t}$},  an alternative family of algorithms is based on a more straightforward optimization that consists \bel{in} minimizing \bel{simultaneously} the sum of the empirical error \bel{\textit{and}} a regularization term that penalizes the \bel{complexity} of the hypothesis \bel{ that belongs to a fixed hypothesis set $\mathbb{H}^{*}$. Therefore, it is natural to choose a hypothesis set $\mathbb{H}^{*}$ with large size}. One way to measure the complexity of a hypothesis is by counting \bel{its number of  parameters}. 
Let us consider $\bm{w}_h \in \R^m$ the set of parameters of a linear hypothesis \bel{$h \in \mathbb{H}^{*}$}. Therefore, hypothesis with few parameters are less complex than hypothesis with more parameters. Given that the optimal number of parameters is usually unknown in practice, a hypothesis \bel{set} with high complexity is usually used. However, such hypothesis is more likely to overfit the data, therefore \bel{fails} to generalize. An intuitive solution to this issue is to constrain the search algorithm to find a hypothesis $h$ with $\bm{w}_h$ that has only few non-zero entries. Setting a subset of the parameters to zero is a way of omitting them, i.e., reducing the complexity of the hypothesis since now it \bel{has less non-zero parameters}. This is known as sparsity \bel{and it} is motivated from signal approximation and compressed sensing \bel{domain}  \cite{Hastie2015book}. In practice, many classes of data are sparse \cite{Hastie2015book}, which means that only few components of the input data are relevant. \bel{Hence, sparse parameters are needed. We note that sparsity and feature selection are two related subjects \cite{Hastie2015book}.}

A straightforward way to constrain the search algorithm to find sparse \bel{solutions} is to count the non-zero entries of the parameters $\bm{w}_h$ which can be done using $L_0$ parameter norm

\begin{equation} 
  \label{eq:eq0-7-0-0}
  L_0(\bm{w}_h) = \norm{\bm{w}_h}_0 = \sum^m_{i=1} \ind_{w_{i_h} \neq 0} \; ,
\end{equation}
\noindent which counts the non-zero elements of $\bm{w}_h$ \cite{Candes2005}.

In terms of computational complexity, constraining the search algorithm using $L_0$ norm was shown to be NP-hard \cite{Natarajan1995}. It was shown in \cite{Candes2005, donoho2004, Hastie2015book} that $L_0$ and $L_p$ norm minimization,
\begin{equation}
  \label{eq:eq0-7-0-1}
  L_p(\bm{w}_h) = \norm{\bm{w}_h}_p = \sum^m_{i=1} \left(\abs{w_{i_h}}^p\right)^{1/p} \; ,
\end{equation}
for $p=1$, has identical solution under some conditions over $\bm{w}_h$. Hence, $L_0$ norm minimization can be relaxed with $L_1$ norm minimization. However, these conditions may be too strong for practical use. Instead, one may consider the sparse solution provided by a relaxed problem for a fixed $p$ with $0 < p < 1$ \cite{chartrand2007, chartrand2009, Chen-tech-report2009, fan2001norm, MouradR2010}. 
Nonetheless, $L_p$ \bel{norm minimization for $0 < p < 1$} is strongly NP-hard \cite{Ge2011}. However, any basic feasible solution of $L_0$ norm minimization is a local minimizer of $L_p$ norm minimization \bel{with} $0< p < 1$ \cite{Ge2011}. This is motivated by the fact that local minimizers are easy to certify and compute \cite{Ge2011}. Although $L_1$ is one of the most common norm used to constrain the hypothesis complexity \cite{Hastie2015book}, other norms with \bel{$p> 1$}, can be used \cite{terlaky1985, xue2000efficient} such as the popular $L_2$ norm \cite{tikhonov1963, Mohri2012bookML}.

To sum up, the regularized optimization problem which is composed of the empirical error and a regularization term, that is typically defined as $L_p(\bm{w}_h)$, can be written as \cite{Mohri2012bookML}
\begin{equation}
    \label{eq:eq0-7}
    h^{REG}_{\mathbb{S}} = \argmin_{h \in \mathbb{H}^{*}} \hat{R}_{\mathbb{S}}(h) + \lambda L_p(\bm{w}_h) \; ,
\end{equation}
\noindent where $\lambda \geq 0$ is a regularization parameter, which can be used to determine the trade-off between empirical error minimization and the model complexity. In practice, $\lambda$ is typically selected using $n$-fold cross validation.
\bigskip

\bel{Although, in the context of machine learning, regularization is best known for reducing the hypothesis complexity \cite{Hastie2015book, Shalev2014bookML}, it can go beyond that to reach the inductive bias definition (Sec.\ref{subsub:inductivebias0}). Therefore, regularization can be defined as any process that allows reducing the generalization error without necessarily reducing the empirical error. This is usually done by introducing prior knowledge that allows narrowing down the hypothesis space to specific subset. This prior knowledge may concern the complexity of the model, the data, or anything related to the task in hand \cite{Goodfellowbook2016}.

In the following, we provide a summary of this section about machine learning and generalization. 

}

}

\subsection{Summary}
\label{sub:summaryML0}
\bel{
In this section, we have seen that learning from data is a crucial part of today's technology. In order to make it work, some important questions must be addressed including what can a machine learn? how many samples are needed to guarantee a better generalization? how to deal with failure of generalization?. In the context of learning theory, we have presented the concept of learnability throughout the PAC learning framework which provides us general guidelines to build good learning algorithms.

We have seen that the generalization error upper bound depends on the number of training samples and the complexity of the hypothesis. Using the ERM learning paradigm to find the best hypothesis that fits the training samples can lead to poor results due to overfitting since ERM takes in consideration only the training samples. A possible and efficient solution to deal with the overfitting issue of the ERM is to restrict the hypothesis search space \cite{Shalev2014bookML}. For instance, one can choose a priori a hypothesis set $\mathbb{H}^{*}$ before seeing the training samples. Such priors are known as \emph{inductive bias}. Restricting the search space by picking such hypothesis set independently from data should ideally be done based on some prior knowledge about the problem to be learned. As an example of inductive bias, we have considered the prior that consists in preferring hypotheses with low complexity which is justified theoretically. As we have seen, a hypothesis with low complexity is in favor of generalization. However, such hypotheses may not be able to reduce the empirical error, i.e., explain the observed data. In the other hand, a hypothesis with high complexity may have low empirical error but it will increase the upper bound of the generalization error leading to overfitting. Therefore, a tradeoff between reducing the empirical error and the hypothesis complexity must be achieved. The search for a hypothesis with low complexity in the context of learning is known as regularization. This definition can be extended to include methods that promote the generalization aspect but without necessarily affecting the hypothesis complexity.}

We provide in Appendix \ref{chap:chapterapp1} more discussions about machine learning (Sec.\ref{sec:applicationsandproblemsapp1}, \ref{sec:definitionsandterminologyapp1}, \ref{sec:learningscenariosapp1}), optimization (Sec.\ref{sec:learningandoptimapp1}), and \bel{generalization} (Sec.\ref{sec:learningandgeneralizationapp1}).

In this thesis, we deal exclusively with neural network models. Therefore, we provide in the next section an introduction to the subject. We provide more details on such models in Appendix \ref{chap:chapterapp2}. If the reader is familiar with neural networks, we recommend skipping toward Sec.\ref{sec:neuralnetsregularization0} where we present different methods to improve the generalization performance of such models.
\section[Introduction to Feedforward Neural Networks]{Introduction to Feedforward Neural Networks}
\label{sec:neuralnetsintro0}

\bel{Artificial neural networks (ANN) are a particular type of parametrized models in machine learning}. We provide in this section an introduction to such models. We cover more technical details on neural networks in Appendix \ref{sub:appendix4fnn}.

While providing a presentation of ANN, \bel{we highlight also important historical key moments in neural networks origins. Extensive tracking of the history of neural networks and deep learning can be found in \cite{SchmidhuberNeuralnets2015}.}

\subsection{Early History}
\label{sub:ehistory0}
In 1943, neurophysiologist Warren McCulloch and mathematician Walter Pitts proposed a mathematical model \cite{McCulloch1943} of neurons in the brain based on threshold logic and demonstrated that combined together they can compute logic functions. They modeled a simple neural network through electrical circuit. Their model lacked a learning mechanism which is important to solve artificial intelligent problems.

In 1949, psychologist Donald Olding Hubb proposed a fundamental work \cite{hebb1949} about the learning process. His hypothesis states that knowledge and learning in the brain occurs primarily through formation and changes of synapses between neurons, known as synaptic plasticity. 

In 1957, psychologist Frank Rosenblat proposed the perceptron \cite{Rosenblatt58} model based on the work of Warren Mcculloch, Walter Pitts \cite{McCulloch1943} and Donald O. Hubb \cite{hebb1949}. Later, he published a book where he described in depth the perceptrons and their related proofs \cite{Rosenblatt1962}.

After 1967, research in neural networks stagnated after the work of Marvin Minsky and Seymour Papert \cite{minsky69perceptrons} who showed the limitation of the perceptrons which are unable to solve problems which are not linearly separable such as the simple XOR problem.

\bel{In the following, we present a brief and formal description of perceptrons.}

\subsection{Perceptron}
\label{sub:perceptron0}
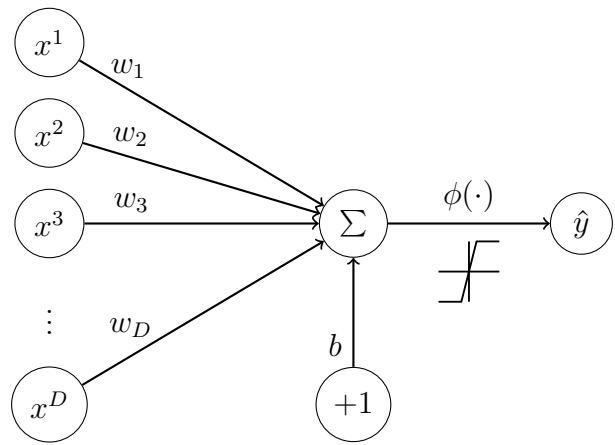
\begin{figure}[!htbp]
  \begin{center}
		\begin{tikzpicture}
  \coordinate (xn) at (0, 0);
  \coordinate (xdots) at (0, 1.2);
  \coordinate (x3) at (0, 2.4);
  \coordinate (x2) at (0, 3.6);
  \coordinate (x1) at (0, 4.8);
  \coordinate (p1) at (4, 0);
  \coordinate (sum) at (4, 2.4);
  \coordinate (y) at (7, 2.4);
  \draw (xn) node[circle, draw] (cxn) {$x^D$};
  \draw (xdots) node (cxdots) {$\vdots$};
  \draw (x3) node[circle, draw] (cx3) {$x^3$};
  \draw (x2) node[circle, draw] (cx2) {$x^2$};
  \draw (x1) node[circle, draw] (cx1) {$x^1$};
  
  \draw (sum) node[circle, draw] (csum) {$\sum$};
  \draw (y) node[circle, draw] (cy) {$\hat{y}$};
  \draw (p1) node[circle, draw] (cp1) {$+1$};
  
  \draw[->, thick] (cx1) -- (csum) node[pos=0.2, above] {$w_1$};
  \draw[->, thick] (cx2) -- (csum) node[pos=0.2, above] {$w_2$};
  \draw[->, thick] (cx3) -- (csum) node[pos=0.2, above] {$w_3$};
  \draw[->, thick] (cxn) -- (csum) node[pos=0.2, above=0.1] {$w_D$};
  
  \draw[->, thick] (csum) -- (cy) node[pos=0.5, above] (nonlinear) {$\phi(\cdot)$};
  
  \draw[->, thick] (cp1) -- (csum) node[pos=0.2, left] {$b$};
  \pic[below=10mm, scale=2] (two) at (nonlinear) {myactivation};
\end{tikzpicture}
	\end{center}
  \caption[\bel{Perceptron model.}]{Perceptron model. (Notation: $x^i$ is the $i^{th}$ component of $\bm{x}$ \bel{the same as $x_i$ in Eq.\ref{eq:eq0-27}}.)}
  \label{fig:fig0-6}
\end{figure}

A perceptron models one simple neuron. Formally, it is a simple function with an argument that is linear with respect to its input,
\begin{equation}
    \label{eq:eq0-27}
    \hat{y} = f(\xvec) = \phi(\xvec \cdot \bm{w} + b) = \phi(\sum_{i=1}^D x_i w_i + b) \; ,
\end{equation}
\noindent where $\xvec \in \R^D$ is an input vector, $\bm{w}$ is a vector of parameters known as weights. $b$ is a scalar parameter known as bias. $\phi(\cdot)$ is called an activation function and is typically nonlinear. If $\phi(\cdot)$ is the Heaviside step function, the perceptron has only two states: $0$ or $1$ which might indicate two different classes. An illustration of a simple perceptron is depicted in Fig.\ref{fig:fig0-6}.

To simplify notations, the weights vector $\bm{w}$ are extended by adding extra component to represent the bias $b$. Then, the input $\xvec$ is also extended by adding an additional component with value of $1$. From now on, we consider the extended notation unless we state the opposite.

Learning consists in searching for the best weights and bias that make $\hat{y}$ close to the target $y$. Rosenblatt training algorithm consists in updating the weights by increasing or decreasing $\bm{w}$ if the output $\hat{y}$ is smaller or greater than the target $y$ as follows,

\begin{equation}
    \label{eq:eq0-28}
    \bm{w} \leftarrow \bm{w} - (\hat{y} - y) \xvec \; ,
\end{equation}
\noindent where $y \in \{0, 1\}$, and $\hat{y}$ is the prediction. This algorithm continues learning as long as the model commits a mistake in classification. It has been shown in \cite{novikoff1962} that in the case of two linearly separable classes, this algorithm converges in a finite number of iterations. However, in the case where the two classes are not linearly separable, the algorithm never converges.

\begin{figure}[!htbp]
  \begin{center}
		\begin{tikzpicture}
  \coordinate (p1) at (0, 0);
  \coordinate (xn) at (0, 1.2);
  \coordinate (xdots) at (0, 2.4);
  \coordinate (x3) at (0, 3.6);
  \coordinate (x2) at (0, 4.8);
  \coordinate (x1) at (0, 6);
  
  \coordinate (sum1) at (3, 0.6);
  \coordinate (sumdots) at (3, 2.4);
  \coordinate (sum2) at (3, 3.6);
  \coordinate (sum3) at (3, 6);
  
  \coordinate (ym) at (6, 0.6);
  \coordinate (ydots) at (6, 2.4);
  \coordinate (y2) at (6, 3.6);
  \coordinate (y1) at (6, 6);
  
  \draw (p1) node[circle, draw] (cp1) {$+1$};
  \draw (xn) node[circle, draw] (cxn) {$x^D$};
  \draw (xdots) node (cxdots) {$\vdots$};
  \draw (x3) node[circle, draw] (cx3) {$x^3$};
  \draw (x2) node[circle, draw] (cx2) {$x^2$};
  \draw (x1) node[circle, draw] (cx1) {$x^1$};
  
  \draw (sum1) node[circle, draw] (csum1) {$\sum$};
  \draw (sumdots) node (csumdots) {$\vdots$};
  \draw (sum2) node[circle, draw] (csum2) {$\sum$};
  \draw (sum3) node[circle, draw] (csum3) {$\sum$};
  
  \draw (ym) node[circle, draw] (cym) {$\hat{y}^M$};
  \draw (ydots) node (cydots) {$\vdots$};
  \draw (y2) node[circle, draw] (cy2) {$\hat{y}^2$};
  \draw (y1) node[circle, draw] (cy1) {$\hat{y}^1$};

  \draw[->, thick] (cx1) -- (csum3) node[pos=0.5, above] {$\bm{W}$};
  \draw[->, thick] (cx1) -- (csum2) node {};
  \draw[->, thick] (cx1) -- (csum1) node {};
  
  \draw[->, thick] (cx2) -- (csum3) node {};
  \draw[->, thick] (cx2) -- (csum2) node {};
  \draw[->, thick] (cx2) -- (csum1) node {};
  
  \draw[->, thick] (cx3) -- (csum3) node {};
  \draw[->, thick] (cx3) -- (csum2) node {};
  \draw[->, thick] (cx3) -- (csum1) node {};
  
  \draw[->, thick] (cxn) -- (csum3) node {};
  \draw[->, thick] (cxn) -- (csum2) node {};
  \draw[->, thick] (cxn) -- (csum1) node {};
  
  \draw[->, thick] (cp1) -- (csum3) node {};
  \draw[->, thick] (cp1) -- (csum2) node {};
  \draw[->, thick] (cp1) -- (csum1) node {};

  \draw[->, thick] (csum3) -- (cy1) node[pos=0.5, above] (nonlinear1) {$\phi(\cdot)$};
  \draw[->, thick] (csum2) -- (cy2) node[pos=0.5, above] (nonlinear2) {$\phi(\cdot)$};
  \draw[->, thick] (csum1) -- (cym) node[pos=0.5, above] (nonlinear3) {$\phi(\cdot)$};
  
  \draw (y2) node (y) [right=10mm] {$\hat{\yvec}$};
  \draw (x3) node (x) [left=10mm] {$\xvec$};

\end{tikzpicture}
	\end{center}
  \caption[\bel{Multiclass perceptron.}]{Multiclass perceptron. (Notation: $x^i$ is the $i^{th}$ component of $\bm{x}$. $\hat{y}^j$ is the $j^{th}$ component of $\hat{\bm{y}}$).}
  \label{fig:fig0-7}
\end{figure}
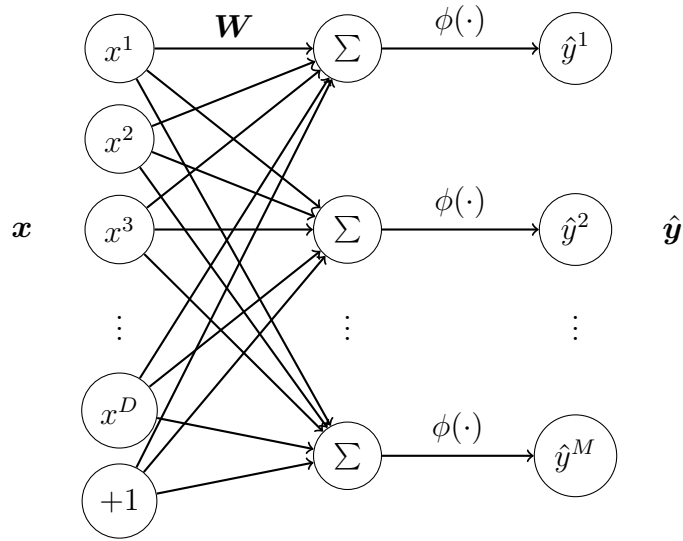

One perceptron can work only with two classes. It is possible to extend a perceptron to work with more than two classes, i.e., $M>2$. This can be done using $M$ perceptrons connected to the same input $\xvec$ but each one has its own weights and bias. By doing so, an output vector is obtained instead of one scalar,

\begin{equation}
    \label{eq:eq0-29}
    \hat{\yvec}: \hat{y}_j = f_j(\xvec), \forall j = 1, \dots, M \; .
\end{equation}
\noindent In this case, the predicted class $k$ is given by the class of the maximum output,

\begin{equation}
    \label{eq:eq0-30}
    k: \hat{y}_k \geq \hat{y}_j, \forall j=1, \dots, M \; .
\end{equation}
Fig.\ref{fig:fig0-7} illustrates an example of multiclass perceptron.

The training of multiclass perceptron is done by applying the peceptron learning rule described previously for each perceptron. Let us use a matrix notation where $\bm{W} \in \R^{(D+1) \times M}$ is the weights of the multiclass perceptron. $\xvec$ is vector of $1 \times (D+1)$\bel{,} $\hat{\yvec}-\yvec$ is a vector of errors with size $1 \times M$, $y_{j=1, \dots, M} \in \{0, 1\}$ is a vector of labels where $1$ indicates the correct class, and $\sum_{j=1}^M y_j = 1$ to make sure that every sample belongs to only one class. The update rule can be written as,
\begin{equation}
    \label{eq:eq0-31}
    \bm{W} \leftarrow \bm{W} - \xvec^{\top} \cdot (\hat{\yvec} - \yvec) \; .
\end{equation}

In 1957, Frank Rosenblatt implemented a multiclass perceptron in custom-built hardware under the name \quotes{Mark I} perceptron with $20\times 20$ inputs and $8$ output classes. Preceding this work by many years, in 1951, Marvin Minsky has implemented a hardware neural network based on memory and rewards named \quotes{SNARC} (Stochastic Neural Analog Reinforcement Calculator). This machine is considered one of the first pioneering works in the field of artificial intelligence.

\bel{In the next paragraph, we present a slightly different approach to train a perceptron based on gradient descent.}

\subsection{Gradient Based Learning}
\label{sub:gradientlearning0}
In 1960, Bernard Widrow and Marcian Hoff developed a linear model \cite{Widrow1960} similar to the perceptron but without the thresholding activation function. This allows using gradient descent method and using the derivatives. They named their model ADALINE for ADAptive LINear Elements. The model is an electrical circuit based on a new circuit called memistor which is a resistor with memory. For training their model, Widrow and Hoff proposed a slightly modified version of the perceptron learning rule. Instead of using the misclassification error, they proposed to use the squared error as \bel{an error measure for} each sample as follows,
\begin{equation}
    \label{eq:eq0-32}
    \ell(f(\xvec), \yvec) = \frac{1}{2} \sum^M_{j=1} (y_j - f_j(\xvec))^2 \; .
\end{equation}
\noindent Therefore, minimizing the mean squared error over all the training samples $N$ can be formulated as follows,

\begin{equation}
    \label{eq:eq0-33-0}
    L(\bm{W}) = \frac{1}{N} \sum^N_{i=1} \ell(f(\xvec^{(i)}), \yvec^{(i)}) \; .
\end{equation}
In this case, $\hat{\yvec}^{(i)} = f(\xvec^{(i)}) = \xvec^{(i)} \cdot \bm{W}$.

To train the ADALINE model, they propose to use gradient descent as follows,
\begin{equation}
    \label{eq:eq0-33}
    \bm{W} \leftarrow \bm{W} - \alpha \frac{\partial L(\bm{W})}{\partial \bm{W}} \; ,
\end{equation}
\noindent where $\alpha$ is a learning rate which controls the speed of convergence. This algorithm suggested to move the parameters weights in the direction that decreases the total error $L(\bm{W})$. This direction is obtained by computing the derivative of the total train loss $L(\bm{W})$ with respect to the weight vector using the chain rule. \bel{Using the linearity of the derivatives of Eq.\ref{eq:eq0-33-0}, we can compute the derivatives of one example $(\xvec, \yvec)$ in Eq.\ref{eq:eq0-32}, then we take the average over all the samples to obtain $\frac{\partial L}{\partial \bm{W}}$, the derivatives of the total loss in Eq.\ref{eq:eq0-33-0}. For one training sample, deriving the loss in Eq.\ref{eq:eq0-32} with respect to a parameter weight gives,}
\begin{equation}
    \label{eq:eq0-34}
    \frac{\partial \ell}{\partial W^{ij}} = \frac{\partial \ell}{\partial \hat{y}^j} \cdot  \frac{\partial \hat{y}^j}{\partial W^{ij}}, \quad \forall i=1, \dots, \bel{D+1}, \forall j=1, \dots, M \quad ,
\end{equation}
\noindent where,
\begin{equation}
    \label{eq:eq0-35}
    \frac{\partial \ell}{\partial \hat{y}^j} = \frac{\partial \frac{1}{2}(\hat{y}^j - y^j)^2}{\partial \hat{y}^j} = \hat{y}^j - y^j \Rightarrow \frac{\partial \ell}{\partial \hat{y}} = \hat{y} - y \; .
\end{equation}
\noindent The notation $x^i$ is the $i^{th}$ component of $\bm{x}$, $\hat{y}^j$ is the $j^{th}$ component of $\hat{\bm{y}}$, and $W^{ij}$ is the \bel{component} at the $i^{th}$ row and the $j^{th}$ coloumn of $\bm{W}$.

\noindent In the case where there is no activation function,
\begin{equation}
    \label{eq:eq0-36}
    \frac{\partial \hat{y}^j}{\partial W^{ij}} = \frac{\partial \left(\sum_k x^k W^{kj}\right)}{\partial W^{ij}} = x^i, \quad \forall i=1, \dots, \bel{D+1}, \forall j=1, \dots, M \; .
\end{equation}
\noindent This gives the delta rule learning algorithm,
\begin{equation}
    \label{eq:eq0-37}
    \bm{W} \leftarrow \bm{W} - \alpha \frac{\partial L}{\partial \bm{W}} = \bm{W} - \alpha \xvec^{\top} \cdot (\hat{\yvec} - \yvec) \; .
\end{equation}
\noindent One can see that the delta rule (Eq.\ref{eq:eq0-37}) is similar to the perceptron learning rule (Eq.\ref{eq:eq0-31}) except for the learning rate $\alpha$.

In the case where there is a differentiable activation function $\hat{y}^j = \phi(h^j = \sum_k x^k W^{kj})$, Eq.\ref{eq:eq0-36} can be developed using chain rule as follows,
\begin{align}
    \frac{\partial \hat{y}^j}{\partial W^{ij}} &= \frac{\partial \hat{y}^j}{\partial h^j} \cdot \frac{\partial h^j}{\partial W^{ij}} \label{eq:eq0-38-0} \\ 
    &=\frac{\partial \phi(h^j)}{\partial h^j} \cdot \frac{\partial \left(\sum_k x^k W^{kj}\right)}{\partial W^{ij}} \label{eq:eq0-38-1} \\
    &=\frac{\partial \phi(h^j)}{\partial h^j} \cdot x^i \label{eq:eq0-38} \; .
\end{align}
\noindent Therefore, the delta rule in this case is,
\begin{equation}
    \label{eq:eq0-39}
    \bm{W} \leftarrow \bm{W} - \alpha \xvec^{\top} \cdot \nabla_{\bm{h}}\phi(\bm{h}) \cdot (\hat{\yvec} - \yvec) \; .
\end{equation}
\noindent In the case of the perceptron, which uses the Heaviside step function, $\nabla_{\bm{h}}\phi(\bm{h})$ is not defined at zero and it is equal to zero everywhere else which makes the application of the delta rule on the perceptron impossible. This led to the use of differentiable functions such \bel{as} the sigmoid functions. \bel{Appendix \ref{subsub:appendix4nonlinearfunctions} covers more details about other activation functions. 

In the following, we present the extension of the perceptron to multilayer perceptron which is a critical change in the history of neural networks. We briefly highlight the historical reasons for such a major change.
}

\subsection{Multilayer Perceptron and Representation Learning}
\label{sub:multilayerperceptron0}
Although perceptrons seemed promising at the beginning, it was quickly shown that they could not be trained to separate every type of classes. In 1969, Marvin Minsky and Seymour Papert published their book \quotes{Perceptrons: An Introduction to Computational Geometry} \cite{minsky69perceptrons} which put an end to perceptrons. In this book, the authors pointed out fundamental limitations of the perceptron. For instance, a single perceptron can not solve an XOR problem. Moreover, they conjectured, mistakenly, that similar results would be found when using multilayer perceptrons. This book caused a significant decline in interest and funding of neural networks research.  This led to an abandonment of connectionism which was the other part of Artificial Intelligence with concurrence with symbolic reasoning which Minsky and Papert were part of. This major criticism participated in starting the AI winter\footnote{Between 1974-1980: AI winter is a period of reduced funding and interest in artificial intelligence research. At this period, AI has experienced several hype cycles, disappointment and criticism, followed by funding cuts. Years later, interest into AI was back.}. Three years later, Stephen Grossborg published a series of work introducing neural networks modeling XOR \cite{grossberg1973c}. In 1987, Minsky and Papert reprinted their book with the name \quotes{Perceptrons - Expanded Edition} \cite{Minsky1988} where some errors of the original book were shown and corrected. Despite this controversy, the reprinted version contains a handwritten dedication to Frank Rosenblatt who did not live to see it. As a side note, Minsky and Rosenblatt knew each other since adolescence. They studied at the same high school with one different year. However, they pursued different paths in AI research. While Minsky promoted symbolism, Rosenblatt promoted connectionism and learning. More on this controversy can be found in \cite{Olazaran1996}.

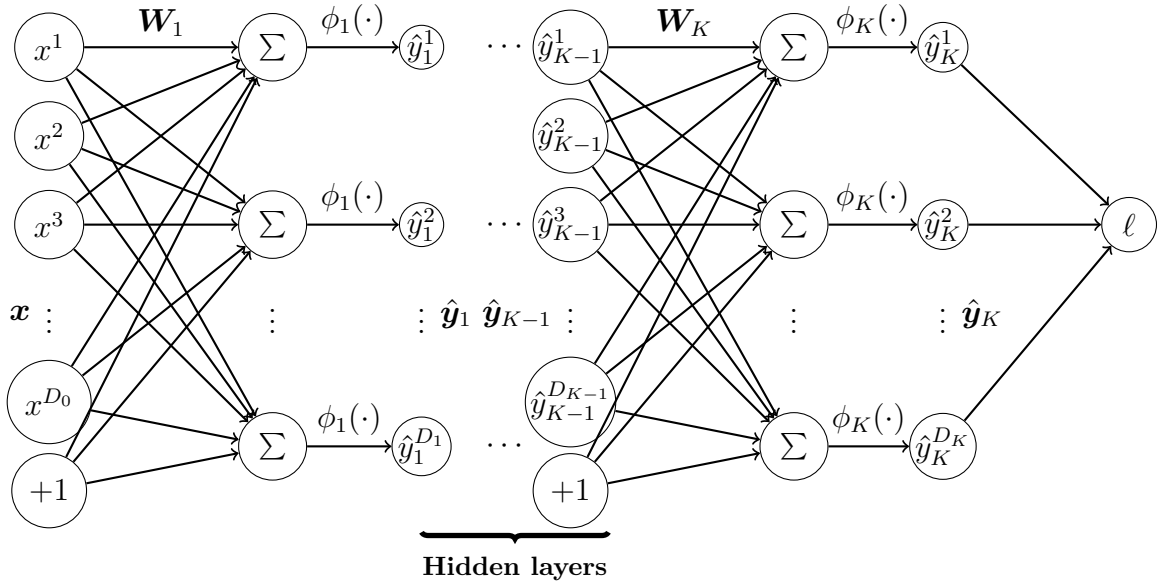
\begin{figure}[!htbp]
  \begin{center}
		\begin{tikzpicture}[scale=0.98]
  \coordinate (p1) at (0, 0);
  \coordinate (xn) at (0, 1.2);
  \coordinate (xdots) at (0, 2.4);
  \coordinate (x3) at (0, 3.6);
  \coordinate (x2) at (0, 4.8);
  \coordinate (x1) at (0, 6);
  
  \coordinate (sum1) at (3, 0.6);
  \coordinate (sumdots) at (3, 2.4);
  \coordinate (sum2) at (3, 3.6);
  \coordinate (sum3) at (3, 6);
  
  \coordinate (ym) at (5, 0.6);
  \coordinate (ydots) at (5, 2.4);
  \coordinate (y2) at (5, 3.6);
  \coordinate (y1) at (5, 6);
  
  \draw (p1) node[circle, draw] (cp1) {$+1$};
  \draw (xn) node[circle, draw] (cxn) {$x^{D_0}$};
  \draw (xdots) node (cxdots) {$\vdots$};
  \draw (x3) node[circle, draw] (cx3) {$x^3$};
  \draw (x2) node[circle, draw] (cx2) {$x^2$};
  \draw (x1) node[circle, draw] (cx1) {$x^1$};
  
  \draw (sum1) node[circle, draw] (csum1) {$\sum$};
  \draw (sumdots) node (csumdots) {$\vdots$};
  \draw (sum2) node[circle, draw] (csum2) {$\sum$};
  \draw (sum3) node[circle, draw] (csum3) {$\sum$};
  
  \draw (ym) node[circle, draw, inner sep=0pt] (cym) {$\hat{y}^{D_1}_1$};
  \draw (ydots) node (cydots) {$\vdots$};
  \draw (y2) node[circle, draw, inner sep=0pt] (cy2) {$\hat{y}^2_1$};
  \draw (y1) node[circle, draw, inner sep=0pt] (cy1) {$\hat{y}^1_1$};

  \draw[->, thick] (cx1) -- (csum3) node[pos=0.5, above] {$\bm{W}_1$};
  \draw[->, thick] (cx1) -- (csum2) node {};
  \draw[->, thick] (cx1) -- (csum1) node {};
  
  \draw[->, thick] (cx2) -- (csum3) node {};
  \draw[->, thick] (cx2) -- (csum2) node {};
  \draw[->, thick] (cx2) -- (csum1) node {};
  
  \draw[->, thick] (cx3) -- (csum3) node {};
  \draw[->, thick] (cx3) -- (csum2) node {};
  \draw[->, thick] (cx3) -- (csum1) node {};
  
  \draw[->, thick] (cxn) -- (csum3) node {};
  \draw[->, thick] (cxn) -- (csum2) node {};
  \draw[->, thick] (cxn) -- (csum1) node {};
  
  \draw[->, thick] (cp1) -- (csum3) node {};
  \draw[->, thick] (cp1) -- (csum2) node {};
  \draw[->, thick] (cp1) -- (csum1) node {};

  \draw[->, thick] (csum3) -- (cy1) node[pos=0.5, above] (nonlinear1) {$\phi_1(\cdot)$};
  \draw[->, thick] (csum2) -- (cy2) node[pos=0.5, above] (nonlinear2) {$\phi_1(\cdot)$};
  \draw[->, thick] (csum1) -- (cym) node[pos=0.5, above] (nonlinear3) {$\phi_1(\cdot)$};
  
  \draw (ydots) node (y) [right=1mm] {$\hat{\yvec}_1$};
  \draw (xdots) node (x) [left=1mm] {$\xvec$};
  
  \draw (cy1) node (cdotsh1) [right=7mm] {$\cdots$};
  \draw (cy2) node (cdotsh2) [right=7mm] {$\cdots$};
  \draw (cym) node (cdotshm) [right=7mm] {$\cdots$};
  \coordinate (p12) at (7, 0);
  \coordinate (xn2) at (7, 1.2);
  \coordinate (xdots2) at (7, 2.4);
  \coordinate (x32) at (7, 3.6);
  \coordinate (x22) at (7, 4.8);
  \coordinate (x12) at (7, 6);
  
  \coordinate (sum12) at (10, 0.6);
  \coordinate (sumdots2) at (10, 2.4);
  \coordinate (sum22) at (10, 3.6);
  \coordinate (sum32) at (10, 6);
  
  \coordinate (ym2) at (12, 0.6);
  \coordinate (ydots2) at (12, 2.4);
  \coordinate (y22) at (12, 3.6);
  \coordinate (y12) at (12, 6);
  
  \draw (p12) node[circle, draw] (cp12) {$+1$};
  \draw (xn2) node[circle, draw, inner sep=0pt] (cxn2) {$\hat{y}^{D_{K-1}}_{K-1}$};
  \draw (xdots2) node (cxdots) {$\vdots$};
  \draw (x32) node[circle, draw, inner sep=0pt] (cx32) {$\hat{y}^3_{K-1}$};
  \draw (x22) node[circle, draw, inner sep=0pt] (cx22) {$\hat{y}^2_{K-1}$};
  \draw (x12) node[circle, draw, inner sep=0pt] (cx12) {$\hat{y}^1_{K-1}$};
  
  \draw (sum12) node[circle, draw] (csum12) {$\sum$};
  \draw (sumdots2) node (csumdots2) {$\vdots$};
  \draw (sum22) node[circle, draw] (csum22) {$\sum$};
  \draw (sum32) node[circle, draw] (csum32) {$\sum$};
  
  \draw (ym2) node[circle, draw, inner sep=0pt] (cym2) {$\hat{y}^{D_K}_K$};
  \draw (ydots2) node (cydots2) {$\vdots$};
  \draw (y22) node[circle, draw, inner sep=0pt] (cy22) {$\hat{y}^2_K$};
  \draw (y12) node[circle, draw, inner sep=0pt] (cy12) {$\hat{y}^1_K$};

  \draw[->, thick] (cx12) -- (csum32) node[pos=0.5, above] {$\bm{W}_K$};
  \draw[->, thick] (cx12) -- (csum22) node {};
  \draw[->, thick] (cx12) -- (csum12) node {};
  
  \draw[->, thick] (cx22) -- (csum32) node {};
  \draw[->, thick] (cx22) -- (csum22) node {};
  \draw[->, thick] (cx22) -- (csum12) node {};
  
  \draw[->, thick] (cx32) -- (csum32) node {};
  \draw[->, thick] (cx32) -- (csum22) node {};
  \draw[->, thick] (cx32) -- (csum12) node {};
  
  \draw[->, thick] (cxn2) -- (csum32) node {};
  \draw[->, thick] (cxn2) -- (csum22) node {};
  \draw[->, thick] (cxn2) -- (csum12) node {};
  
  \draw[->, thick] (cp12) -- (csum32) node {};
  \draw[->, thick] (cp12) -- (csum22) node {};
  \draw[->, thick] (cp12) -- (csum12) node {};

  \draw[->, thick] (csum32) -- (cy12) node[pos=0.5, above] (nonlinear12) {$\phi_K(\cdot)$};
  \draw[->, thick] (csum22) -- (cy22) node[pos=0.5, above] (nonlinear22) {$\phi_K(\cdot)$};
  \draw[->, thick] (csum12) -- (cym2) node[pos=0.5, above] (nonlinear32) {$\phi_K(\cdot)$};

  \draw (ydots2) node (yy2) [right=1mm] {$\hat{\yvec}_{K}$};
  \draw (xdots2) node (xx2) [left=1mm] {$\hat{\yvec}_{K-1}$};
  
  \coordinate (L) at (14.5, 3.6);
  \draw (L) node[circle, draw] (cL) {$\ell$};
  
  \draw[->, thick] (cy12) -- (cL);
  \draw[->, thick] (cy22) -- (cL);
  \draw[->, thick] (cym2) -- (cL);

  \coordinate (fictif) at (5, 0);
  \draw [thick,decoration={brace,mirror,raise=0.55cm}, decorate, yshift=2ex, line width=2pt]
(fictif.west) -- (cp12.east) node [pos=0.5,anchor=north,yshift=-0.7cm] {\footnotesize
\textbf{Hidden layers}};

\end{tikzpicture}
	\end{center}
  \caption[\bel{Multilayer perceptron.}]{Multilayer perceptron. (Notation: $D_k, k=1, \cdots, K$ is the dimension of the output of the layer $k$. $D_0=D$ is the dimension of the input $\bm{x}$ of the network. $D_K=M$ is the dimension of the output $\hat{\bm{y}}$ of the network, i.e., $M$. $x^i$ is the $i^{th}$ component of $\bm{x}$. $\hat{y}^j_k$ is the $j^{th}$ component of the output representation $\hat{\bm{y}}_k$ at the layer $k$).}
  \label{fig:fig0-8}
\end{figure}

Despite this pessimism toward perceptrons, the book of Minsky and Papert provided new insights and research directions to improve them. The main result is that the perceptron fails in many recognition tasks not because of the learning algorithm but because of its lack of \fromont{represention} the required knowledge about the task to be solved. The authors stated that no machine can learn to recognize an object unless it possesses, at least potentially, some scheme for representing the object. In the case of the perceptron, Minsky and Papert pointed out that if there is a layer of simple perceptron-like hidden units which can recode the input pattern into an internal representation, there is always a recoding in this hidden representation that can support any required mapping from the input to the output. We note that at that time, few networks use this technique such as MADALINE \cite{Winter1988} which has a different training algorithm than the perceptron rule referred to as MRII algorithm for MADALINE Rule II. MADALINE consists basically of two sequential layers where each one is composed of multiple ADALINE neurons \cite{Widrow1960} that are followed by a threshold function. The use of threshold functions prevents MADALINE from using gradient based training algorithms. MRII algorithm uses instead the principle of minimal disturbance \cite{Ridgway1962} where the network parameters are disturbed whenever there is a mistake in the output. MADALINE is an extension of the two-layer network of Ridgway \cite{Ridgway1962} which is based on two layers: an adaptive layer that contains multiple ADALINE neurons followed each by a threshold function; then a fixed logic layer that takes the output of the previous ADALINEs as input to provide the final output. The logic layer is a simple logic function: \quotes{AND}, \quotes{OR}, and majority vote. MADALINE goes further by implementing such logic functions using ADALINE neurons and provides a learning algorithm \bel{for} a multilayer network. Earlier to that, particularly in the beginning of the $60^{\prime}$s, we find Gamaba machines \cite{Gamba1961} as a two perceptrons machine where the output of the first one is fed to the second one. The main issue at the time is that there was no strong learning algorithm to learn networks with hidden units. Moreover, the computation power required by such networks exceeds what was available at the moment. Neural networks field had to wait until the arrival of the backpropagation algorithm (Appendix \ref{subsub:appendix4backprop}) for training multilayer networks. An illustration of a multilayer perceptron is depicted in Fig.\ref{fig:fig0-8}.

We discuss in Appendix \ref{sub:appendix4fnn} more details on neural networks including backpropagation algorithm (Sec.\ref{subsub:appendix4backprop}), nonlinear activations (Sec.\ref{subsub:appendix4nonlinearfunctions}), universal approximation properties and depth (Sec.\ref{subsub:appendix4univapproxtheoremanddepth}), and other neural architectures (Sec.\ref{subsub:appendix4otherneuralarchitectures}).

\bel{Neural network domain kept struggling in the late of $90^{\prime}$s and the beginning of $2000^{\prime}$s for many reasons including the lack of data, practical issues in optimization algorithms, and most importantly the lack of computation power which slowed down research. It was until around $2006$ that neural network field entered a new era which moved neural network models from shallow, i.e., few layers, to deep, i.e., many layers, which led to a spiking success in the history of neural networks. Such success has attracted the research community to start working again on such models. Started in the beginning of $2000^{\prime}$s, the expression \quotes{\emph{deep learning}}\footnote{\bel{The term \quotes{\emph{deep learning}} was introduced to the machine learning community by Rina Dechter in 1986 \cite{Dechter86}, and to artificial neural networks by Igor Aizenberg \cite{Aizenberg2000, Gomez2005CRN} in 2000.}} was coined to broadly describe neural based models that use many layers to learn hierarchical representations. Aside from the advances in optimization algorithms that pushed deep learning field forward, such success could not be done without: the modern computational power that speeds up drastically the training and inference of deep models and the availability of massive supervised data in many domains such as computer vision, natural language processing, and voice recognition. In the next paragraph, we present a modern version and current advances in deep learning field.}

\FloatBarrier

\subsection{Deep Learning: from Late $60^{\prime}$s to Today}
\label{sub:deeplearning0}
\bel{
Deep learning, also known as deep structural learning or hierarchical learning, is part of a broader family of machine learning algorithms based on learning data representations where learning can be achieved through supervised, semi-supervised or unsupervised approaches.

\cite{Deng2014DLM} defines deep learning as a class of machine learning algorithms that: \begin{itemize*} \item use a cascade of multiple layers of nonlinear processing units for feature extraction and transformation. In most cases, each layer uses the output of the previous layer as input. \item learning such layers is done in a supervised and/or unsupervised fashion. \item each layer is seen as an abstraction level of representation. Stacking the layers forms a hierarchy of concepts that are built up from lower levels  toward more abstract ones \cite{Bengio2013RLR, bengio09}. The assumption underlying distributed representations is that observed data are generated by the interaction of layered factors \cite{Hinton1986PRALLEL, Bengio2013RLR}. Deep learning adds the assumption that such layers of factors correspond to levels of abstraction or composition. Varying number of layers and layer sizes can provide different degrees of abstraction \cite{Bengio2013RLR}. \end{itemize*}

\cite{SchmidhuberNeuralnets2015} introduce the notion of \emph{credit assignment path (CAP)} to define what is a \emph{shallow} and \emph{deep} model. The CAP is defined as a chain of transformations from the input to the output to describe potentially causal connections between the input and the output. For instance, feedforward neural networks have a CAP with a depth equals to the number of hidden layers plus one (the output layer). For recurrent networks (Appendix \ref{subsub:appendix4otherneuralarchitectures}), the CAP depth is potentially unlimited \cite{SchmidhuberNeuralnets2015}. Although there is no universal agreement upon threshold of CAP depth to divide shallow learning from deep learning, most researchers agree that deep learning involves a CAP depth greater than 2 which has been shown to be a universal approximation in a sense that it can emulate any function (Appendix \ref{subsub:appendix4univapproxtheoremanddepth}).

Neural networks trained by the \emph{Group Method of Data Handling} (GMDH) \cite{ivakhnenko1965, ivakhnenko1967} were perhaps the first deep learning models based on feedforward multilayer perceptron. Although, the units of their networks may have polynomial activation functions implementing \emph{Kolmogorov-Gabor polynomials} \cite{ivakhnenko1971} to introduce nonlinearity. Such activation functions are different than the other widely used nowadays. Their deep networks are incrementally trained then pruned. \cite{ivakhnenko1971} describe a network with 8 layers. A presentation of \emph{Kolmogorov-Gabor polynomials} can be found in \cite{wang2005self}.

Aside from deep GMDH networks, the \emph{Neocognitron} \cite{Fukushima1979neocognitron}, by Fukushima, was maybe the first artificial neural network that deserves the name \emph{deep}. It was the first to incorporate the neurophysiological insights about the visual cortex found around the 60$^{\prime}$s \cite{wiesel1959, Hubel62} into a learning framework. \emph{Neocognitron} introduced convolution layers composed of a set of convolution operators parametrized with a weights, under the form of rectangular matrix, that are duplicated over the 2D input through a shifting process. The same model introduced subsampling\footnote{\bel{A subsampling layer is an average pooling layer performs down-sampling by dividing the input into rectangular pooling regions and computing the average values of each region.}}, known also as downsampling, to promote a certain insensitivity to small shifts in the 2D input image. \emph{Neocognitron} is very similar to the feedforward, gadient-based, and backpropagation-based convolutional neural networks. However, Fukushima did not set the weights by backpropagation but by local Winner-Take-All-based unsupervised learning rule\footnote{\bel{Winner-Take-All (WTA) is a computational principle applied in computational models of neural networks by which neurons in a layer compete with each other for activation. In a classical setup, only the neuron with the highest activation stays active while all the other neurons shut down. However, other variations may allow more than one neuron to be active. In the theory of artificial neural networks, WTA networks are a case of competitive learning in recurrent neural networks. Output nodes in the network mutually inhibit each other, while simultaneously activating themselves through reflexive connections \cite{Grossberg1982, OsterDL09neco}.}} \cite{Fukushima2013b} or by pre-wiring. Therefore, deep learning issues did not matter (Appendix \ref{subsub:appendix4backprop}). For down-sampling, Fukushima used spatial averaging \cite{fukushima1980, Fukushima2011} instead of max-pooling\footnote{\bel{A max-pooling layer is a layer that performs down-sampling by dividing the input into rectangular pooling regions and computing the maximum values of each region.}} that is well known in modern convolutional networks. 

In 1989, backpropagation algorithm (Appendix .\ref{subsub:appendix4backprop}) was successfully applied to \emph{Neocognitron}-like model with weights sharing and convolutional layers \cite{leCun1989, LeCun90, lecun98gradient} (Fig.\ref{fig:fig0-6-00-00-lenet}). The purpose of the application is to recognize handwritten zip codes on mail. Their algorithm required 3 days of training. 

\begin{figure}[!htbp]
  \begin{center}
		\includegraphics[scale=0.8]{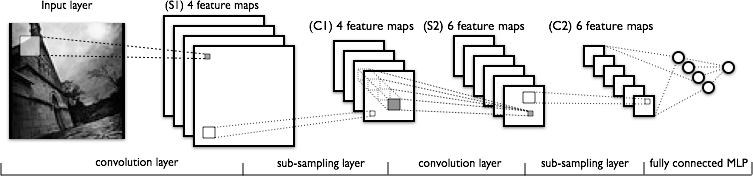}
	\end{center}
  \caption[\belx{\emph{Lenet} convolutional network.}]{Architecture of \emph{lenet} convolutional network \cite{lecun98gradient} which is composed of: two convolutional layers followed by a pooling layer each; followed by two dense layers. (Source: Deep Learning Tutorial: \href{http://deeplearning.net/tutorial/lenet.html}{http://deeplearning.net/tutorial/lenet.html}).}
  \label{fig:fig0-6-00-00-lenet}
\end{figure}

Inspired by \emph{Neocognitron}, \emph{Cresceptron} was born in 1992 \cite{weng1992} which adapts its topology during training. Instead of using local subsampling or WTA methods \cite{fukushima1980, Schmidhuber89cs}, the \emph{Cresceptron} introduced, for the first time, max-pooling layers. A more complex version of \emph{Cresceptron} was proposed which includes blurring layers to improve object location tolerance \cite{weng1997}.



By the late of the $80^{\prime}$s, experiments had shown that traditional deep forward networks or recurrent networks (Appendix \ref{subsub:appendix4backprop}) are hard to train using backpropagation. It was until 1991, that this issue was understood and the reason for such issue is now known by the name of the vanishing/explosion of the gradient (Appendix \ref{subsub:appendix4backprop}). Unsupervised pre-training was a major tool to deal with such an issue in feedforward networks and recurrent networks as well \cite{ballard1987modular, schmidhuber1992}. Long Short-term memory (LSTM) recurrent networks helped as well avoiding gradient problems and allow training very deep learning models \cite{Hochreiter1997LSM, SchmidhuberNeuralnets2015}.

Closely related works to \cite{ballard1987modular} have appeared in post-2000. For instance, in 2006, \cite{hinton06, Hinton2007, bengio12} showed that many layered feedforward networks could be effectively trained by pre-training one layer at a time, treating each layer as an unsupervised restricted Boltzmann machine, then fine-tune the whole network using supervised backpropagation. This is referred to as training of deep belief networks. Similar ideas were proposed in \cite{rifaiVMGB11, salah2011} which helped training deep networks.

For a long time, the slow computation power stepped in the way of training deep neural models \cite{SchmidhuberNeuralnets2015, Deng2014DLM}. Advances in hardware renewed the interest in such domain. In 2009, Nvidia\footnote{\bel{Nvidia Corporation is an American technology company that designs graphics processing units (GPUs) for the gaming, cryptocurrency, and professional markets, as well as system on a chip units (SoCs) for the mobile computing and automotive market. Website: \href{http://www.nvidia.com/}{http://www.nvidia.com}}} was involved in what was called the \quotes{\emph{big bang}} of deep learning. Their GPUs have speed up drastically the training of deep networks and allowed going deeper in a reasonable time. In particular, GPUs are well-suited for the matrix/vector operations involved in machine learning \cite{Steinkrau2005UGM, Ciresan2010DBS, Raina2009LDU}. Specialized hardware and optimization algorithms can be used for efficient processing \cite{szegpus2017}.


Aside from the increase of computational power that speeds up training deep networks and the increase of the available supervised data in certain applications, which make training deep architectures practical \cite{SchmidhuberNeuralnets2015}, there have been, in the last few years, many advances in optimization approaches that helped: \begin{itemize*} \item  speeding the training by improving gradient descent approach such as momentum \cite{sutskever1, polyak1964some}, Adagrad \cite{duchi1, Dean2012xx}, Adadelta \cite{zeiler1}, Adam \cite{KingmaB14, dozat2016incorporating}, and possibly the use of second order optimization such as Hessian-free optimization \cite{Moller93, Martens10, Martens2011hessfreeICML}. \item improving the generalization and avoid overfitting such as dropout \cite{SrivastavaDropout2013, srivastava14a} and batch normalization \cite{IoffeICML15}. \item  avoiding the vanishing of the gradient by introducing new activation functions that do not saturate such as rectifier \cite{NairHicml10, AISTATS2011GlorotBB11, GoodfellowWMCB13ICML} (Appendix \ref{subsub:appendix4nonlinearfunctions}). \end{itemize*}

Nowadays, different deep neural models including feedforward and recurrent networks have been successfully applied to different tasks including: computer vision \cite{ciresan12, krizhevsky12}, speech recognition \cite{Deng2014DLM, SchmidhuberNeuralnets2015}, natural language processing and machine translation \cite{Socheretal2013, ShenHGDM14cikm}, visual art processing\footnote{\bel{\quotes{\emph{DeepDream}} is a computer vision program created by Google engineer Alexander Mordvintsev which uses a convolutional network to find and enhance patterns in images via algorithmic pareidolia, thus creating a dream-line hallucinogenic appearance in the deliberately over-processed image.}} \cite{szegedyLJSRAEVR14, SimonyanVZ13}, recommendation systems \cite{WangW14acmm, OordDS13nips}, image restoration \cite{UlyanovVL17, ZhangZGZ17cvpr}, social network filtering \cite{NguyenAOI17corrfiltering}, bioinformatics and drug design \cite{Chicco2014DAN, Sathyanarayana2016JMIR}, where they have produced results comparable and in some cases superior to human experts \cite{ciresan12, krizhevsky12}.

In the last four years, generative models based on neural networks have been a hot topic particularly using Generative Adversarial Network (GANs) \cite{GoodfellowNIPS2014} (Appendix \ref{subsub:appendix4otherneuralarchitectures}) which is an area of deep learning that is growing rapidly. Hinton, one of the founders of neural networks, has recently proposed a new architecture named  \quotes{\emph{capsules}} \cite{SabourFH17, capsulese2018matrix} in an attempt to solve a convolutional network issue related to its lack of taking in consideration the spatial relations between the parts of an object. Deep reinforcement learning is another breakthrough of deep learning models that is making a big step to improve artificial intelligence and get computers to learn like humans, without explicit instructions \cite{Arulkumaran2017corr, Andersen2018corr, MnihKSRVBGRFOPB15Nature}.  Many research teams focus now on developing systems capable of learning how to play \emph{ATARI} video games using only pixels as data input \cite{mnihatari2013, HosuR16corr}. Autonomous vehicles driving is also an area where deep reinforcement learning is making progress \cite{Fridman2018corr, chi2017corrdeeptesla, yu2016deep}. Deep reinforcement learning has been used to learn \emph{Go}\footnote{\bel{\emph{Go} is an abstract strategy board game for two players, in which the aim is to surround more territory than the opponent.}} game well enough to beat a professional \emph{Go} player \cite{SilverHMGSDSAPL16}.

\cite{SchmidhuberNeuralnets2015, Deng2014DLM} provide a detailed and extensive presentation of the history and applications of deep learning methods.

We mention that most of the success of neural networks today is due to the depth of their architectures and not the width of their layers \cite{Goodfellowbook2016} (Fig.\ref{fig:fig0-6-00-01-resnet}). We cover this aspect with more details in Appendix \ref{subsub:appendix4univapproxtheoremanddepth}. \cite{pmlrpandey14} provide a discussion on the question of which one to use: deep or wide learning mechanism?

\begin{figure}[!htbp]
\begin{center}
\vspace{.5em}
\includegraphics[scale=1.]{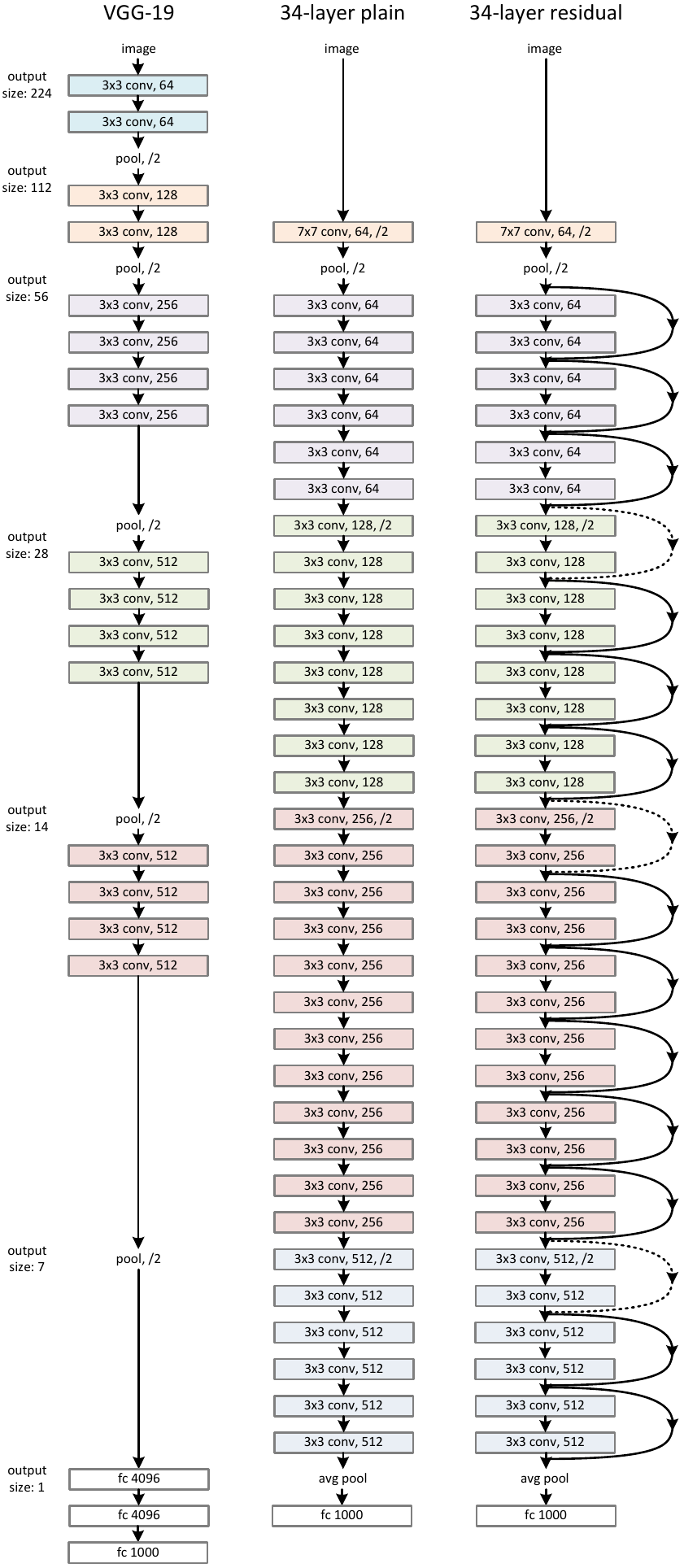}
\end{center}
\caption[\belx{Example of \emph{Deep Residual} network architectures for image recognition.}]{Example of \emph{Deep Residual} network architectures for image recognition. \emph{Left}: the VGG-19 model \cite{SimonyanZ14aCORR}. \emph{Middle}: a plain network with 34 parameter layers. \emph{Right}: a residual network with 34 parameter layers. (Credit: \cite{heZRS16})}
\label{fig:fig0-6-00-01-resnet}
\vspace{-1em}
\end{figure}

\FloatBarrier

\textbf{Deep Learning Issues and Criticism}
\bigskip

Although deep learning models have achieved high performance in many applications, they still display problematic and questionable behaviors such as classifying unrecognizable images as belonging to a familiar category of ordinary images or misclassifying small perturbed images that have been correctly classified \cite{Geortzel2015, NguyenYC15cvpr, SzegedyZSBEGF13CORR}. As deep learning models move from the laboratory to the real world, such misbehavior can cause a serious security threat. For instance, an attacker can add a specific type of noise to an image so that the human eye does not notice it, but it will cause the neural network to produce a completely wrong prediction. Such attacks are known by \quotes{\emph{adversarial attacks}}. A neural network can be trained to minimize its error over such attacks \cite{SzegedyZSBEGF13CORR, GoodfellowSS14CORR}. Another serious and old issue that is known to the research community is that neural networks memorize chunks of training data \cite{hansel1992memorization, ArpitJBKBKMFCBL17ICML, carlini2018memorization}. This phenomenon is not well-understood. Language models are probably the most vulnerable type of models at the moment while it is harder over images \cite{carlini2018memorization}. This shows again how deep learning models are vulnerable to information leakage. Using particular search algorithms, an attacker can retrieve sensitive data such as text messages, emails, medical data, etc. An important work that changed the research community understanding of the generalization aspects in deep learning models has suggested that \quotes{brute-force memorization} may be part of an effective learning strategy for deep neural networks on real data, implying that generalization and memorization are not necessarily opposed \cite{zhangBHRV16}. However, Not all the community seems to agree that neural networks memorize the training data \cite{krueger2017deep}. There may be ways to get around the memorization issue. The researchers recommend developers to use \quotes{\emph{differential privacy learning algorithms}}\footnote{\bel{In cryptography, \emph{Differential privacy} is a mathematical definition for the privacy loss that results to individuals when their private information is used in the creation of a data product. It aims to provide means to maximize the accuracy of queries from statistical databases while minimizing the chances of identifying its records \cite{DworkMNS06tcc, Dwork11cacm, DworkR14fttcs}.}} \cite{Abadi2016DLD, PapernotAEGT16corr} which are currently applied in big companies that often deal with private information of the users which is mostly textual. Another way is to scramble and randomize private information so that it is difficult to reproduce.

Neural network field has seen many criticism, and still do, since its birth \cite{minsky69perceptrons, Minsky1988}. The main criticism concerns the lack of theoretical foundations. Deep learning models are often looked at as a \quotes{\emph{black box}}, with most confirmations are done empirically rather than theoretically. Modern critics such as Gary Marcus, pointed out that deep learning should be looked at as a step towards realizing strong artificial intelligence, and not as an entire solution \cite{marcus2018corr, marcus2003algebraic}. Despite the power and the success of deep learning methods, they still lack much of the functionalities needed for attaining the goal \fromont{of} artificial intelligence such as their lack of representing causal relationships, logic inference, and integrating abstract knowledge \cite{marcus2018corr, marcus2003algebraic}. Hopefully, such criticism will lead to improvements in neural networks domain.
}

\subsection{Summary}
\label{sub:summaryNN0}
\bel{We have presented in this section a brief, technical, and historical introduction to neural networks from the early 40$^{\prime}$s to its modern version under the name of \emph{deep learning}. We have presented also some of the last advances in the field.

As we saw earlier, adding more hidden layers to encode the input is one possible way to allow perceptrons to perform better on different tasks. This idea has led to what is known today as deep models, i.e., models composed of many hidden layers. This makes learning representations within neural networks an important aspect to obtain a better generalization error \cite{bengioetlecun2007, bengio2013corr}. However, increasing the depth of a network leads to an increase of the number of its parameters which requires a large number of training data to fit the model and avoid overfitting. Unfortunately, in practice, one usually deals with small datasets which causes deep models to overfit. Different \bel{approaches} were proposed in the literature to deal with such issue using variant regularization methods. We present the most common used methods in the following section (Sec.\ref{sec:neuralnetsregularization0}). \belxx{As it is impractical to cite most the approaches in this chapter, we refer the reader to  \cite{Goodfellowbook2016} where most neural networks regularization techniques are covered,} while we continue our discussion of neural networks in Appendix \ref{sub:appendix4fnn}.
}
\section[Improving Neural Networks Generalization]{Improving Neural Networks Generalization}
\label{sec:neuralnetsregularization0}
Training neural networks and particularly deep architectures is known to be difficult \cite{glorot10, Goodfellowbook2016}. Moreover, deep architectures are known to overfit the data particularly when using few training samples.
Regularization is one of the commonly known solutions to deal with the overfitting issue by providing tools to reduce the generalization error. We describe in this section selected methods to regularize neural networks.
 
 In Sec.\ref{sub:modelselectionandregularization0}, we provided the definition of the regularization as any process that allows reducing the generalization error without necessarily reducing the error over the training samples. In machine learning, and for long time, regularization consists generally in reducing the model complexity using $L_p$ parameters norm penalty. However, with the advances in machine learning, other approaches have been proposed to reduce the generalization error. Nevertheless, such methods do not necessarily reduce the model complexity. We attempt in this section to separate these two approaches, even though the frontier between them is still vague:
 \begin{itemize}
   \item Explicit regularization where the aim is to explicitly reduce the model complexity.
   \item Implicit regularization where the aim is to reduce the generalization error. However, the model complexity may or may not be reduced.
 \end{itemize}

\subsection[Explicit Regularization: Explicit Complexity Reduction]{Explicit Regularization: Explicit Complexity \\ Reduction}
\label{sub:explicityreg0}
Many regularization approaches are based on limiting the capacity of the models by adding a parameter norm penalty to the training objective function $J$. Let us denote the regularized objective function by $\widetilde{J}$
\begin{align}
    \label{eq:eq0-68}
    \widetilde{J}(\bm{\theta}; \bm{X}, \yvec) = J(\bm{\theta}; \bm{X}, \yvec) + \alpha \Omega(\bm{\theta}) ,\quad \Omega(\bm{\theta}) = \frac{1}{p} \lVert \bm{\theta} \rVert^{p}_p \; .
\end{align}
\noindent In this \bel{context}, we provide the study for the case where $p \in \{1, 2\}$. $\alpha \in [0, \infty)$ is a hyperparameter that weights the relative contribution of the norm penalty term, $\Omega$, to the standard training objective function $J$. Large values of $\alpha$ result in more regularization. This approach of regularization is referred to as $L_p$ norm which was widely used in machine learning with a variety of models. Although, $L_p$ parameters norm regularization is not specific to neural networks.

When considering the $L_p$ norm for neural networks regularization \cite{krizhevsky12}, one typically chooses a parameter norm penalty that penalizes only the weights vector $\bm{w}$ at each layer and leaves the biases unregularized \cite{Goodfellowbook2016}. It is desirable to use a different $\alpha$ per layer. However, \bel{as this can make it computationally expensive to compute each optimal value, one may consider using the same $\alpha$ across all the layers \cite{Goodfellowbook2016}}.

\subsubsection[$L_2$ Parameters Norm]{$L_2$ Parameters Norm}
\label{subsub:l20}

The $L_2$ norm penalty is commonly known as weight decay. This regularization approach drives the weights closer to the origin by adding a regularization term
\begin{equation}
    \label{eq:eq0-69}
    \Omega(\bm{\theta}) = \frac{1}{2} \lVert \bm{w} \rVert^2_2 \; ,
\end{equation}
\noindent to the objective function. It is also known in other communities as ridge regression or Tikhonov regularization \cite{tikhonov1963}. The impact of using the $L_2$ parameter norm as a regularization is to shrink the components of $\bm{w}$ where the entries that do not contribute to reduce the objective function are shrunk to have \emph{nearly} zero magnitude, while the rest of the entries are slightly reduced. More technical details are provided in Appendix \ref{para:appendix4l2norm}.

\subsubsection[$L_1$ Parameters Norm]{$L_1$ Parameters Norm}
\label{subsub:l10}

As we mentioned in Sec.\ref{sub:modelselectionandregularization0}, $L_1$ parameters norm minimization is an approximation to the $L_0$ parameters norm. 
 Formally, it is defined as
\begin{equation}
    \label{eq:eq0-87}
    \Omega(\theta) = \lVert \bm{w} \rVert_1 = \sum_i |w_i| \; .
\end{equation}
Compared to $L_2$ norm, $L_1$ norm promotes sparsity by providing a solution with subset of the entries set \emph{exactly} to zero while the rest of the entries are shrunk toward zero. The sparsity aspect of $L_1$ regularization plays an important role in feature selection in machine learning \cite{Hastie2015book, Tibshirani94regressionshrinkage}. More technical details on $L_1$ parameters norms and a comparaison with $L_2$ parameters norm are provided in Appendix \ref{para:appendix4l1norm}.

\subsection{Implicit Regularization}
We discuss here some approaches that are used to reduce the generalization error of neural network without necessarily reducing their complexity. However, some of these methods may have an effect on the model complexity.

\subsubsection{Early Stopping}
\label{subsub:earlystopping0}
Early stopping assumes that minimizing iteratively an objective function $J$ is stopped before it reaches its minimum. If the model's capacity exceeds the optimal capacity, early stopping may prevent overfitting.

\begin{figure}[!htbp]
  \centering
  \includegraphics[scale=0.5]{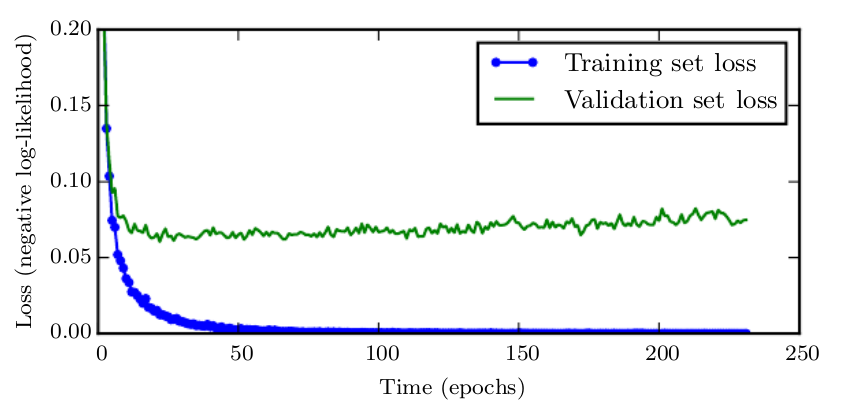}
  \caption[\bel{Training and validation learning curves.}]{Learning curves showing how the negative log-likelihood loss changes over epochs. In this example, a maxout network is trained over MNIST. One can observe how the validation set average loss begins to increase again, forming an asymmetric U-shape curve while the training objective loss keeps decreasing. (Credit: \cite{Goodfellowbook2016})}
  \label{fig:fig0-17}
\end{figure}

When training such models with over capacity, one often observes that the training error decreases steadily through the learning epochs, while the validation set error begins to raise again at some point (Fig.\ref{fig:fig0-17}). This means that one can obtain a model with a better validation set error and hopefully a better test set error by returning the model's parameters at the point with the lowest validation set error. In practice, this can be achieved by keeping a copy of the model's parameters at each time a better validation set error is found.

Early stopping is one of the most commonly form of regularization in training neural networks \cite{Goodfellowbook2016}. This is due to both its effectiveness and simplicity. It can be seen as an efficient hyperparameter selection algorithm that requires one run of the training process to find the best value when to stop training while most hyperparameters need multiple runs.  Besides its training time reduction, it provides a regularization aspect without adding a penalty to the objective cost. However, early stopping requires evaluating the model over the validation set periodically which may slow down the training. Moreover, it requires storing a copy of the model's parameters.

We mentioned that early stopping is a form of regularization. In this section,  we provide an intuition to this idea while formal demonstration is presented in Appendix \ref{subsub:appendix4earlystopping} following the work of \cite{bishop1995ICANN, sjoberg1995}.

Consider a network with weights initialized from a distribution with zero mean. This means that initially most of the neurons provide a value close to zero for most the input vector. Thus, the network can be seen as a linear nilpotent operator with a very low complexity. During the training process, the weights are more likely to increase their magnitude which gradually increase the complexity of the network. Early stopping comes to prevent this increase of magnitude. Therefore, it serves as a method of a weight decay. Also, it prevents the network of becoming too complex \cite{bishop1995ICANN, sjoberg1995}.

\subsubsection{Data Augmentation}
\label{sub:dataaugmentation0}
One way to promote generalization in machine learning is to train the model using a large number of training samples. However, in practice, the amount of training data is limited. A possible solution to deal with this issue is to create new data using the original one, mix them up and feed them to the model as a large training set \cite{niyogi98, Baird90, Ciresan2010, Simard2003}. This approach assumes the generation of new training samples using some invariant parametric transformation function $g(\xvec; \bm{\theta})$ applied to the existing object $(\xvec, \yvec)$. The invariance of this function is considered with respect to the sample target $\yvec$. This technique can be applied on a wide range of tasks. For instance in classification, when considering the input as an image, the label of the sample is invariant to a number of transformations applied on the input image such as translation, rotation, scaling, elastic deformation, contrast, etc.

Generating new training samples is a simple way to incorporate prior knowledge about the problem into the learning algorithm \cite{niyogi98}, which is the aim of most regularization approaches. For instance, when presenting a sample with different rotation angles and the same label, we are indicating to the learning algorithm that the label is invariant to rotation, thus, we introduce a domain knowledge. Generating new samples is also motivated by the bias-variance decomposition (Appendix \ref{sub:appendix4biasvariance}) where increasing the size of the training set will reduce the variance, but keep the bias the same. However, adding new samples does not reduce the complexity of the model.

Although, data augmentation might be helpful to improve generalization, it must be carried out with care. For example, in classification task, some transformations may change the class of the sample. For instance, in optical character recognition tasks, the model is required to recognize the difference between \quotes{n} and \quotes{u}, \quotes{b} and \quotes{d}. Therefore, $180^{\circ}$ rotation and horizontal flip are not appropriate for all characters.

Neural networks are known to be \fromont{sensitive to noise} \cite{TangE10ICML}. One way to improve their robustness is by training them with random noise applied to the input \cite{SIETSMA1991}. This can be seen as data augmentation. Adding noise to the input is a well known mechanism for some unsupervised learning methods \cite{vincent2008, vincent10}. One can also carefully add noise to the hidden units \cite{PooleSG14CORR} which can be seen as an augmentation of the data at multiple levels of abstraction.

Depending on the number of generated samples, data augmentation may increase the training time \cite{krizhevsky12}. For models that need the whole training samples at once, augmenting the training data can raise the issue of the memory. However, when training neural networks, one needs only few samples at once. This allows to generate \bel{examples} as much as one wants. Moreover, one can generate new samples on the fly without the need to store them.

\subsubsection{\bel{Multi-task Learning}}
\label{sub:multitasklearning0}
\bel{
Multi-task learning (MTL) is a learning scenario where multiple learning tasks are solved at the same time, while exploiting commonalities and differences across tasks. This can result in improving the learning efficiency and prediction accuracy of the task-specific models, when compared to training models separately \cite{caruana97ML, Baxter2000MIB, Thrun96islearning, pmlrv22romera12}.

\cite{caruana97ML} define MTL as an approach to inductive transfer\footnote{\bel{\emph{Inductive transfer} or \emph{transfer learning}
 is a research problem in machine learning that focuses on storing knowledge gained while solving one problem and applying it to different but related problem.}} that improves generalization by using domain information contained in the training signals of related tasks as an inductive bias. This is achieved by learning tasks in parallel while using a shared representation. Therefore, the learned representation for a task can be helpful to learn other tasks. Using an MLT framework introduces a bias in the model selection in order to prefer hypotheses that explain more than one task. This shows to improve the generalization of the model and prevents its overfitting since it is required to solve many tasks at once which makes it less likely to overfit one of the tasks. Hence, MTL approaches are considered as a regularization. We mention that MTL framework does not only concern deep learning algorithms but a broad learning algorithms in machine learning \cite{ZhangY17aa}. Aside from designing MTL algorithms, there are many works that study the theoretical aspects of MTL and their generalization bounds \cite{Baxter2000MIB, BenDavidGS02kdd, bendavid2003springer, BenDavidB08ml}. For instance, \cite{Baxter2000MIB} showed that the generalization bounds can be improved through an MTL framework due to the parameters sharing which prevents overfitting. This holds when some assumptions about the statistical relationship between the different tasks are valid, meaning that there is something shared across some of the tasks.

In an MTL framework, a task can be any general learning task such as supervised task, unsupervised task, semi-supervised task, reinforcement learning task, or multi-view learning tasks. \cite{ZhangY17aa} provide a recent and detailed survey that contains different approaches of MTL algorithms.

In deep learning, MTL is a common approach and goes back to the $90^{\prime}$s \cite{Caruana93icml}. It is generally applied by sharing the hidden layers between all the tasks, while keeping several task-specific output layers (Fig.\ref{fig:fig0-18-00-000-00-mtl}). This is known as \emph{hard parameter sharing}. Another MTL scenario consists in considering each task has its own model. However, the different models are encouraged to have similar parameters by constraining the distance between the parameters to be small \cite{DuongCBC15acl, YangH16acorr}. This is referred to as \emph{soft parameter sharing}. 
Nevertheless, some works escape such traditional MTL schemes. \cite{Long015acorr} improve upon hard parameter sharing by placing matrix priors on each level of dense layers, which allow to learn the relationship between tasks. \cite{LuKZCJF17cvpr} propose a bottom-up approach that starts with a thin network and dynamically widens it greedily during training using a criterion that promotes grouping of similar tasks. \cite{MisraSGH16cvpr} propose \emph{cross-stitch networks}. The process starts with two separate models just as in soft parameter sharing. Then, the authors use what is referred to as \emph{cross-stitch} units to allow the model to determine in what way the task-specific networks leverage the knowledge of the other task by learning a linear combination of the output of the previous layers. Such units are placed after the pooling and dense layers. \cite{RuderBAS17corr} propose \emph{sluice networks}, a generalization of \emph{cross-stitch networks}. Other innovations in MTL exist and depend on the task in hand or they are inspired from non-neural MTL frameworks \cite{SogaardG16acl, HashimotoXTS17emnlp, KendallGC17corr, YangH16corr}.

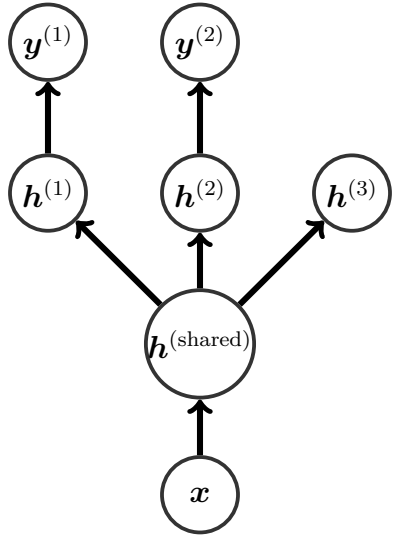
\begin{figure}[!ht]
  \begin{center}
    \begin{tikzpicture}
\coordinate (x) at (0, 0);
\coordinate (hsh) at (0, 2);
\coordinate (h1) at (-2, 4);
\coordinate (h2) at (0, 4);
\coordinate (h3) at (2, 4);
\coordinate (y1) at (-2, 6);
\coordinate (y2) at (0, 6);

\draw (x) node[circle, draw=black!80,  inner sep=0pt, minimum size=1cm, line width=0.5mm] (cx) {$\bm{x}$};

\draw (hsh) node[circle, draw=black!80,  inner sep=0pt, minimum size=1cm, line width=0.5mm] (chsh) {$\bm{h}^{(\text{shared})}$};

\draw (h2) node[circle, draw=black!80,  inner sep=0pt, minimum size=1cm, line width=0.5mm] (ch2) {$\bm{h}^{(2)}$};

\draw (h1) node[circle, draw=black!80,  inner sep=0pt, minimum size=1cm, line width=0.5mm] (ch1) {$\bm{h}^{(1)}$};

\draw (h3) node[circle, draw=black!80,  inner sep=0pt, minimum size=1cm, line width=0.5mm] (ch3) {$\bm{h}^{(3)}$};

\draw (y1) node[circle, draw=black!80,  inner sep=0pt, minimum size=1cm, line width=0.5mm] (cy1) {$\bm{y}^{(1)}$};

\draw (y2) node[circle, draw=black!80,  inner sep=0pt, minimum size=1cm, line width=0.5mm] (cy2) {$\bm{y}^{(2)}$};

\draw[->, thick, line width=0.8mm] (cx) -- (chsh) node {};
\draw[->, thick, line width=0.8mm] (chsh) -- (ch1) node {};
\draw[->, thick, line width=0.8mm] (chsh) -- (ch2) node {};
\draw[->, thick, line width=0.8mm] (chsh) -- (ch3) node {};
\draw[->, thick, line width=0.8mm] (ch1) -- (cy1) node {};
\draw[->, thick, line width=0.8mm] (ch2) -- (cy2) node {};
\end{tikzpicture}
  \end{center}
  \caption[\bel{Common architecture for multi-task learning in neural networks.}]{\bel{Typical architecture for multi-task learning in neural networks where the tasks share a common input but involve different output targets. The low layers, $\bm{h}^{\text{shared}}$, can be shared among the different tasks while task-specific layers, $\bm{h}^{(1)}$ and $\bm{h}^{(2)}$, can be learned on top of the shared representation. The underlying assumption is that there exists a common pool of factors that explain the variations of $\bm{x}$, while each task is associated with a subset of such factors. In an unsupervised task, it is possible to pool top-level factors, $\bm{h}^{(3)}$, to be associated with none of the output task. Such factors may explain some of the variations of the input $\bm{x}$, but they are irrelevant to predict the targets $\bm{y}^{(1)}$ and $\bm{y}^{(2)}$. (Reference: \cite{Goodfellowbook2016})}}
  \label{fig:fig0-18-00-000-00-mtl}
\end{figure}

\FloatBarrier

MTL framework is an adequate choice where we are interested in obtaining predictions for multiple tasks at the same time. Such scenarios are common for instance in finance or economics forecasting, where we might want to predict the value of many possibly related variables, or in bioinformatics where we might want to predict symptoms for multiple diseases simultaneously. In the case of drug discovery, where tens or hundreds of active compounds may be predicted, MTL increases the prediction accuracy with the increase of the number of tasks \cite{RamsundarKRWKP15corr}. Although MTL is constructed to improve the performance of many tasks at once, in some situations we only care about the performance of one task, that we call \emph{main task}, whereas, the other tasks, named \emph{auxiliary tasks}, are not important in the inference time. They are useful only during training in order to prevent the main task from overfitting. Using such auxiliary tasks as an MTL framework is a classical choice. \belx{In the case of using $T$ auxiliary tasks, denoted $f_j(\cdot)$, along with a main task $f(\cdot)$, a standard optimization formula can be cast as follows. Let us consider $\bm{\theta}_{MTL}=\{\bm{\theta}_{sh}, \bm{\theta}, \bm{\theta}_j, \cdots, \bm{\theta}_T\}$ a set of parameters of the whole MTL framework, where $\bm{\theta}_{sh}$ is a shared set of parameters among all tasks, $\bm{\theta}$ is the set of parameters of the main task, and $\bm{\theta}_j$ is the set of parameters associated with the $j^{th}$ auxiliary task. We note $\mathcal{C}(\cdot, \cdot)$ as the cost of the main task, while $\mathcal{C}_{j}(\cdot, \cdot)$ is the cost of the $j^{th}$ auxiliary task. We consider that the same input $\bm{x}_i$ is fed to each task, while a different label $\bm{y}^j_i$ is associated to it depending on the task $j$. The training set has $N$ samples, with $\fromont{\mathbb{D}_j}=\{(\bm{x}_i, \bm{y}_i^j)\}_{i=1}^{N}$ a training set of the $j^{th}$ auxiliary task, and $\mathbb{D}=\{(\bm{x}_i, \bm{y}_i)\}_{i=1}^{N}$ is a training set for the main task. Each auxiliary task $j$ is weighted using a coefficient $\lambda_j$. The optimization consists in solving
\begin{equation}
  \label{eq:eq0-140-mtl-example}
  \argmin_{\bm{\theta}_{MTL}} \sum_{i=1}^N\mathcal{C}(f(\bm{x}_i; \bm{\theta}_{sh}, \bm{\theta}), \bm{y}_i) + \sum_{j=1}^{T} \lambda_j \mathcal{C}_j(f_{j}(\bm{x}_i; \bm{\theta}_{sh}, \bm{\theta}_j), \bm{y}_i^j) \; .
\end{equation}

Optimizing Eq.\ref{eq:eq0-140-mtl-example} can be done easily in parallel using stochastic gradient descent. However, in practice, alternating between tasks seems to work better \cite{collobert08ICML, zhang14ECCV}.
}

Auxiliary tasks have been used in different setups in an MTL framework. For instance, \cite{caruana97ML} use tasks that predict different characteristics of the road in order to predict the steering direction in a self-driving car. \cite{zhang14ECCV} use head pose estimation and facial attributes inference as auxiliary tasks to predict facial landmarks. 
 \cite{GaninJMLRv172016} use an adversarial loss in a domain adaptation\footnote{\bel{\emph{Domain adaptation} is a field associated with machine learning and transfer learning. It consists in a learning scenario where we aim at learning from a source data distribution a well performing model on a different, but related, target data distribution \cite{BridleC90nips, BenDavid2010da, CrammerKW08jmlr}. For instance, in spam filtering task, domain adaptation consists in adapting a model from one use (the source distribution) to a new user who receives significantly different email, i.e., the target distribution.}} scenario in order to constrain the model to build internal representations that do not distinguish between domains. In some cases, one can encode some prior knowledge as an auxiliary task, mostly to learn better representations. This is known as \emph{hints} and it is the old name of an MTL framework \cite{suddarth90NN, abuMostafa90, abuMostafab92}. Reconstructing the input \cite{bel16} or/and the output \cite{belharbiNeurocomp2017} data can be used as well as auxiliary tasks.

Although auxiliary tasks are helpful in an MTL framework, it is still unclear what tasks should be used in practice. Finding a good auxiliary task is often based on the assumption that the auxiliary task should be related to the main task somehow and that it should be helpful for the prediction of the main task. However, the relatedness of two tasks is still unclear. \cite{caruana97ML} define two tasks to be related if they use the same features to make a decision. \cite{Baxter2000MIB} argue that related tasks share a common optimal hypothesis class, i.e., have the same inductive bias. \cite{bendavid2003springer} propose that two tasks are related if the data for both tasks can be generated from a fixed probability distribution using a set of transformations. \cite{XueLCK07jmlr} define that two tasks are related if their classification boundaries, i.e., parameters, are close. Despite the theoretical lack of our understanding to task relatedness, such concept is not binary but a spectrum. Allowing the models to learn what to share with each task might be a way to build better MTL frameworks and make better use of related or loosely related tasks \cite{ZhangY17aa}.
}

\subsubsection{\bel{Transfer Learning}}
\label{sub:transferlearning0}
\bel{
Since transfer learning is not specific to deep learning methods, we decide to provide a general background on transfer learning as a learning paradigm in machine learning. Then, we detail its applications and how to achieve it in a deep learning based model.

\bigskip
\textbf{Transfer Learning: Background}
\bigskip

Transfer learning (TL) is the improvement of learning in a new task through the transfer of knowledge from a related task that has already been learned \cite{Olivas2009HRM, Aggarwal2014DCA, Pan2010STL}. Common machine learning algorithms traditionally address isolated tasks such as classification, regression, and clustering, etc, under the assumption that training and test data are draw from the same feature space and the same distribution. When the distribution changes, most statistical models need to be rebuilt from scratch using newly collected data. In many real world applications, it is expensive or impossible to re-collect the needed data and rebuild the models. TL domain attempts to change this by developing methods to transfer knowledge learned in one or more \emph{source tasks} and use it to improve learning a related \emph{target task} (Fig.\ref{fig:fig0-18-00-000-00-tl}). Techniques that enable knowledge transfer represent progress toward making machine learning as efficient as human learning. Extensive and detailed study of TL, its applications, and issues can be found in \cite{Pan2010STL, Aggarwal2014DCA, Olivas2009HRM}. 

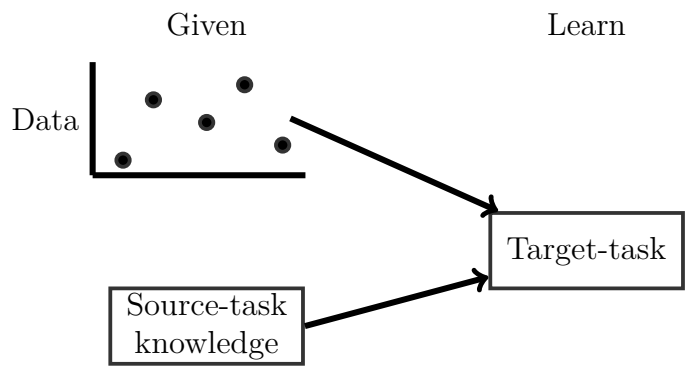
\begin{figure}[!htbp]
  \begin{center}
		\begin{tikzpicture}
\coordinate (zz) at (0, 0);  
\coordinate (stk) at (-2, -1);
\coordinate (cp) at (-3.5, 1); 
\coordinate (cpr) at (-0.7, 1);
\coordinate (cpt) at (-3.5, 2.5);

\coordinate (tt) at (3, 0);

\coordinate (p1) at (-3.1, 1.2);
\coordinate (p2) at (-2.7, 2);
\coordinate (p3) at (-2, 1.7);
\coordinate (p4) at (-1.5, 2.2);
\coordinate (p5) at (-1., 1.4);
\coordinate (pinv) at (-1, 1.8);
\coordinate (g) at (-2, 3);
\coordinate (l) at (3, 3);


\draw (stk) node[rectangle, text width=2.5cm, align=center, draw=black!80,  inner sep=0.5pt, minimum size=1cm, line width=0.5mm] (cstk) {Source-task knowledge};

\draw[-, thick, line width=0.8mm] (cp) -- (cpr) node {};
\draw[-, thick, line width=0.8mm] (cp) -- (cpt) node[pos=0.5, left] {Data};

\draw (tt) node[rectangle, text width=2.5cm, align=center, draw=black!80,  inner sep=0.5pt, minimum size=1cm, line width=0.5mm] (ctt) {Target-task};

\draw (p1) node[circle, draw=black!80, fill=black, inner sep=0pt, minimum
size=5pt, line width=0.5mm] (cp1) {};
\draw (p2) node[circle, draw=black!80, fill=black, inner sep=0pt, minimum size=5pt, line width=0.5mm] (cp2) {}; size=5pt, line width=0.5mm] (cp1) {};
\draw (p3) node[circle, draw=black!80, fill=black, inner sep=0pt, minimum size=5pt, line width=0.5mm] (cp3) {};
\draw (p4) node[circle, draw=black!80, fill=black, inner sep=0pt, minimum size=5pt, line width=0.5mm] (cp4) {};
\draw (p5) node[circle, draw=black!80, fill=black, inner sep=0pt, minimum size=5pt, line width=0.5mm] (cp5) {};

\draw (pinv) node[circle, draw=white!80, inner sep=0pt, minimum size=5pt, line width=0.5mm, fill opacity=0.0] (cpinv) {};

\draw[->, thick, line width=0.8mm] (cpinv) -- (ctt) node {};
\draw[->, thick, line width=0.8mm] (cstk.east) -- (ctt) node {};

\node at (g) {Given};
\node at (l) {Learn};
\end{tikzpicture}
	\end{center}
  \caption[\bel{Transfer learning scheme.}]{\bel{Transfer learning is a machine learning algorithm with an additional source of information apart from the standard training data: knowledge extracted from one or more related tasks. (Reference: \cite{Olivas2009HRM})}}
  \label{fig:fig0-18-00-000-00-tl}
\end{figure}

The study of TL can be motivated biologically by the fact that people can intelligently apply knowledge learned previously to solve new problems faster with better solutions \cite{ellis1965transfer, woodworth1901influence}. The fundamental motivation for TL in the field of machine learning is the need for lifelong machine learning methods that retain and reuse previously learned knowledge.

TL techniques have been applied in different learning algorithms including inductive learning and reinforcement learning \cite{Olivas2009HRM, Aggarwal2014DCA, Pan2010STL}. In this section, we focus on its application to the former.

The goal of TL is to improve learning in the target task by leveraging knowledge from the source task. \cite{Olivas2009HRM} refer to three common measures by which TL might improve learning, illustrated in Fig.\ref{fig:fig0-18-00-000-00-tlbenefits}: \begin{itemize*}
\item The initial performance achievable in the target task using only the transferred knowledge, before any further learning is done, compared to the initial performance of a model with a fresh start. \item The amount of time it takes to fully learn the target task given the transferred knowledge compared to the amount of time to learn it from scratch. \item The final performance level achievable in the target task compared to the final level without transfer.
\end{itemize*}

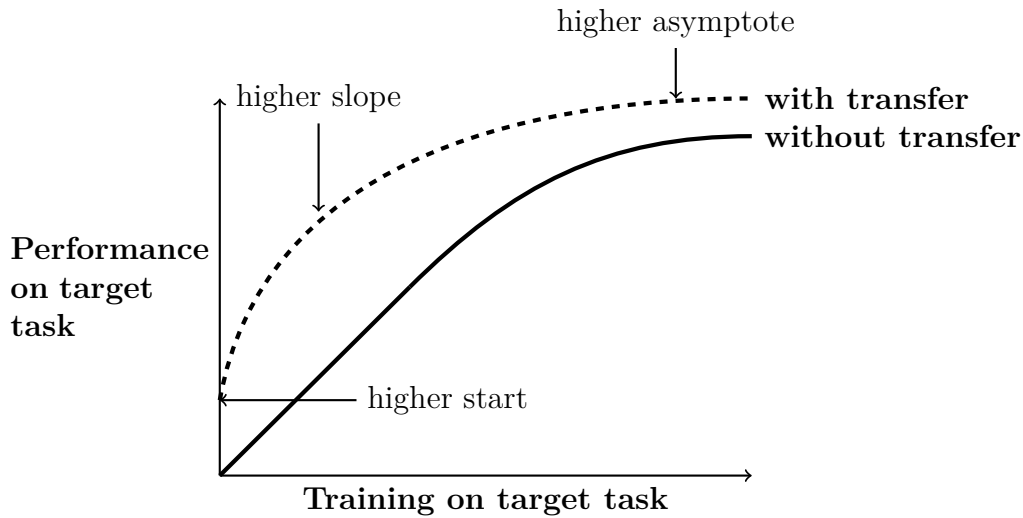
\begin{figure}[!ht]
  \begin{center}
		\begin{tikzpicture}
\coordinate (center1) at (0, 0);
\coordinate (xneg) at (0, 0);
\coordinate (xpos) at (7, 0);
\coordinate (ypos) at (0, 5);
\coordinate (yneg) at (0, 0);
\coordinate (wtlh) at (2.5, 2.5);
\coordinate (wtl) at (7, 4.5);

\coordinate (wdtl0) at (0, 1);
\coordinate (wdtlend) at (7, 5.);

\coordinate (ar1) at (2, 1);

\coordinate (ar2) at (1.3, 5);
\coordinate (ar3) at (1.3, 3.5);

\coordinate (ar4) at (6, 6);
\coordinate (ar5) at (6, 5.);

\draw[->, thick] (xneg) -- (xpos) node[pos=0.5, below] (xaxis) {\textbf{Training on target task}};
\draw[->, thick] (yneg) -- (ypos) node [pos=0.5, left, text width=2.6cm] {\textbf{Performance on target task}};

\draw[color=black, ultra thick] (xneg) to[out=45,in=225] (wtlh) node[right, color=black]  {};

\draw[color=black, ultra thick] (wtlh) to[out=45,in=180] (wtl) node[right, color=black]  {\textbf{without transfer}};

\draw[color=black, dashed, ultra thick] (wdtl0) to[out=80,in=180] (wdtlend) node[right, color=black]  {\textbf{with transfer}};

\node[right of=ar1]  (textar1) {higher start};
\draw[->, thick] (textar1) -- (wdtl0);

\node at (ar2)  (textar2) {higher slope};
\draw[->, thick] (textar2) -- (ar3);

\node at (ar4)  (textar4) {higher asymptote};
\draw[->, thick] (textar4) -- (ar5);
%


\end{tikzpicture}
	\end{center}
  \caption[\bel{Three ways in which transfer learning might improve learning.}]{\bel{Three ways in which transfer learning might improve learning the target task. (Reference: \cite{Olivas2009HRM})}}
  \label{fig:fig0-18-00-000-00-tlbenefits}
\end{figure}

It is important to distinguish the difference between TL and multi-task learning discussed in Sec.\ref{sub:multitasklearning0}, where several tasks are learned simultaneously (Fig.\ref{fig:fig0-18-00-000-00-mtlvstl}). Multi-task learning is clearly closely related to TL, but it does not involve designated source and target tasks; instead the learning algorithm receives several tasks at once. In contrast, in TL, the learning algorithm in the source task knows nothing about the target task. Moreover, TL aims at boosting the performance of the target domain by using the source domain data, i.e., its knowledge. 

\begin{figure}[!ht]
  \begin{center}
		\begin{tikzpicture}
\coordinate (zz) at (0, 0);  
\coordinate (st) at (-6, 0);
\coordinate (tt) at (-2, 0);

\coordinate (t1) at (2, 1);
\coordinate (t2) at (2, -1);

\coordinate (t3) at (6, 1);
\coordinate (t4) at (6, -1);


\draw (st) node[rectangle, text width=2.5cm, align=center, draw=black!80,  inner sep=0.5pt, minimum size=1cm, line width=0.5mm] (cst) {Source task};

\draw (tt) node[rectangle, text width=2.5cm, align=center, draw=black!80,  inner sep=0.5pt, minimum size=1cm, line width=0.5mm] (ctt) {Target task};

\draw (t1) node[rectangle, text width=2.5cm, align=center, draw=black!80,  inner sep=0.5pt, minimum size=1cm, line width=0.5mm] (ct1) {Task 1};

\draw (t2) node[rectangle, text width=2.5cm, align=center, draw=black!80,  inner sep=0.5pt, minimum size=1cm, line width=0.5mm] (ct2) {Task 2};

\draw (t3) node[rectangle, text width=2.5cm, align=center, draw=black!80,  inner sep=0.5pt, minimum size=1cm, line width=0.5mm] (ct3) {Task 3};

\draw (t4) node[rectangle, text width=2.5cm, align=center, draw=black!80,  inner sep=0.5pt, minimum size=1cm, line width=0.5mm] (ct4) {Task 4};

\draw[->, thick, line width=0.8mm] (cst) -- (ctt) node {};
\draw[<->, thick, line width=0.8mm] (ct1) -- (ct2) node {};
\draw[<->, thick, line width=0.8mm] (ct1) -- (ct3) node {};
\draw[<->, thick, line width=0.8mm] (ct3) -- (ct4) node {};
\draw[<->, thick, line width=0.8mm] (ct4) -- (ct2) node {};
\draw[<->, thick, line width=0.8mm] (ct1) -- (ct4) node {};
\draw[<->, thick, line width=0.8mm] (ct2) -- (ct3) node {};
\end{tikzpicture}
	\end{center}
  \caption[\bel{Transfer learning vs. multi-task learning.}]{\bel{\emph{left}: in transfer learning, the information flows in one direction only, from the source task toward the target task. \emph{right}: in multi-task learning, information can flow freely among all tasks. (Reference: \cite{Olivas2009HRM})}}
  \label{fig:fig0-18-00-000-00-mtlvstl}
\end{figure}
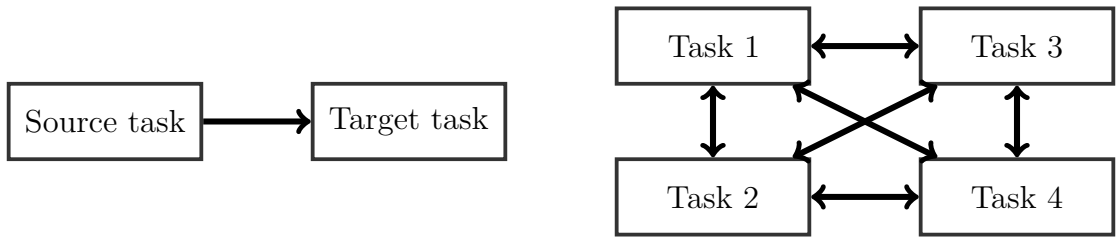

In the case of inductive learning methods \cite{mitchell1997}, the target-task inductive bias is chosen or adjusted based on the source-task knowledge  \cite{Olivas2009HRM}. It is usually concerned with improving the speed with which a model is learned, or with improving its generalization capability. The way this is done varies depending on which inductive is used to learn the source and target tasks. Some TL algorithms narrow down the hypothesis space, limiting the possible hypotheses, or remove search steps from consideration. Other methods broaden the space, allowing the search to discover more complex hypotheses, or add new search steps \cite{Baxter2000MIB, ThrunM95ijcai, mihalkova2006transfer}.


\cite{Aggarwal2014DCA, Pan2010STL} provide an extensive categorization of TL techniques. 
In the case of inductive learning, four possible ways of achieving a knowledge transfer: \begin{itemize*}
\item Transferring knowledge of instances: where certain parts of the source data are reused together with the labeled data in the target task \cite{DaiYXY07icml, JiangZ07acl}. \item Transferring knowledge of representations: in this case, the aim is to find a good representation in source and target tasks which is similar to finding common features in a multi-task learning scheme (Sec.\ref{sub:multitasklearning0}). Depending on the available labeled or unlabeled data in the source task, TL in this category can be done either in a supervised way by using an MTL framework to learn low-dimensional representations that are shared across the tasks  \cite{ArgyriouEP06nips, ArgyriouMPY07nips} or in unsupervised way to learn higher level representations \cite{RainaBLPN07icml, WangM08icml}. \item Transferring knowledge of parameters: the underlying assumption in this approach is that individual models for related tasks should share some parameters such as in SVMs \cite{EvgeniouP04kdd} , ensemble learning \cite{GaoFJH08kdd}, or prior distributions of hyper-parameters such as in Bayesian frameworks \cite{LawrenceP04icml, BonillaCW07nips}. \item Transferring relational knowledge: this approach deals with the data that is non-i.i.d. and can be represented by multiple relations, such as networked data and social network data. Its aim is to transfer the relationship among data from a source domain to  a target domain. Statistical relational learning techniques are required to deal with such context \cite{MihalkovaHM07aaai, mihalkova2008transfer}.
\end{itemize*}

An important issue in TL is to recognize its limitations. Different works address theoretical aspects of TL such as its generalization bounds \cite{MahmudR07nips, BenDavidBCP06nips, BlitzerCKPW07nips, BenDavid2010da, germain2017pac}. Although the generalization bounds are slightly different, under some conditions, the bound consists generally in two terms: the first is the error bound of the model on the source domain, the second is the bound on the distance between the source and the target domains, i.e., the distance between their marginal probability distributions which explains the relatedness of both domains. The aspect of relatedness of domains is an open issue in TL. When transferring knowledge between unrelated domains, \emph{negative transfer} may happen: a case where the performance of the model in the target domain decrease when applying TL compared to its performance without TL. Often, when applying TL from a domain to another, it is necessary to \emph{map} the characteristics/feature/representation of one task to another. In much cases, this is achieved by hand. Other methods have been propose to perform automatic \emph{mapping} between domains. We intentionally skip the discussion of such important aspects in this thesis \fromont{in order to stay focused on its main subject}. \cite{Olivas2009HRM, Aggarwal2014DCA, Pan2010STL} provide further details on theses subjects.

In the following, we present some aspects of using TL in deep learning methods.

\bigskip
\textbf{Transfer Learning in Deep Learning}
\bigskip

TL is a learning paradigm that can be applied to different machine learning models including deep learning. Most success in deep learning methods in academic research or industry has been driven by the use of large models that have be trained using a huge amount of supervised data \cite{Goodfellowbook2016}. In real life applications, nor the large amount of labeled data, nor the computation power required for training are usually available to conduct such large scale learning. We often encounter a situation where only few training samples are available in a particular target domain of interest. TL seems to be an alternative way to classical supervised learning in order to obtain better performance on the target domain by leveraging knowledge extracted from a source domain with abundant training samples.

We present in this section two main approaches to apply TL in deep learning, both of them are based on representation learning paradigm: either through using pre-trained models, or through learning domain-invariant representations.

\begin{itemize}[leftmargin=*, wide=\parindent]
  \item The application of convolutional neural networks (CNNs) has seen a large success, mostly in computer vision tasks \cite{Goodfellowbook2016}. In such models, low convolutional layers \fromont{in the network} tend to capture low level image features such as edges, while higher convolutional layers tend to capture more complex and task dependent details such as body parts, faces, and more compositional patterns \cite{YosinskiCBL14nips, ZeilerF14eccv}. To perform TL in a target domain in computer vision tasks using CNNs, it is common \cite{belharbi2017, RazavianASC14cvpr, Jiang2015} to use off-the-shelf pre-trained CNNs on ImageNet \cite{imagenet09}. In practice, this is achieved by reusing the pre-trained convolutional layers, preferably low layers \cite{YosinskiCBL14nips}, and adding on top of it new layers specific to the target task. Then, the new network, as a whole or partially, is trained over the target data. A practical issue in such approach is that, often, the CNN model is over-parametrized for the target task and it may cause a slow in running speed mainly due to unnecessary computations. A practical solution is to drop useless filters through a pruning process \cite{li2016pruning, molchanov2016pruning, park2016faster} while tolerating slight reduction of the performance. While such reuse of pre-trained low layers of CNNs has seen large success in computer vision, pre-trained models have limited use in natural language processing (NLP) \cite{RamachandranLL17emnlp, MikolovSCCD13nips, Kim14fcorr}. Low layers of an NLP model tend to learn task specific aspects such as syntax, which can not be helpful to perform cross domain adaptation \cite{JozefowiczVSSW16corr}. What is needed in NLP is features that capture more fine-grained rules which are located at the top layers \cite{Goodfellowbook2016, JozefowiczVSSW16corr}. While object recognition may be a prototype task that is shared among most computer vision tasks, language modeling \cite{JozefowiczVSSW16corr} may be the closest analogy in NLP, where in order to predict the next word, a model needs: to possess knowledge of how a language is structured, understand what words likely are related to and likely to follow each other, and to model long-term dependencies. Such aspects may be shared among different tasks in NLP.

A model trained on ImageNet seems to capture details about the way animals and objects are structured and composed which is generally relevant when dealing with images. As such, the classification task on ImageNet seems to be a good proxy for general computer vision problems, as the same knowledge that is required to excel in it is also relevant for many other computer vision tasks. A similar assumption is used to motivate the use of generative model, that is, when training generative models, it is assumed that the ability to generate realistic images, for instance, requires an understanding of the underlying structure of images. Such knowledge about the structure can be used in different task where the structure is relevant. Such assumption relies itself on the premise that all images lie on a low-dimensional manifold, i.e., that there is some underlying structure to images that can be extracted by a model. \cite{RadfordMC15corr} indicate that such a structure might indeed exist, and demonstrate it by generating realistic images describing transitions points in an image.

Unsupervised layer-wise pre-training technique \cite{Goodfellowbook2016, bengio07, hinton06} is another practical example of TL in deep learning. In such learning approach, the source task consists in learning in an unsupervised and incremental way hidden representations that disentangle the variation factors of the input data throughout reconstruction of the raw data.

\item The second method of using TL in deep learning consists in learning domain-invariant representations. Creating representations that do not change based on the domain is very interesting since it should capture the variations of the data independently of the domain. This is less expensive and more feasible for non-vision tasks than generating representations that are useful for all tasks. In such scenario, only unlabeled data of each domain are needed in order to create domain-invariant representations. Such representations are generally learned using stacked denoising auto-encoders and have seen success in NLP \cite{GlorotBB11icml, ChenXWS12icml} as well as in computer vision \cite{ZhuangCLPH15ijcai}. Other approaches encourage the data representation in different domains to be more similar and avoid domain-specific representations \cite{daumeiii2007ACLMain, BousmalisTSKE16nips, GaninJMLRv172016, TzengCVPR2017}.
\end{itemize}

TL is perhaps a potential learning paradigm that may allow breakthrough of deep learning techniques in a large number of small-data settings, which is the case in most real life applications.  We hope that it will get more attention in deep learning research community in order to democratize the use of such powerful models.
}

\subsubsection{Parameter Sharing: Particular Case of Convolutional Networks and Auto-encoders}
\label{sub:parametersharing0}
\bel{As we mentioned earlier in Sec.\ref{sub:explicityreg0}, regularization can be obtained by introducing prior knowledge of the domain or of the model's architecture into the way of estimating the parameters values. For instance, one may introduce some prior knowledge about the dependency between the values of different parameters. In pattern recognition applied to images, one can assume that an elementary pattern such as small corners may appear multiple times in an image. Therefore, it make sense to use the same sensor, i.e., parameter, all over the image in order to detect the pattern, instead of using different sensors at different locations. This is referred to as parameter sharing which can be seen as a regularization \cite{Goodfellowbook2016}. The first obvious advantage of parameter sharing over regularizing the parameters to be close to some predefined ideal parameters is the significant reduction of the model size in terms of memory use. Moreover, some models have more natural way, in certain applications, to use shared parameters such as the case of convolutional networks \cite{Fukushima1979neocognitron, LeCun89}.}

\begin{figure}[!ht]
  \centering
  \includegraphics[scale=0.3]{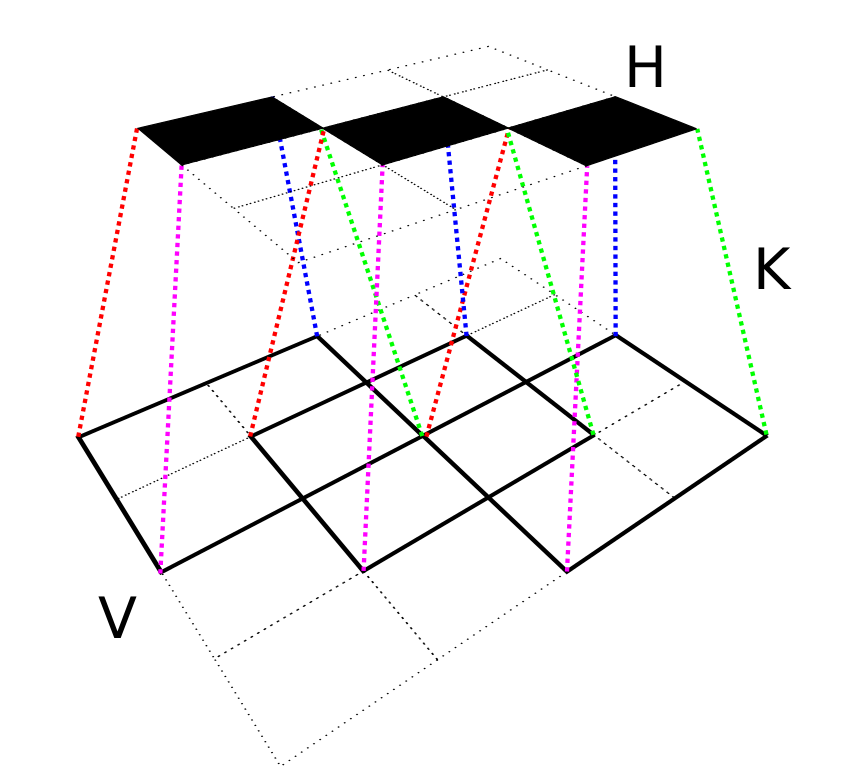}
  \caption[\bel{A $2$-dimensional convolution layer.}]{A $2$-dimensional convolution layer. The $2 \times 2$ filter $\bm{K}$ is applied to the $4 \times 4$ input $\bm{V}$ in order to get a $3 \times 3$ output $\bm{H}$. The number of weights is reduced from $4 \times 4 \times 3 \times 3 = 144$ to $4$. (Credit: \cite{Demyanovtehsis2015})}
  \label{fig:fig0-18-1}
\end{figure}

Convolutional networks were motivated by the neurophysiological insights from \cite{wiesel1959, Hubel62} that showed that simple and complex cells, which are found in the cat's visual cortex, fire in response to certain properties of visual sensory of inputs such as the orientation of edges. This led to convolutional networks as we know them today where the receptive field of \bel{a} convolutional unit with given weight, which is typically a square filter, is shifted step by step across a 2 dimensional array of input values such as an image (Fig.\ref{fig:fig0-18-1}). The resulting 2D array of a subsequent activation events of this unit can be provided as inputs to higher level units, and so on. Usually, there are many filters at one representation level where each one learns to respond to specific properties. In the other hand, sharing filters is also motivated by some properties of the input signal. For instance, natural images have many statistical properties that are invariant to translation\bel{: a picture of a car remains a picture of car even if it is translated one pixel to any direction}. Convolutional networks can take this property into account by sharing parameters across different location\bel{s} of the image where the same feature is computed over different positions in the input. This means we can find a car with the same car detector even if we moved the car slightly. Another example that concerns low level features such as edges. One filter can learn to detect a specific type of edges with different angle. It is intuitive to apply the same filter across the whole input in order to find similar edges. Parameter sharing has enabled convolutional networks to dramatically lower the number of \bel{parameters} (Fig.\ref{fig:fig0-18-1}) and to significantly increase network size without requiring a corresponding increase \bel{of} the number of training samples.

Auto-encoders \cite{bengio06NIPS, aurelio07} employ as well parameter sharing between the encoder and the decoder layer. However, instead of using the same weight of the encoder, the decoder uses its transpose. This has the advantage to reduce the number of parameters of the auto-encoder. Depending on using a non-linearity in both layers, this type of parameter sharing may prevent the auto-encoder from learning linear transformation similarly to principal component analysis (PCA) method \cite{bengio13}. Furthermore, \bel{sharing} the parameters between the encoder and the decoder may have come \bel{originally} from Restricted Boltzmann Machines (RBMs) \cite{smolensky86} where the same weight and its transpose are used to infer the hidden and the visible states.

\FloatBarrier

\subsubsection{Dropout}
\label{subsub:dropout0}
\bel{Dropout \cite{SrivastavaDropout2013, srivastava14a} provides a computationally inexpensive but powerful method for regularizing a broad family of models and prevent their overfitting. The main idea of dropout consists in randomly omitting a subset of neurons during training for each sample by setting the output of these neurons to zero. This can prevent the remaining neurons from co-adaptation in which a feature extractor is only helpful in the context of several other specific feature detectors. Instead, dropout pushes each neuron to learn to detect a feature that is generally helpful to produce the correct answer independently of the internal context, i.e., the presence or the absence of other features. During the test phase, a scaling of the neuron output is necessary.

In practice, dropout is performed as follows. Let consider a multilayer perceptron with the layers: $\yvec_0, \cdots, \yvec_M$. Then, the dropout on the layer $\yvec_i$ of the size $N$ can be described as follows:
\begin{enumerate}
    \item For each training case, generate a binary vector mask $\bm{\mu}$ of length $N$, where each element is sampled from the Bernoulli distribution with probability $0< p < 1$.
    \item On the forward pass, multiply the values of $\yvec_i$ by $\bm{\mu}$.
    \item On the backward pass, multiply the gradients $\bm{dy}_i$ by $\bm{\mu}$.
    \item During the test phase, multiply all values of $\yvec_i$ by $p$.
\end{enumerate}
Fig.\ref{fig:fig0-18-00-000-00-dropout} illustrates the application of dropout across all the layers of a feedforward network.

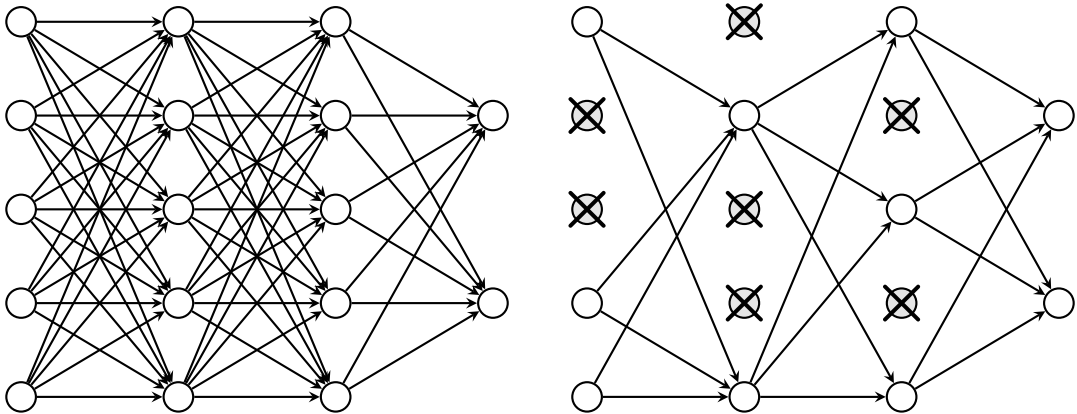
\begin{figure}[!ht]
  \begin{center}
		\begin{tikzpicture}
	\node[circle, draw, thick] (i1) {};
	\node[circle, draw, thick, above=2em of i1] (i2) {};
	\node[circle, draw, thick, above=2em of i2] (i3) {};
	\node[circle, draw, thick, below=2em of i1] (i4) {};
	\node[circle, draw, thick, below=2em of i4] (i5) {};
	
	\node[circle, draw, thick, right=4em of i1] (h1) {};
	\node[circle, draw, thick, right=4em of i2] (h2) {};
	\node[circle, draw, thick, right=4em of i3] (h3) {};
	\node[circle, draw, thick, right=4em of i4] (h4) {};
	\node[circle, draw, thick, right=4em of i5] (h5) {};
	
	\node[circle, draw, thick, right=4em of h1] (hh1) {};
	\node[circle, draw, thick, right=4em of h2] (hh2) {};
	\node[circle, draw, thick, right=4em of h3] (hh3) {};
	\node[circle, draw, thick, right=4em of h4] (hh4) {};
	\node[circle, draw, thick, right=4em of h5] (hh5) {};
	
	\node[circle, draw, thick, right=4em of hh2] (o1) {};
	\node[circle, draw, thick, right=4em of hh4] (o2) {};
	
	\draw[-stealth, thick] (i1) -- (h1);
	\draw[-stealth, thick] (i1) -- (h2);
	\draw[-stealth, thick] (i1) -- (h3);
	\draw[-stealth, thick] (i1) -- (h4);
	\draw[-stealth, thick] (i1) -- (h5);
	\draw[-stealth, thick] (i2) -- (h1);
	\draw[-stealth, thick] (i2) -- (h2);
	\draw[-stealth, thick] (i2) -- (h3);
	\draw[-stealth, thick] (i2) -- (h4);
	\draw[-stealth, thick] (i2) -- (h5);
	\draw[-stealth, thick] (i3) -- (h1);
	\draw[-stealth, thick] (i3) -- (h2);
	\draw[-stealth, thick] (i3) -- (h3);
	\draw[-stealth, thick] (i3) -- (h4);
	\draw[-stealth, thick] (i3) -- (h5);
	\draw[-stealth, thick] (i4) -- (h1);
	\draw[-stealth, thick] (i4) -- (h2);
	\draw[-stealth, thick] (i4) -- (h3);
	\draw[-stealth, thick] (i4) -- (h4);
	\draw[-stealth, thick] (i4) -- (h5);
	\draw[-stealth, thick] (i5) -- (h1);
	\draw[-stealth, thick] (i5) -- (h2);
	\draw[-stealth, thick] (i5) -- (h3);
	\draw[-stealth, thick] (i5) -- (h4);
	\draw[-stealth, thick] (i5) -- (h5);
	
	\draw[-stealth, thick] (h1) -- (hh1);
	\draw[-stealth, thick] (h1) -- (hh2);
	\draw[-stealth, thick] (h1) -- (hh3);
	\draw[-stealth, thick] (h1) -- (hh4);
	\draw[-stealth, thick] (h1) -- (hh5);
	\draw[-stealth, thick] (h2) -- (hh1);
	\draw[-stealth, thick] (h2) -- (hh2);
	\draw[-stealth, thick] (h2) -- (hh3);
	\draw[-stealth, thick] (h2) -- (hh4);
	\draw[-stealth, thick] (h2) -- (hh5);
	\draw[-stealth, thick] (h3) -- (hh1);
	\draw[-stealth, thick] (h3) -- (hh2);
	\draw[-stealth, thick] (h3) -- (hh3);
	\draw[-stealth, thick] (h3) -- (hh4);
	\draw[-stealth, thick] (h3) -- (hh5);
	\draw[-stealth, thick] (h4) -- (hh1);
	\draw[-stealth, thick] (h4) -- (hh2);
	\draw[-stealth, thick] (h4) -- (hh3);
	\draw[-stealth, thick] (h4) -- (hh4);
	\draw[-stealth, thick] (h4) -- (hh5);
	\draw[-stealth, thick] (h5) -- (hh1);
	\draw[-stealth, thick] (h5) -- (hh2);
	\draw[-stealth, thick] (h5) -- (hh3);
	\draw[-stealth, thick] (h5) -- (hh4);
	\draw[-stealth, thick] (h5) -- (hh5);

	\draw[-stealth, thick] (hh1) -- (o1);
	\draw[-stealth, thick] (hh1) -- (o2);
	\draw[-stealth, thick] (hh2) -- (o1);
	\draw[-stealth, thick] (hh2) -- (o2);
	\draw[-stealth, thick] (hh3) -- (o1);
	\draw[-stealth, thick] (hh3) -- (o2);
	\draw[-stealth, thick] (hh4) -- (o1);
	\draw[-stealth, thick] (hh4) -- (o2);
	\draw[-stealth, thick] (hh5) -- (o1);
	\draw[-stealth, thick] (hh5) -- (o2);
	

	
	\node[circle, draw, thick, black, fill=black!10, right=7em of hh1] (i1) {};
	\node[circle, draw, thick, black, fill=black!10, above=2em of i1] (i2) {};
	\node[circle, draw, thick, above=2em of i2] (i3) {};
	\node[circle, draw, thick, below=2em of i1] (i4) {};
	\node[circle, draw, thick, below=2em of i4] (i5) {};
	
	\node[black] (icr) at (i1) {$\mathlarger{\mathlarger{\mathlarger{\mathlarger{\mathlarger{\bm{\times}}}}}}$};
	\node[black] (icr) at (i2) {$\mathlarger{\mathlarger{\mathlarger{\mathlarger{\mathlarger{\bm{\times}}}}}}$};
	
	\node[circle, draw, thick, black, fill=black!10, right=4em of i1] (h1) {};
	\node[circle, draw, thick, right=4em of i2] (h2) {};
	\node[circle, draw, thick, black, fill=black!10, right=4em of i3] (h3) {};
	\node[circle, draw, thick, black, fill=black!10, right=4em of i4] (h4) {};
	\node[circle, draw, thick, right=4em of i5] (h5) {};
	
	\node[black] (icr) at (h1) {$\mathlarger{\mathlarger{\mathlarger{\mathlarger{\mathlarger{\bm{\times}}}}}}$};
	\node[black] (icr) at (h3) {$\mathlarger{\mathlarger{\mathlarger{\mathlarger{\mathlarger{\bm{\times}}}}}}$};
	\node[black] (icr) at (h4) {$\mathlarger{\mathlarger{\mathlarger{\mathlarger{\mathlarger{\bm{\times}}}}}}$};
	
	\node[circle, draw, thick, right=4em of h1] (hh1) {};
	\node[circle, draw, thick, black, fill=black!10, right=4em of h2] (hh2) {};
	\node[circle, draw, thick, right=4em of h3] (hh3) {};
	\node[circle, draw, thick, black, fill=black!10, right=4em of h4] (hh4) {};
	\node[circle, draw, thick, right=4em of h5] (hh5) {};
	
	\node[black] (icr) at (hh2) {$\mathlarger{\mathlarger{\mathlarger{\mathlarger{\mathlarger{\bm{\times}}}}}}$};
	\node[black] (icr) at (hh4) {$\mathlarger{\mathlarger{\mathlarger{\mathlarger{\mathlarger{\bm{\times}}}}}}$};
	
	\node[circle, draw, thick, right=4em of hh2] (o1) {};
	\node[circle, draw, thick, right=4em of hh4] (o2) {};
	
	\draw[-stealth, thick] (i3) -- (h2);
	\draw[-stealth, thick] (i3) -- (h5);
	\draw[-stealth, thick] (i4) -- (h2);
	\draw[-stealth, thick] (i4) -- (h5);
	\draw[-stealth, thick] (i5) -- (h2);
	\draw[-stealth, thick] (i5) -- (h5);
	
	\draw[-stealth, thick] (h2) -- (hh1);
	\draw[-stealth, thick] (h2) -- (hh3);
	\draw[-stealth, thick] (h2) -- (hh5);
	\draw[-stealth, thick] (h5) -- (hh1);
	\draw[-stealth, thick] (h5) -- (hh3);
	\draw[-stealth, thick] (h5) -- (hh5);
	
	\draw[-stealth, thick] (hh1) -- (o1);
	\draw[-stealth, thick] (hh1) -- (o2);
	\draw[-stealth, thick] (hh3) -- (o1);
	\draw[-stealth, thick] (hh3) -- (o2);
	\draw[-stealth, thick] (hh5) -- (o1);
	\draw[-stealth, thick] (hh5) -- (o2);

\end{tikzpicture}
	\end{center}
  \caption[\bel{Dropout applied to a neural network.}]{\bel{\emph{left}: a standard neural network with two hidden layers. \emph{right}: An example of thinned network produced by applying dropout to the network on the left with $p=0.5$ across all the layers. Crossed units have been dropped. (Reference: \cite{srivastava14a})}}
  \label{fig:fig0-18-00-000-00-dropout}
\end{figure}

Dropout can be interpreted as a biological behavior. It pressures the hidden unit \fromont{to} be able to perform well regardless of which other hidden units are available. Hidden units must be ready to be swapped and interchanged between the subnetworks. Dropout \cite{srivastava14a} was inspired by an idea from biology: sexual reproduction, which involves swapping genes between two different organisms, creates evolutionary pressure for genes to become not just good but readily swapped between different organisms. Such genes and such features are robust to changes in their environment because they are not able to incorrectly adapt to unusual features of any one organism. Dropout thus regularizes each hidden unit to be merely a good feature but a feature that is good in many contexts. Other aspects of dropout with more details can be found in \cite{Goodfellowbook2016}.

As a second interpretation, dropout can be seen as a method of making bagging \cite{Breiman1996} practical for ensembles of very many large neural networks \cite{WardeFarleyGCB2014ICLR, Goodfellowbook2016}. However, this seems impractical when each model is a large neural network, since training and evaluating such networks is costly in term of runtime and memory. Dropout provides a hack to perform bagging on an ensemble of exponentially many neural networks.

Applying dropout to a neural network during training consists into sampling a \quotes{thinned} network from it. The thinned network consists of all the units that survived dropout. A neural network with $N$ units, can be seen as an ensemble of $2^N$ possible thinned network that share weights. For each representation of each training case, a new thinned network is sampled and trained. Therefore, training a neural network with dropout can be seen as training a collection of $2^N$ subnetworks with extensive weight sharing, where each subnetwork gets trained rarely, if at all. The first hack in dropout is to use a binary mask as a way of approximating sampling a subnetwork from the total neural network. The mask of each unit is sampled independently from the others. The probability of sampling a mask with value $1$ (causing a unit to be included in the subnetwork) is a hyperparameter $p$ fixed before the training begins. Typical choice for input unit is $0.8$ and for the hidden unit is $0.5$ \cite{Goodfellowbook2016, SrivastavaDropout2013, srivastava14a}.

The second hack involved in dropout is at the inference time. Instead of averaging a bench of subnetworks, dropout infers the prediction by evaluating only one model: the model with all units, but with the weights going out of unit $i$ multiplied by the probability of including unit $i$ \cite{SrivastavaDropout2013, srivastava14a, Goodfellowbook2016}. The motivation of this modification is to capture the right expected value of the output from that unit at the test phase. This is referred to as weight scaling inference rule. There is not yet any theoretical argument for the accuracy of this approximation inference rule in deep nonlinear networks, but, empirically, it performs very well \cite{SrivastavaDropout2013, srivastava14a, Goodfellowbook2016}.  Theoretical demonstration is provided only for $p=\frac{1}{2} = 0.5$ \cite{SrivastavaDropout2013, srivastava14a, Goodfellowbook2016}.

While dropout showed satisfying improvements on a large number of tasks, it also has its limitations. Due to randomness in the architecture, it increases the convergence time. Considering dropout as a regularization technique, it reduces the effective capacity of a model \cite{Goodfellowbook2016}. To offset this effect, one must increase the size of the model to compensate the missed units. However, this comes at the cost of a larger model and longer training time. Using larger model may allow the model to remember the dropout noise, which makes it worse. In this case, using other regularization techniques may be better. When extremely few labeled training examples are available, dropout is less effective \cite{srivastava14a}. On very large datasets, the obtained improvement is negligible, so for computational reasons it is recommended to not use it \cite{Goodfellowbook2016}. It was shown \cite{wagerNIPS2013} that when applying dropout to linear regression, it is equivalent to $L_2$ weight decay, with a different weight decay coefficient for each input feature. The magnitude of each feature's weight decay coefficient is determined by its variance. Similar results hold for other linear models. For deep models, dropout is not equivalent to weight decay \cite{Goodfellowbook2016}.

Dropout has motivated other stochastic approaches to training exponentially large ensembles of models that share weights such as DropConnect \cite{WanZZLFICML13}, and stochastic pooling \cite{matthew2013ICLR}. So far, dropout remains the most widely used implicit ensemble method.
}

\subsubsection{Batch Normalization}
\label{subsub:batchnormalization0}
Batch normalization \cite{IoffeICML15} is one of the most recent innovations in deep learning. Its primary purpose is to improve the optimization speed of deep neural networks. It is considered as a reparameterization of the model in a way that introduces both additive and multiplicative noise to the hidden units during training time. This noise can have a regularization effect and sometimes it makes dropout \cite{SrivastavaDropout2013, srivastava14a} unnecessary.

\emph{Covariance shift} \cite{Shimodaira2000} is a well known issue in machine learning. The problem raises when the input distribution of a model changes. For instance, when the test distribution is different than the train distribution, the model will perform poorly. For a long time, it has been  known \cite{LeCun1998backprop, Wiesler2011} that the neural network training converges faster if its input are whitened, i.e. linearly transformed to have zero mean and unit variances, and decorrelated.

\bel{D}eep \bel{neural} models consist in the composition of several layers. The computed gradient indicates how to update the parameters of each layer assuming that the other layers do not change. However, in practice, all the parameters are updated at once. When the updates are done, unexpected results can happen because many functions composed together have changed simultaneously. In other words, the stream of information in this hidden layer changes constantly. One possible way to deal with this is to consider higher level of interactions between layers, i.e., their parameters, such as second or higher order optimization. However, such optimization for deep models is impractical due to the computational cost. Batch normalization came up with a solution in order to stabilize the hidden distributions and thus prevent what is known as \emph{internal covariance shift} issue. Its strategy consists in maintaining the distribution at each hidden unit \bel{fixed}, i.e., normalized. 

\bel{Given an input minibatch}, let $\bm{H}$ \bel{be} a matrix that contains the activations of the layer to be normalized, with the activations for each example appearing in a row of the matrix. To normalize $\bm{H}$, it is replaced with
\begin{equation}
    \label{eq:eq0-141}
    \bm{H}^{\prime} = \frac{\bm{H} - \bm{\mu}}{\bm{\sigma}} \; ,
\end{equation}
\noindent where $\bm{\mu}$ is a vector containing the mean of each unit and $\bm{\sigma}$ is a vector containing the standard deviation of each unit. The calculation here are based on broadcasting the vector $\bm{\mu}$ and the vector $\bm{\sigma}$ to be applied to every row of the matrix $\bm{H}$. Within each row, the arithmetic is element-wise. Therefore, $H_{i,j}$ is normalized by subtracting $\mu_j$ and dividing by $\sigma_j$. The rest of the network then operates on $\bm{H}^{\prime}$ instead of $\bm{H}$. $\bm{\mu}$ and $\bm{\sigma}$ are estimated over each minibatch during train time
\begin{equation}
    \label{eq:eq0-142}
    \bm{\mu} = \frac{1}{m} \sum_{i} \bm{H}_{i, :} \; ,
\end{equation}

\begin{equation}
    \label{eq:eq0-143}
    \bm{\sigma} = \sqrt{\delta + \frac{1}{m} \sum_{i} (\bm{H} - \bm{\mu})^2_{i, :}} \; ,
\end{equation}
\noindent where $\delta$ is a small positive value to avoid numerical instability. The most important thing is that the backpropagation is performed through $\bm{\mu}$ and $\bm{\sigma}$ to compute the updates which means that the computed gradient will not change the units distributions which is a crucial innovation in this approach. Previous approaches \cite{Wiesler2014ICASSP, raikoPMLR2012, PoveyZK14CORR, DesjardinsNIPS2015} had involved adding penalties to the cost function to encourage units to have normalized activation statistics or involved intervening to re-normalize unit statistics after each gradient descent step. The former approach usually resulted in imperfect normalization and the latter usually resulted in significant waste of time, as the learning algorithm repeatedly proposed changing the mean and variance, and the normalization step repeatedly undid this change. Batch normalization reparametrizes the model to make some units always \bel{standardized} by definition.

At the test time, $\bm{\mu}$ and $\bm{\sigma}$ may be replaced by running averages that were collected during train time. This allows the model to be evaluated on a single sample at a time without the need to re-estimating these statistics that \bel{depend} on an  entire minibatch.
\bigskip

\bel{Normalizing} the mean and standard deviation of a unit can reduce the expressive power of the network that contains that unit. To maintain the expressive power of the network, it is common to replace the batch of hidden unit activation $\bm{H}$ with
\begin{equation}
    \label{eq:eq0-144}
    \gamma \mathbf{H}^{\prime} + \beta \; ,
\end{equation}
\noindent rather than simply the normalized $\bm{H}^{\prime}$. $\gamma$ and $\beta$ are learned parameters that allow the new variable to have any possible mean and standard deviation depending on the optimization problem. Setting the mean as a learnable parameter makes it easier than estimate it stochastically based on a minbatch.

\subsubsection{Unsupervised and Semi-supervised Learning}
\label{subsub:representationlearning0}
In machine learning, learning better representations usually leads to better generalization \cite{bengioetlecun2007, bengio2013corr}. In neural networks field, representations learning has been a hot topic for \fromont{a} long time and still is \cite{Goodfellowbook2016}. The main idea is to learn a model that provides adequate representation\bel{s} for the task in hand. In standard supervised learning \bel{setup}, learning such representation\bel{s} require\bel{s} large number of labeled data which may not be available. A possible solution to deal with this is to use unlabeled data either alone or combined with labeled data. Semi-supervised learning refers to a context where \bel{the learning is based on both: labeled and unlabeled data}. Unsupervised learning refers to a learning context where only data without labels are used. 

In semi-supervised learning, the main idea is to use unlabeled data to discover the underlying input domain distribution in \bel{an} unsupervised way. In most  cases, the supervised task partially shares its parameters with an unsupervised task. This may be seen as a regularization of the supervised task where the cost is constrained by the unsupervised cost \cite{erhan10whydoes, bengio07}. Most importantly, the unsupervised task introduces a generalization and prevents the supervised task from overfitting. This can be achieved through sharing general representations \cite{Goodfellowbook2016}. For instance, let us consider an unsupervised task that learns representation of trucks such as car, bus, motor-cycle. On the other hand, let us consider a supervised task that learns to recognize each of the previous trucks. One of the possible learned features in the unsupervised task is the concept of a \quotes{wheel} and more complicated feature \quotes{counting the number of wheels}. Learning both these tasks can be helpful because they are important features in the supervised task that aims at distinguishing between the different trucks. This type of unsupervised learning can be very helpful in the case where only few labeled data is available with a large number of unlabeled data \cite{Goodfellowbook2016}.

\bel{Since 2006, many works showed that unsupervised learning techniques can help to efficiently train deep feedforward networks \cite{hinton06SCI, hinton06NC, bengio06NIPS} using multilayer perceptrons (MLP). Such methods are mostly based on layer-wise pre-training where layers are pre-trained sequentially using a reconstruction criterion}. This allows to train Deep Belief Networks \cite{hinton06SCI, hinton06NC} using RBMs and deep feedforward networks using different variants of auto-encoders \cite{aurelio07, RanzatoBL2007, salah2011}. Pre-training techniques can be seen as a regularization to the supervised task where it allows to learn intermediate feature functions that provide general features. It may also be seen as a better initialization of the weights by pushing them toward better regions in the parameters space. Thus, it avoids local minima. More insights on the unsupervised learning can be found in \cite{erhan10whydoes}. \cite{lerouge15} provide a technical description of pre-training approach and auto-encoders.

Recently, there was an attempt to pre-train convolutional networks in an unsupervised way through reconstruction \cite{Masci2011, ZhangDNK17CORR, DosovitskiySTB17PAMI}, however these methods are still facing open issues such as the deconvolution.

Pre-training technique has been abandoned in the last few years due to its greedy approach, the \bel{availability} of large supervised data, and also to the improvements \bel{and accessibility to less greedy regularization} techniques. One can also mention its lack of  efficient stopping criterion to end the pre-training phase. Moreover, we did not see new advances in auto-encoders aspects in the last years. Today, learning better representations is achieved using \bel{a large} number of supervised data and more complicated/deep models such as convolutional networks \cite{krizhevsky12, Sermanet2013CVPR, farabet13}. 

\bel{In this thesis, we focus more on regularizing neural networks through learning better representations using unsupervised learning, prior knowledge, or transfer learning.}

\subsection{Summary}
\label{sub:summaryRNN0}
\bel{
Throughout this section, we have presented different variant of tools that are commonly used to prevent the overfitting of neural networks. We have divided such approaches into two categories: methods that reduce the model complexity, and methods that promote the model generalization without necessarily affecting its complexity.
}

\section[Conclusion]{Conclusion}
\label{sec:conclusion0}
We have presented in this first chapter \bel{of this thesis} an introduction to machine learning with a focus on the generalization aspect (Sec.\ref{sec:machinel0}), followed with an introduction to neural networks (Sec.\ref{sec:neuralnetsintro0}) and their overfitting issue. Finally, we closed this chapter by presenting some selected approaches used to improve the generalization of such models and alleviate their overfitting (Sec.\ref{sec:neuralnetsregularization0}). Such methods may aim directly at reducing the network complexity \bel{or aim directly at improving the generalization performance of the model}.

In this thesis, we tackled the overfitting issue of neural networks by focusing more on \bel{representation learning} within the model, particularly when using few labeled training samples. We performed the regularization of learning representations in neural networks either through  learning input/output representations using unsupervised approach \cite{belharbiCAP2015, belharbiESANN2016, belharbiNeurocomp2017}, incorporating prior knowledge on learning internal representations \cite{belharbiIEEETNNLS2017}, or using transfer learning \cite{BELHARBICBM2017}. All these contributions were made under the angle of tackling the overfitting issue of deep neural networks when dealing with small training sets which is the case in real life applications. We can divide our contributions into three different approaches:
\begin{itemize}
  \item Approach based on unsupervised learning (\autoref{chap:chapter4}): This provides an access to large unsupervised samples and  allows to learn the structure of the data without the need to large labeled data. This leads to improvements in the network performance \cite{belharbiCAP2015, belharbiESANN2016, belharbiNeurocomp2017}.
  \item Approach based on prior knowledge about the task (\autoref{chap:chapter5}): The key idea is to exploit a prior belief about the distribution of the internal representation, in the case of a classification task, which is: samples within the same class should have the same internal representation. Incorporating this prior knowledge into the network training allows improving its generalization while using small dataset for training \cite{belharbiIEEETNNLS2017}.
  \item Approach based on transfer learning (\autoref{chap:chapter6}): This idea consists in training a deep model on a large labeled data over a specific task $t_1$. Then, one takes a subset of the learned parameters and use them, combined with new parameters, to learn a second task $t_2$ which has few training samples. Usually, only the parameters that learn low representations are used. Such parameters seem to be common among different tasks, particularly in computer vision. This allows to obtain models with large capacity partially trained and use them over small datasets with success \cite{BELHARBICBM2017}.
\end{itemize}
The following chapters describe \bel{each of our contributions.}

\chapter[Deep Neural Networks Regularization for Structured Output \texorpdfstring{\\}{ } Prediction]{Deep Neural Networks Regularization for Structured Output Prediction}
\label{chap:chapter4}

\ifpdf
    \graphicspath{{Chapter4/Figs/Raster/}{Chapter4/Figs/PDF/}{Chapter4/Figs/}}
\else
    \graphicspath{{Chapter4/Figs/Vector/}{Chapter4/Figs/}}
\fi

\makeatletter
\def\input@path{{Chapter4/}}
\makeatother

\section{Prologue}

\noindent\emph{\underline{Article Details}:}
\begin{itemize}
    \item \textbf{Deep Neural Networks Regularization for Structured Output Prediction}. Soufiane Belharbi, Romain Hérault\footnote{\label{note1}Authors with equal contribution.}, Clément Chatelain\footnoteref{note1},  and Sébastien Adam. Neurocomputing Journal, 281C:169-177, 2018.
\end{itemize}

\noindent\emph{\underline{Other Related Publications}:}
\begin{itemize}
  \item \textbf{Learning Structured Output Dependencies Using Deep Neural Networks}. Soufiane Belharbi, Clément Chatelain, Romain Hérault, and Sébastien Adam. Deep Learning Workshop in the $32^{nd}$ International Conference on Machine Learning (ICML), 2015.
  \item \noindent \textbf{A Unified Neural Based Model For Structured Output Problems}. Soufiane Belharbi, Clément Chatelain, Romain Hérault, and Sébastien Adam. Conférence Francophone sur l'Apprentissage Automatique (CAP), 2015.
  \item \textbf{Deep Multi-Task Learning with Evolving Weights}. Soufiane Belharbi, Romain Hérault, Clément Chatelain, and Sébastien Adam. European Symposium on Artificial Neural Networks (ESANN), 2016.
  \item \textbf{Pondération Dynamique dans un Cadre Multi-Tâche pour Réseaux de Neurones Profonds}. Soufiane Belharbi, Romain Hérault, Clément Chatelain, and Sébastien Adam. Reconnaissance des Formes et l'Intelligence Artificielle (RFIA) (Session spéciale: Apprentissage et vision), 2016.
\end{itemize}

\bigskip

\noindent\emph{\underline{Context}:}
\\
We provide in this chapter our first contribution which concerns regularization of neural network in the context of structured output problems. Structured output problems is a set of problems where the output variable $\y$ is multi-dimensional and structural relations exist between its components. In this work, we aim at providing a regularization framework that makes use of unsupervised learning on both input $\X$ and more importantly on output $\Y$. Learning the input and output distribution allows to speed the training of the neural network and improves its generalization error. Moreover, this allows exploiting unlabeled inputs and/or label only outputs.

This work comes after a series of related works \cite{belharbiCAP2015, belharbiICMLWSDL2015, belharbiESANN2016, sbelharbiRFIA2016}. Our framework is composed of three tasks: two unsupervised tasks and a main supervised task. Our first proposition \cite{belharbiCAP2015, belharbiICMLWSDL2015} of this framework had a sequential optimization scheme. Although it shows significant improvements, it still has issues with respect to the optimization schedule which is difficult to tune and can easily lead to overfitting for the unsupervised tasks. Our first attempt to fix this issue was through the works \cite{belharbiESANN2016, sbelharbiRFIA2016} where we proposed a parallel optimization technique of pre-training \cite{hinton06SCI, hinton06NC, bengio06NIPS, aurelio07} instead of sequential one in hope to avoid overfitting and to reduce the number of hyperparameters one needs to setup. Through the work \cite{belharbiESANN2016}, we succeed to show that parallel optimization in this context of multi-tasking is better than sequential one. Later on, we extend this parallel optimization setup to our framework for solving structured output problems where the three tasks are optimized at once which led to this work \cite{belharbiNeurocomp2017}. Evaluated on a facial landmark detection problem, it allows to improve the generalization of the network and add more speed to their training. Furthermore, we show experimentally the possibility to use label-only data in an unsupervised way to improve more the generalization. We present this final work \cite{belharbiNeurocomp2017} as a chapter of this thesis under the form of one contribution. This chapter contains the original paper as it was accepted in Neurocomputing journal with slight adaptation of notation.

\bigskip

\noindent\emph{\underline{Contributions}:}
\\
The contribution of this paper is to provide a parallel strategy to optimize the framework proposed initially in \cite{belharbiCAP2015, belharbiICMLWSDL2015} to solve structured output problems. The proposed strategy showed a significant improvement in the generalization of the network. Furthermore, we showed the possibility to use label-only data in an unsupervised fashion over the output which allowed to improve more the generalization.

\newpage


\section[Introduction]{Introduction}
\label{sec:introduction4}
\begin{figure}[!htbp]
     \centering
     \includegraphics[width=0.2\linewidth]{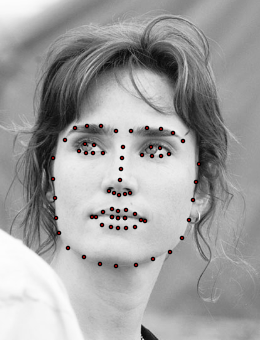}
     \includegraphics[width=0.2\linewidth]{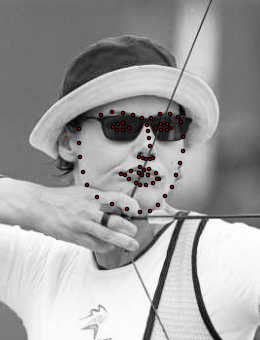}
     \includegraphics[width=0.2\linewidth]{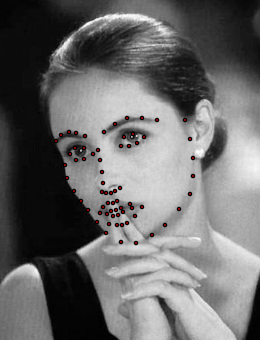}
     \includegraphics[width=0.2\linewidth]{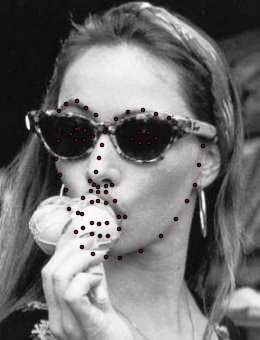}
     \caption[\bel{Examples of facial landmarks from LFPW training set.}]{Examples of facial landmarks from LFPW \citep{belhumeur11} 
   training set.}
     \label{fig:fig4-0}
\end{figure}

In \fromont{the} machine learning field, the main task usually consists in learning general regularities over the input space in order to provide a specific output. Most of machine learning applications aim at predicting a single value: a label for classification or a scalar value for regression. Many recent applications address challenging problems where the output lies in a multi-dimensional space describing discrete or continuous variables that are most of the time  interdependent. 
A typical example is speech recognition, where the output label is a sequence of characters which are interdependent, following the statistics of the considered language.
These dependencies generally constitute a regular structure such as a sequence, a string, a tree or a graph. As it provides constraints that may help the prediction, this structure should be either discovered if unknown, or integrated in the learning algorithm using prior assumptions. The range of applications that deal with structured output data is large. One can cite, among others, image labeling \cite{farabet13, longSD15, nohHH15, ronnebergerFB15, ZhangDG17, HoffmanWYD16, LiUBBSB16, SohnLY15}, statistical natural language processing (NLP) \cite{JaderbergSVZ14b2014, och03, sleator95, schmid94}, bioinformatics \cite{jones99, syed09}, speech processing \cite{rabiner89, zen09} and handwriting recognition \cite{graves2009novel,Stuner16}. Another example which is considered in the evaluation of our proposal in this paper is the facial landmark detection problem. The task consists in predicting the coordinates of a set of keypoints given the face image as input (Fig.\ref{fig:fig4-0}). The set of points are interdependent throughout geometric relations induced by the face structure. Therefore, facial landmark detection can be considered as a structured output prediction task.

One main difficulty in structured output prediction is the exponential number of possible configurations of the output space. From a statistical point of view, learning to predict accurately high dimensional vectors requires a large amount of data where in practice we usually have limited data. In this article we propose to consider structured output prediction as a representation learning problem, where the model must i) capture the discriminative relation between $\x$ (input) and $\y$ (output), and ii) capture the interdependencies laying between the variables of each space by efficiently modeling the input and output  distributions. We address this modelization through a regularization scheme for training neural networks. Feedforward neural networks lack exploiting the structural information between the $\y$ components. Therefore, we incorporate in our framework an unsupervised task which aims at discovering this hidden structure. The advantage of doing so is there is no need to fix beforehand any prior structural information. The unsupervised task learns it on itself.

Our contributions is a multi-task framework dedicated to train feedforward neural networks models for structured output prediction. We propose to combine unsupervised tasks over the input and output data in parallel with the supervised task. This parallelism can be seen as a regularization of the supervised task which helps it to generalize better. Moreover, as a second contribution, we demonstrate experimentally the benefit of using the output labels $\y$ without their corresponding inputs $\x$. In this work, the multi task framework is instantiated using auto-encoders \cite{vincent10, bengio07} for both representations learning and exploiting  unlabeled data (input) and label-only data (output). We demonstrate the efficiency of our proposal over a real-world facial landmark detection problem.

The rest of the paper is organized as follows. Related works about structured output prediction is proposed in section \ref{sec:relatedw4}. Section \ref{sec:mtl4} presents the proposed formulation and its optimization details. Section \ref{sec:impl4} describes the instantiation of the formulation using a deep neural network. Finally, section \ref{sec:expes4} details the conducted experiments including the datasets, the evaluation metrics and the general training setup. Two types of experiments are explored: with and without the use of unlabeled data. Results are presented and discussed for both cases.

\section[Related Work]{Related work}
\label{sec:relatedw4}
We distinguish two main categories of methods for structured output prediction.  For a long time, graphical models have showed a large success in different applications involving 1D and 2D signals. Recently, a new trend has emerged based on deep neural networks.

\subsection{Graphical Models Approaches}
\label{sub:graphicalmodels4}
Historically, graphical models are well known to be suitable for learning structures. One of their main strength is an easy integration of explicit structural constraints and prior knowledge directly into the model's structure. They have shown a large success in modeling structured data thanks to their capacity to capture dependencies among relevant random variables. For instance, Hidden Markov Models (HMM) framework has a large success in modeling sequence data. HMMs make an assumption that the output random variables are supposed to be independent which is not the case in many real-world applications where strong relations are present. Conditional Random Fields (CRF) have been proposed to overcome this issue, thanks to its capability to learn large dependencies of the observed output data. These two frameworks are widely used to model structured output data represented as a 1-D sequence \cite{elyacoubi02,rabiner89,bikel99,lafferty01}. Many approaches have also been proposed to deal with 2-D structured output data as an extension of HMM and CRF. \cite{nicolas06} propose a Markov Random Field (MRF) for document image segmentation. \cite{szummer04} provide an adaptation of CRF to 2-D signals with hand drawn diagrams interpretation. Another extension of CRF to 3-D signal is presented in \cite{tsechpenakis07} for 3-D medical image segmentation. Despite the large success of graphical models in many domains, they still encounter some difficulties. For instance, due to their inference computational cost, graphical models are limited to low dimensional structured output problems. Furthermore, HMM and CRF models are generally used with discrete output data where few works address the regression problem \cite{noto12,fridman93}.

\subsection{Deep Neural Networks Approaches}

More recently, deep learning based approaches have been widely used to solve structured output prediction, especially proposed for image labeling problems. Deep learning domain provides many different architectures. Therefore, different solutions were proposed depending on the application in hand and what is expected as a result.

In image labeling task (also known as semantic segmentation), one needs models able to adapt to the large variations in the input image. Given their large success in image processing related tasks \cite{krizhevsky12}, convolutional neural networks is a natural choice. Therefore, they have been used as the core model in image labeling problems in order to learn the relevant features. They have been used either combined with simple post-processing in order to calibrate the output \cite{ciresanGGS12} or with more sophisticated models in structure modeling such as CRF \cite{farabet13} or energy based models \cite{ningDLPBB05}. Recently, a new trend has emerged, based on the application of convolution  \cite{longSD15, ronnebergerFB15} or deconvolutional \cite{nohHH15} layers in the output of the network which goes by the name of fully convolutional networks and showed successful results in image labeling. Despite this success, these models does not take in consideration the output representation.

In many applications, it is not enough to provide the output prediction, but also its probability. In this case, Conditional Restricted Boltzmann Machines, a particular case of neural networks and probabilistic graphical models have been used with different training algorithms according to the size of the plausible output configurations  \cite{mnihLH11}. Training and inferring using such models remains a difficult task. In this same direction, \cite{belangerM16} tackle structured output problems as an energy minimization through two feed-forward networks. The first is used for feature extraction over the input. The second is used for estimating an energy by taking as input the extracted features and the current state of the output labels. This allows learning the interdependencies within the output labels. The prediction is performed using an iterative backpropagation-based method with respect to the labels  through the second network which remains computationally expensive. Similarly, Recurrent Neural Networks (RNN) are a particular architecture of neural networks. They have shown a great success in modeling sequence data and outputing sequence probability for applications such as Natural Language Processing (NLP) tasks \cite{liu2014, sutskeverVL14, auliGQZ13} and speech recognition \cite{gravesJ14}. It has also been used for image captioning \cite{karpathyL15}. However, RNN models doe not consider explicitly the output dependencies.

In \cite{lerouge15}, our team proposed the use of auto-encoders in order to learn the output distribution in a pre-training fashion with application to image labeling with promising success. The approach consists in two sequential steps. First, an input and output pre-training is performed in an unsupervised way using autoencoders. Then, a finetune is applied on the whole network using supervised data. While this approach allows incorporating prior knowledge about the output distribution, it has two main issues. First, the alteration of a network output layer is critical and must be performed carefully. Moreover, one needs to perform multiple trial-error loops in order to set the autoencoder's training hyper-parameters. The second issue is overfitting. When pre-training the output auto-encoder, there is actually no information that indicates if the pre-training is helping the supervised task, nor when to stop the pre-training.

The present work proposes a general and easy to use multi-task training framework for structured output prediction models. The input and the output unsupervised tasks are embedded into a regularization scheme and learned in parallel with the supervised task. The rationale behind is that the unsupervised tasks should provide a generalization aspect to the main supervised task and should limit overfitting. This parallel transfer learning which includes an output reconstruction task constitutes the main contribution of this work. In structured output context, the role of the output task is to learn the hidden structure within the original output data, in an unsupervised way. This can be very helpful in models that do not consider the relations between the components of the output representation such as feedforward neural networks. We also show that the proposed framework enables to use labels without input in an unsupervised fashion and its effect on the generalization of the model. This can be very useful in applications where the output data is abundant such as in a speech recognition task where the output is ascii text which can be easily gathered from Internet. In this article, we validate our proposal on a facial landmark prediction problem over two challenging public datasets (LFPW and HELEN). The performed experiments show an improvement of the generalization of deep neural networks and an acceleration of their training.

\section[Multi-task Training Framework for Structured Output Prediction]{Multi-task Training Framework for Structured Output Prediction}
\label{sec:mtl4}
Let us consider a training set $\mathbb{D}$ containing examples with both features and targets $(\xvec,\yvec)$,  features without target~$(\xvec,\_)$, and targets without features~$(\_,\yvec)$.
Let us consider a set $\mathbb{F}$ which is the subset of $\mathbb{D}$ containing examples with at least features~$\xvec$, a set $\mathbb{L}$ which is the subset of  $\mathbb{D}$ containing examples with at least targets~$\yvec$, and a set $\mathbb{S}$ which is the subset of  $\mathbb{D}$ containing examples with both features~$\xvec$ and targets~$\yvec$.
One can note that all examples in $\mathbb{S}$ are also in $\mathbb{F}$ and in $\mathbb{L}$.

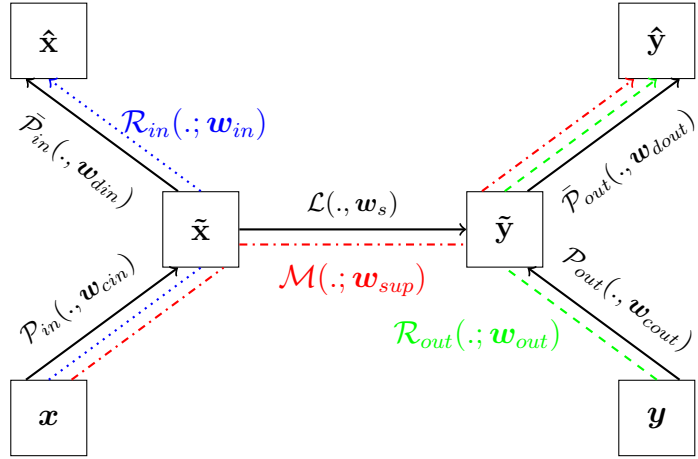
\begin{figure}[h!]
	\begin{center}
		\begin{tikzpicture}

\tikzstyle{state} = [rectangle,draw=black,minimum width=10mm,minimum height=10mm];

\coordinate (csw) at (0,0);
\coordinate (cne) at (80mm,50mm);
\coordinate (cse) at (csw-|cne);
\coordinate (cnw) at  (cne-|csw);
\path (csw)--(cse) coordinate[pos=0.25] (axp) coordinate[pos=0.75] (ayp);
\path (cse)--(cne) coordinate[pos=0.5] (omid);

\node[state] (x) at (csw) {$\xvec$};
\node[state] (xproj) at (omid-|axp) {$\xTilde$};
\node[state] (xest) at (cnw) {$\xHat$};

\node[state] (y) at (cse) {$\yvec$};
\node[state] (yproj) at (omid-|ayp) {$\yTilde$};
\node[state] (yest) at (cne) {$\yHat$};

\path[postaction={transform canvas={xshift=-3mm},draw,->,thick}] (x.north)--(xproj.south) node[pos=0.5,sloped,above,yshift=3mm,xshift=-3mm] {\footnotesize $\mathcal{P}_{in}(.,\bm{w}_{cin})$};
\path[draw,thick,dotted,blue] (x.north)--(xproj.south);
\path[postaction={transform canvas={xshift=3mm},draw,thick,dashdotted,red}] (x.north)--(xproj.south);

\path[postaction={transform canvas={xshift=-3mm},draw,->,thick}] (xproj.north)--(xest.south) node[pos=0.5,sloped,below,yshift=-3mm,xshift=-3mm] {\footnotesize $\bar{\mathcal{P}}_{in}(.,\bm{w}_{din})$};
\path[draw,thick,dotted,blue,->] (xproj.north)--(xest.south) node[pos=0.6,right,blue] {$\mapIn(.; \bm{w}_{in})$};

\path[postaction={transform canvas={xshift=3mm},draw,->,thick}] (y.north)--(yproj.south) node[pos=0.5,sloped,above,yshift=3mm,xshift=3mm] {\footnotesize $\mathcal{P}_{out}(.,\bm{w}_{cout})$};
\path[draw,thick,dashed,green] (y.north)--(yproj.south) node[pos=0.4,left=3mm,green] {$\mapOut(.; \bm{w}_{out})$};

\path[postaction={transform canvas={xshift=3mm},draw,->,thick}] (yproj.north)--(yest.south) node[pos=0.5,sloped,below,yshift=-3mm,xshift=3mm] {\footnotesize $\bar{\mathcal{P}}_{out}(.,\bm{w}_{dout})$};
\path[draw,thick,dashed,green,->] (yproj.north)--(yest.south) ;
\path[postaction={transform canvas={xshift=-3mm},draw,thick,dashdotted,red,->}] (yproj.north)--(yest.south);

\draw[->,thick] (xproj)--(yproj) node[pos=0.5,above] {\footnotesize $\mathcal{L}(.,\bm{w}_{s})$};
\path[postaction={transform canvas={yshift=-2mm},draw,thick,dashdotted,red}] (xproj)--(yproj) node[pos=0.5,below=3mm,red] {$\mapSup(.; \bm{w}_{sup})$};


\end{tikzpicture}
	\end{center}
	\caption[\bel{Proposed MTL framework.}]{Proposed MTL framework. Black plain arrows stand for intermediate functions, blue dotted arrow for input auxiliary task $\mathcal{R}_{in}$, green dashed arrow for output auxiliary task $\mathcal{R}_{out}$, and red dash-dotted arrow for the main supervised task $\mathcal{M}$.}
	\label{fig:fig4-0.1}
\end{figure}

\begin{description}
	
\item[Input task]

The input task $\mathcal{R}_{in}$ is an unsupervised reconstruction task which aims at learning global and more robust input representation based on the original input data $\xvec$.
This task projects the input data $\xvec$ into an intermediate representation space $\xproj$ through a coding function $\mathcal{P}_{in}$, known as encoder. Then, it attempts to recover the original input by reconstructing $\xest$ from $\xproj$ through a decoding function $\bar{\mathcal{P}}_{in}$, known as decoder
\begin{equation}
  \label{eq:eq4-00}
\xest=\mathcal{R}_{in}\left(\xvec; \bm{w}_{in}\right) =\bar{\mathcal{P}}_{in}\left(\xproj=\mathcal{P}_{in}\left(\xvec;\bm{w}_{cin}\right);\bm{w}_{din}\right) \; ,
\end{equation}

\noindent where $\bm{w}_{in}=\{\bm{w}_{cin}, \bm{w}_{din}\}$.
The decoder parameters $\bm{w}_{din}$ are proper to this task however the encoder parameters $\bm{w}_{cin}$ are shared with the main task (see Fig.\ref{fig:fig4-0.1}). This multi-task aspect will attract, hopefully, the shared parameters in the parameters space toward regions that build more general and robust input representations and avoid getting stuck in local minima. Therefore, it promotes generalization. This can be useful to start the training process of the main task.

The training criterion for this task is given by

\begin{equation}
  \label{eq:eq4-2}
  J_{in}(\mathbb{F}; \bm{w}_{in}) = \frac{1}{\card{\mathbb{F}}}\sum\limits_{\xvec \in
    \mathbb{F}}\cost_{in}(\mathcal{R}_{in}(\xvec; \bm{w}_{in}) , \xvec) \; ,
\end{equation}
\noindent where $\cost_{in}(\cdot, \cdot)$ is an unsupervised learning cost which can be computed on all the samples with features (i.e. on $\mathbb{F}$). Practically, it can be the mean squared error.

\item[Output task]

The output task $\mathcal{R}_{out}$ is an unsupervised reconstruction task which has the same goal as the input task.
Similarly, this task projects the output data $\yvec$ into an intermediate representation space $\yproj$ through a coding function $\mathcal{P}_{out}$, i.e. a coder. Then, it attempts to recover the original output data by reoncstructing  $\yest$ based on $\yproj$ through a decoding function $\bar{\mathcal{P}}_{out}$, i.e. a decoder. In structured output data, $\yproj$ can be seen as a code that contains many aspect of the original output data $\yvec$, most importantly, its hidden structure that describes the global relation between the components of $\yvec$. This hidden structure is discovered in an unsupervised way without  priors fixed beforehand which makes it simple to use. Moreover, it allows using labels only (without input $\xvec$) which can be helpful in tasks with abundant output data such as in speech recognition task (Sec.\ref{sec:relatedw4})
\begin{equation}
  \label{eq:eq4-2-1}
\yest=\mathcal{R}_{out}\left(\yvec; \bm{w}_{out}\right) =\bar{\mathcal{P}}_{out}\left(\yproj=\mathcal{P}_{out}\left(\yvec;\bm{w}_{cout}\right);\bm{w}_{dout}\right) \; .
\end{equation}

\noindent where $\bm{w}_{out} = \{\bm{w}_{cout}, \bm{w}_{dout}\}$.
In the opposite of the input task, the encoder parameters $\bm{w}_{cout}$ are proper to this task while the decoder parameters $\bm{w}_{dout}$ are shared with the main task (see Fig.\ref{fig:fig4-0.1}).

The training criterion for this task is given by
\begin{equation}
  \label{eq:eq4-3}
  J_{out}(\mathcal{L}; \bm{w}_{out}) = \frac{1}{\card{\mathbb{L}}}\sum\limits_{\yvec \in \mathbb{L}}\cost_{out}(\mathcal{R}_{out}(\yvec;
  \bm{w}_{out}) , \yvec) \; ,
\end{equation}
\noindent where $\cost_{out}(\cdot, \cdot)$ is an unsupervised learning cost which can be computed on all the samples with labels (i.e. on $\mathbb{L}$), typically, the mean squared error.

\item[Main task]

The main task is a supervised task that attempts to learn the mapping function $\mathcal{M}$ between features $\xvec$ and labels $\yvec$.
In order to do so, the first part of the mapping function is shared with the encoding part $\mathcal{P}_{in}$ of the input task and the last part is shared with the decoding part $\bar{\mathcal{P}}_{out}$ of the output task. The middle part $\mathcal{L}$ of the mapping function $\mathcal{M}$ is specific to this task
\begin{equation}
  \label{eq:eq4-4}
\yest=\m\left(\xvec; \bm{w}_{sup}\right) =\bar{\mathcal{P}}_{out}\left(\mathcal{L}\left(\mathcal{P}_{in}\left(\xvec;\bm{w}_{cin}\right);\bm{w}_{s}\right);\bm{w}_{dout}\right) \; .
\end{equation}
\noindent where  $\bm{w}_{sup}=\{\bm{w}_{cin}, \bm{w}_{s}, \bm{w}_{dout}\}$. Accordingly, $\bm{w}_{cin}$ and $\bm{w}_{dout}$ parameters are respectively shared with the input and output tasks.

Learning this task consists in minimizing its learning criterion $J_s$, 
\begin{equation}
  \label{eq:eq4-1}
  J_{s}(\mathbb{S}; \bm{w}_{sup}) = \frac{1}{\card{\mathbb{S}}}
  \sum\limits_{(\xvec, \yvec) \in \mathbb{S}}\cost_{s}(\mathcal{M}(\xvec; \bm{w}_{sup}) , \yvec) \; ,
\end{equation}
\noindent where $\cost_{s}(\cdot, \cdot)$ can be the mean squared error.
\end{description}

\bigskip

\FloatBarrier

As a synthesis, our proposal is formulated as a multi-task learning framework (MTL) \cite{caruana97ML}, which gathers a main task and two secondary tasks. This framework is illustrated in Fig. \ref{fig:fig4-0.1}.

Learning the three tasks is performed in parallel.
This can be translated in terms of training cost as the sum of the corresponding costs.
Given that the tasks have different importance, we weight each cost using a corresponding importance weight $\lambda_{sup}$, $\lambda_{in}$ and $\lambda_{out}$ respectively for the supervised, the input and output tasks.
Therefore, the full objective of our framework can be written as
\begin{equation}
  \label{eq:eq4-5-0}
  J(\mathbb{D}; \bm{w})  =  \lambda_{sup} \times J_{s}(\mathbb{S}; \bm{w}_{sup})
  + \lambda_{in} \times J_{in}(\mathbb{F}; \bm{w}_{in})
  +\lambda_{out} \times J_{out}(\mathbb{L}; \bm{w}_{out})\; ,
\end{equation}
\noindent where $\bm{w} = \{ \bm{w}_{cin}, \bm{w}_{din}, \bm{w}_{s}, \bm{w}_{cout}, \bm{w}_{dout}\}$ is the complete set of parameters of the framework.

Instead of using fixed importance weights that can be difficult to optimally set, we \fromont{adapt} them through the learning epochs. In this context, Eq. \ref{eq:eq4-5-0} is modified as follows
\begin{align}
\label{eq:eq4-5}
J(\mathbb{D}; \bm{w})  &=  \lambda_{sup}(t) \times J_{s}(\mathbb{S}; \bm{w}_{sup}) \nonumber \\
& + \lambda_{in}(t) \times J_{in}(\mathbb{F}; \bm{w}_{in})
+\lambda_{out}(t) \times J_{out}(\mathbb{L}; \bm{w}_{out})\; ,
\end{align}
\noindent where $t \ge 0$ indicates the learning epochs. Our motivation to \fromont{adapt} the importance weights is that we want to use the secondary tasks to start the training and avoid the main task to get stuck in local minima early in the beginning of the training by moving the parameters towards regions that generalize better. Then, toward the end of the training, we drop the secondary tasks by annealing their importance toward zero because they are no longer necessary for the main task. The early stopping of the secondary tasks is important in this context of mult-tasking as shown in \cite{zhang14ECCV} otherwise, they will overfit, therefore, they will harm the main task. The main advantage of Eq.\ref{eq:eq4-5} is that it allows an interaction between the main supervised task and the secondary tasks. Our hope is that this interaction will promote the generalization aspect of the main task and prevent it from overfitting.

\section{Implementation}
\label{sec:impl4}

In this work, we implement our framework throughout a deep neural network.
The main supervised task is performed using a deep neural network (DNN) with $K$ layers.
Secondary reconstruction tasks are carried out by auto-encoders (AE): the input task is achieved using an AE that has $K_{in}$ layers in its encoding part, with an encoded representation of the same dimension as $\xproj$. Similarly, the output task is achieved using an AE that has $K_{out}$ layers in its decoding part, with an encoded representation of the same dimension as $\yproj$.
At least one layer must be dedicated in the DNN to link $\xproj$ and $\yproj$ in the intermediate spaces. Therefore, $K_{in} + K_{out} < K$.

Parameters $\bm{w}_{in}$ are the parameters of the whole input AE, $\bm{w}_{out}$ are the parameters of the whole output AE and $\bm{w}_{sup}$ are the parameters of the main neural network (NN).
The encoding layers of the input AE are tied to the first layers of the main NN, and the decoding layers of the output AE are in turn tied to the last layers of the main NN.
If $\bm{w}_i$ are the parameters of layer $i$ of a neural network, then $\bm{w}_1$ to $\bm{w}_{K_{in}}$ parameters of the input AE are shared with $\bm{w}_1$ to $\bm{w}_{K_{in}}$ parameters of the main NN.
Moreover, if $\bm{w}_{-i}$ are the parameters of last minus $i-1$ layer of a neural network, then parameters $\bm{w}_{-K_{out}}$ to $\bm{w}_{-1}$ of the output AE are shared with the parameters  $\bm{w}_{-K_{out}}$ to $\bm{w}_{-1}$ of the main NN.

During training, the loss function of the input AE is used as $J_{in}$, the loss function of the output AE is used as $J_{out}$, and the loss function of the main NN is used as $J_{s}$.

Optimizing Eq.\ref{eq:eq4-5} can be performed using Stochastic Gradient Descent. In the case of task combination, one way to perform the optimization is to alternate between the tasks when needed \cite{collobert08ICML, zhang14ECCV}. In the case where the training set does not contain unlabeled data, the optimization of Eq.\ref{eq:eq4-5} can be done in parallel over all the tasks. When using unlabeled data, the gradient for the whole cost can not be computed at once. Therefore, we need to split the gradient for each sub-cost according to the nature of the samples at each mini-batch.
For the sake of clarity, we illustrate our optimization scheme in Algorithm \ref{alg:alg4-0} using on-line training (i.e. training one sample at a time). Mini-batch training can be performed in the same way.

\begin{algorithm}
    \caption{Our training strategy for one epoch}
    \label{alg:alg4-0}
    \resizebox{1.\textwidth}{!}{%
\begin{minipage}{1.2\textwidth}
    \begin{algorithmic}[1]
      \State $\mathbb{D}$ is the shuffled training set. $B$ a sample.
      \For{$B$ in $\mathbb{D}$}
        \If{$B$ contains $\xvec$}
        	\State \texttt{Update $\bm{w}_{in}$}: Make a gradient step toward $\lambda_{in} \times J_{in}$ using $B$ (Eq.\ref{eq:eq4-2}).
        \EndIf
        \If{$B$ contains $\yvec$}
        \State \texttt{Update $\bm{w}_{out}$}: Make a gradient step toward $\lambda_{out} \times J_{out}$ using $B$ (Eq.\ref{eq:eq4-3}).
        \EndIf
        \State \texttt{\# parallel parameters update}
        \If{$B$ contains $\xvec$ and $\yvec$}
        \State \texttt{Update $\bm{w}$}: Make a gradient step toward $J$ using $B$
        (Eq.\ref{eq:eq4-5}).
        \EndIf
	\State \texttt{Update $\lambda_{sup}$, $\lambda_{in}$ and $\lambda_{out}$}.
      \EndFor
    \end{algorithmic}
    \end{minipage}
}
\end{algorithm}

\section[Experiments]{Experiments} 
\label{sec:expes4}
We evaluate our framework on a facial landmark detection problem which is typically a structured output problem since the facial landmarks are spatially  inter-dependent. 
Facial landmarks are a set of key points on human face images as shown in Fig.  \ref{fig:fig4-0}. Each key point is defined by the  coordinates $(x,y)$ in the image ($(x, y) \in \R^2$). The number of landmarks is dataset or application dependent.

It must be emphasized here that the purpose of our experiments in this paper was not to outperform the state of the art in facial landmark detection but to show that learning the output dependencies helps improving the performance of DNN on that task. Thus, we will compare a model with/without input and output training. \cite{zhang14} use a cascade of neural networks. In their work, they provide the performance of their first global network. Therefore, we will use it as a reference to compare our performance (both networks \fromont{have the same number of layers}) except they use larger training dataset.

We first describe the datasets followed by a description of the evaluation metrics used in facial landmark problems. Then, we present the general setup of our experiments followed by two types of experiments: without and with unlabeled data. An opensource implementation of our MTL deep instantiation is available online\footnote{\url{https://github.com/sbelharbi/structured-output-ae}}.

\subsection{Datasets}

\mbox{} We have carried out our evaluation over two challenging public datasets
for facial landmark detection problem: LFPW \cite{belhumeur11} and HELEN
\cite{le12}.

\textbf{LFPW dataset} consists of 1132 training images and 300 test images
taken under unconstrained conditions (in the wild) with large variations in the
pose, expression, illumination and with partial occlusions
(Fig.\ref{fig:fig4-0}). This makes the facial point detection a challenging task
on this dataset. From the initial dataset described in LFPW
\cite{belhumeur11}, we use only the 811 training images and the 224 test
images provided by the ibug website\footnote{300 faces in-the-wild challenge
 \url{http://ibug.doc.ic.ac.uk/resources/300-W/}}. Ground truth annotations of
68 facial points are provided by \cite{sagonas13}. We divide the available
training samples into two sets: validation set (135 samples) and training set
(676 samples).

\begin{figure}[!htbp]
 \centering
 \includegraphics[height=0.2\linewidth]{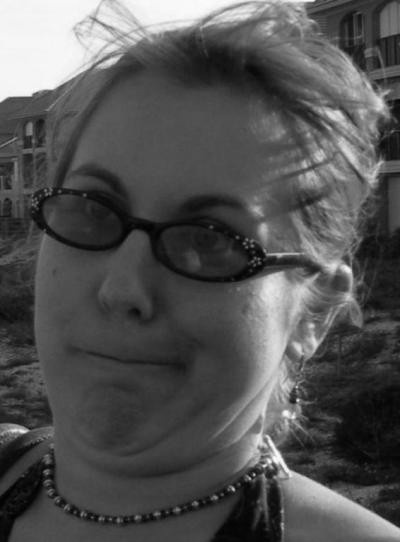}
 \includegraphics[height=0.2\linewidth]{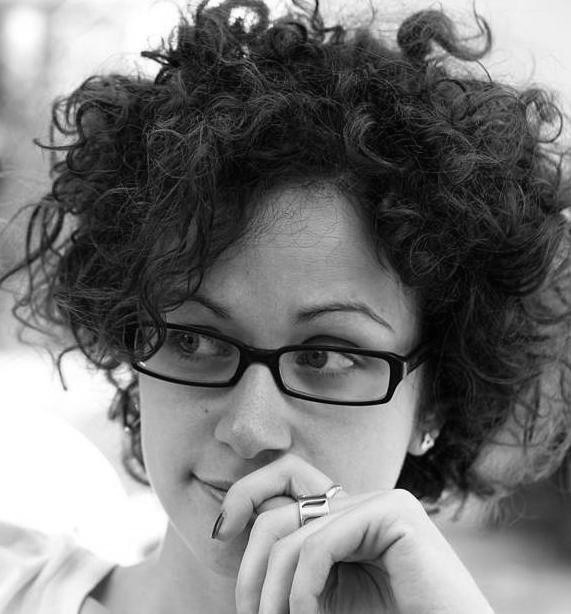}
 \includegraphics[height=0.2\linewidth]{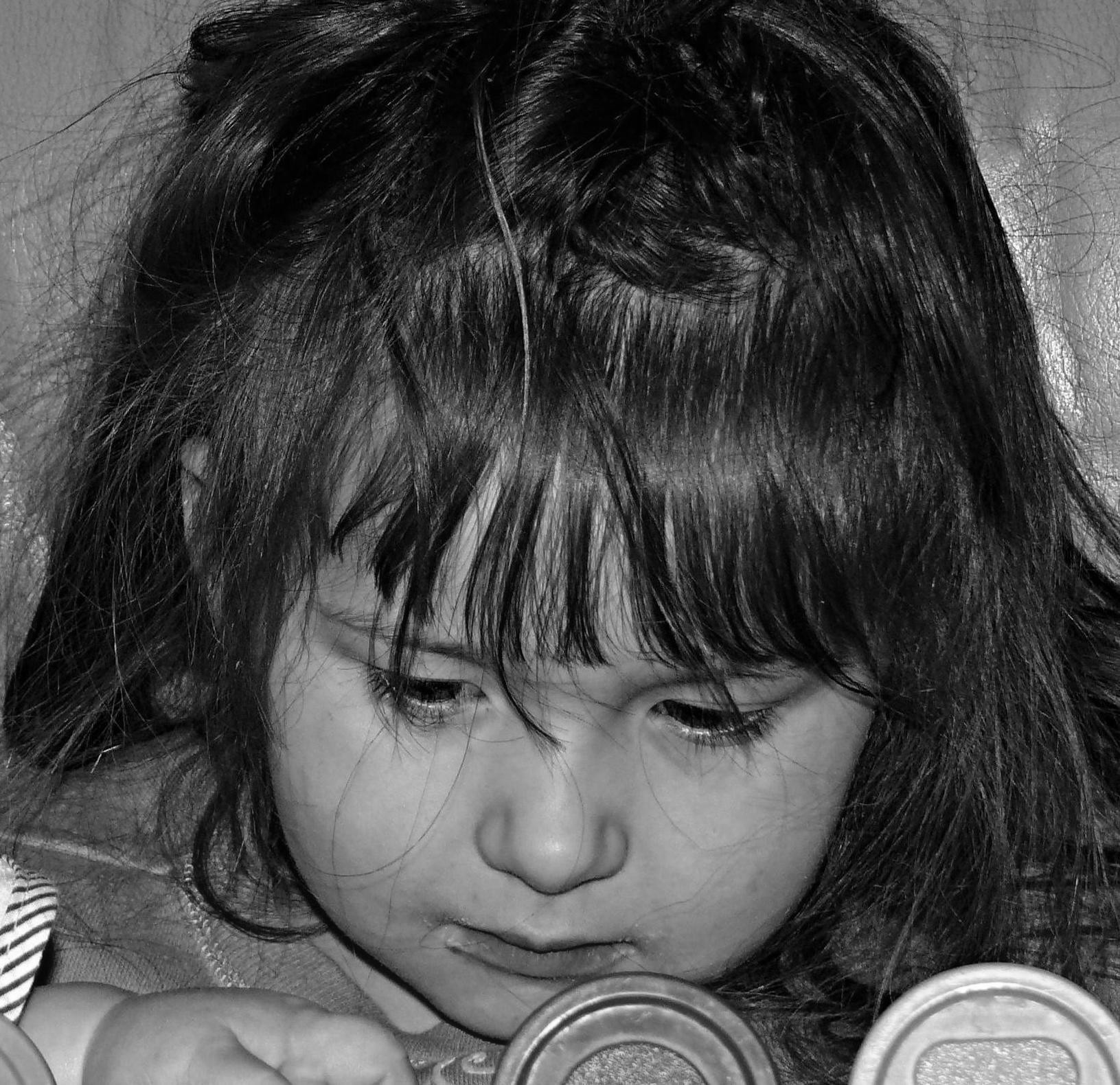}
 \includegraphics[height=0.2\linewidth]{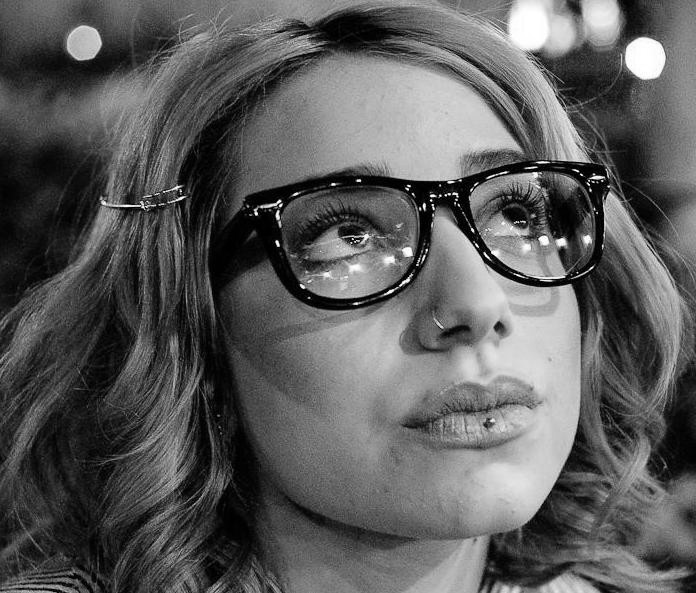}
 \caption[\bel{Samples from HELEN dataset.}]{Samples from HELEN \cite{le12} dataset.}
 \label{fig:fig4-1}
 \end{figure}

\textbf{HELEN dataset} is similar to LFPW dataset, where the images have
been taken under unconstrained conditions with high resolution and collected
from Flikr using text queries. It contains 2000 images for training, and 330
images for test. Images and face bounding boxes are provided by the
same site as for LFPW. The ground truth annotations are provided by
\cite{sagonas13}.  Examples of dataset are shown in Fig.\ref{fig:fig4-1}.

All faces are cropped into the same size ($50 \times 50$) and pixels are normalized in [0,1]. The facial landmarks are normalized into [-1,1].

\subsection{Metrics}
In order to evaluate the prediction of the model, we use the standard metrics
used in facial landmark detection problems.

The Normalized Root Mean Squared Error (NRMSE) \cite{cristinacce06}
(Eq.\ref{eq:eq4-22}) is the Euclidean distance between the predicted shape and the
ground truth normalized by the product of the number of points in the shape and
the inter-ocular distance $D$ (distance between the eyes pupils of the ground
truth),
\begin{equation}
  \label{eq:eq4-22}
NRMSE(s_p, s_g) = \frac{1}{N*D} \sum^N_{i=1} ||s_{pi} - s_{gi}||_2 \enspace ,
\end{equation}
where $s_p$ and $s_g$ are the predicted and the ground truth shapes,
respectively. Both shapes have the same number of points $N$.  $D$ is the
inter-ocular distance of the shape $s_g$.

Using the NMRSE, we can calculate the Cumulative Distribution Function for a
specific NRMSE ($CDF_{NRMSE}$) value (Eq.\ref{eq:eq4-33}) overall the database,
\begin{equation}
  \label{eq:eq4-33}
CDF_x = \frac{card(NRMSE \le x)}{n} \enspace ,
\end{equation}
where $card(.)$ is the cardinal of a set. $n$ is the total number of images.

The $CDF_{NRMSE}$ represents the percentage of images with error less or equal
than the specified NRMSE value. For example a $CDF_{0.1}=0.4$ over a test set
means that $40\%$ of the test set images have an error less or equal than
$0.1$. A CDF curve can be plotted according to these $CDF_{NRMSE}$ values by
varying the value of $NRMSE$.

These are the usual evaluation criteria used in facial landmark detection
problem. To have more numerical precision in the comparison in our experiments,
we calculate the Area Under the CDF Curve (AUC), using only the NRMSE range
[0,0.5] with a step of $10^{-3}$.

\subsection{General training setup}

To implement our framework, we use
\begin{inparaitem}[-]
\item a DNN with four layers $K=4$ for the main task;
\item an input AE with one encoding layer $K_{in}=1$ and one decoding layer;
\item an output AE with one encoding layer and one decoding layer $K_{out}=1$.
\end{inparaitem}
Referring to Fig.\ref{fig:fig4-0.1}, the size of the input representation $\xvec$ and estimation $\xest$ is $2500=50 \times 50$;
the size of the output representation $\yvec$ and estimation $\yest$ is $136= 68 \times 2$, given the 68 landmarks in a 2D plane;
the dimension of intermediate spaces $\xproj$ and $\yproj$ have been set to 1025 and 64 respectively;
finally, the hidden layer in the $\mathcal{L}$ link between  $\xproj$ and $\yproj$ is composed of 512 units.
The size of each layer has been set using a validation procedure on the LFPW validation set.

Sigmoid activation functions are used everywhere in the main NN and in the two AEs, except for the last layer of the main NN and the tied last layer of output AE which use a hyperbolic tangent activation function to suite the range $[-1, \;1]$ for the output $y_i \in \yvec$.

We use the same architecture through all the experiments for the different
training configurations. To distinguish between the multiple configurations we
set the following notations:
\begin{enumerate}
\item \textbf{MLP}, a DNN for the main task with no concomitant training;
\item \textbf{MLP + in}, a DNN with input AE parallel training;
\item \textbf{MLP + out}, a DNN with output AE parallel training;
\item \textbf{MLP + in + out}, a DNN with both input and output reconstruction secondary tasks.
\end{enumerate}
We recall that the auto-encoders are used only during the training phase. In the test phase, they are dropped. Therefore, the final test networks have the same architecture in all the different configurations.

Beside these configurations, we consider the mean shape (the average of the $\yvec$ in the training data) as a simple predictive model. For each test image, we predict the same estimated mean shape over the train set.

To clarify the benefit of our approach, all the configurations must start from the same initial weights to make sure that the obtained improvement is due to the training algorithm, not to the random initialization.

For the input reconstruction tasks, we use a denoising auto-encoder with a
corruption level of $20\%$ for the first hidden layer. For the output
reconstruction task, we use a simple auto-encoder. To avoid overfitting, the auto-encoders are trained using $L_2$ regularization with a weight decay of $10^{-2}$.

In all the configurations, the update of the parameters of each task (supervised and unsupervised) is performed using Stochastic Gradient Descent with momentum
\cite{sutskever1} with a constant momentum coefficient of $0.9$. We use
mini-batch size of 10. The training is performed for 1000 epochs with a learning rate of $10^{-3}$.

\begin{figure}[!htbp]
 \centering
 \includegraphics[scale=0.5]{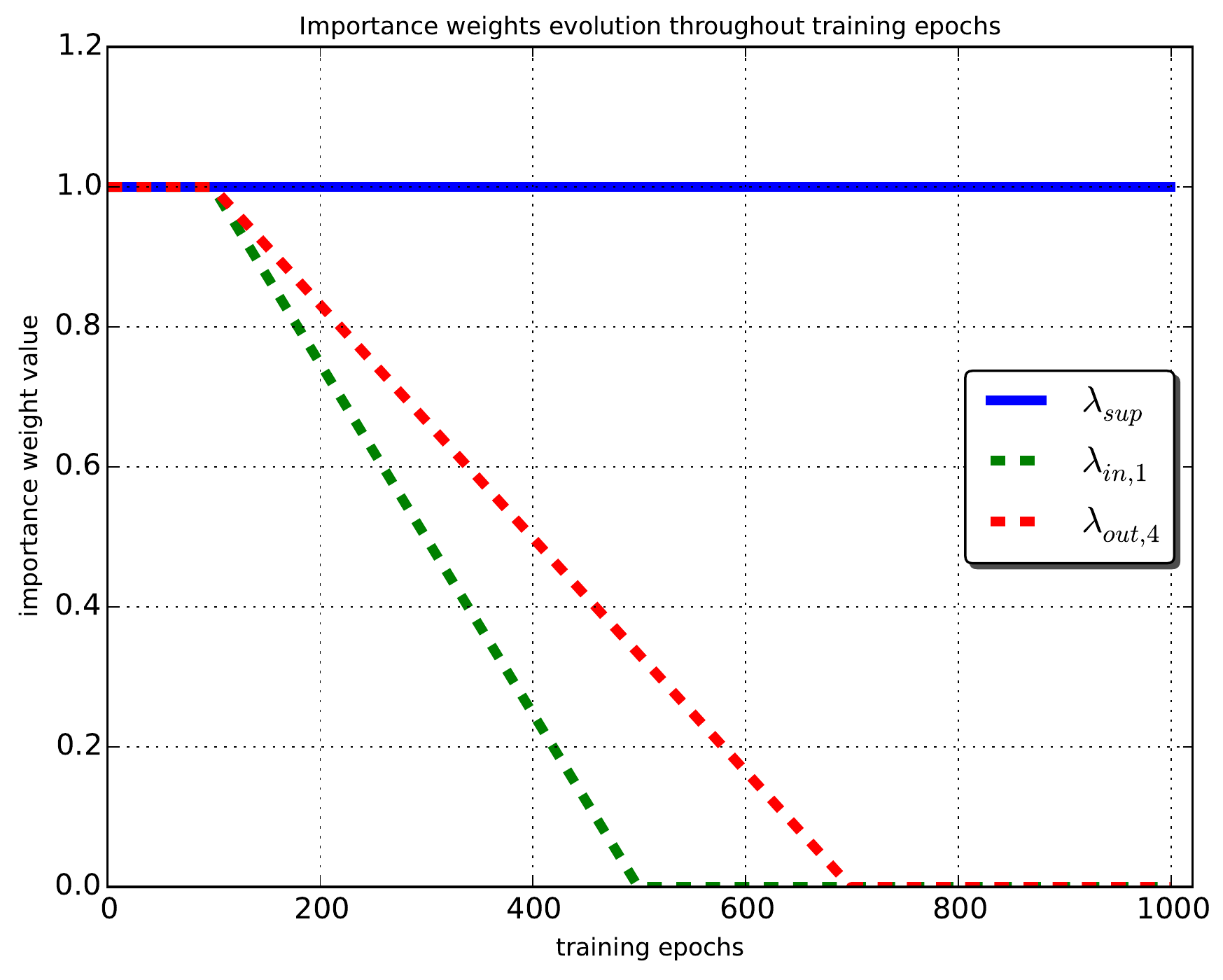}
 \caption{Linear \fromont{adaptation} of the importance weights during training.}
 \label{fig:fig4-000}
\end{figure}

In these experiments, we propose to use a simple linear \fromont{adaptation} scheme for the importance weights $\lambda_{sup}$ (supervised task), $\lambda_{in}$ (input task) and  $\lambda_{out}$ (output task). We retain the \fromont{adaptation scheme} proposed in \cite{bel16},  and presented in Fig.\ref{fig:fig4-000}.

The hyper-parameters (learning rate, batch size, momentum coefficient, weight decay, the importance weights) have been optimized on the LFPW validation set. We apply the same optimized hyper-parameters for HELEN dataset.

Using these configurations, we perform two types of experiments: with and
without unlabeled data. We present in the next sections the obtained results.

\subsubsection{Experiments with fully labeled data}
\label{subsub:naug4}
In this setup, we use the provided labeled data from each set in a classical way. For LFPW set, we use the 676 available samples for training and 135 samples for validation. For HELEN set, we use 1800 samples for training and 200 samples for validation. 

In order to evaluate the different configurations, we first calculate the Mean Squared Error (MSE) of the best models found using the validation during the training. Column 1 (no unlabeled data) of Tab.\ref{tab:tab4-1}, \ref{tab:tab4-2} shows the MSE over the train and valid sets of LFPW and HELEN datasets, respectively. Compared to an MLP alone, adding the input training of the first hidden layer slightly reduces the train and validation error in both datasets. Training the output layer also reduces the train and validation error, with a more important factor. Combining the input train of the first hidden layer and output train of the last layer gives the best performance. We plot the tracked MSE over the train and valid sets of HELEN dataset in Fig.\ref{fig:fig4-4:fig4-2}, \ref{fig:fig4-4:fig4-3}. One can see that the input training reduces slightly the validation MSE. The output training has a major impact over the training speed and the generalization of the model which suggests that output training is useful in the case of structured output problems. Combining the input and the output training improves even more the generalization. Similar behavior was found on LFPW dataset.

At a second time, we evaluate each configuration over the test set of each datasets using the $CDF_{0.1}$ metric. The results are depicted in Tab.\ref{tab:tab4-3}, \ref{tab:tab4-4} in the first column for LFPW and HELEN datasets, respectively. Similarly to the results  previously found over the train and validation set, one can see that the joint training (supervised, input, output) outperforms all the other configurations in terms of $CDF_{0.1}$ and AUC. The CDF curves in Fig.\ref{fig:fig4-4} also confirms this result. Compared to the global DNN in \citep{zhang14} over LFPW test set, our joint trained MLP performs better (\citep{zhang14}: $CDF_{0.1}=65\%$, ours: $CDF_{0.1}=69.64\%$), despite the fact that their model was trained using larger supervised dataset (combination of multiple supervised datasets beside LFPW).

\begin{figure}[!htbp]
 \centering
 \includegraphics[width=30mm,height=30mm]{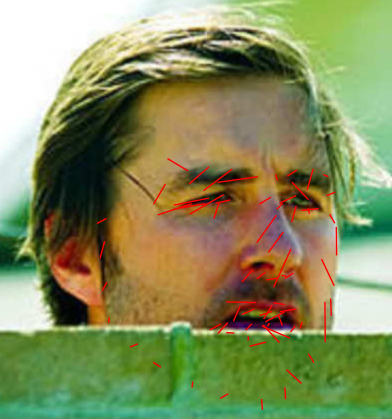}
 \includegraphics[width=30mm,height=30mm]{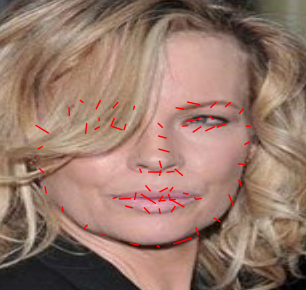}
 \includegraphics[width=30mm,height=30mm]{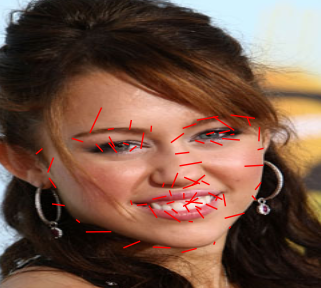} \\
 \includegraphics[width=30mm,height=30mm]{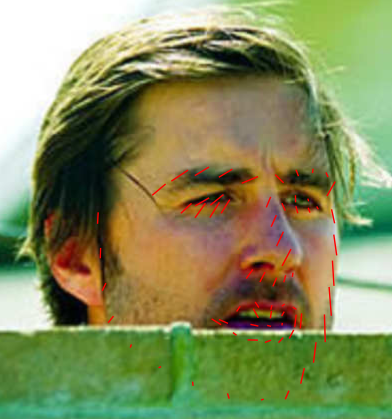}
 \includegraphics[width=30mm,height=30mm]{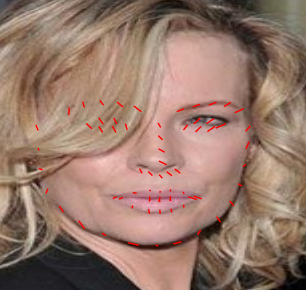}
 \includegraphics[width=30mm,height=30mm]{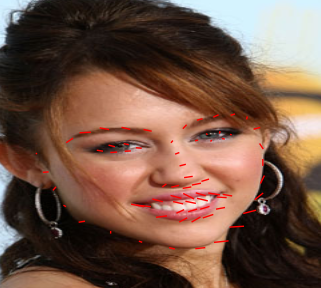}
 \caption[\bel{Examples of prediction on LFPW test set.}]{Examples of prediction on LFPW test set. For visualizing errors, red segments have been drawn between ground truth and predicted landmark. Top row: MLP. Bottom row: MLP+in+out. (no unlabeled data)}
 \label{fig:4-figlfpw}
 \end{figure}

\begin{figure}[!htbp]
 \centering
 \includegraphics[width=30mm,height=30mm]{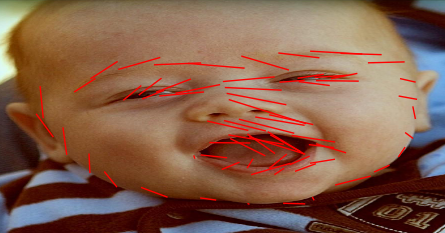}
 \includegraphics[width=30mm,height=30mm]{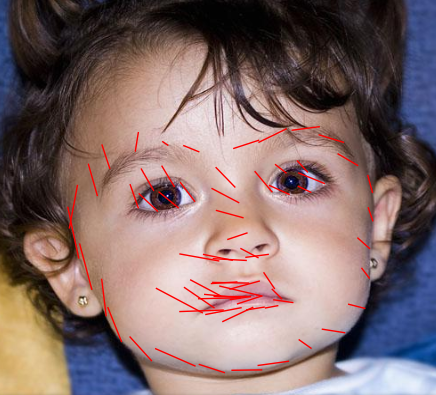}
 \includegraphics[width=30mm,height=30mm]{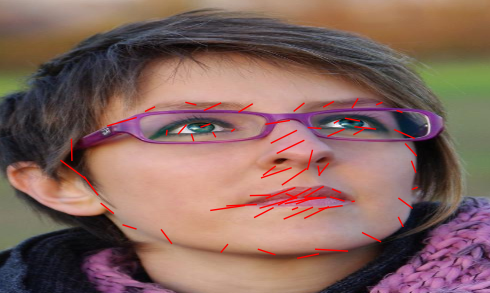} \\
 \includegraphics[width=30mm,height=30mm]{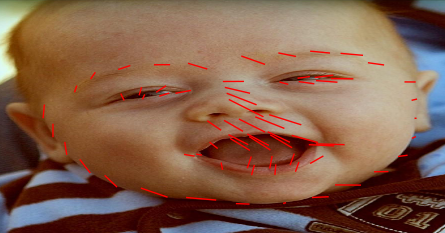}
 \includegraphics[width=30mm,height=30mm]{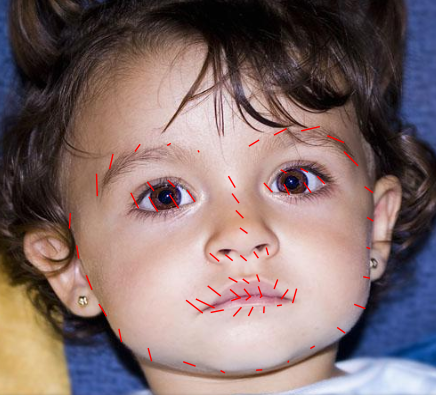}
 \includegraphics[width=30mm,height=30mm]{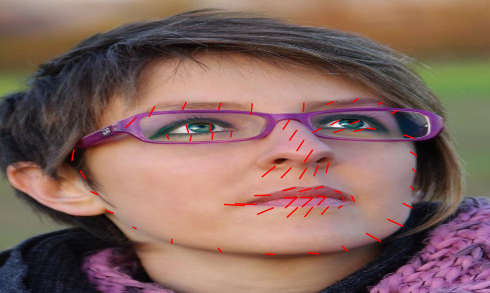}
 \caption[\bel{Examples of prediction on HELEN test set.}]{Examples of prediction on HELEN test set. Top row: MLP. Bottom row: MLP+in+out. (no unlabeled data)}
 \label{fig:4-fighelen}
 \end{figure}

\begin{figure}[!htbp]
 \centering
 \begin{subfigure}{1.\textwidth}
 \centering
 \includegraphics[scale=0.65]{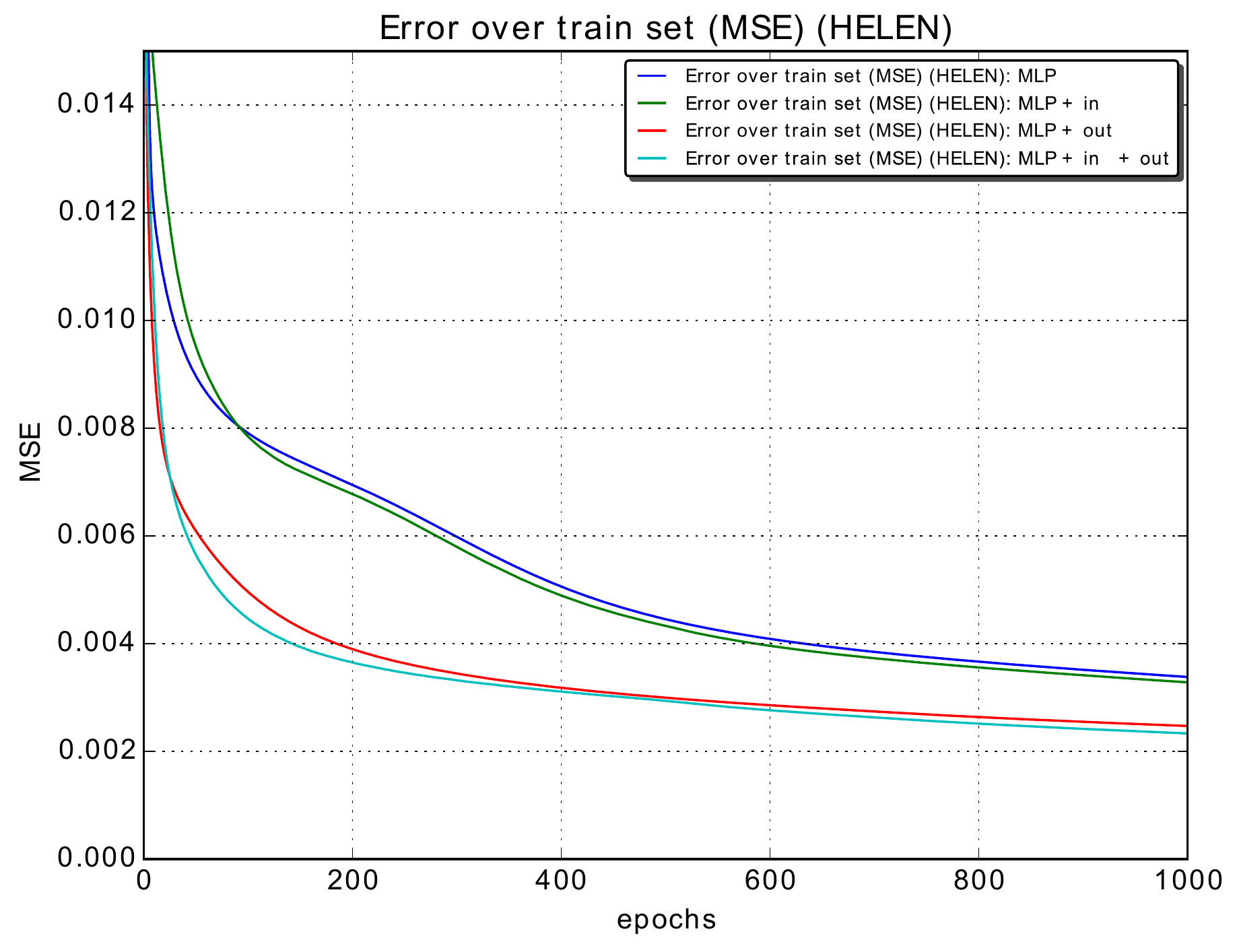}
 \caption{}
 \label{fig:fig4-4:fig4-2}
 \end{subfigure}%
 \\
 ~
 \begin{subfigure}{1.\textwidth}
 \centering
 \includegraphics[scale=0.65]{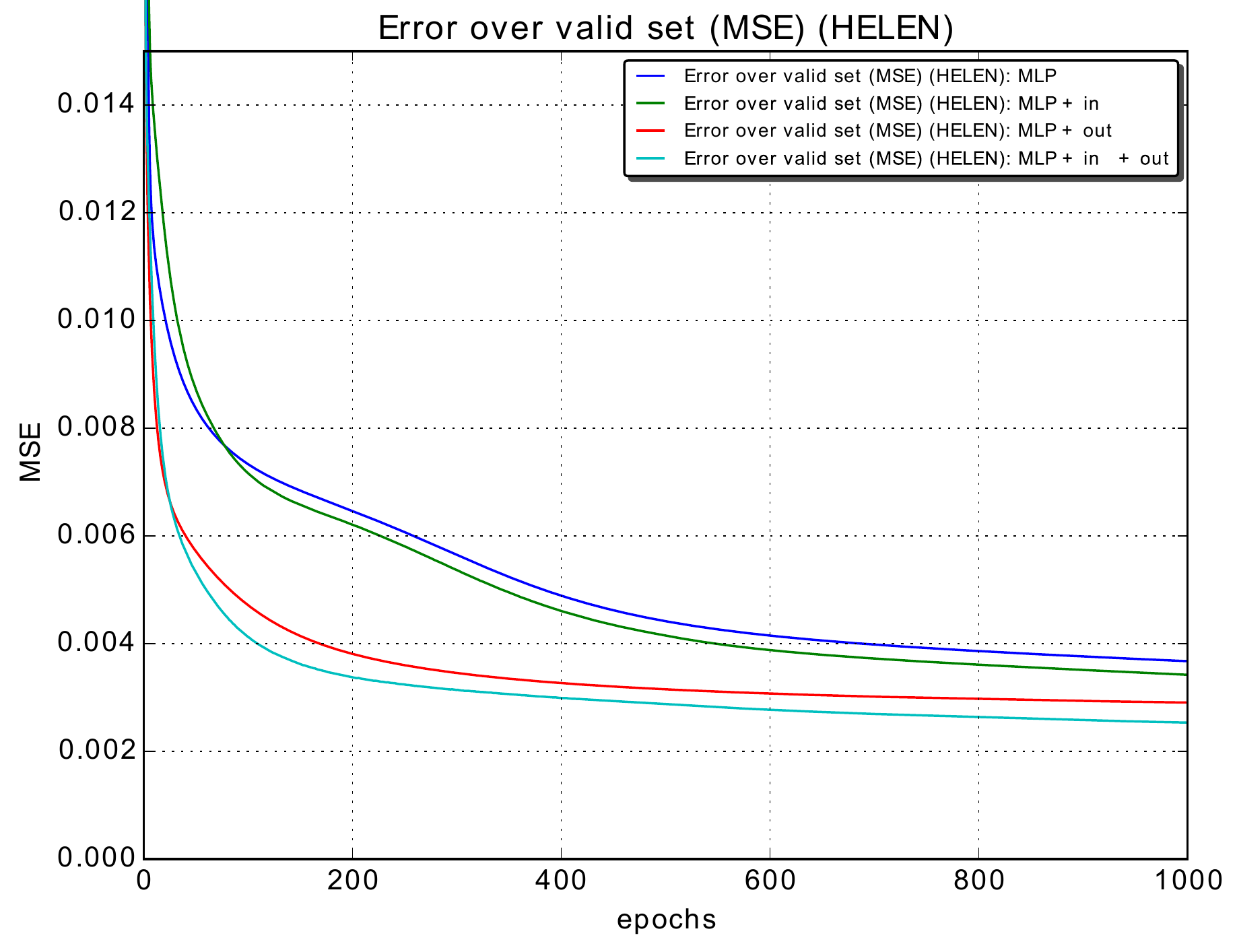}
 \caption{}
 \label{fig:fig4-4:fig4-3}
 \end{subfigure}
 \caption[\bel{MSE during training epochs over HELEN train.}]{MSE during training epochs over HELEN train (a) and valid (b)
   sets using different training setups for the MLP.}
 \label{fig:fig4-5}
\end{figure}

\begin{figure}[!htbp]
 \centering
 \begin{subfigure}{1.\textwidth}
 \centering
 \includegraphics[scale=0.63]{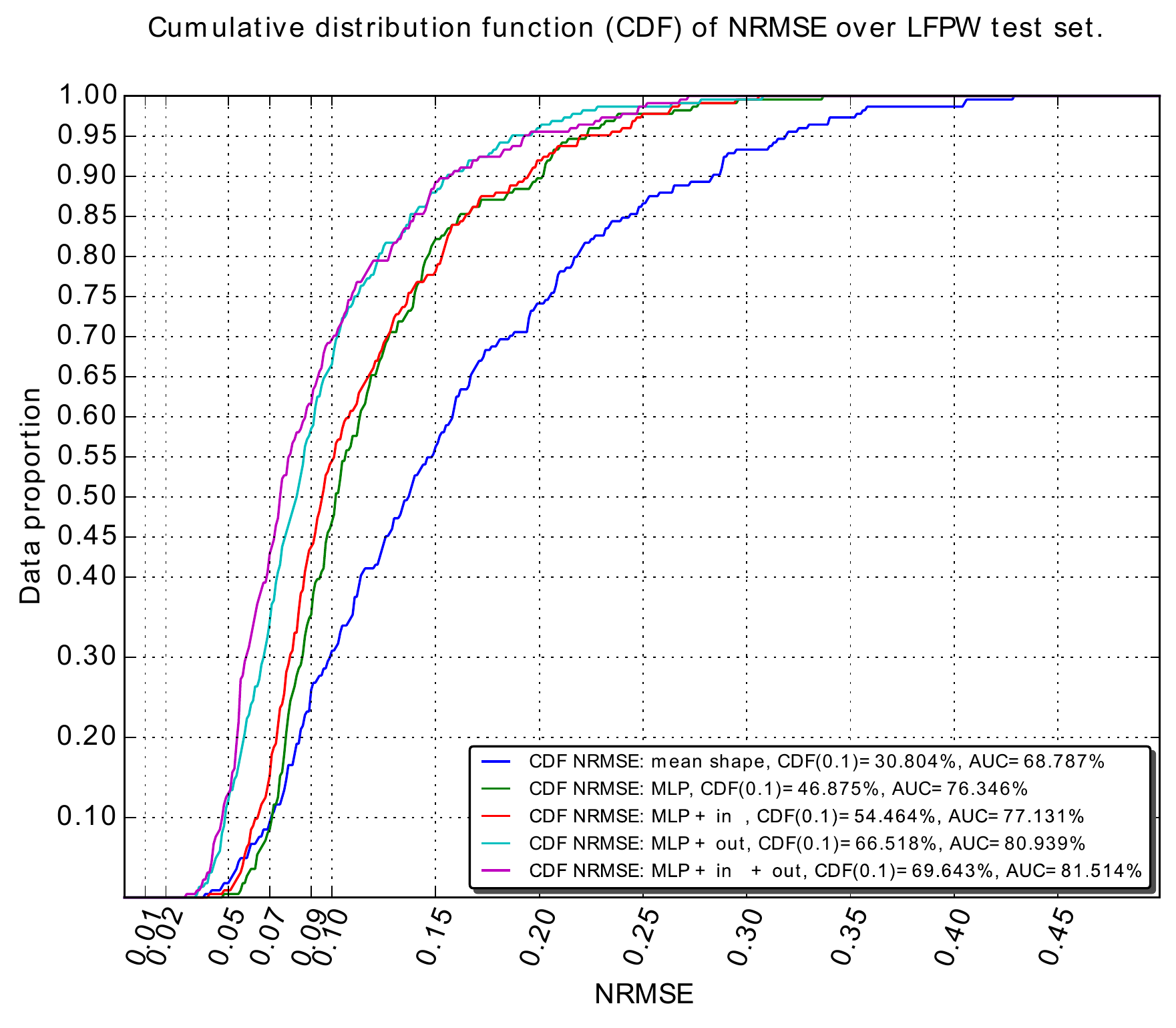}
 \caption{}
 \label{fig:fig4-4:fig4-0}
 \end{subfigure}%
 \\
 ~
 \begin{subfigure}{1.\textwidth}
 \centering
 \includegraphics[scale=0.63]{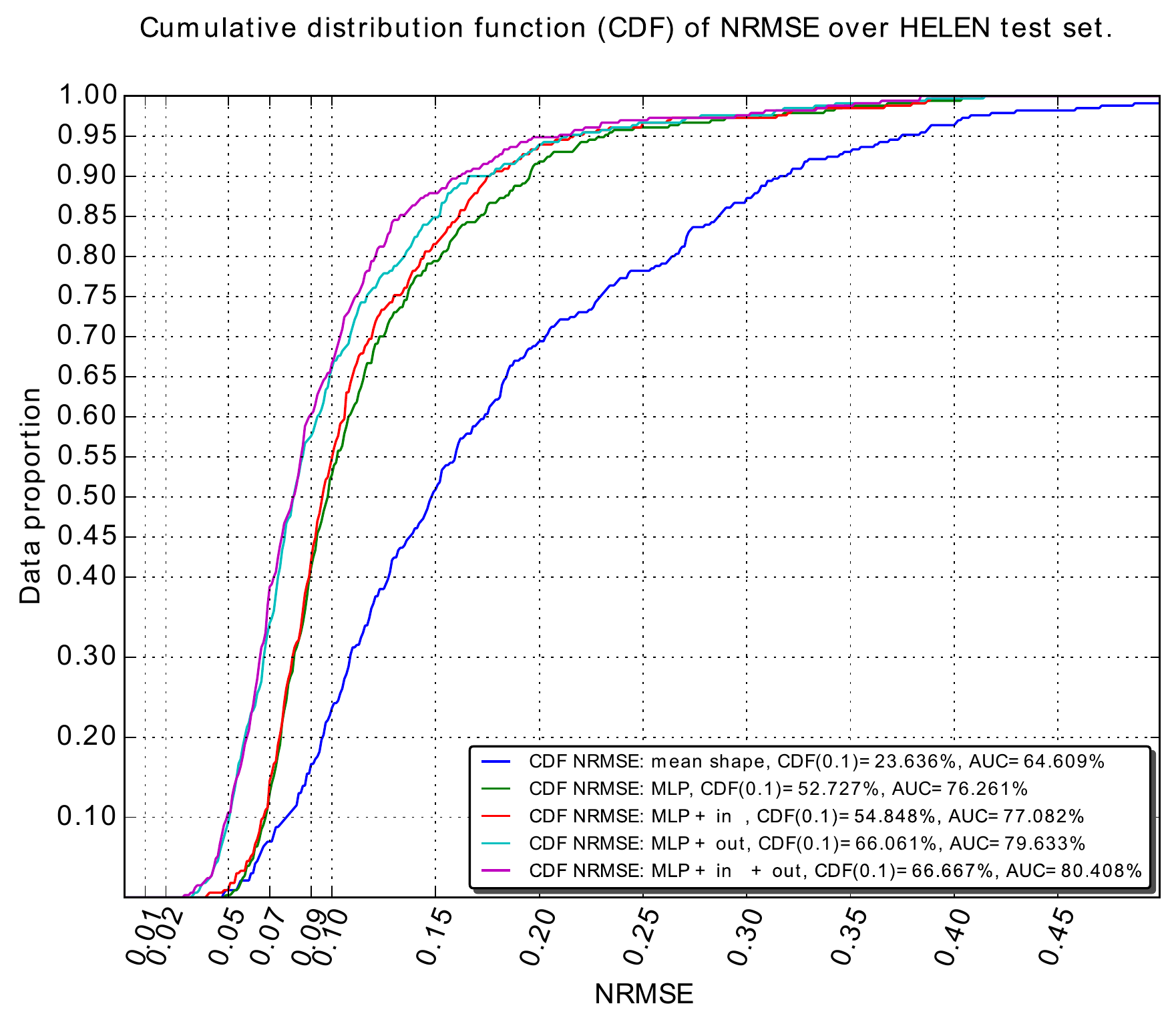}
 \caption{}
 \label{fig:fig4-4:fig4-1}
 \end{subfigure}
 \caption[\bel{CDF curves.}]{CDF curves of different configurations on: (a) LFPW, (b) HELEN.}
 \label{fig:fig4-4}
\end{figure}

An illustrative result of our method is presented in Fig.\ref{fig:4-figlfpw}, \ref{fig:4-fighelen} for LFPW and HELEN using an MLP and MLP with input and output training.

\begin{table}[!htbp]
 \centering
 \caption[\bel{MSE over LFPW.}]{MSE over LFPW: train and valid sets, at the end of training with and
   without unlabeled data.}
 \label{tab:tab4-1}
 \resizebox{0.9\textwidth}{!}{
\sisetup{detect-weight=true,detect-inline-weight=math}
\begin{tabular}{l|c|c||c|c|}
  \cline{2-5}
  &\multicolumn{2}{|c||}{\textbf{No unlabeled data}} 
&\multicolumn{2}{|c|}{\textbf{With unlabeled data}}\\
  \cline{2-5}
  &\multicolumn{1}{|c|}{MSE train}&\multicolumn{1}{|c||}{MSE valid}&\multicolumn{1}{|c|}{MSE train}&\multicolumn{1}{|c|}{MSE valid}\\
              \hline
  \multicolumn{1}{ |l | } {\textbf{Mean 
                  shape}}&
              \num{7.74d-3} &\num{8.07d-3}&\num{7.78d-3}&\num{8.14d-3}\\
  \hline
  \hline
  \multicolumn{1}{ |l | } {\textbf{MLP}}&
              \num{3.96d-3}&\num{4.28d-3}&-&-\\
  \hline
              \hline
  \multicolumn{1}{ |l | } {\textbf{MLP + in}}&
              \num{3.64d-3}&\num{3.80d-3}&\num{1.44d-3}&\num{2.62d-3}\\
  \hline
  \multicolumn{1}{ |l | } {\textbf{MLP + out}}&
              \num{2.31d-3}&\num{2.99d-3}&\num{1.51d-3}&\num{2.79d-3}\\
  \hline
  \multicolumn{1}{ |l | } {\textbf{MLP + in + out}}&
              \textbf{\num{2.12d-3}}&\textbf{\num{2.56d-3}}&\textbf{\num{1.10d-3}}&\textbf{\num{2.23d-3}}\\
  \hline
\end{tabular}
}
\end{table}

 \begin{table}[!htbp]
 \centering
 \caption[\bel{MSE over HELEN.}]{MSE over HELEN: train and valid sets, at the end of training with and
   without data augmentation.}
 \label{tab:tab4-2}
 \resizebox{0.9\textwidth}{!}{
  \sisetup{detect-weight=true,detect-inline-weight=math}
\begin{tabular}{l|c|c||c|c|}
  \cline{2-5}
  &\multicolumn{2}{|c||}{\textbf{Fully labeled data only }} 
&\multicolumn{2}{|c|}{\textbf{Adding unlabeled or label-only data}}\\
  \cline{2-5}
  &\multicolumn{1}{|c|}{MSE train}&\multicolumn{1}{|c||}{MSE valid}&\multicolumn{1}{|c|}{MSE train}&\multicolumn{1}{|c|}{MSE valid}\\
              \hline
  \multicolumn{1}{ |l | } {\textbf{Mean 
                  shape}}&
              \num{7.59d-3}&\num{6.95d-3}&\num{7.60d-3}&\num{.95d-3}\\
  \hline
  \hline
  \multicolumn{1}{ |l | } {\textbf{MLP}}&
              \num{3.39d-3}&\num{3.67d-3}&-&-\\
  \hline
              \hline
  \multicolumn{1}{ |l | } {\textbf{MLP + in}}&
              \num{3.28d-3}&\num{3.42d-3}&\num{2.31d-3}&\num{2.81d-3}\\
  \hline
  \multicolumn{1}{ |l | } {\textbf{MLP + out}}&
              \num{2.48d-3}&\num{2.90d-3}&\num{2.00d-3}&\num{2.74d-3}\\
  \hline
  \multicolumn{1}{ |l | } {\textbf{MLP + in + out}}&
             \textbf{\num{2.34d-3}}&\textbf{\num{2.53d-3}}&\textbf{\num{1.92d-3}}&\textbf{\num{2.40d-3}}\\
    \hline
\end{tabular}
      }
\end{table}


\begin{table}[!htbp]
 \centering
 \caption[\bel{\textbf{AUC} and $\mathbf{CDF_{0.1}}$ performance over LFPW test set.}]{\textbf{AUC} and $\mathbf{CDF_{0.1}}$ performance over LFPW test dataset with and without unlabeled data.}
 \label{tab:tab4-3}
 \resizebox{.8\textwidth}{!}{
\begin{tabular}{c|c|c||c|c|}
  \cline{2-5}
  &\multicolumn{2}{|c||}{\textbf{Fully labeled data only}} 
&\multicolumn{2}{|c|}{\textbf{Adding unlabeled or label-only data}}\\
  \cline{2-5}
  &\textbf{AUC}&$\mathbf{CDF_{0.1}}$&\textbf{AUC}&$\mathbf{CDF_{0.1}}$\\
              \hline
  \multicolumn{1}{ |l | } {\textbf{Mean 
                  shape}}&
              68.78\%&30.80\%&77.81\%&22.33\%\\
  \hline
  \hline
  \multicolumn{1}{ |l | } {\textbf{MLP}}&
              76.34\%&46.87\%&-&-\\
  \hline
              \hline
  \multicolumn{1}{ |l | } {\textbf{MLP + in}}&
              77.13\%&54.46\%&80.78\%&67.85\%\\
  \hline
  \multicolumn{1}{ |l | } {\textbf{MLP + out}}&
              80.93\%&66.51\%&81.77\%&67.85\%\\
  \hline
  \multicolumn{1}{ |l | } {\textbf{MLP + in + out}}&
              \textbf{81.51\%}&\textbf{69.64\%}&\textbf{82.48\%}&\textbf{71.87\%}\\
  \hline
\end{tabular}
   }
\end{table}

\begin{table}[!htbp]
 \centering
 \caption[\bel{\textbf{AUC} and $\mathbf{CDF_{0.1}}$ performance over HELEN test set.}]{\textbf{AUC} and $\mathbf{CDF_{0.1}}$ performance over HELEN test dataset with and without unlabeled data.}
 \label{tab:tab4-4}
 \resizebox{.8\textwidth}{!}{
\begin{tabular}{c|c|c||c|c|}
  \cline{2-5}
  &\multicolumn{2}{|c||}{\textbf{Fully labeled data only}} 
&\multicolumn{2}{|c|}{\textbf{Adding unlabeled or label-only data}}\\
  \cline{2-5}
  &\textbf{AUC}&$\mathbf{CDF_{0.1}}$&\textbf{AUC}&$\mathbf{CDF_{0.1}}$\\
              \hline
  \multicolumn{1}{ |l | } {\textbf{Mean 
                  shape}}&
              64.60\%&23.63\%&64.76\%&23.23\%\\
  \hline
  \hline
  \multicolumn{1}{ |l | } {\textbf{MLP}}&
              76.26\%&52.72\%&-&-\\
  \hline
              \hline
  \multicolumn{1}{ |l | } {\textbf{MLP + in}}&
              77.08\%&54.84\%&79.25\%&63.33\%\\
              \hline
  \multicolumn{1}{ |l | } {\textbf{MLP + out}}&
              79.63\%&66.60\%&80.48\%&65.15\%\\
  \hline
  \multicolumn{1}{ |l | } {\textbf{MLP + in + out}}&
              \textbf{80.40\%}&\textbf{66.66\%}&\textbf{81.27\%}&\textbf{71.51\%}\\
    \hline
\end{tabular}
    }
\end{table}

\subsubsection{Data augmentation using unlabeled data or label-only data}
In this section, we experiment our approach when adding unlabeled data (input and output). Unlabeled data (i.e. image faces without the landmarks annotation) are abundant and can be found easily for example from other datasets or from the Internet which makes it practical and realistic. In our case, we use image faces from another dataset.

In the other hand, label-only data (i.e. the landmarks annotation without image faces) are more difficult to obtain because we usually have the annotation based on the image faces. One way to obtain accurate and realistic facial landmarks without image faces is to use a 3D face model as a generator. We use an easier way to obtain facial landmarks annotation by taking them from another dataset.

In this experiment, in order to add unlabeled data for LFPW dataset, we take all the image faces of HELEN dataset (train, valid and test) and vice versa for HELEN dataset by taking all LFPW image faces as unlabeled data. The same experiment is performed for the label-only data using the facial landmarks annotation. We summarize the size of each train set in Tab.\ref{tab:tab4-5}..

\begin{table}[!htb]
\centering
\caption{Size of augmented LFPW and HELEN train sets.}
\label{tab:tab4-5}
\resizebox{\textwidth}{!}{
\begin{tabular}{|c|c|c|c|}
  \hline
  Train set / size of& Supervised data& Unsupervised input $\xvec$ & Unsupervised
  output $\yvec$\\
  \hline
  LFPW&676&2330&2330\\
  \hline
  HELEN&1800&1035&1035\\
  \hline
\end{tabular}
}
\end{table}
We use the same validation sets as in Sec.\ref{subsub:naug4} in order to have a fair comparison. The MSE are presented in the second column of Tab.\ref{tab:tab4-1}, \ref{tab:tab4-2} over LFPW and HELEN datasets. One can see that adding unlabeled data decreases the MSE over the train and validation sets. Similarly, we found that the input training along with the output training gives the best results. Identically,
these results are translated in terms of $CDF_{0.1}$ and AUC over the test sets (Tab.\ref{tab:tab4-3}, \ref{tab:tab4-4}). All these results suggest that adding unlabeled input and output data can improve the generalization of our framework and the training speed.
\FloatBarrier

\section[Conclusion]{Conclusion}
\label{sec:conclusion4}
In this paper, we tackled structured output prediction problems as a representation learning problem. We have proposed a generic multi-task training framework as a regularization scheme for structured output prediction models. It has been instantiated through a deep neural network model which learns the input and output distributions using auto-encoders while learning the supervised task $\X \to \Y$. Moreover, we explored the possibility of using the output labels $\y$ without their corresponding input data $\x$ which showed more improvement in the generalization. Using a parallel scheme allows an interaction between the main supervised task and the unsupervised tasks which helped preventing the overfitting of the main task.

We evaluated our training method on a facial landmark detection task over two public datasets. The obtained results showed that our proposed regularization scheme improves the generalization of neural networks model and speeds up their training. We believe that our approach provides an alternative for training deep architectures for structured output prediction where it allows the use of unlabeled input and label of the output data. 

As a future work, we plan to \fromont{adapt} automatically the importance weights of the tasks. For that and in order to better guide their \fromont{adaptation}, we can consider the use of different indicators based on the training and the validation errors instead of the learning epochs only. Furthermore, one may consider other kind of models instead of simple auto-encoders in order to learn the output distribution. More specifically, generative models such as variational and adversarial auto-encoders \cite{MakhzaniSJG15CORR} could be explored.

\section*{Acknowledgement}
This work has been partly supported by the grant ANR-11-JS02-010 LeMon and the grant ANR-16-CE23-0006 \quotes{Deep in France}.

\chapter[Neural Networks Regularization Through Class-wise Invariant\texorpdfstring{\\}{ } Representation Learning]{Neural Networks Regularization Through Class-wise Invariant Representation Learning}
\label{chap:chapter5}
\ifpdf
    \graphicspath{{Chapter5/Figs/Raster/}{Chapter5/Figs/PDF/}{Chapter5/Figs/}}
\else
    \graphicspath{{Chapter5/Figs/Vector/}{Chapter5/Figs/}}
\fi

\makeatletter
\def\input@path{{Chapter5/}}
\makeatother

\section{Prologue}

\noindent\emph{\underline{Article Details}:}
\begin{itemize}
    \item \textbf{Neural Networks Regularization Through Class-wise Invariant Representation Learning}. Soufiane Belharbi, Clément Chatelain, Romain Hérault, and Sébastien Adam. \textbf{Under review}, \emph{2017}.
\end{itemize}

\bigskip

\noindent\emph{\underline{Context}:}
\\
Neural network models, particularly deep models, have seen a large success in different applications. However, training such models requires a large number of training data which are not available in many real world applications. Despite this issue, one would like to be able to use deep neural networks using only few samples which is the challenge that we tackle in this contribution.

Not far away from the time of writing this work, there was an urge toward learning unsupervised representations. Many of these works suggested using different approach of regularizations to learn better representations. This motivated us to explore a different and intuitive approach to regularize supervised learning of neural networks, especially in the case where only few training samples are available. In this work, we present a new approach to regularize neural networks when trained over few data for classification task. The key idea here is the use of a prior belief about the internal representation within a neural network. The idea simply states that samples within the same class should have the same representation. We formulate this idea as a cost function, under the form of a dissimilarity measure, which we integrate with the training cost to be minimized. We note that our regularization requires supervised samples.

Empirical results over different classification tasks showed improvements of the generalization error especially in the case where only few training samples are available. Moreover, we showed an intriguing behavior in learning intermediate representations within neural networks which is: the hidden layers do not tend, on its own, to learn invariant features. This confirms that the propagated classification error does not necessarily train the hidden layers to learn meaningful and understandable features. This brings us one step further to make neural network more understandable and shed more light on the comprehension of the information stream within the layers to gain more control over it.

We present this work \cite{belharbiIEEETNNLS2017} as a chapter of this thesis under the form of one contribution. This chapter contains the original paper as it was submitted to Neural Networks journal with slight adaptation.
\bigskip

\noindent\emph{\underline{Contributions}:}
\\
The contribution of this work is to provide a new regularization approach for training supervised neural networks for \fromont{a} classification task by guiding learning the internal representation of the network using a prior belief. This new approach showed to be more useful when only few training samples are available.

\newpage


\section[Introduction]{Introduction}
\label{sec:introduction5}
For a long time, it has been understood in the field of deep learning that building a model by stacking multiple levels of non-linearity is an efficient way to achieve good performance on complicated artificial intelligence tasks  such as vision \cite{ krizhevsky12, SimonyanZ14aCORR, szegedyLJSRAEVR14, heZRS16} or natural language processing \cite{CollobertWeston2008, WestonRC2008, Kim14, Graves13}. The rationale behind this statement is the hierarchical learned representations throughout the depth of the network which circumvent the need of extracting handcrafted features.

For many years, the non-convex optimization problem of learning a neural network has prevented going beyond one or two hidden layers. In the last decade, deep learning has seen a breakthrough with efficient training strategies of deeper architectures \cite{hinton06NC, RanzatoPCL2006, bengio06NIPS}, and a race toward deeper models has began \cite{krizhevsky12, SimonyanZ14aCORR, szegedyLJSRAEVR14, heZRS16}. This urge to deeper architectures was due to (i) large progress in optimization, (ii) the powerful computation resources brought by GPUs\footnote{Graphical Processing Units.} and (iii) the availability of huge datasets such as ImageNet \cite{imagenet09} for computer vision problems. However, in real applications, few training samples are usually available which makes the training of deep architectures difficult. Therefore, it becomes necessary to provide new learning schemes for deep networks to perform better using few training samples. 

A common strategy to circumvent the lack of annotated data is to exploit extra informations related to the data, the model or the application domain, in order to guide the learning process. This is typically carried out through regularization which can rely for instance on data augmentation, $L_2$ regularization \cite{tikhonov77}, dropout\cite{srivastava14a}, unsupervised training \cite{hinton06NC, RanzatoPCL2006, bengio06NIPS, rifaiVMGB11, salah2011, bel16}, shared parameters \cite{leCun1989, riesenhuber99, fukushimaM82},  etc. 

Our research direction in this work is to provide a new regularization framework to guide the training process in a supervised classification context. The framework relies on the exploitation of prior knowledge which has already been used in the literature to train and improve models performance when few training samples are available  \cite{mitchell1997, scholkopf2001, heckerman1995, niyogi98, krupka2007, yu2007, wu2004, YuSJ10Neurocomp}.

Indeed, prior knowledge can offer the advantage of more consistency, better generalization and fast convergence using less training data by guiding the learning process \cite{mitchell1997}. By using prior knowledge about the target function, the learner has a better chance to generalize from \fromont{few} data \cite{mitchell1997, abuMostafa90, abuMostafab92,abuMostafa93}. For instance, in object localization such as part of the face, knowing that the eyes are located above the nose and the mouth can be helpful. One can exploit this prior structure about the data representation: to constrain the model architecture, to guide the learning process, or to post-process the model's decision.

In classification task, although it is difficult to define what makes a representation  good, two properties are inherent to the task: Discrimination , i.e., representations must allow to separate samples of distinct classes. Invariance, i.e., representations must allow to obtain robust decision despite some variations of input samples. Formally, given two samples $\xvec^{(1)}$ and $\xvec^{(2)}$, a representation function ${\Gamma(\cdot)}$ and a decision function $\Psi(\cdot)$;  when ${\xvec^{(1)} \approx \xvec^{(2)}}$ , we seek invariant representations that provide ${\Gamma(\xvec^{(1)}) \approx \Gamma(\xvec^{(2)})}$, leading to smooth decision ${\Psi(\Gamma(\xvec^{(1)})) \approx\Psi(\Gamma(\xvec^{(2)}))}$. In this work, we are interested in the invariance aspect of the representations. This definition can be extended to more elaborated transformations such as rotation, scaling, translation, etc. However, in real life there are many other transformations which are difficult to formalize or even enumerate. Therefore, we extend in this work the definition of the invariant representations to the class membership, where samples within the same class should have the same representation. At a representation level, this should generate homogeneous and tighter clusters per class.

\begin{figure*}
	\centering
\resizebox{\textwidth}{!}{\input{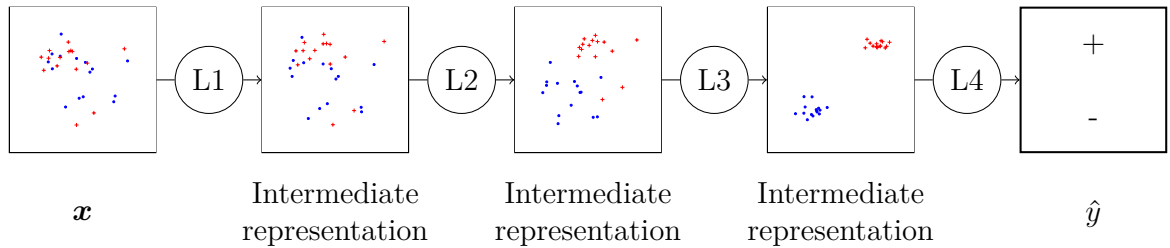}}
\caption[\bel{Input/Hidden representations in an MLP.}]{Input/Hidden representations of samples from an artificial dataset along  4 layers of a MLP. Each representation is projected into a 2D space. }
\label{fig:datarep5}
\end{figure*}

In the training of neural networks, while the output layer is guided by the provided target, the hidden layers are left to the effect of the propagated error from the output layer without a specific target. Nevertheless, once the network \fromont{is} trained, examples may form (many) modes on hidden representations, i.e. outputs of hidden layers, conditionally to their classes. Most notably, on the penultimate representation before the decision stage, examples should agglomerate in distinct clusters according to their label as seen on Figure \ref{fig:datarep5}.
From the aforementioned prior perspective about the hidden representations, we aim in this work to provide a learning scheme that promotes the hidden layers to build representations which are class-invariant and thus agglomerate in restricted number of modes. By doing so, we constrain the network to build invariant intermediate representations per class with respect to the variations in the input samples without explicitly specifying these variations nor the transformations that caused them. 

We express this class-invariance prior as an explicit criterion combined with the classification training criterion. It is formulated as a dissimilarity between the representations of each pair of samples within the same class. The average dissimilarity over all the pairs of all the classes is considered to be minimized. To the best of our knowledge, none has used this class membership to build invariant representations. Our motivation in using this prior knowledge, as a form of regularization, is to be able to train deep neural networks and obtain better generalization error using less training data. We have conducted different experiments over MNIST benchmarck using two models (multilayer perceptrons and convolutional networks) for different classification tasks. We have obtained results that show important improvements of the model's generalization error particularly when trained with few samples.

The rest of the paper is organized as follows: Sec.\ref{sec:related5} presents related works for invariance learning  in neural networks. We present our learning framework in Sec.\ref{sec:method5} followed by a discussion of the obtained results in Sec.\ref{sec:exps5}.

\section[Related Work]{Related Work}
\label{sec:related5}
Learning general invariance, particularly in deep architectures, is an attractive subject where different approaches have been proposed. The rational behind this framework is to ensure the invariance of the learned model toward the variations of the input data. In this section, we describe three kinds of approaches of learning invariance within neural networks. Some of these methods were not necessarily designed to learn invariance however we present them from the invariance perspective. For this description, $f$ is the target function to be learned.

\begin{description}

\item [\emph{Invariance through data transformations}:]
\hfill \\
It is well known that generalization performance can be improved by using larger quantity of training samples. Enlarging the number of samples can be achieved by generating new samples through the application of small random transformations such as rotation, scaling, random noise, etc \cite{Baird90, Ciresan2010, Simard2003} to the original examples. Incorporating such transformed data within the learning process has shown to be helpful in generalization \cite{niyogi98}. \cite{abuMostafa90} proposes the use of prior information about the behavior of $f$ over perturbed examples using different transformations where $f$ is constrained to be invariant over all the samples generated using these transformations. While data transformations successfully incorporate certain invariance into the learned model, they remain limited to some predefined and well known transformations. Indeed, there are many other transformations which are either unknown or difficult to formalize.

\item [\emph{Invariance through model architectures}:]
\hfill \\
In some neural network models, the architecture implicitly builds a certain type of invariance. For instance, in convolutional networks \cite{leCun1989, riesenhuber99, fukushimaM82}, combining layers of feature extractors using weight sharing with local pooling of the feature maps introduces some degree of translation invariance \cite{ranzato-cvpr-07, lee2009}. These models are currently state of the art strategies for achieving invariance in computer vision tasks. However, it is unclear how to explicitly incorporate in these models more complicated invariances such as large angle rotation and complex illumination. Moreover, convolutional and max-pooling techniques are somewhat specialized to visual and audio processing, while deep architectures are generally task independent.

\item [\emph{Invariance through analytical constraints}:]
\hfill \\
Analytical invariance consists in adding an explicit penalty term to the training objective function in order to reduce the variations of $f$ or its sub-parts when the input varies. This penalty is generally based on the derivatives of a criterion related to $f$ with respect to the input.
For instance, in unsupervised representation learning, \cite{rifaiVMGB11} introduces a penalty for training auto-encoders which encourages the intermediate representation to be robust to small changes of the input around the training samples, referred to as contractive auto-encoders. This penalty is based on the Frobenius norm of the first order derivative of the hidden representation of the auto-encoder with respect to the input. Later, \cite{salah2011} extended the contractive auto-encoders by adding another penalty using the norm of an approximation of the second order derivative of the hidden representation with respect to the input. The added term penalizes curvatures and thus favors smooth manifolds. \cite{shixiang2014} exploit the idea that solving adversarial examples is equivalent to increase the attention of the network to small perturbation for each example. Therefore, they propose a layer-wise penalty which creates flat invariance regions around the input data using the contractive penalty proposed in \cite{rifaiVMGB11}. \cite{simVicLeCDen92, simard1993} penalize the derivatives of $f$ with respect to perturbed inputs using simple distortions in order to ensure local invariance to these transformations.
Learning invariant representations through the penalization of the derivatives of the representation function $\Gamma(\cdot)$ is a strong mathematical tool. However, its main drawback is that the learned invariance is local and is generally robust toward small variations.

\bigskip

Learning invariance through explicit analytical constraints can also be found in metric learning. For instance, \cite{chopraHL05, hadsellCL06} use a contrastive loss which constrains the projection in the output space as follows: input samples annotated as similar must have close (adjacent) projections and samples annotated as dissimilar must have far projections. In the same way, Siamese networks \cite{bromley1993} proceed in learning similarity by projecting input points annotated as similar to be adjacent in the output space. This approach of analytical constraints is our main inspiration in this work, where we provide a penalty that constrains the representation function $\Gamma(\cdot)$ to build similar representation for samples from the same class, i.e., in a supervised way. 
\end{description}
\bigskip

In the following section, we present our proposal with more details.

\section{Proposed Method}
\label{sec:method5}

In deep neural networks, higher layers tend to learn the most abstract features. We would like that samples of the same class have the same features. In order to do so, we add a penalty to the training criterion of the network to constrain the intermediate representations to be class-invariant. We first describe our regularization framework by providing basic definitions and our training criterion. Then, we discuss three measures of invariance studied in this work followed by the implementation of our framework.


\subsection{Model Decomposition}

Let us consider a parametric mapping function for classification: ${\mathcal{M}(.; \bm{\theta}): \X \to \Y}$, represented here by a neural network model, where $\X$ is the input space and $\Y$ is the label space.
This neural network is arbitrarily decomposed into two parametric  sub-functions
\begin{enumerate}
  \item ${\Gamma(\cdot; \bm{\theta}_{\Gamma}): \X \to \Z}$, a representation function parameterized with the set $\bm{\theta}_{\Gamma}$. This sub-function projects an input sample $\xvec$ into a representation space $\Z$.
  \item ${\Psi(\cdot;\bm{\theta}_{\Psi}): \Z \to \Y}$, a decision function parameterized with the set $\bm{\theta}_{\Psi}$. It performs the classification decision over the representation space $\Z$.
\end{enumerate}

The network decision function can be written as follows
\begin{equation}
\label{eq:eq5-00}
\mathcal{M}(\xvec^{(i)};\bm{\theta}) = \Psi(\Gamma(\xvec^{(i)};\bm{\theta}_{\Gamma}); \bm{\theta}_{\Psi}) \; ,
\end{equation}
\noindent where  ${\bm{\theta}=\{\bm{\theta}_{\Gamma}, \bm{\theta}_{\Psi}\}}$.

\begin{figure*}[htbp!]
	\centering
	\resizebox{\textwidth}{!}{

\tikzset{%
  block/.style    = {draw, thick, rectangle, minimum height = 4em,
    minimum width = 6em}
}

\begin{tikzpicture}[auto, thick, >=triangle 45]
\draw
    node at (0,0) (xin) {}
    node [block, right=20mm of xin.east, anchor=west] (l1) {$Layer_1$}
    node [block, right=5mm of l1.east, anchor=west] (l2) {$Layer_2$}
    node [block, right=5mm of l2.east, anchor=west] (l3) {$Layer_3$}
    node [block, right=20mm of l3.east, anchor=west] (l4) {$Layer_4$}
    node [right=50mm of l4.east, anchor=west ] (yout) {}
    ;
    
    \path (l3)--(l4)  node[pos=0.5]{$\Gamma(\xvec;\bm{\theta}_\Gamma)$};

    \draw[-](xin) -- node {} (l1);
    \draw[-](l1) -- node {} (l2);
    \draw[-](l2) -- node {} (l3);
    \draw[-](l3) -- node {} (l4);
    \draw[->](l4) -- node {} (yout);
  
    \coordinate[above left=3mm of l1.north west] (gammanw);
    \coordinate[below right=3mm of l3.south east] (gammase);
    \draw [color=gray,thick, dotted] (gammanw) rectangle (gammase);
 	\node[above=1mm of gammanw, anchor=south west]  (gammalabel) {$\Gamma(\cdot;\bm{\theta}_\Gamma)$};
  
    \coordinate[above left=3mm of l4.north west] (psinw);
	\coordinate[below right=3mm of l4.south east] (psise);
	\draw [color=gray,thick, dotted] (psinw) rectangle (psise);
	\node[above=1mm of psinw, anchor=south west]  {$\Psi(\cdot;\bm{\theta}_\Psi)$};
 
  \coordinate[above left=3mm of gammalabel.north west] (mnw);
  \coordinate[below right=3mm of psise] (mse);
  \draw [color=gray,thick, dashed] (mnw) rectangle (mse);
  \node[above=1mm of mnw, anchor=south west]  {$\mathcal{M}(\cdot;\bm{\theta}=\{\bm{\theta}_{\Gamma},\bm{\theta}_{\Psi}\})= \Psi(\Gamma(.;\bm{\theta}_{\Gamma});\bm{\theta}_{\Psi})$};
  
  \path (mnw|-xin)--(xin) node[pos=0.5,above] {$\xvec$};
  \path (mse|-yout)--(yout) node[pos=0.5,above] {$\mathcal{M}(\xvec;\bm{\theta})=\Psi(\Gamma(\xvec;\bm{\theta}_\Gamma);\bm{\theta}_\Psi)$};

\end{tikzpicture}}
	\caption[\bel{Decomposition of the neural network.}]{Decomposition of the neural network $\mathcal{M}(\cdot)$ into a representation function $\Gamma(\cdot)$ and a decision function $\Psi(\cdot)$.}
	\label{fig:fig5-0-1}
\end{figure*}
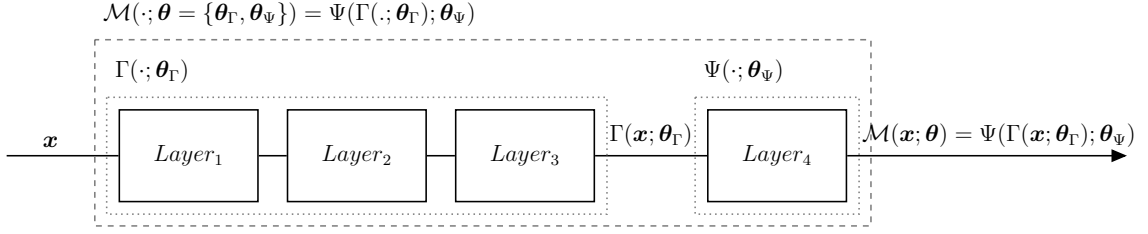

Such a possible  decomposition of a neural network with $K=4$ layers is presented in Fig.\ref{fig:fig5-0-1}. Here, the decision function $\Psi(\cdot)$ is composed of solely the output layer while the rest of the hidden layers form the representation function $\Gamma(\cdot)$.

\subsection{General Training Framework}

In order to constrain the intermediate representations $\Gamma(\cdot)$ to form clusters over all the samples within the same class we modify the training loss by adding a regularization term. Thus, the training criterion $J$ is composed of the sum of two terms. The first term $J_{sup}$ is a standard supervised term which aims at reducing the classification error. The second and proposed regularization term $J_{H}$ is a hint penalty that aims at constraining the intermediate representations of samples within the same class to be similar. 
By doing so, we constrain $\Gamma(\cdot)$ to lean invariant representations with respect to the class membership of the input sample.

\textbf{Proposed Hint Penalty}

Let ${\mathbb{D} = \{(\xvec^{(i)}, y^{(i)})\}}$ be a training set for classification task with $S$ classes and $N$ samples; $(\xvec^{(i)}, y^{(i)})$ denotes an input sample and its label. Let $\mathbb{D}_s$ be the sub-set of $\mathbb{D}$ that consists in all the examples of class $s$, i.e. $\mathbb{D}_s=\{(\xvec, y) \in \mathbb{D}\enspace s.t. \enspace y=s\}$.
By definition, $\mathbb{D}=\bigcup\limits_{s=1}^S\mathbb{D}_s$.
For the sake of simplicity,  even if $\mathbb{D}$ and  $\mathbb{D}_s$ contains tuples of (feature,target),  $\xvec$ represents only the feature part in the notation $\xvec \in \mathbb{D}$.

Let $\xvec^{(i)}$ be an input sample. We want to reduce the dissimilarity over the space $\Z$ between the projection of $\xvec^{(i)}$ and the projection of every sample ${\xvec^{(i)} \in \mathbb{D}_s}$ with $j\neq{i}$. For this sample $\xvec^{(i)}$, our hint penalty can be written as follows

\begin{equation}
  \label{eq:eq5-0}
  J_{h}(\xvec^{(i)}; \bm{\theta}_{\Gamma}) = \frac{1}{|\mathbb{D}_s|-1}\sum_{\substack{{\xvec^{(j)} \in \mathbb{D}_s}\\{j \neq i}}} \mathcal{C}_h(\Gamma(\xvec^{(i)};\bm{\theta}_{\Gamma}), \Gamma(\xvec^{(j)}; \bm{\theta}_{\Gamma})) \; ,
\end{equation}
\noindent where $\mathcal{C}_{h}(\cdot, \cdot)$ is a loss function that measures how much two projections in $\Z$ are dissimilar and $|\mathbb{D}_s|$ is the number of samples in $\mathbb{D}_s$.

\begin{figure*}[htbp!]
\centering
\resizebox{\textwidth}{!}{

\tikzset{%
  block/.style    = {draw, thick, rectangle, minimum height = 4em,
    minimum width = 6em}
}
\begin{tikzpicture}[auto, thick, >=triangle 45]

\draw
node at (0,0) (xin) {}
node [block, right=25mm of xin.east, anchor=west] (l1) {$Layer_1$}
node [block, right=5mm of l1.east, anchor=west] (l2) {$Layer_2$}
node [block, right=5mm of l2.east, anchor=west] (l3) {$Layer_3$}
node [block, right=20mm of l3.east, anchor=west] (l4) {$Layer_4$}
node [right=25mm of l4.east, anchor=west ] (yout) {}
;
\path (l3)--(l4)  coordinate[pos=0.5] (gammai) node[pos=0.5]{$\Gamma(\xvec^{(i)};\bm{\theta}_{\Gamma})$};

\draw[-](xin) -- node {} (l1);
\draw[-](l1) -- node {} (l2);
\draw[-](l2) -- node {} (l3);
\draw[-](l3) -- node {} (l4);
\draw[->](l4) -- node {} (yout);

\coordinate[above left=3mm of l1.north west] (gammanw);
\coordinate[below right=3mm of l3.south east] (gammase);
\draw [color=gray,thick, dotted] (gammanw) rectangle (gammase);
\node[above=1mm of gammanw, anchor=south west]  (gammalabel) {$\Gamma(\cdot;\bm{\theta}_{\Gamma})$};

\coordinate[above left=3mm of l4.north west] (psinw);
\coordinate[below right=3mm of l4.south east] (psise);
\draw [color=gray,thick, dotted] (psinw) rectangle (psise);
\node[above=1mm of psinw, anchor=south west]  {$\Psi(\cdot;\bm{\theta}_{\Psi})$};

\coordinate[above left=3mm of gammalabel.north west] (mnw);
\coordinate[below right=3mm of psise] (msei);
\draw [color=gray,thick, dashed] (mnw) rectangle (msei);
\node[above=1mm of mnw, anchor=south west]  {$\mathcal{M}(\cdot;\bm{\theta}=\{\bm{\theta}_{\Gamma},\bm{\theta}_{\Psi}\})= \Psi(\Gamma(.;\bm{\theta}_{\Gamma});\bm{\theta}_{\Psi})$};

\path (mnw|-xin)--(xin) node[pos=0.5,above] {$\xvec^{(i)} \in \mathbb{D}_s$};
\path (msei|-yout)--(yout) node[pos=0.5,above] {$\mathcal{M}(\xvec^{(i)};\bm{\theta})$};

\node[right=0mm of yout.east, anchor=west] (csup) {$\underset{\bm{\theta}=\{\bm{\theta}_{\Gamma},\bm{\theta}_{\Psi}\}}{\min}\enspace \mathcal{C}_{sup}(\mathcal{M}(\xvec^{(i)};\bm{\theta}), y^{(i)})$};

\draw
node at (0,-45mm) (xin) {}
node [block, right=25mm of xin.east, anchor=west] (l1) {$Layer_1$}
node [block, right=5mm of l1.east, anchor=west] (l2) {$Layer_2$}
node [block, right=5mm of l2.east, anchor=west] (l3) {$Layer_3$}
node [block, right=20mm of l3.east, anchor=west] (l4) {$Layer_4$}
node [right=25mm of l4.east, anchor=west ] (yout) {}
;
\path (l3)--(l4)  coordinate[pos=0.5] (gammaj) node[pos=0.5,below] {$\Gamma(x_j;\bm{\theta}_{\Gamma})$};

\draw[-](xin) -- node {} (l1);
\draw[-](l1) -- node {} (l2);
\draw[-](l2) -- node {} (l3);
\draw[-](l3) -- node {} (l4);
\draw[->](l4) -- node {} (yout);

\coordinate[above left=3mm of l1.north west] (gammanw);
\coordinate[below right=3mm of l3.south east] (gammase);
\draw [color=gray,thick, dotted] (gammanw) rectangle (gammase);
\node[above=1mm of gammanw, anchor=south west]  (gammalabel) {$\Gamma(\cdot)$};

\coordinate[above left=3mm of l4.north west] (psinw);
\coordinate[below right=3mm of l4.south east] (psise);
\draw [color=gray,thick, dotted] (psinw) rectangle (psise);
\node[above=1mm of psinw, anchor=south west]  {$\Psi(\cdot)$};

\coordinate[above left=3mm of gammalabel.north west] (mnwj);
\coordinate[below right=3mm of psise] (mse);
\draw [color=gray,thick, dashed] (mnwj) rectangle (mse);
\node[above=1mm of mnwj, anchor=south west]  {Replica of $\mathcal{M}(\cdot)$};

\path (mnwj|-xin)--(xin) node[pos=0.5,above] {$\forall \xvec^{(j)} \in \mathbb{D}_s$} node[pos=0.5, below] {$j \neq{i}$};
\path (mse|-yout)--(yout) node[pos=0.5,above] {$\mathcal{M}(\xvec^{(j)};\bm{\theta})$};

\coordinate  (hinti) at (gammai|-msei);
\coordinate  (hintj) at (gammaj|-mnwj);
\path (hinti)--(hintj) coordinate[pos=0.5] (hint);
\node[] (ch) at (hint) {$\underset{\bm{\theta}_{\Gamma}}{\min} \enspace \mathcal{C}_h(\Gamma(\xvec^{(i)};\bm{\theta}_{\Gamma}) ,\Gamma(\xvec^{(j)};\bm{\theta}_{\Gamma}))$};
\draw[->] (gammai) -- (ch);
\draw[->] (gammaj) -- (ch);

\end{tikzpicture}}
\caption[\bel{Constraining the intermediate representations of an MLP.}]{Constraining the intermediate learned representations to be similar over a decomposed network $\mathcal{M}(\cdot)$ during the training phase.}
\label{fig:fig5-00}
\end{figure*}
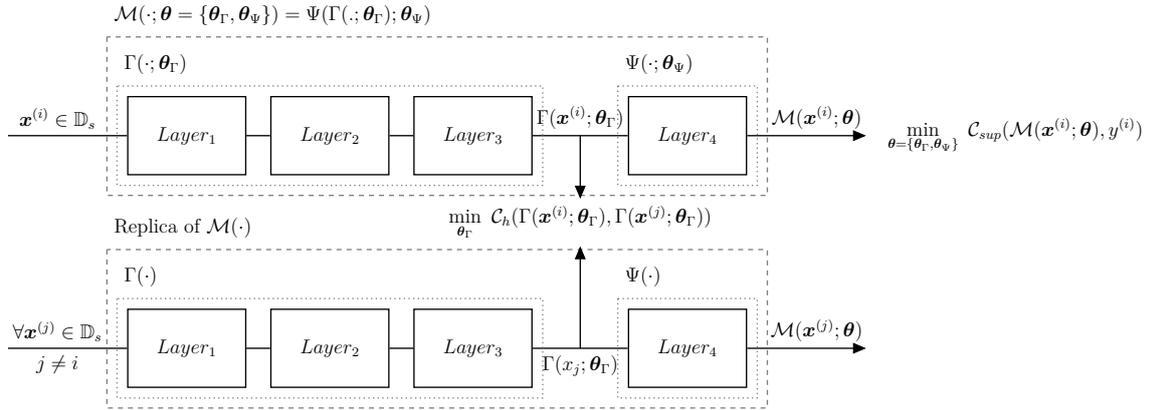

Fig.\ref{fig:fig5-00} illustrates the procedure to measure the dissimilarity in the intermediate representation space $\Z$ between two input samples $\xvec^{(i)}$ and $\xvec^{(j)}$ with the same label.
Here, we constrained only one hidden layer to be invariant. Extending this procedure for multiple layers is straightforward. It can be done by applying a similar constraint over each concerned layer.

\FloatBarrier

\textbf{Regularized Training Loss}

The full training loss can be formulated as follows
\begin{equation}
  \label{eq:eq5-12}
  J(\mathbb{D};\bm{\theta}) =  \underbrace{ \frac{\gamma}{N} \sum_{(\xvec^{(i)},y^{(i)}) \in \mathbb{D}} \mathcal{C}_{sup}(\Psi(\Gamma(\xvec^{(i)}; \bm{\theta}_{\Gamma}); \bm{\theta}_{\Psi}), y^{(i)})}_{\text{Supervised loss } J_{sup}} 
  + \underbrace{ \frac{\lambda}{S} \sum_{s=1}^{S} \frac{1}{|\mathbb{D}_s|} \sum_{\xvec^{(i)} \in \mathbb{D}_s} J_{h}(\xvec^{(i)}; \bm{\theta}_{\Gamma})}_{\text{Hint penalty } J_{H}} \; ,
\end{equation}

\noindent where $\gamma$  and $\lambda$ are regularization weights, $\mathcal{C}_{sup}(\cdot, \cdot)$ the classification loss function.
If one use a  dissimilarity measure $\mathcal{C}_h(\cdot, \cdot)$ in $J_h$ that is symmetrical such as typically a distance, summations in the term $J_H$ could be rewritten to prevent the same sample couple to appear twice.

Eq.\ref{eq:eq5-12} shares a similarity with the contrastive loss \cite{chopraHL05, hadsellCL06, bromley1993}. This last one is composed of two terms. One term constrains the learned model to project similar inputs to be closer in the output space. In Eq.\ref{eq:eq5-12}, this is represented by the hint term. In \cite{chopraHL05, hadsellCL06, bromley1993}, to avoid collapsing all the inputs into one single output point, the contrastive loss uses a second term which projects dissimilar points far from each other by at least a minimal distance. In Eq.\ref{eq:eq5-12}, the supervised term prevents, implicitly, this collapsing by constraining the extracted representations to be discriminative with respect to each class in order to minimize the classification training error.

\subsection{Implementation and Optimization Details}
\label{implandoptim}

In the present work, we have chosen the cross-entropy as the classification loss $\mathcal{C}_{sup}(\cdot, \cdot)$.

In order to quantify how much two representation vectors in $\Z$, $\bm{a}, \bm{b} \in \Z \subset \R^{V}$, are dissimilar we proceed using a distance based approach for $\mathcal{C}_h(\cdot, \cdot)$.
We study three different measures: the squared Euclidean distance (SED),

\begin{equation}
	\label{eq:eq5-2}
	\mathcal{C}_h(\bm{a}, \bm{b}) = \lVert {\bm{a}} - {\bm{b}}\rVert^2_2 =  \sum_{v=1}^V ({a}_{ v} - {b}_{v})^2\enspace,
\end{equation}

\noindent the normalized Manhattan distance (NMD),

\begin{equation}
	\label{eq:eq5-22}
	\mathcal{C}_h({\bm{a}}, {\bm{b}}) = \frac{1}{V}\sum_{v=1}^V |{a}_{v} - {b}_{v}|\enspace,
\end{equation}

\noindent and the angular similarity (AS),

\begin{equation}
\label{eq:eq5-23}
\mathcal{C}_h({\bm{a}}, {\bm{b}}) = \arccos \left(\frac{\langle {\bm{a}}, {\bm{b}}\rangle}{\lVert {\bm{a}}\rVert_2 \; \lVert {\bm{b}}\rVert_2}\right) \enspace.
\end{equation}

Minimizing the loss function of Eq.\ref{eq:eq5-12} is achieved using Stochastic Gradient Descent (SGD). Eq.\ref{eq:eq5-12} can be seen as multi-tasking where two tasks represented by the supervised term and the hint term are in concurrence. One way to minimize Eq.\ref{eq:eq5-12} is to perform a parallel optimization of both tasks by adding their gradient. Summing up the gradient of both tasks can lead to issues mainly because both tasks have different objectives that do not steer necessarily in the same direction. In order to avoid these issues, we propose to separate the gradients by alternating between the two terms at each mini-batch which showed to work well in practice \cite{caruana97ML, weston2012, collobert08ICML, bel16}. Moreover, we use two separate optimizers where each term has its own optimizer. By doing so, we make sure that both gradients are separated.

On a large dataset, computing all the dissimilarity measures in $J_H$ in Eq.\ref{eq:eq5-12} over the whole training dataset is computationally expensive due to the large number of pairs. Therefore, we propose to compute it only over the mini-batch presented to the network. Consequently, we need to shuffle the training set $\mathbb{D}$ periodically in order to ensure that the network has seen almost all the possible combinations of the pairs. We describe our implementation in Alg.\ref{alg:alg5-0}.

\begin{algorithm}[htbp!]
    \caption{Our training strategy}
    \label{alg:alg5-0}
    \begin{algorithmic}[1]
      \State $\mathbb{D}$ is the 
      training set. $B_s$ a mini-batch. $B_r$ a mini-batch of all the possible pairs in $B_s$ (Eq.\ref{eq:eq5-12}). $OP_s$ an optimizer of the supervised term. $OP_r$ an optimizer of the dissimilarity term. \texttt{max\_epochs}: maximum epochs. $\gamma, \lambda$ are regularization weights.
      \For{i=1..\texttt{max\_epoch}}
      \State Shuffle $\mathbb{D}$. Then, split it into mini-batches.
      \For{$(B_s, B_r)$ in $\mathbb{D}$}
        \State Make a gradient step toward $J_{sup}$ using $B_s$ and \par \hskip\algorithmicindent $OP_s$. (Eq.\ref{eq:eq5-12})
        \State Make a gradient step toward $J_H$ using $B_h$ and \par \hskip\algorithmicindent $OP_r$. (Eq.\ref{eq:eq5-12})
      \EndFor
      \EndFor
    \end{algorithmic}
\end{algorithm}

\FloatBarrier

\section[Experiments]{Experiments}
\label{sec:exps5}
In this section, we evaluate our regularization framework for training deep networks on a classification task as described in Sec.\ref{sec:method5}. In order to show the effect of using our regularization on the generalization performance, we will mainly compare the generalization error of a network trained with and without our regularizer on different benchmarks of classification problems.

\subsection[Classification Problems and Experimental Methodology]{Classification Problems and Experimental \\ Methodology}

\begin{figure}[!htbp]
\centering
\includegraphics[scale=0.3]{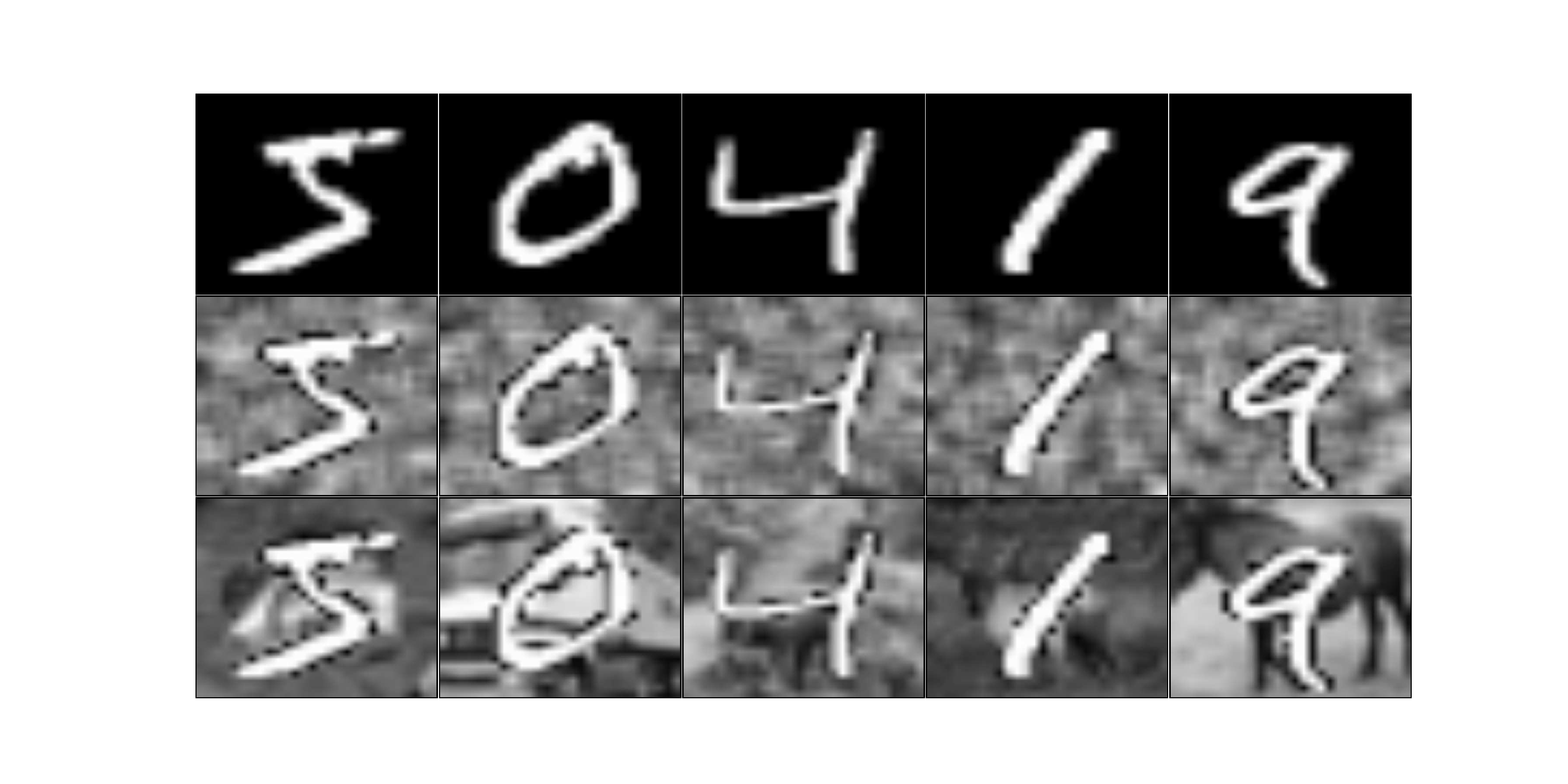}
\caption[\bel{Samples from training set of each benchmark.}]{Samples from training set of each benchmark. \emph{Top row}: \emph{mnist-std} benchmark. \emph{Middle row}: \emph{mnist-noise} benchmark. \emph{Bottom row}: \emph{mnist-img} benchmark.}
\label{fig:figsamples5}
\end{figure}

In our experiments, we consider three classification problems. We start by the standard MNIST digit dataset. Then, we complicate the classification task by adding different types of noise. We consider the three following problems:

\begin{itemize}
\item The standard MNIST digit classification problem with $\mathit{50000}$, $\mathit{10000}$ and $\mathit{10000}$ training, validation and test set. We refer to this benchmark as \emph{mnist-std}. (Fig.\ref{fig:figsamples5}, top row).
\item MNIST digit classification problem where we use a background mask composed of a random noise followed by a uniform filter. The dataset is composed of $\mathit{100000}$, $\mathit{20000}$ and $\mathit{50000}$ samples for train, validation and test set. Each set is generated from the corresponding set in the benchmark \emph{mnist-std}. We refer to this benchmark as \emph{mnist-noise}. (Fig.\ref{fig:figsamples5}, middle row).
\item MNIST digit classification problem where we use a background mask composed of a random picture taken from CIFAR-10 dataset \cite{krizhevsky09learningmultiple}. This benchmark is composed of $\mathit{100000}$ samples for training built upon $\mathit{40000}$ training samples of CIFAR-10 training set, $\mathit{20000}$ samples for validation built upon the rest of CIFAR-10 training set (i.e. $\mathit{10000}$ samples) and $\mathit{50000}$ samples for test built upon the $\mathit{10000}$ test samples of CIFAR-10. We refer to this benchmark as \emph{mnist-img}. (Fig.\ref{fig:figsamples5}, bottom row).
\end{itemize}

All the images are $28 \times 28$ gray-scale values scaled to $[0, 1]$. In order to study the behavior of our proposal where we have few training samples, we use different configurations for the training set size. We consider four configurations where we take only $\mathit{1000}$, $\mathit{3000}$, $\mathit{5000}$, $\mathit{50000}$ or $\mathit{100000}$ training samples from the whole available training set. We refer to each configuration by $\mathit{1k}$, $\mathit{3k}$, $\mathit{5k}$, $\mathit{50k}$ and $\mathit{100k}$ respectively. For the benchmark \emph{mnist-std}, only the configurations $\mathit{1k}$, $\mathit{3k}$, $\mathit{5k}$ and $\mathit{50k}$ are considered.

For all the experiments, we consider the two following neural network architectures:
\begin{itemize}
\item Multilayer perceptron with 3 hidden layers followed by a classification output layer. We use the same architecture as in \cite{devriesESANN2016} which is $1200-1200-200$. This model is referred to as \emph{mlp}.
\item LeNet convolutional network \cite{lecun98gradient}, which is well known in computer vision tasks, (with similar architecture to LeNet-4) with 2 convolution layers with 20 and 50 filters of size $5 \times 5$, followed by a dense layer of size $500$, followed by a classification output layer. This model is referred to as \emph{lenet}.
\end{itemize}
Each model has three hidden layers, we refer to each layer from the input toward the output layer by: $h_1, h_2$ and $h_3$ respectively. The output layer is referred to as $h_4$. When using our hint term, we refer to the model by \emph{mlp + hint} and \emph{lenet + hint} for the \emph{mlp} and \emph{lenet} models respectively.

Each experiment is repeated $7$ times. The best and the worst test classification error cases are discarded. We report the \emph{mean} $\pm$ \emph{standard deviation} of the validation (vl) and the test (tst) classification error of each benchmark. Models without regularization are trained for $\mathit{2000}$ epochs. All the models regularized with our proposal are trained for $\mathit{400}$ epochs which we found enough to converge and find a better model over the validation set. All the trainings are performed using stochastic gradient descent with an adaptive learning rate applied using AdaDelta \cite{zeiler12}, with a batch size of $\mathit{100}$.

\textbf{Technical Details}:

\begin{itemize}
\item We found that layers with bounded activation functions such as the logistic sigmoid or the hyperbolic tangent function are more suitable when applying our hint term. Applying the regularization term over a layer with unbounded activation function such as the Relu \cite{NairHicml10} did not show an improvement.
\item In practice, we found that setting $\gamma=1, \lambda=1$ works well.
\end{itemize}
The source code of our implementation is freely available \footnote{\url{https://github.com/sbelharbi/learning-class-invariant-features}}.

\subsection{Results}
As we have described in Sec.\ref{sec:method5}, our hint term can be applied at any hidden layer of the network. In this section, we perform a set of experiments in order to have an idea \fromont{about} which one is more adequate to use our regularization. To do so, we trained the \emph{mlp} model for classification task over the benchmark \emph{mnist-std} using different configurations with and without regularization. The regularization is applied for one hidden layer at a time $h_1, h_2$ or $h_3$. We used the squared Euclidean distance (Eq.\ref{eq:eq5-2}) as a dissimilarity measure. The obtained results are presented in Tab.\ref{tab:tab5-17}.

\begin{table*}[!htbp]
    \centering
  \resizebox{1.\textwidth}{!}{
   \begin{tabular}{|l||c|c||c|c||c|c||c|c|}
		\hline
        Model/train data size&\multicolumn{2}{|c||}{$\mathit{1k}$}&\multicolumn{2}{c||}{$\mathit{3k}$}&\multicolumn{2}{c||}{$\mathit{5k}$}&\multicolumn{2}{c|}{$\mathit{50k}$}\\        
                \hline

                \multicolumn{1}{c||}{}
                &vl&tst&
                vl&tst&
                vl&tst&
                vl&tst\\
                \cline{2-9}
                \cline{2-9}
                \multicolumn{1}{c||}{}&\multicolumn{8}{|c|}{\textbf{\emph{mlp}}}\\
                \cline{2-9}
                \cline{2-9}
                \multicolumn{1}{c||}{}&
                $10.49\pm0.031$&
                $11.24\pm0.050$&
                $6.69\pm0.039$&
                $7.17\pm0.010$&
                $5.262\pm0.030$&
                $5.63\pm0.126$&
                $\bm{1.574\pm0.016}$&
                $\bm{1.66\pm0.016}$\\
                \cline{2-9}
                \cline{2-9}
                \multicolumn{1}{c||}{}&\multicolumn{8}{|c|}{\textbf{\emph{mlp + hint}}}\\
                \cline{2-9}
                \hline
                $h_3$&
                $\bm{8.80\pm0.093}$&
                $\bm{9.50\pm0.093}$&
                $\bm{5.81\pm0.104}$&
                $\bm{6.24\pm0.069}$&
                $\bm{4.74\pm0.065}$&
                $\bm{5.05\pm0.035}$&
                $1.67\pm0.043$&
                $1.73\pm0.080$\\
                \hline
                \hline
                $h_2$&
                $11.48\pm0.081$&
                $12.32\pm0.090$&
                $6.72\pm0.031$&
                $7.29\pm0.038$&
                $5.33\pm0.031$&
                $5.84\pm0.030$&
                $1.88\pm0.043$&
                $1.97\pm0.071$\\
                \hline
                \hline
                $h_1$&
                $12.15\pm0.043$&
                $12.74\pm0.189$&
                $6.75\pm0.041$&
                $7.26\pm0.049$&
                $5.35\pm0.028$&
                $5.87\pm0.050$&
                $1.83\pm0.033$&
                $1.95\pm0.025$\\
                \hline
	\end{tabular}
        }
	\caption[\bel{Mean $\pm$ standard deviation error over validation and test sets.}]{Mean $\pm$ standard deviation error over validation and test set of the benchmark \emph{mnist-std} using the model \emph{mlp} and the SED as dissimilarity measure over the different hidden layers: $h_1, h_2, h_3$. (\textbf{bold font indicates lowest error.})}
	\label{tab:tab5-17}
\end{table*}

From Tab.\ref{tab:tab5-17}, \fromont{it seems that the proposed method decreases systematically the performance when used in layers 1, and 2 in the configuration $1k$.}
This may be explained by the fact that low layers in neural networks tend to learn low representations which are \emph{shared} among high representations. This means that these representations are not ready yet to discriminate between the classes. Therefore, they can not be used to describe each class separately. This makes our regularization inadequate at these levels because we aim at constraining the representations to be similar within each class while these layers are incapable to deliver such representations. Therefore, regularizing these layers may hamper their learning. As a future work, we think that it would be beneficial to use at low layers a regularization term that constrains the representations of samples within different classes be dissimilar such as the one in the contrastive loss \cite{chopraHL05, hadsellCL06, bromley1993}.

In the case of regularizing the last hidden layer $h_3$, we notice from Tab.\ref{tab:tab5-17} an important improvement in the classification error over the validation and the test set in most configurations. This may be explained by the fact that the representations at this layer are more abstract, therefore, they are able to discriminate the classes. Our regularization term constrains these representations to be tighter by re-enforcing their invariance which helps in generalization. Therefore, applying our hint term over the last hidden layer makes more sense and supports the idea that high layers in neural networks learn more abstract representations. Making these discriminative representations invariant helps the linear output layer in the classification task. For all the following experiments, we apply hint term over the last hidden layer.
Moreover, one can notice that our regularization has less impact when adding more training samples. For instance, we reduced the classification test error by: $1.74\%$, $0.92\%$ and $0.58\%$ in the configurations $1k$, $3k$ and $5k$. This suggests that our proposal is more efficient in the case where few training samples are available. However, this does not exclude using it for large training datasets as we will see later (Tab.\ref{tab:tab5-18}, \ref{tab:tab5-20}). We believe that this behavior depends mostly on the model's capacity to learn invariant representations. For instance, from the invariance perspective, convolutional networks are more adapted, conceptually, to process visual content than multilayers perceptrons.

\bigskip

In another experimental setup, we investigated the effect of the measure used to compute the dissimilarity between two feature vectors as described in Sec.\ref{implandoptim}. To do so, we applied our hint term over the last hidden layer $h_3$ using the measures SED, NMD and AS over the benchmark \emph{mnist-std}. The obtained results are presented in Tab.\ref{tab:tab5-18}. These results show that the squared Euclidean distance performs significantly better than the other measures and has more stability when changing the number of training samples ($\mathit{1k}$, $\mathit{3k}$, $\mathit{5k}$, $\mathit{50k}$) or the model (\emph{mlp}, \emph{lenet}).

\begin{table*}[!htbp]
    \centering
  \resizebox{1.\textwidth}{!}{
   \begin{tabular}{|l||c|c||c|c||c|c||c|c|}
		\hline
        Model/train data size&\multicolumn{2}{|c||}{$\mathit{1k}$}&\multicolumn{2}{c||}{$\mathit{3k}$}&\multicolumn{2}{c||}{$\mathit{5k}$}&\multicolumn{2}{c|}{$\mathit{50K}$}\\        
                \hline
                \multicolumn{1}{c||}{}
                &vl&tst&
                vl&tst&
                vl&tst&
                vl&tst\\
                \cline{2-9}
                \multicolumn{1}{c||}{}
                &\multicolumn{8}{|c|}{\textbf{MLP}}\\
                \hline{2-9}
                \emph{mlp}&
                $10.49\pm0.031$&
                $11.24\pm0.050$&
                $6.69\pm0.039$&
                $7.17\pm0.010$&
                $5.262\pm0.030$&
                $5.63\pm0.126$&
                $1.574\pm0.016$&
                $1.66\pm0.016$\\
                \hline
                \hline
                \emph{mlp + hint} (SED)&
                $\bm{8.80\pm0.093}$&
                $\bm{9.50\pm0.093}$&
                $\bm{5.81\pm0.104}$&
                $\bm{6.24\pm0.069}$&
                $\bm{4.74\pm0.065}$&
                $\bm{5.05\pm0.035}$&
                $1.67\pm0.043$&
                $1.73\pm0.080$\\
                \hline
                \hline
                \emph{mlp + hint} (NMD)&
                $10.32\pm0.028$&
                $10.92\pm0.094$&
                $6.69\pm0.075$&
                $7.22\pm0.059$&
                $5.34\pm0.035$&
                $5.79\pm0.045$&
                $1.44\pm0.020$&
                $1.47\pm0.020$\\
                \hline
                \hline
                \emph{mlp + hint} (AS)&
                $10.27\pm0.068$&
                $10.71\pm0.123$&
                $6.52\pm0.044$&
                $6.89\pm0.013$&
                $4.96\pm0.041$&
                $5.25\pm0.051$&
                $\bm{1.37\pm0.023}$&
                $\bm{1.37\pm0.025}$\\
                \hline
                \multicolumn{1}{c||}{}
                &\multicolumn{8}{|c|}{\textbf{Lenet}}\\
                \hline
                \hline
                \emph{lenet}&
                $6.25\pm0.016$&
                $7.27\pm0.033$&
                $3.65\pm0.085$&
                $4.02\pm0.073$&
                $2.62\pm0.031$&
                $2.90\pm0.058$&
                $1.31\pm0.028$&
                $1.23\pm0.024$\\
                \hline
                \hline
                \emph{lenet + hint} (SED)&
                $\bm{4.54\pm0.150}$&
                $5.05\pm0.115$&
                $\bm{2.70\pm0.124}$&
                $\bm{2.85\pm0.082}$&
                $\bm{2.06\pm0.113}$&
                $\bm{2.37\pm0.105}$&
                $\bm{0.97\pm0.087}$&
                $\bm{1.04\pm0.060}$\\
                \hline
                \hline
                \emph{lenet + hint} (NMD)&
                $6.70\pm0.040$&
                $\bm{4.60\pm0.065}$&
                $3.85\pm0.032$&
                $4.30\pm0.036$&
                $2.87\pm0.045$&
                $3.14\pm0.035$&
                $1.99\pm0.043$&
                $2.075\pm0.079$\\
                \hline
                \hline
                \emph{lenet + hint} (AS)&
                $6.72\pm0.024$&
                $7.66\pm0.024$&
                $3.86\pm0.049$&
                $4.26\pm0.049$&
                $2.80\pm0.033$&
                $3.12\pm0.021$&
                $1.75\pm0.123$&
                $1.97\pm0.063$\\
                \hline
	\end{tabular}
        }
	\caption[\bel{Mean $\pm$ standard deviation error over validation and test sets.}]{Mean $\pm$ standard deviation error over validation and test set of the benchmark \emph{mnist-std} using different dissimilarity measures (SED, NMD, AS) over the layer $h_3$. (\textbf{bold font indicates lowest error.})}
	\label{tab:tab5-18}
\end{table*}

\bigskip

In another experiment, we evaluated the benchmarks \emph{mnist-noise} and \emph{mnist-img}, which are more difficult compared to \emph{mnist-std}, using the model \emph{lenet} which is more suitable to process visual content. Similarly to the previous experiments, we applied our regularization term over the last hidden layer $h_3$ using the SED measure. The results depicted in Tab.\ref{tab:tab5-20} show again that using our proposal improves the generalization error of the network particularly when only few training samples are available. For example, our regularization allows to reduce the classification error over the test set by $2.98\%$ and by $4.16\%$ over the benchmark \emph{mnist-noise} and \emph{mnist-img}, respectively when using only $\mathit{1k}$ training samples.

\begin{table*}[!htbp]
    \centering
  \resizebox{1.\textwidth}{!}{
   \begin{tabular}{|l||c|c||c|c||c|c||c|c|}
		\hline
        Model/train data size&\multicolumn{2}{|c||}{$\mathit{1k}$}&\multicolumn{2}{c||}{$\mathit{3k}$}&\multicolumn{2}{c||}{$\mathit{5k}$}&\multicolumn{2}{c|}{$\mathit{100k}$}\\        
                \hline
                \multicolumn{1}{c||}{}
                &vl&tst&
                vl&tst&
                vl&tst&
                vl&tst\\
                \cline{2-9}
                \multicolumn{1}{c||}{}
                &\multicolumn{8}{|c|}{\textbf{\emph{mnist-noise}}}\\
                \hline
                \emph{lenet}&
                $9.62\pm0.123$&
                $10.72\pm0.116$&
                $5.95\pm0.059$&
                $6.39\pm0.032$&
                $4.92\pm0.036$&
                $5.11\pm0.012$&
                $1.90\pm0.020$&
                $2.011\pm0.018$\\
                \hline
                \hline
                \emph{lenet + hint}&
                $\bm{7.12\pm0.200}$&
                $\bm{7.74\pm0.148}$&
                $\bm{4.09\pm0.130}$&
                $\bm{4.62\pm0.059}$&
                $\bm{3.53\pm0.117}$&
                $\bm{3.98\pm0.167}$&
                $\bm{1.60\pm0.107}$&
                $\bm{1.64\pm0.116}$\\
                \hline
                \multicolumn{1}{c||}{}
                &\multicolumn{8}{|c|}{\textbf{\emph{mnist-img}}}\\
                \hline
                \emph{lenet}&
                $13.88\pm0.114$&
                $15.34\pm0.124$&
                $8.34\pm0.030$&
                $8.66\pm0.024$&
                $6.64\pm0.057$&
                $6.46\pm0.033$&
                $2.53\pm0.080$&
                $2.55\pm0.007$\\
                \hline
                \hline
                \emph{lenet + hint}&
                $\bm{10.30\pm0.425}$&
                $\bm{11.18\pm0.290}$&
                $\bm{6.19\pm0.281}$&
                $\bm{6.61\pm0.212}$&
                $\bm{5.37\pm0.358}$&
                $\bm{5.65\pm0.310}$&
                $\bm{2.15\pm0.105}$&
                $\bm{2.21\pm0.032}$\\
                \hline
	\end{tabular}
        }
	\caption[\bel{Mean $\pm$ standard deviation error over validation and test sets.}]{Mean $\pm$ standard deviation error over validation and test set of the benchmarks \emph{mnist-noise} and \emph{mnist-img} using \emph{lenet} model (regularization applied over the layer $h_3$). (\textbf{bold font indicates lowest error.})}
	\label{tab:tab5-20}
\end{table*}

\bigskip

Based on the above results, we conclude that using our hint term in the context of classification task using neural networks is helpful in improving their generalization error particularly when only few training samples are available. This generalization improvement came at the price of an extra computational cost due the dissimilarity measures between pair of samples. Our experiments showed that regularizing the last hidden layer using the squared Euclidean distance give better results. More generally, the obtained results confirm that guiding the learning process of the intermediate representations of a neural network can be helpful to improve its generalization.

\subsection{On Learning Invariance within Neural Networks}
\begin{figure*}[!htbp]
\centering
\includegraphics[scale=0.3]{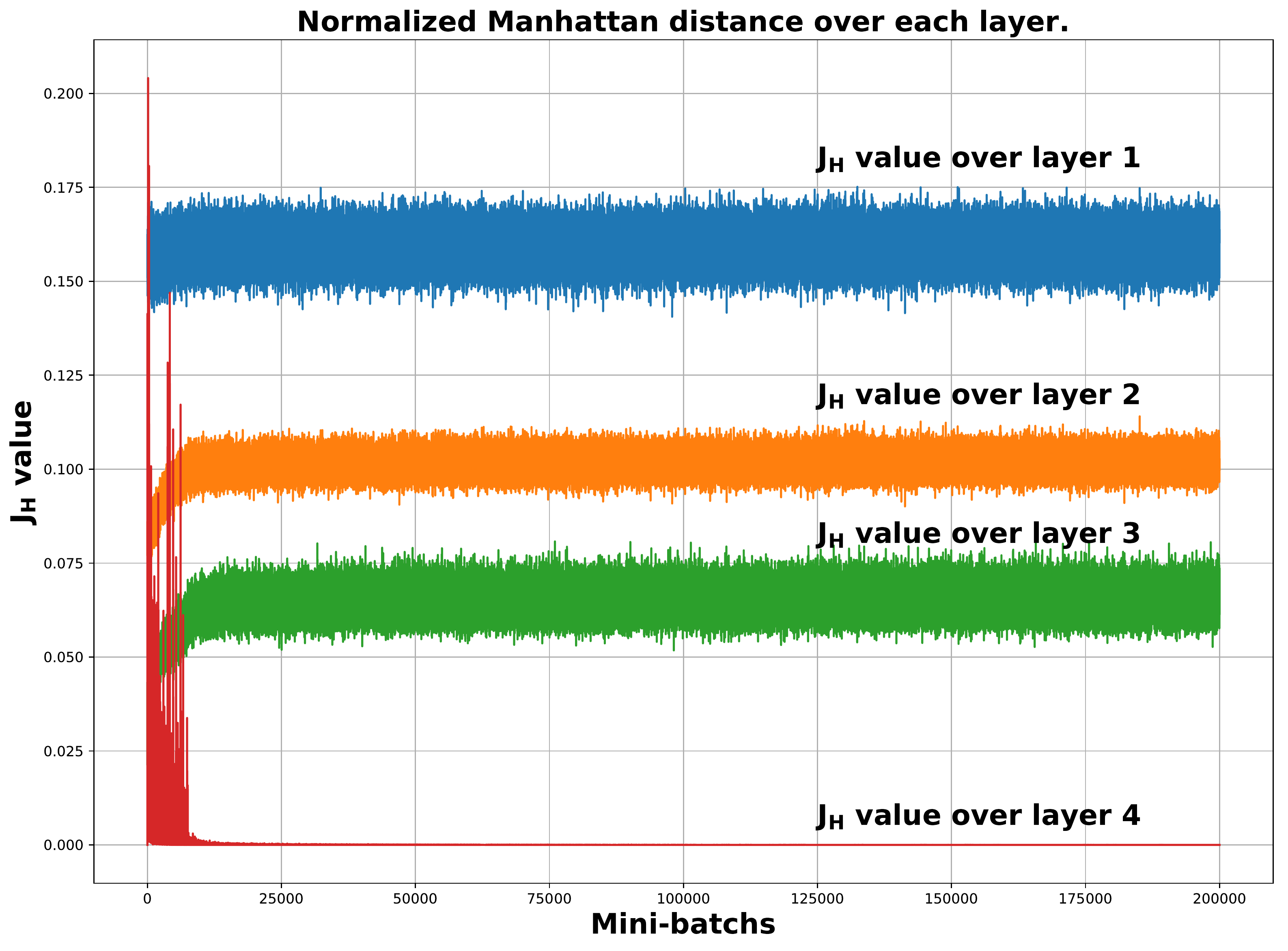}
\caption[\bel{Measuring the hint term of the different layers.}]{Measuring the hint term $J_H$ of Eq.\ref{eq:eq5-12} over the training set within each layer (simultaneously) of the \emph{mlp} over the train set of \emph{mnist-std} benchmark for a binary classification task: the digit \quotes{1} against the digit \quotes{7}.}
\label{fig:figmeasuremlpandlenet5}
\end{figure*}

We show in this section an intriguing property of the learned representations at each layer of a neural network from the invariance perspective. For this purpose and for the sake of simplicity, we consider a binary classification case of the two digits \quotes{1} and \quotes{7}. Furthermore, we consider the \emph{mlp} model over the \emph{lenet} in order to be able to measure the features invariances over all the layers. We trained the \emph{mlp} model over the benchmark \emph{mnist-std} where we used all the available training samples of both digits. The model is trained without our regularization. However, we tracked, at each layer and at the same time, the value of the hint term $J_H$ in Eq.\ref{eq:eq5-12} over the training set using the normalized Manhattan distance as a dissimilarity measure. This particular dissimilarity measure allows comparing the representations invariance between the different layers due to the normalization of the measure by the representations dimension. The obtained results are depicted in Fig.\ref{fig:figmeasuremlpandlenet5} where the x-axis represents the number of mini-batches already processed and the y-axis represents the value of the hint term $J_H$ at each layer. Low value of $J_H$ means high invariance (better case) whereas high value of $J_H$ means low invariance.

In Fig.\ref{fig:figmeasuremlpandlenet5}, we note two main observations:
\begin{itemize}
\item The value of the hint term $J_H$ is reduced through the depth of the network which means that the network learns more invariant representations at each layer in this order: layer 1, 2, 3, 4. This result supports the idea that abstract representations, which are known to be more invariant, are learned toward the top layers.
\item At each layer, the network does not seem to learn to improve the invariance of the learned representations by reducing $J_H$. It appears that the representations invariance is kept steady all along the training process. Only the output layer has learned to reduce the value of $J_H$ term because minimizing the classification term $J_{sup}$ reduces automatically our hint term $J_H$. This shows a flaw in the back-propagation procedure with respect to learning intermediate representations. Assisting the propagated error through regularization can be helpful to guide the hidden layers to learn more suitable representations.
\end{itemize}

These results show that relying on the classification error propagated from the output layer does not necessarily constrain the hidden layers to learn better representations for classification task. Therefore, one would like to use different prior knowledge to guide the internal layers to learn better representations which is our future work. Using these guidelines can help improving neural networks generalization especially when trained with few samples.

\FloatBarrier

\section[Conclusion]{Conclusion}
\label{sec:conclusion5}
We have presented in this work a new regularization framework for training neural networks for classification task. Our regularization constrains the hidden layers of the network to learn class-wise invariant representations where samples of the same class have the same  representation. Empirical results over MNIST dataset and its variants showed that the proposed regularization helps neural networks to generalize better particularly when few training samples are available which is the case in many real world applications.

Another result based on tracking the representation invariance within the network layers confirms that neural networks tend to learn invariant representations throughout staking multiple layers. However, an intriguing observation is that the invariance level does not seem to be improved, within the same layer, through learning. We found that the hidden layers tend to maintain a certain level of invariance through the training process.

All the results found in this work suggest that guiding the learning process of the internal representations of a neural network can be helpful to train them and improve their generalization particularly when few training samples are available. Furthermore, this shows that the classification error propagated from the output layer does not necessarily train the hidden layers to provide better representations. This encourages us to explore other directions to incorporate different prior knowledge to constrain the hidden layers to learn better representations in order to improve the generalization of the network and be able to train it with less data.

\section*{Acknowledgment}

This work has been partly supported by the grant ANR-16-CE23-0006 \quotes{Deep in France} and benefited from computational means from 
CRIANN, the contributions of which are greatly appreciated.

\chapter[Application:
Spotting L3 Slice in CT Scans using Deep Convolutional Network and Transfer Learning]{Application:\texorpdfstring{\\}{ } Spotting L3 Slice in CT Scans using Deep Convolutional Network and Transfer Learning}
\label{chap:chapter6}
\ifpdf
    \graphicspath{{Chapter6/Figs/Raster/}{Chapter6/Figs/PDF/}{Chapter6/Figs/}}
\else
    \graphicspath{{Chapter6/Figs/Vector/}{Chapter6/Figs/}}
\fi

\makeatletter
\def\input@path{{Chapter6/}}
\makeatother

\section{Prologue}

\noindent\emph{\underline{Article Details}:}
\begin{itemize}
    \item \textbf{Spotting L3 Slice in CT Scans using Deep Convolutional Network and Transfer Learning}. Soufiane Belharbi, Clément Chatelain\footnote{\label{note3}Authors with equal contribution.}, Romain Hérault\footnoteref{note3}, Sébastien Adam, Sébastien Thureau, Mathieu Chastan, and Romain Modzelewski. \emph{Computers in Biology and Medicine, 87: 95-103 (2017)}.
\end{itemize}

\bigskip

\noindent\emph{\underline{Context}:}
\\
We saw previously that the generalization error is bounded by two terms: a training error term and a complexity term (Sec.\ref{sub:modelselectionandregularization0}). Moreover, we saw that these two terms are antagonist. Therefore, one needs to strike a balance between these two terms in order to get a better generalization error. Moreover, we concluded that in order to well train models with high capacity, one needs large number of training samples. In this work, we provide a real life application of an idea that allows us to \quotes{cheat}, i.e., train a model with high capacity using only few samples.

In the last years, many neural network models have seen large success in many tasks such as pattern recognition, particularly deep convolution networks, e.g., Alexnet \cite{krizhevsky12}, VGG16 \cite{SimonyanZ14aCORR}, VGG19 \cite{SimonyanZ14aCORR}, Googlenet (Inception V1)  \cite{szegedy15}, which were trained on enormous corpus of labeled data such as ImageNet \cite{imagenetcvpr09}. This success has attracted many people and motivated the use of such models. However, training such models is time consuming and most importantly requires millions of labeled data. Luckily, the authors of the original models have made available the parameters of the trained models. Many researchers started experimenting using these parameters and adjusting them for their own tasks following a transfer learning paradigm. This idea started to spread to different applications such as character recognition \cite{Jiang2015,Ciresan2012}, signature identification \cite{Hafemann2016} and medical imaging \cite{chest-transfer,lung-transfer}. The idea consists in
\begin{enumerate*} \item taking low layers of the pre-trained network, \item plug them into a new network, \item stack on top random fully connected layers, and finally \item train the whole network on the new task \end{enumerate*}. Using this transfer learning approach, we applied a deep convolutional network to a medical domain problem that lacks data.

We present this work \cite{BELHARBICBM2017} as a chapter of this thesis under the form of one single contribution. This chapter contains the original paper as it was accepted in Computers in Biology and Medicine journal.

\bigskip

\emph{Technical context:}
\\
This work came as a part of a project developed in the clinic \quotes{\emph{Rouen Henri Becquerel Center}} to analyze 3D Computed Tomography (CT) scans. The idea consists in locating a particular vertebra slice (the third lumbar vertebra, i.e., L3) in the 3D CT scan. Then, perform an imagery analysis on it \cite{lerouge15}. Our task is to locate the L3 slice. In this work, we provide a complete automated system to locate the L3 slice in a 3D CT scan without any assumptions on which part of the patient's body is covered by the scan.

\bigskip

\noindent\emph{\underline{Contributions}:}
\\
The contribution of this work is to provide a complete automated system to locate the third lumbar vertebra in a 3D CT scan. The system was validated on a real world data. This work shows that transfer learning can be helpful in the case where only few training samples are available. Moreover, it shows the possibility to apply deep neural networks, particularly convolutional neural networks, on medical images. Furthermore, the provided system is a generic solution which can be used to locate any organ of the patient's body, providing the necessary data.

\newpage
%

\section[Introduction]{Introduction}
\label{sec:introduction6}
\begin{figure}[!htbp]
\centering
\begin{tikzpicture}
\begin{scope}[x={(-0.7cm,0.4cm)}, y={(.9cm,.2cm)}, z={(0cm,0.5cm)}]
	\node[canvas is yx plane at z=0,transform shape,opacity=0.3] at (0,0) {\includegraphics[width=5cm]{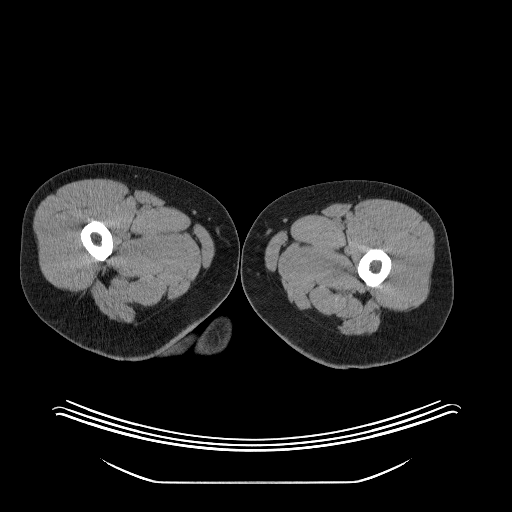}};
	\node[canvas is yx plane at z=0.5,transform shape, opacity=0.5] at (0,0) {\includegraphics[width=5cm]{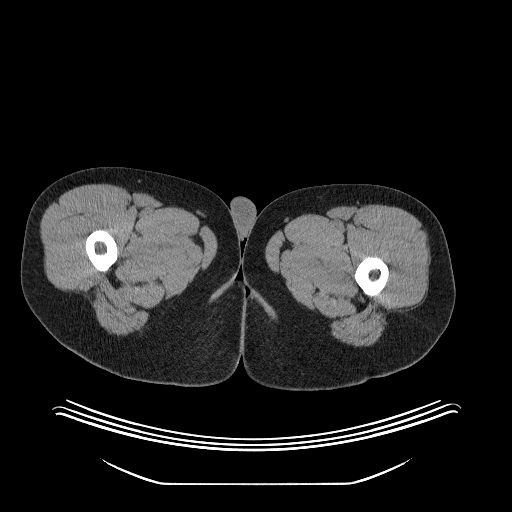}};
	\node[canvas is yx plane at z=1,transform shape, opacity=0.5] at (0,0) {\includegraphics[width=5cm]{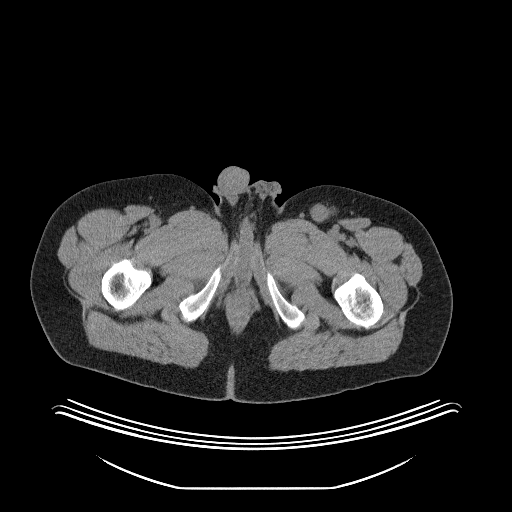}};
	\node[canvas is yx plane at z=1.5,transform shape, opacity=0.5] at (0,0) {\includegraphics[width=5cm]{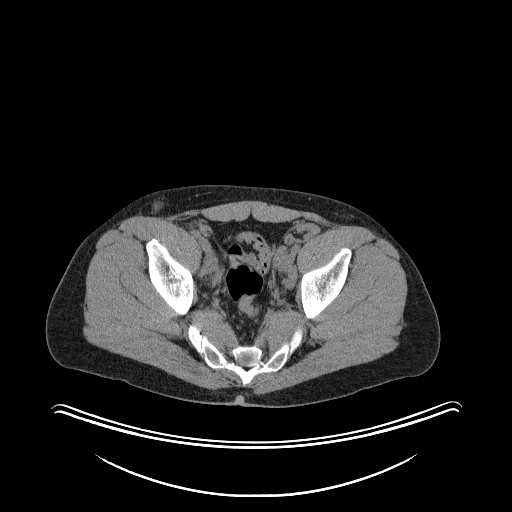}};
	\node[canvas is yx plane at z=2,transform shape, opacity=0.5] at (0,0) {\includegraphics[width=5cm]{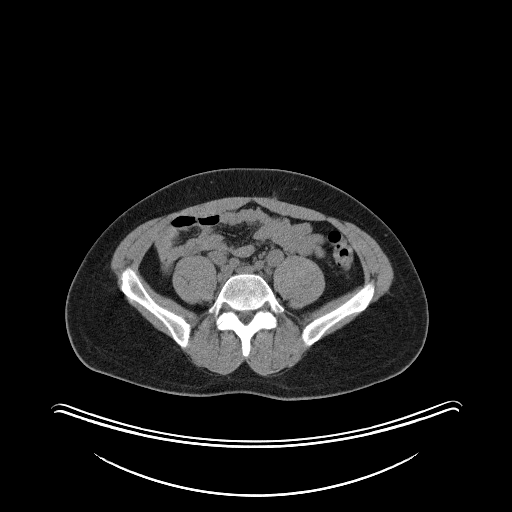}};
	\node[canvas is yx plane at z=2.5,transform shape, opacity=1, shift={(3,-1.5)}] (l3) at (0,0) {\includegraphics[width=5cm]{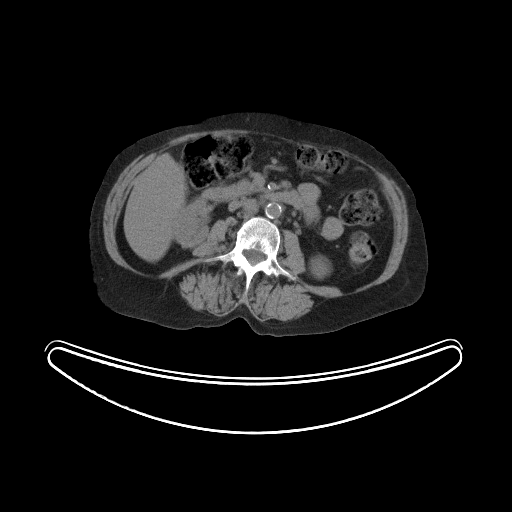}};
	\node[canvas is yx plane at z=3,transform shape, opacity=0.5] at (0,0) {\includegraphics[width=5cm]{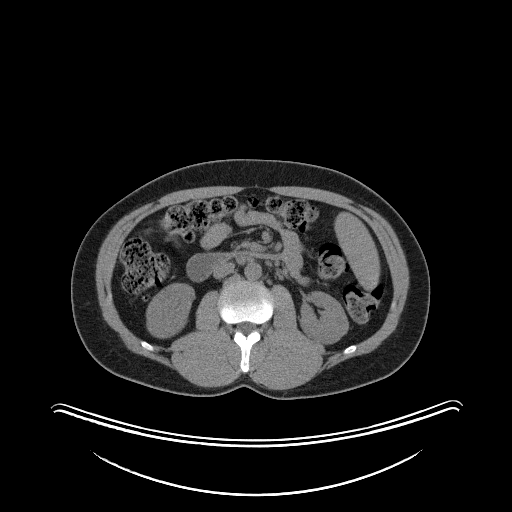}};
	\node[canvas is yx plane at z=3.5,transform shape, opacity=0.5] at (0,0) {\includegraphics[width=5cm]{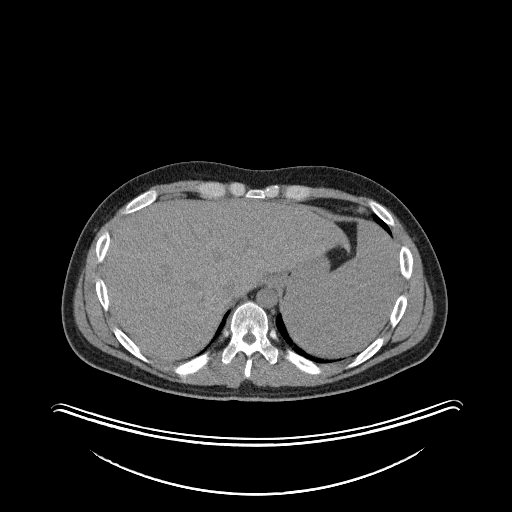}};
	\node[canvas is yx plane at z=4,transform shape, opacity=0.5] at (0,0) {\includegraphics[width=5cm]{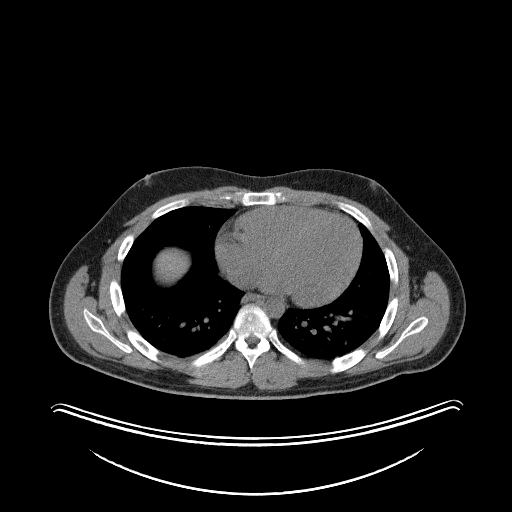}};
	\node[canvas is yx plane at z=4.5,transform shape, opacity=0.5] at (0,0) {\includegraphics[width=5cm]{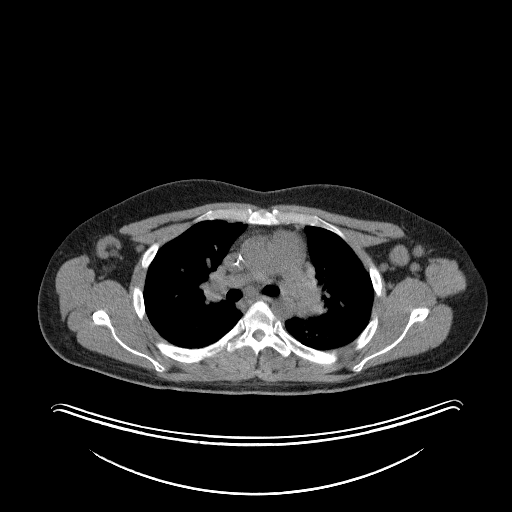}};
	\node[canvas is yx plane at z=5,transform shape, opacity=0.5] at (0,0) {\includegraphics[width=5cm]{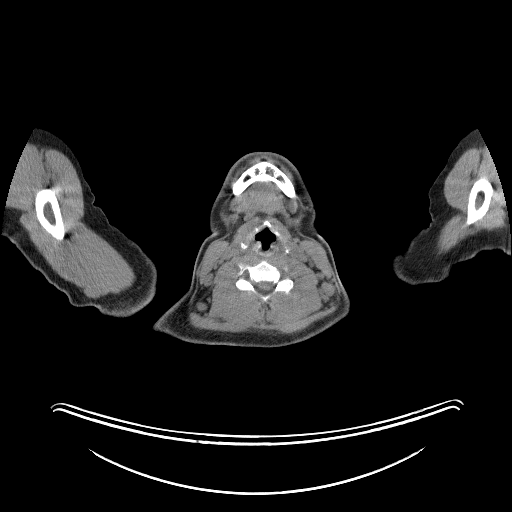}};
	\node[canvas is yx plane at z=5.5,transform shape, opacity=0.5] at (0,0) {\includegraphics[width=5cm]{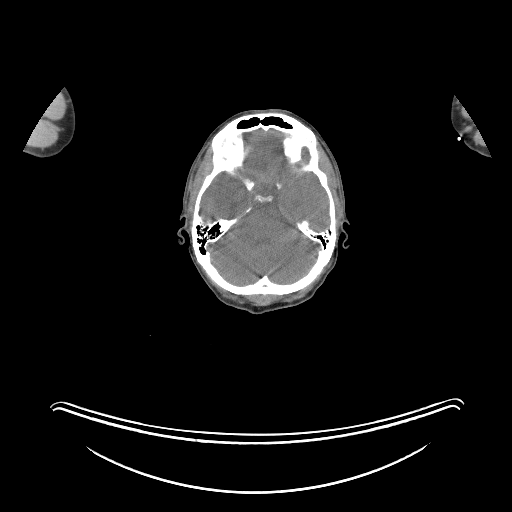}};
	\node[canvas is yx plane at z=6,transform shape, opacity=0.5] at (0,0) {\includegraphics[width=5cm]{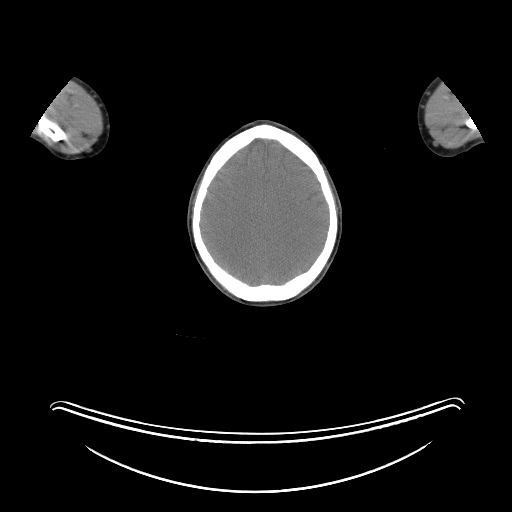}};
\end{scope}

\node[below=of l3,anchor=north] (l3label) {L3 slice};
\draw[->,thick] (l3)--(l3label);

\end{tikzpicture}

   \caption{Finding the L3 slice within a whole CT scan.}
   \label{fig:6-problem}
\end{figure}

In recent years, there has been an increasing interest in the analysis of body composition for estimating patient outcomes in many pathologies. 
For instance, sarcopenia (loss of muscle), visceral and subcutaneous obesity are known prognostic factors in cancers \cite{Martin2013,Yip2015}, cardiovascular diseases \cite{Atkins2014} and surgical procedures \cite{Peng2011,Kaido2013}. 
Body composition can also be used to improve individual nutritional care and chemotherapy dose calculation \cite{Gouerant2013,Lanic2014}. 
It is usually assessed by CT and Magnetic Resonance Imaging (MRI). 
Moreover, It has been shown that the composition of the third lumbar vertebra (L3) slice is a good estimator of the whole body measurements  \cite{Mitsiopoulos1998, Shen2004}. 
To assess the patient's body composition, radiologists usually have to manually find the corresponding L3 slice in the whole CT exam (spotting step, see Figure \ref{fig:6-problem}), and then to segment the fat and muscle on a dedicated software platform (segmentation step). 
These two operations take more than 5 minutes for an experienced radiologist and are prone to errors.  
Therefore, there is a need for automating these two tasks.

\begin{figure}[!htbp]
\begin{center}
   \includegraphics[width=0.45\textheight]{l3}
   \includegraphics[width=0.45\textheight]{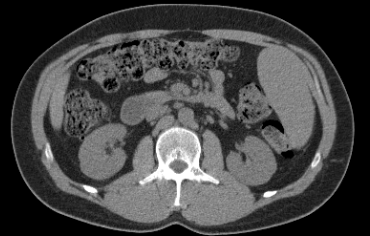}
   \end{center}
   \caption[\bel{Two slices from the same patient.}]{Two slices from the same patient: a L3 (up) and a non L3 (L2) (down). The similar shapes of both vertebrae prevent from taking a robust decision given a single slice.}
   \label{fig:6-l3pasl3}
\end{figure}

The segmentation step has been extensively addressed in the literature among the
medical imaging community \cite{pham2000current,mcinerney1996deformable}. Dedicated approaches for L3 slice have been proposed such as atlas
based methods \cite{Chung2009} or deep learning \cite{lerouge15}. On the
other hand, to the best of our knowledge, the automatic spotting of a specific
slice within the whole CT scan has not been investigated in the literature. 
The spotting task is particularly challenging since it has to handle:
\begin{itemize}
  \item The intrinsic variability in the patient's anatomy (genders, ages, morphologies or medical states).
  \item The various acquisition/reconstruction protocols (low/high X-rays dose,
    slice thickness, reconstruction filtering, enhanced/non enhanced contrast agent).
  \item The arbitrary field-of-view scans, displaying various anatomical regions.
  \item The strong similarities between the L3 slice and other slices, due to the repetitive nature of vertebrae (Fig.\ref{fig:6-l3pasl3}).
\end{itemize}

In the literature, spotting tasks are often achieved using ad hoc approaches such as registration which are not suitable for high variability problems \cite{Glocker2014,Cunliffe2015}. In particular, a 3D registration on a whole CT scan would require a large amount of computation at decision time \cite{Savva2016}.  Here, we suggest a more generic strategy based on machine learning in order to handle high variability context, while maintaining a fast decision process.

In this work, spotting a slice within a CT scan is tackled as a regression
problem, where we try to estimate the slice position height. An efficient processing flow is proposed, including a Convolutional Neural Network (CNN) learned using transfer learning. Our approach tackles the classical issues faced in medical image analysis:
the data representation issue is addressed using Maximum Intensity Projection (MIP); the variability of the shapes in CT scans is handled using a CNN; and the lack of annotated data is circumvented using transfer learning.

The article is organized as follows: Section \ref{sec:relatedw6} presents the related work and the general framework for applying machine learning for L3 detection in a CT scan. Section \ref{sec:model6} presents the proposed approach and describes each stage of the whole processing flow. Section \ref{sec:exps6} describes the experiments and the obtained results.

\section[Related Work]{Related Work}
\label{sec:relatedw6}
Machine learning approaches provide generic and flexible systems, provided
enough annotated data is available.  From a machine learning perspective, the
localization of the L3 slice given a whole CT scan can either be considered as a slice-classification problem, a sequence labeling problem or a regression problem. Let us now consider these three options.

\begin{description}
\item The classification paradigm consists of deciding for each slice of the whole CT scan whether the L3 vertebra is present or not. However, the repetitive nature of individual
  vertebra induces a similarity between the L3 slice and its neighbors, which
  prevents to efficiently classify an isolated slice without any
  context (see Fig. \ref{fig:6-l3pasl3}). This explains why even experienced radiologists need to browse the CT scan to infer the relative position and precisely identify the L3 slice. To the best of our knowledge, the classification paradigm has not been used in the literature to detect the L3 slice within a whole CT scan. 
\item The sequence labeling paradigm consists of estimating the label (L1, L2, etc.) of every slice of a complete CT scan, then, choose the one that is more likely to correspond to the L3. The advantage of this approach is that the decision is globally taken on the whole CT scan by analyzing the dependencies between the slices. This kind of approach has been recently investigated for labeling the vertebrae of complete spine images \cite{Ghosh2011,Golodetz2009,MichaelKelm2013,
    Glocker2013,Kadoury2011,Glocker2012,Huang2009,Ma2013,Oktay2011}. The dependencies are modeled using graphical models, such as Hidden Markov Models
  (HMMs) \cite{Glocker2012} or Markov Random Fields (MRFs) \cite{Kadoury2011}. A full review of the spine labelization methods can be found in \cite{Major13}. The major drawback of sequence labeling approaches is that they require a fully annotated learning database where every slice of the CT scan is labeled, which is very time consuming. Such a dataset is proposed by \cite{Glocker2014}, but this dataset cannot be easily exploited for our problem since
\begin{inparaenum}[i)]
\item  the data are cropped images of the whole spine, and
\item it contains only 224 CT scan.
\end{inparaenum}  
\item The regression problem consists of directly estimating \fromont{a real value that indicates } the L3 slice \fromont{position (i.e., the number of the slice)}
  given the whole CT scan, in a spotting fashion. Like the previous paradigm, it
  has the advantage of performing a global decision by taking into account the
  dependencies within the entire exam. Another major advantage of a spotting
  approach is that it does not require a full labeling of the exams. Indeed, the
  only annotation needed for learning such a model is the L3 position within the
  whole exam. For radiologists, this annotation is more lightweight than a full
  annotation and may lead to creating large datasets easily. 
\end{description}

In this work, we retain the third paradigm and propose a machine learning
approach for spotting the L3 slice in heterogeneous arbitrary field-of-view CT scans.
To the best of our knowledge, this is the first time that slice spotting is addressed as a machine learning regression problem.  

Usually, traditional machine learning methods exploit generic hand-designed features which are fed to a learning model with the assumption that they are suitable for describing the image. To achieve high accuracy, usually one ends up combining many types of features which require extensive computation, more time and large memory size. Ideally, it would be better if the model is capable of learning on its own task-dependent features.

Deep neural networks (DNN) are a specific category of models in machine learning
which are capable of learning on their own hierarchical features based on the raw
image. Convolutional neural networks (CNN) are a particular type of DNN which
gained a large reputation in computer vision due to their high performance for
many tasks on natural scene images \cite{Szegedy13, Erhan14, Shaoqing15,
  krizhevsky12}.

In the last years, the use of machine learning, in general, and using CNN, in particular, has grown in various medical domains such as cancer diagnosis \cite{Roth14, Urban14}, segmentation \cite{Huang13,Havaei15,Lai2015} or histological \cite{histo-cnn} and drusen identification \cite{drusen-cnn}.
In all these works, the authors are faced with a common issue which is the lack of annotated data. Although extremely powerful, CNN architectures require a huge amount of data to avoid the \quotes{learning by heart} phenomenon, also known as overfitting in machine learning. The classical techniques to limit these issues are dropout, data augmentation or the use of regularization. All these technical tricks are exploited in \cite{Lai2015}, but the lack of data is still a limitation to train such large models. 
Recently, a more efficient way has been proposed to circumvent the lack of annotated data in vision. This method consists of exploiting models that have been pre-trained on a huge amount of annotated data on another task and is known as \quotes{transfer learning}.  

In this work, we explore the idea of using a CNN model for the localization of the L3 slice using transfer learning. A full description of our approach is presented in section \ref{sec:model6}.

\section[Proposed Approach]{Proposed Approach}
\label{sec:model6}
Using a CNN for solving the L3 detection task formulated as a regression problem (see fig. \ref{fig:6-problem}) is not straightforward, and requires the alleviation of some constraints which are inherent to the medical domain and to the data that is being processed
\begin{inparaenum}[(i)]
\item Training a CNN on 3D data such as CT scans requires very large computing and memory resources that can even exceed the memory limit of most accelerator cards, while such cards are essential for learning a CNN in a reasonable time;
\item Training a CNN requires fixed size inputs, while the size of the CT scans can vary from one exam to another because of an arbitrary field of view; 
\item Training a CNN requires a large amount of labeled data.
\end{inparaenum}

\begin{figure}[!htbp]
\centering
   \input{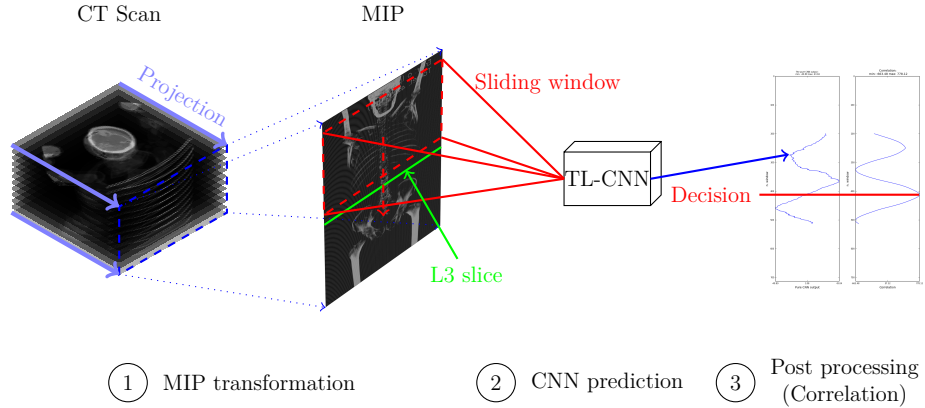}
   \caption[\bel{System overview.}]{System overview describing the three important stage of our approach : MIP transformation, TL-CNN prediction, and post processing.}
   \label{fig:6-decision}
\end{figure}

In this paper, we propose to overcome these limitations by using the approach depicted in figure \ref{fig:6-decision}. In this approach, the CT scan is first converted into another representation using Maximum Intensity Projection (MIP), in order to reduce the dimension of the input from 3D to 2D, without loss of important information. Then, the MIP image is processed in a sliding window fashion to be fed to a CNN with a fixed-size input. This CNN is trained with Transfer Learning (TL-CNN) to solve the requirement of a large amount of labeled examples. Once the trained TL-CNN has computed its prediction for each position of a sliding window, the resulting prediction sequence is processed in order to estimate the final L3 position in the full CT scan. The following subsections detail the three important contributions of the proposed system.  

\subsection{MIP Transformation}
\label{sub:MIP6}
Ideally, one can use the raw 3D scan image to feed the CNN. If $N$ is the number of slices of the arbitrary field of view CT scan, the input size is $512^2\times N$. For example, a CT-scan with 1000 slices represents 262M inputs. However, the input size of CNN models strongly impacts their number of parameters. Therefore it would require a very large number of training samples to efficiently learn the CNN. Thus, in the case of few training samples, using the 3D scan directly as an input is not efficient. We believe that the patient's skeleton carries enough visual information in order to detect the L3.

For these reasons, we propose to use a different data representation which
focuses on the patient's skeleton and dramatically reduces the size of the input space. This representation is based on a frontal Maximum Intensity Projection (MIP) \cite{wallis89, Wallis91, wallis92}. The idea is to project a line from a frontal view of the
CT scan and retain the maximum intensity over all the voxels that fall into that line. We experimented using different views such as frontal and lateral views, as well as their combination but they did not work well as compared to the frontal view alone. 

Since the slice thickness can vary within the same scan and the voxels are not squared, the projection often generates a distorted MIP. Visually, this gives an unrealistic image where the skeleton is shrunk or enlarged. The cause of this distortion is that, often, the resulting pixel from the projection does not correspond to one voxel. Often, one voxel can be represented by more than one pixel. In order to obtain an equal correspondence (i.e. one pixel corresponds to one voxel), we resize (normalize) the 2D MIP image using an estimated ratio $r$ and average slice thickness $s$ where $r$ represents the number of pixels corresponding to one voxel (slice).

\begin{figure}[!htbp]
  \centering
  \includegraphics[height=8.6cm]{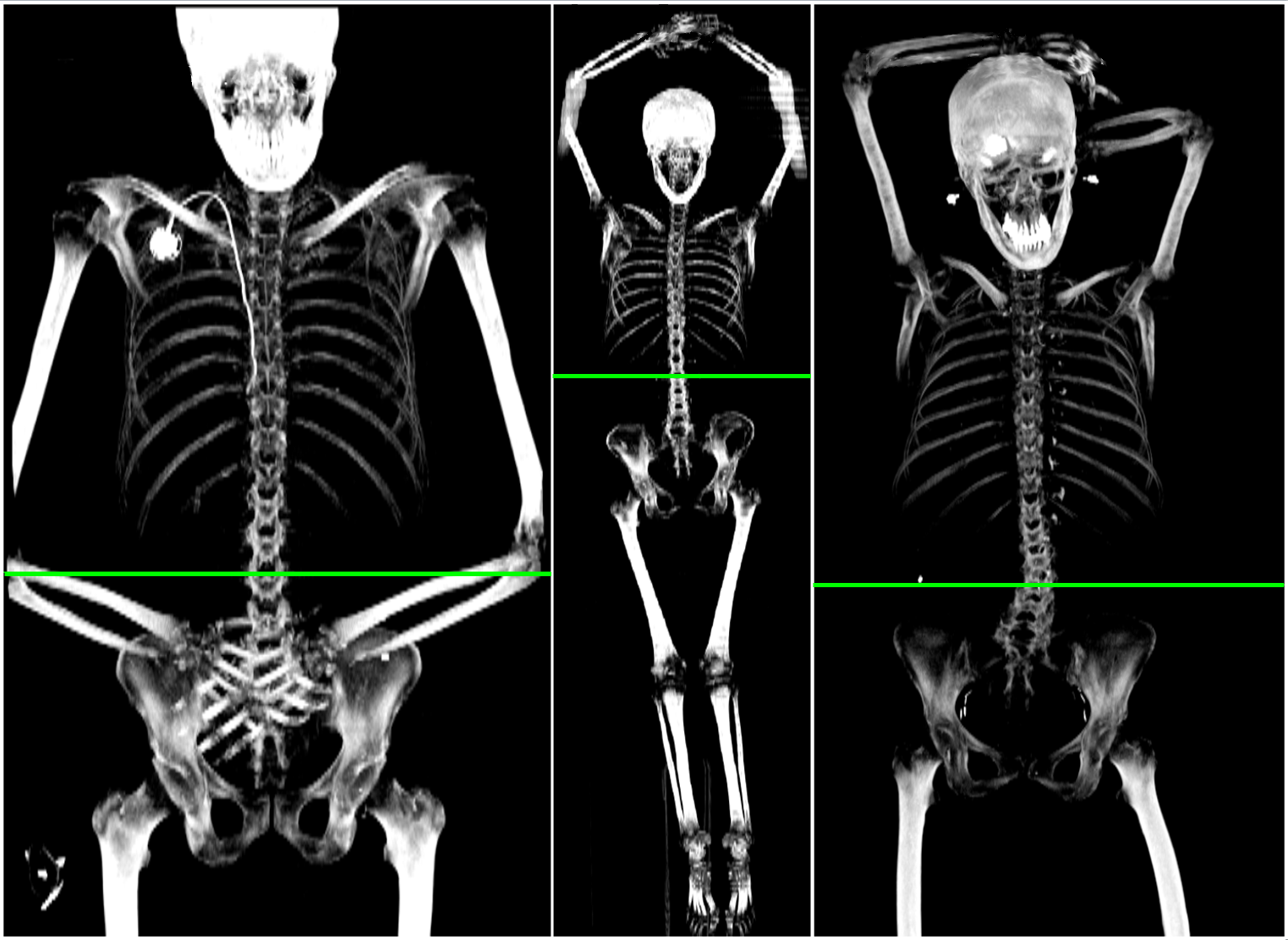}
\caption[\bel{Examples of normalized frontal MIP}]{Examples of normalized frontal MIP images with the L3 slice
    position.}
\label{fig:fig6-2}
\end{figure}

Fig.\ref{fig:fig6-2} shows an example of a normalized frontal MIP image. The MIP transformation reduces the input size from $512^2\times N$ to $512\times N$.

\FloatBarrier

\subsection{Learning the TL-CNN}
\label{sub:learning6}
Convolutional neural networks (CNN) are particular architecture of neural networks.
Their main building block is a convolution layer that performs a non-linear filtering operation.
This convolution can be viewed as a feature extractor applied identically over a plane.
The values of the convolution kernel constitute the layer parameters. 
Several convolution layers can be stacked to extract hierarchical features, where each layer builds a set of features from the previous layer.
After the convolutional layers, fully connected layers can be stacked to perform the adequate task such as the classification or the regression.

In the learning phase, both parameters of convolutional layers and fully connected layers are optimized according to a loss function.
The optimization of these huge number of parameters is generally performed using stochastic gradient descent method. This process requires a very large number of training samples.

\begin{figure}[!htbp]
  \begin{center}
\includegraphics[scale=0.5]{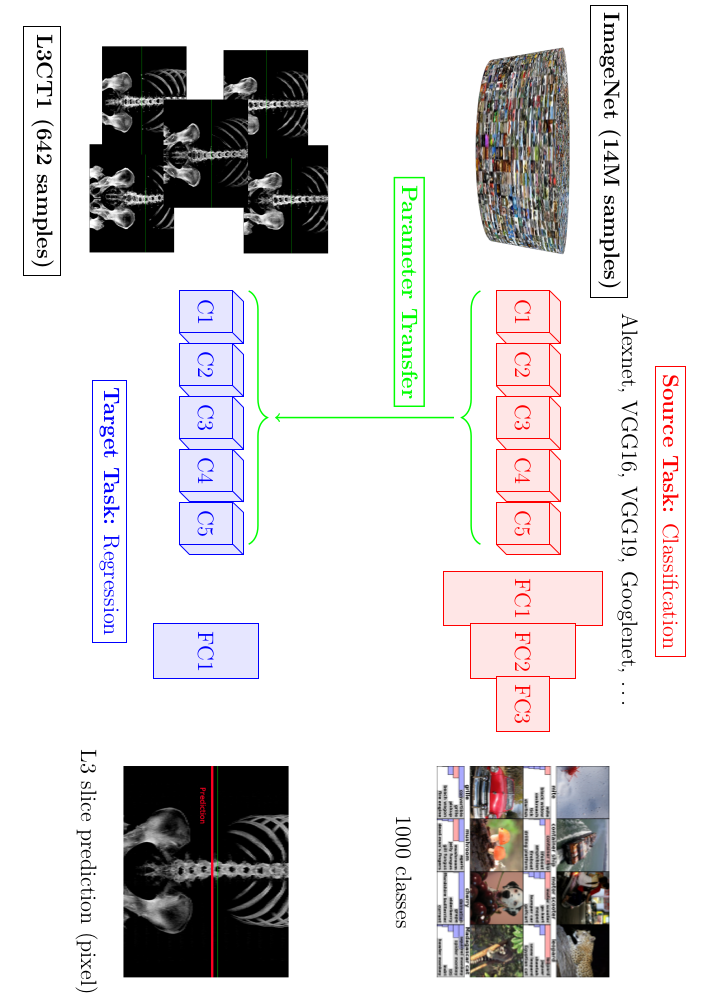}
\end{center}
	\caption[\bel{System overview.}]{System overview. Layers $C_i$ are Convolutionnal layers, while $FC_i$ denote Full Connected layers. Convolution parameters of previously learnt ImageNet classifier are used as initial values of corresponding L3 regressor layers to overcome the lack of CT examples.}
	\label{fig:6-overview}
\end{figure}

Recently, there has been a growing interest in the exploration of transfer learning methods to overcome the lack of training data.
Transfer learning consists in adapting models, trained for different task, to the task in hand (target).
It has been applied with success for various applications such as character recognition \cite{Jiang2015,Ciresan2012}, signature identification \cite{Hafemann2016} or medical imaging \cite{chest-transfer,lung-transfer}.
All these contributions exploit CNN architectures which have been pre-trained on computer vision problems, where huge labeled datasets exist.
In this framework, the weights of the convolutional layers are initialized with the weights of a pre-trained CNN on another dataset, and then fine-tuned to fit the target application.
The fine-tuning starts by transferring only the weights of the convolutional layers from a pre-trained network to the target network.
Then, randomly initialized fully connected layers are stacked over the pre-trained convolutional layers and the optimization process is performed on the whole network.
This transfer learning framework carried out for our application is illustrated by Figure \ref{fig:6-overview} .

A well-known difficulty when using the transfer learning paradigm is to fit the data to the input size of the pre-trained architecture. Since the size of the normalized MIP images varies from one patient to another, two solutions can be considered.
The first one consists of resizing the whole scan to a given fixed size.
This solution is straightforward but it dramatically impacts the image quality and the output precision.
The second solution consists in decomposing the input MIP into a set of fixed-size windows with a sampling strategy.
In this paper, we adopt the second approach which enables to preserve the initial quality of the image data.

\FloatBarrier

When sampling windows from the MIP image, two sets of window images can be produced.
The first one is made of windows containing the L3, and the other one is made of windows without the L3.
This raises the question whether the windows without L3 should be present or not in the CNN learning dataset.
As we propose a regression approach, adding the non-L3 images in the learning dataset would imply that the CNN learns (and outputs in the decision stage) the offset of the L3 with respect to the current window.
Obviously, this offset can be very difficult to learn, particularly if the current window is far from the L3 position.
Thus, we have decided to include only the windows containing the L3 in the learning dataset.

Thus, for building the training dataset, we sample all the possible windows of height $H$ such that the L3 position is in the support $[-a, +a]$ where 0 denotes the center of the window.
This leads to $2a+1$ possible windows from each MIP image to be included in the training set.
All windows from all MIP are then shuffled: it is highly improbable that two neighboring windows from the same MIP will appear next to each other in the optimization procedure. 

\subsection{Decision Process using a Sliding Window over the MIP Images}
\label{subsec:dec6}

\begin{figure}[!htbp]
	\centering
    \input{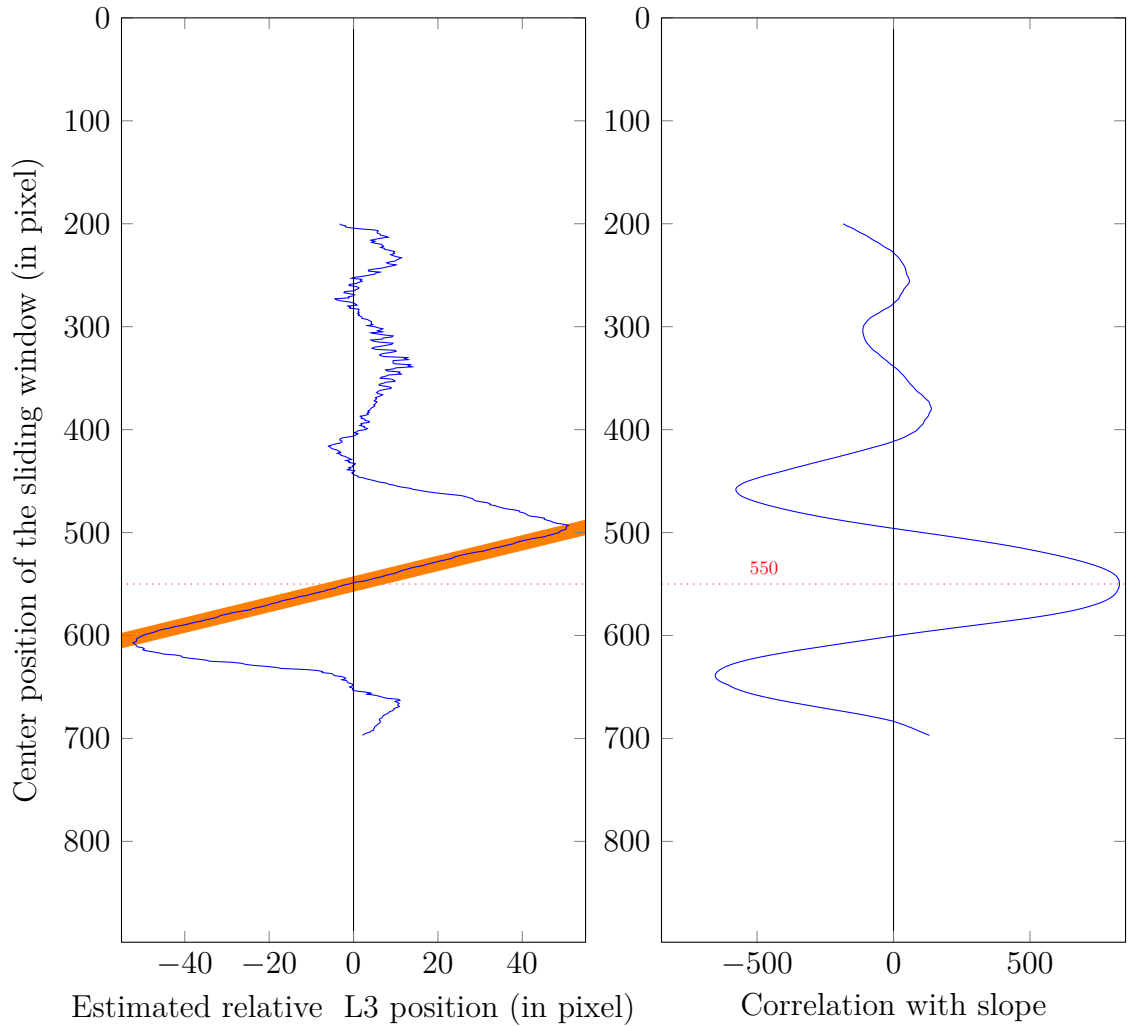}
    \caption[\bel{CNN output and post-processing.}]{[left]: CNN output sequence obtained for $H=400$ and $a=50$ on a test CT scan. The sequence contains the typical straight line of slope $-1$ centered on the L3 (the theoretical line is plotted in orange), surrounded by random values. [right]: correlation between the CNN output sequence and the theoretical slope. We retain the maximum of correlation as an estimation of the L3 position.}
	\label{fig:6-slidingoutput}
\end{figure}

A sliding window procedure is applied at the decision phase on the entire MIP image, leading to a sequence of relative L3 position predictions.
Such a sequence is illustrated in the left of figure \ref{fig:6-slidingoutput}.

In this sequence, one can observe two distinct behaviors depending on the presence of the L3 in the corresponding window:
\begin{inparaenum}[i)]
\item If the L3 is not in the window, the CNN tends to output random values since it has been trained only on images containing L3.
This behavior is illustrated in Figure \ref{fig:6-slidingoutput} at the beginning and (less clearly) at the end of the sequence. 
\item If the L3 is within the window, the CNN is expected to predict (correctly) the relative L3 position within the window.
Since the L3 position is fixed in the MIP and the window slides line by line on the region of interest, the true relative L3 position should decrease one by one.
In consequence, the CNN output should evolve linearly along the sequence of windows, leading to a noisy straight line with a slope of $-1$\footnote{\fromont{$\frac{\Delta y}{\Delta x} = -1$ is the slope of the line where $\Delta x$ is the moved distance (slided distance caused by moving the window down which is always positive). If we move the window from line $x_1$ to line $x_2 = x_1 + s$ where $s$ is the stride (i.e., how many lines we move the window down). $y$ is the relative prediction inside the window. If the network predicts $y_1$ at the window sampled at $x_1$, therefore, we expected that when we slide the window down by $s$ lines, the relative prediction should move by $-s$. Therefore, $y_2 = y_1 - s$ which means that $\Delta y= -s$. Therefore, we find that $\frac{\Delta y}{\Delta x} = \frac{-s}{s} = -1$.}}.
The noise may come from local imprecision or error on an individual slide.
This behavior can be observed in figure \ref{fig:6-slidingoutput} between offset $500$ and $600$, and it is highlighted with a theoretical orange line.
\end{inparaenum}

Therefore, at decision stage, the L3 position can be estimated through the localization of the middle of this particular straight segment.
This estimation can easily be achieved by searching the maximum of a simple correlation between the sequence and the expected slope.
This procedure, illustrated at the bottom of Fig. \ref{fig:6-slidingoutput}, easily filters out boundary windows which do not contain the L3, and shows robustness by averaging several predictions of the CNN.

\section[Experimental Protocol]{Experimental Protocol}
\label{sec:exps6}

\subsection{CT Exams Database Description}
\label{sub:ctdataset6}

In order to validate the proposed approach, a database named L3CT1 has been collected\footnote{This dataset is available on demand, please contact the corresponding author}.  The main part of the dataset is composed of 642 CT exams from different patients. All patients were included in this study after being informed of the possible use of their images in a retrospective research. The institutional ethical board of the Rouen Henri Becquerel Center approved this study \footnote{IRB Number 1604B.}. The CT exams show a high heterogeneity of patients in terms of anatomy, sex, cancer pathologies, position and properties of the reconstructed CT images: $4$ scanner models (PET/CT modalities) and $2$ manufacturer, acquisition protocols (low dose acquisition ($100$ to $120$ kV) and modulated mAs along the body) axial field of view (FOV) ($400$ to $500$ mm), reconstruction algorithms (Filtered Back  Projection (FBP) or iterative reconstruction) and slice thickness ($2$ to $5$ mm).

On each CT scan, the L3 slice was located by an expert radiologist on a dedicated software \cite{Lanic2014}, providing the annotation for the position of the L3 through its distance in (mm) from the first slice in the scan (top). 

Moreover, 43 supplementary CT scans have been annotated by the same radiologist and 3 other experts, in order to evaluate the variability of annotations among experts.

To be as reproducible and precise as possible, detailed guidelines were given to all radiologists for annotation.

From all the scans, frontal MIP images have been computed using the process described in \ref{sub:MIP6}. This results in a set of 642 images of constant width (512 pixels) and variable height, varying from 659 to 1862 pixels. Fig \ref{fig:fig6-2} shows some examples of frontal MIP images extracted from three patients of the L3CT1 database.

\subsection{Datasets Preparation}
\label{sub:datasetprep6}

The first step consists in splitting the dataset into 5 folds, in order to allow a cross-validation procedure. The split is applied at the patient level, in order to prevent that a given CT-scan provides windows in different sets (learning, validation, test), what should lead to biased results. Moreover, due to variable slice thickness in the dataset, we make sure when dividing the dataset to obtain stratified folds. Thus, we end up with the same number of samples from each slice thickness in each set. 

Once the MIP images folds have been generated, learning, validation and test windows are sampled as explained in section \ref{subsec:dec6}, where the value of $a$ has been experimentally set to $a=50$ using a cross validation procedure. 
For the validation set, in order to speed up the training, we take only 300 random windows from different patients. 

\subsection{Neural Networks Models}
\label{sub:NNM6}

In order to conduct our experiments, two types of convolutional neural
networks have been compared:
\begin{itemize}
\item \textbf{Homemade CNN (CNN4):} We have designed and trained a CNN from scratch, with specific architecture of four convolutional layers followed by a fully connected output layer. In each convolution layer, a horizontal max-pooling is performed. We found in practice that vertical max-pooling distorts the target position. The number of kernels that we used in the four convolution layers are $[10, 3, 3, 5]$, with respective sizes $[5, 7, 9, 3]$. The hyper-parameters of our CNN were tuned on the validation set \cite{bengio12}. We refer to our model as $CNN4$.

\item \textbf{Pre-trained CNNs:} In our study, we have collected a set of pre-trained convolutional neural networks over ImageNet dataset  \cite{imagenetcvpr09}: Alexnet \cite{krizhevsky12}, VGG16 \cite{SimonyanZ14aCORR}, VGG19 \cite{SimonyanZ14aCORR}, Googlenet (Inception V1)  \cite{szegedy15}\footnote{The weights of Googlenet were obtained from:   \url{https://gist.github.com/joelouismarino/a2ede9ab3928f999575423b9887abd14}, and the weights of the rest of the models were obtained from  \url{https://github.com/heuritech/convnets-keras}}. The models are created using the library Keras \cite{chollet2015keras}. For each model, we keep only the convolutional layers which are considered as shared perception layers that may be used for different tasks. On top of that, we add one fully connected layer to be specialized in our specific task (i.e. L3 detection). Our experiments have shown that adding more fully connected layers does not improve the results.

The input of pre-trained models is supposed to be an RGB image (i.e. a 3D matrix), while in the other hand, our sampled windows are 2D matrix. In order to match the required input, we duplicate the 2D matrix in each color channel. Then, each channel is normalized using its mean from the ImageNet Dataset.
    
We use $L_2$ regularization for training all the models with value of $\lambda =
10^{-3}$, except for Googlenet where we used the original regularization values.

\end{itemize}

\section[Results]{Results}
\label{sec:results6}
\subsection{Data View: Frontal Vs. Lateral}
\label{sub:dataview6}

The use of the MIP representation allows us to access to different views of the CT scan, such as the frontal and lateral views (other views with different angles are possible). In order to choose the best view, we re-train a VGG16 model with one fully connected layer using different input views. We recall that the input of the VGG16 is an image with 3 plans. We experimented three configurations. In the first and second cases, we repeat the frontal and lateral views, respectively, in the three input channels. In the last case, we mixed the frontal and the lateral view. The motivation behind the combination of the views is that each view will provide an additional information (hopefully complementary) that will help the model to decide. The sampling margin of the windows is done over the range $[-50, +50]$. Tab.\ref{tab:tab6-000} shows that using frontal view alone is more suitable. One possible explanation of this results is that the frontal view contains more structural context (ribs, pelvis) which helps to locate the L3 slice, in the opposite of the lateral view. Combining lateral and frontal views gave better results than lateral alone but worse than frontal alone. One may think that lateral view adds noise to the frontal view.

\begin{table}[!htbp]
  \centering
  {
	\begin{tabular}{|>{\color{black}}l|>{\color{black}}c|}
		\hline
		View&VGG16\\
                \hline
                &Error $m_c$ (slices)\\
                \hline
                Frontal &$\bm{1.71\pm1.59}$\\
                \hline
                Lateral &$4.29\pm14.90$\\
                \hline
                Frontal Lateral Frontal &$1.89\pm2.05$\\
                \hline
	\end{tabular}
    }
	\caption[\bel{Test error.}]{Test error (mean $\pm$ standard deviation) over the test set of fold 0, expressed in slices, using VGG16 model with frontal and lateral views.}
	\label{tab:tab6-000}
\end{table}

\subsection{Detection Performance}
\label{sub:dectectionperg6}

All the models described in section \ref{sub:NNM6} have been evaluated in a cross validation procedure on the L3CT1 dataset by computing the prediction error. The prediction error for one CT scan is computed as the absolute difference between the prediction $y_{pred}$ and the target $y$: $ e = |y - y_{pred}| $. The error is expressed in slices. We report the mean and the standard deviation of the test error $(\mu_e, \sigma_e)$, respectively in the form $\mu_e \pm \sigma_e$, over the entire test set. Obtained results are reported in Tab.\ref{tab:tab6-15-4}.

For the sake of comparison, we used Random Forest Regression (RF) \cite{breiman2001, ho1995} as a regressor instead of our CNN. As in most pattern recognition problems, we need to extract input features to train our Random Forest Regression. Local Binary Patterns (LBP) features have shown to be very efficient in many computer vision tasks \cite{ojala2002}, especially in medical imaging \cite{nanni2010}. Therefore, we have retained this feature descriptor. To extract the LBP features we used a number of neighbors of $8$ and a radius of $3$ which creates an input feature vector with dimension of $2^8=256$. From each sampled window, we extract LBP features. We investigated different number of trees: $10$, $100$ and $500$. The obtained results showed that random forests do not perform well over this task. We report in Tab.\ref{tab:tab6-15-4} the results using $500$ ($RF500$) trees which are in the same order of performance compared the other cases (i.e. $10$ and $100$ trees).

\begin{table}[!htbp]
    \centering
\resizebox{\linewidth}{!}{%
\tabcolsep=3pt
    \begin{tabular}{|l||>{\color{black}}c||c|c|c|c|c|}
		\hline
        &RF500& CNN4 & Alexnet & VGG16 & VGG19 & Googlenet \\        
                \hline
                \hline
                fold 0&
                $7.31\pm6.52$&
                $2.85\pm2.37$&
                $2.21\pm2.11$&
                $2.06\pm4.39$&
                $1.89\pm1.77$&
                $1.81\pm1.74$ \\
                \hline
                fold 1&
                $11.07\pm11.42$&
                $3.12\pm2.90$&
                $2.44\pm2.41$&
                $1.78\pm2.09$&
                $1.96\pm2.10$&
                $3.84\pm12.86$ \\
                \hline
                fold 2&
                $13.10\pm13.90$&
                $3.12\pm3.20$&
                $2.47\pm2.38$&
                $1.54\pm1.54$&
                $1.65\pm1.73$&
                $2.62\pm2.52$\\
                \hline
                fold 3&
                $12.03\pm14.34$&
                $2.98\pm2.38$&
                $2.42\pm2.23$&
                $1.96\pm1.62$&
                $1.76\pm1.75$&
                $2.22\pm1.79$ \\
                \hline
                fold 4&
                $8.99\pm7.83$&
                $1.87\pm1.58$&
                $2.69\pm2.41$&
                $1.74\pm1.96$&
                $1.90\pm1.83$&
                $2.20\pm2.20$ \\
                \hline
                \hline
                Average&
                $10.50\pm10.80$&
                $2.78\pm 2.48$&
                $2.45\pm 2.42$&
                $\bm{1.82\pm2.32}$&
                $1.83\pm 1.83$&
                $2.54\pm 4.22$ \\
                \hline
	\end{tabular}
}
	\caption[\bel{Error expressed in slice over all the folds.}]{Error expressed in slice over all the folds using different models: RF500, CNN4 (Homemade model), and Alexnet/VGG16/VGG19/GoogleNet (Pre-trained models).}
	\label{tab:tab6-15-4}
\end{table}
\FloatBarrier

From Tab.\ref{tab:tab6-15-4}, one can see that pre-trained models perform
 better than our homemade CNN4 with an improvement of about $35\%$\footnote{\fromont{${(2.78-1.82)/2.78 \approx 0.3453 \approx 35\%}$.}}. In particular, VGG16 showed the best results by an average error of $1.82 \pm 2.32$ followed by VGG19 with $1.83 \pm 1.83$. This result confirms the strong benefit of transfer learning between two different tasks. Moreover, it shows that the convolutional layers can be shared as a perception tool between different tasks with slight adaptation. On the other hand, this illustrates the capability for modeling such task using the pre-trained models. 

\subsection{Processing Time Issues}
\label{sub:processingtime6}

One must mention that the price we paid in order to reach the performance mentioned above is to increase the complexity of the model. In Table \ref{tab:tab6-15-03}, we present the number of parameters of each model and the average required time for the prediction of the L3 slice. We observe that VGG16 contains approximately 264 times more parameters than CNN4.  Beside the required memory for such models, the real paid cost is the evaluation time during the test phase. Computed on a GPU (Tesla K40), VGG16 requires an average of $13.28$ seconds per CT scan while our CNN4 only needs $4.46$ second per CT scan.

\begin{table}[!htbp]
  \centering
    {
    \resizebox{0.98\columnwidth}{!}{%
	\begin{tabular}{|l|>{\color{black}}r|c|}
		\hline
		&Number of parameters& Average forward pass time (seconds/CT scan)\\
                \hline
                CNN4&55,806&$04.46$\\
                \hline
                Alexnet&2,343,297&$06.37$\\
                \hline
                VGG16&14,739,777&$13.28$\\
                \hline
                VGG19&20,049,473&$16.02$\\
                \hline
                Googlenet&6,112,051&$17.75$\\
                \hline
	\end{tabular}
  }
        }
	\caption[\bel{Number of parameters vs. processing time.}]{Number of parameters for different models and average
          forward pass time per CT scan.}
	\label{tab:tab6-15-03}
\end{table}

An important factor which affects the evaluation time in these experiments is the number of windows processed by the CNN for a given CT scan. Thus, it is possible to dramatically reduce the computation time by shifting the window by a bigger value than 1 pixel. An experimental evaluation of this strategy with VGG16 has shown that a good compromise between processing time and performance could be obtained for a shift value up to $6$ pixels without affecting the localization precision. This sub-sampling reduces the evaluation time from $13.28$ seconds/CT scan to $2.36$ seconds/CT scan and moved the average localization error from $1.82 \pm 2.32$ slices to $1.91 \pm 2.69$ slices, respectively. This shows the robustness of the proposed correlation post-processing.

\subsection{Comparison with Radiologists}
\label{sub:compareradiologists6}

In order to further assess the performance of the proposed approach, an extra set of 43 CT scans was used for test. This particular dataset was annotated by the same radiologist who annotated L3CT1 dataset and also by three other experts. Each annotation was performed at two different times, in order to evaluate the intra-annotator variability. We refer to both annotations by the same expert by \textit{Review 1} and \textit{Review 2}.

Obtained results are illustrated in Tab.\ref{tab:tab6-16}. It compares the error made by CNN models with those made by the radiologists, using the radiologist who annotated the L3CT1 dataset as reference. These results corroborate the results provided in Table \ref{tab:tab6-15-4} since VGG16 is better than CNN4 with an improvement of about $35\%$ in average for both reviews. The results also demonstrate that radiologists are in average more precise than automatic models with an improvement of about $50\%$. However, they also show that there exists some variabilities among radiologist annotations and even an intra-annotator variability. This latter is visible in Tab. \ref{tab:tab6-16} since computed errors for automatic systems vary between both reviews while the automatic system gives the same output, showing that reference values have changed. This illustrates the difficulty of the task of precisely locating the L3 slice and the interest of CNN which does not change its prediction.

\begin{table}[!htbp]
  \centering
  {
    \resizebox{\linewidth}{!}{%
	\begin{tabular}{|l||c|c|c|c|c|}
		\hline
		Errors (slices) /
                operator&CNN4&VGG16&Ragiologist
                $\#1$&Radiologist $\#2$ &Radiologist $\#3$\\
                \hline
                Review1 &$2.37 \pm 2.30$&$1.70 \pm 1.65$&$0.81 \pm 0.97$&$0.72 \pm 1.51$&$0.51 \pm 0.62$\\
                \hline
                Review2 &$2.53 \pm 2.27$&$1.58 \pm 1.83$& $0.77 \pm 0.68$&$0.95 \pm 1.61$&$0.86 \pm 1.30$\\
                \hline
	\end{tabular}
    }
    }
	\caption[\bel{Comparison of the performance.}]{Comparison of the performance of both the automatic systems and radiologists. The L3 annotations given by the reference radiologist vary between the two reviews.}
	\label{tab:tab6-16}
\end{table}

\FloatBarrier

\section[Conclusion]{Conclusion}
\label{sec:conclusion6}

In this paper, we proposed a new and generic pipeline for spotting a particular slice in a CT scan. In our work, we applied our approach to the L3 slice, but it can easily be generalized to other slices, provided a labeled dataset is available.    

First, the CT scan is converted into a frontal Maximum Intensity Projection (MIP) image. Afterwards, this representation is processed in a sliding window fashion to be fed to a CNN which is trained using Transfer Learning. In the test phase, all the predictions concerning the position of the L3 within the sliding windows are merged into a robust post-processing stage to take the final decision about the position of the L3 slice in the full CT scan.

Obtained results show that the approach is efficient to precisely detect the target slice. Using a fine-tuned VGG16 network coupled with an adequate decision strategy, the average error is under 2 slices where experienced radiologists can provide annotations that differ of about 1 slice. The computing time is within an acceptable range for clinical applications, and can be further reduced by (i) increasing the shift value (ii) adapting the network architecture by pre-training smaller networks over ImageNet, for example, which has not been studied in this work (iii) and \fromont{prune} the final trained CNN by dropping the less important filters. Recently, pruning CNNs \cite{li2016pruning, molchanov2016pruning, park2016faster} has seen a lot of attention in order to deploy large CNNs on devices with less computation resource. We are currently working on this idea to speedup more the computation.

This contribution confirms the interest of using machine learning and more particularly deep learning in medical problems. One of the main reasons deep learning is not popular in medical domain is the lack of training data. Pre-training the networks over other large dataset will strongly alleviate this problem and encourage the use of such efficient models.

\section*{Acknowledgement}
This work has been partly supported by the grant ANR-11-JS02-010 LeMon and the grant ANR-16-CE23-0006 \quotes{Deep in France}.

%
%
%

\begin{generalconclusion}
\markboth{General Conclusion and Perspectives}{}
\addcontentsline{toc}{chapter}{General Conclusion and Perspectives}
\fromont{In this thesis, we have tackled the overfitting issue in neural network models particularly in learning scenarios where only few labeled data are available. Regularization is the most common used approach to deal with such issue. In the literature, different regularization methods have been proposed. While each regularization method tackles the overfitting issue differently, we can distinguish a class of methods that uses representation learning as a fundamental mechanism such as dropout, sparse representations, unsupervised/semi-supervised learning, tangent propagation and manifold learning. Following the success of such methods, we address in this thesis the overfitting of neural networks by proposing three different regularization methods based on representation learning paradigm, where each method is adapted to the task in hand. Such methods were designed and validated mainly in learning scenarios where only few labeled data are available which is a challenging task since most successful neural networks are trained using very large number of samples that reaches easily millions. In most real life applications, such large number of samples is not available. However, one still wants to exploit the performance of such models.

In the following, we present a brief description of each of our proposals along with their respective perspectives.}

\bigskip
\begin{description}
\item[\fromont{1. Unsupervised learning for structured output problems}] \hfill \\
In the first contribution, we addressed structured output problems, i.e., a mapping $\X \to \Y$ where the output is multidimensional and where there are some relations among its components. In this contribution, we proposed \fromont{ to use the unsupervised learning paradigm to learn/discover the hidden structure in the output data. To do so, we proposed  a multi-task framework which is composed of} a supervised task and two unsupervised tasks; the first unsupervised task learns the input distribution data while the second one learns the output distribution. Explicit incorporating learning the output data structure into the network learning, which is rarely used, has shown to speedup its training, and more importantly, to improve its generalization. Moreover, learning could be achieved in an unsupervised way where the network discovers on its own the underlying structure that may help to perform accurate prediction. Therefore, no need to a supervised intervention to specify what type of relations the network should learn. Furthermore, it allows using unlabeled data and labels only data which showed to help further more improving the generalization error. The application of our framework to facial landmark detection problem \fromont{showed} a speedup of neural networks training and an improvement of their generalization performance.


Although our framework has shown improvements, it still can be improved. For instance, the \fromont{adaptation} scheme of the importance weights of the three tasks can be achieved differently. Instead of \fromont{adapting} them in terms of training epochs, we can consider an automatic scheme that uses different indicators based on the training and  validation errors. Furthermore, one may consider another type of models to learn the output distribution instead of simple auto-encoders. Generative models, such as generative auto-encoders and adversarial auto-encoders \cite{MakhzaniSJG15CORR}, seem a good choice to start with. This will lead to probabilistic outputs in \fromont{ the case where} multiple decisions are required.

\item[2. \fromont{The use of prior knowledge for classification}] \hfill \\
In the second contribution, neural networks regularization is achieved through the use of prior knowledge about \fromont{the internal representations of the network for a classification task}. Prior knowledge can be helpful, \fromont{in terms of generalization, when dealing with few training data. Since the prior explains a decision rule that helps in generalization, it can guide the learning process to choose a better solution to avoid overfitting.} More precisely, to deal with a classification task, we have proposed to integrate the following prior knowledge about the internal representation within  a neural network: \quotes{samples within the same class should have the same internal representation}. In this contribution, our suggestion consists in formulating this prior knowledge as a penalty which is added to the supervised cost in order to be minimized. The proposed penalty constrains the hidden representations to be class-wise invariant. Empirical evidence has showed that incorporating such prior knowledge helps in improving the network generalization when trained with few samples.

Moreover, this work has shown that such class-wise invariance is learned with the increase of depth, i.e., the more the network is deep the more the internal representation is class-wise invariant. However, tracking this invariance in a network did not show its improvement during the learning. This means that the backpropagation algorithm does not always learns understandable/reasonable internal representations and one can do better using prior knowledge. 
\fromont{In this work, we exploited only the invariance property of the internal representation in a neural network. Such property creates compact classes where samples within the same class are close to each other. As an extension to this work, which showed promising results, we plan to add a discrimination term that constrains the classes to be far from each other which is an important property in a classification task. For instance, we are planning to use non-linear discriminant analysis tool to learn efficient internal representations. Moreover, we plan to address the issue of multi-modality to choose which pairs are important for optimization. Many works in linear discriminant analysis showed that considering the distance between samples with a uniform importance degrades the performance \cite{Sugiyama2007, flamary17}. Giving high importance to particular pairs leads to better performance. Based on such results, we plan to use a probabilistic framework that provides dynamically a probabilistic importance to each pair. Such probabilistic framework is inspired from ideas in optimal transport \cite{peyre17}. Moreover, we plan to tackle the optimization problem when dealing with multi-task learning where we have the following setup: a main task, and a set of secondary tasks. In such setup, we usually end up with unbalanced gradients between the main task and the secondary tasks. For instance, in low layers in a neural network (the ones close to its input), the supervised (main) gradient is low compared to the secondary tasks (at the same layer). In practice, this scenario usually leads to degrading the performance of the network. We plan to use a suitable normalization of the gradients to avoid such issue.}

\item[3. \fromont{Transfer learning and medical domain}] \hfill \\
In our last contribution, we have presented a real-life application to deal with the lack of labeled data in medical domain. The idea consists in using a network with high capacity, such as convolutional networks, which was pre-trained over a different task with abundant data. Then, a subset of its parameters responsible for feature building are extracted and re-used into a new fresh network. This learning procedure falls into the transfer learning paradigm where learned knowledge from a source task is transfered to a target task. This gives more advantage to the target task to start with the parameters of a complex network partially trained. Therefore, only few data are required to adapt the network to fit the target task.

We applied this learning process to localize the third lumbar vertebra, i.e., L3, over a 3D CT scan. Instead of using the 3D CT scan as a raw input data, we used its frontal projection to obtain a 2D image. This allows reducing processing time. To locate the target, we slide our trained model over the whole CT scan, and perform a prediction over each window. This makes the prediction independent of which part of the patient's body is covered by the scan. Then, we perform a post-processing using a correlation to detect where the network spikes to localize the target. The framework was designed to be task independent, i.e., it could be used to localize any other organ.

The obtained results from this contribution showed the interest of applying transfer learning in machine learning and exploiting pre-trained deep architectures. However, as a real world application, this work raised an issue which is the running time. Running large networks such as VGG in a production software where many images are processed requires a lot of computation power which is not available in most cases, at least in the clinic. We note that such high complex models are over-parameterized with respect to our task. This motivated us to explore a solution to speed up the computation. Recently, a trend of speeding up convolutional networks, especially when using transfer learning, has raised. This trend is based on pruning a subset of the filters from the network using different strategies \citep{li2016pruning, molchanov2016pruning, park2016faster}. We developed the necessary code for pruning the models used in this work using minimal norm, i.e., filters with low norm are pruned. However, such approach has yet to be evaluated. We recommended the developers team of the clinic to pursue this path. Another possible way to speedup the computation is to reverse the procedure by including the computation power as a model selection criterion. The idea consists in finding a model with complexity that allows running it over average computation machine in acceptable time. Next, this model is pre-trained over a dataset with enough data. Then, its pre-trained filters are extracted and re-used as described above. The advantage of this approach is that we are able to control the model size which is directly involved into the required computation power.
\end{description}
\belxx{
\bigskip
{\begin{center} \Large \textbf{Perspectives} \end{center}}
\bigskip
  \fromont{We} have presented in this thesis different approaches of using inductive bias to improve the generalization of neural networks. Such inductive bias is based on representation learning paradigm that is one the main reasons of the success of neural networks. In the following, \fromont{based on the promising results, and the extensive literature study conducted in this thesis, we} present \fromont{two main} research directions that \fromont{we believe they will improve the generalization of neural networks and help us build more understandable neural networks}:
  
  \begin{description}
    
  \item[1. Prior knowledge and domain knowledge to improve deep learning] \hfill \\
  Although deep learning based on neural networks has shown impressive results in different domains, \fromont{we} believe that sooner or later, the performance of such models will reach a saturation regime. A situation where  adding more data will not improve their performance, even though such models are known to be incredibly eager for data. For now, the hype that we see in the use of deep learning is just a start. This issue is the result of neural network limitations. \fromont{From our understanding that is based on the reviewed literature and the history of neural networks, we concluded} that neural networks are not that \emph{intelligent} as \fromont{ machine learning models}. They learn through a mechanical process by repeating over and over the same process until they \emph{memorize} patterns of data. We note that other machine learning models, if not all \fromont{of them}, do the same thing. Humans seem to learn way faster and smarter. One of the main reasons is their heavy use of \emph{prior knowledge} \cite{dubey2018investigating} which allows them to learn new tasks in a short time with less effort and less data. \fromont{We} think that using generic prior knowledge in training neural networks is inevitable in order to introduce some intelligence aspects into their learning/behavior. Learning in such framework will more likely require less training samples. One can go further by using domain knowledge, and benefit from the knowledge of experts. Therefore, neural networks need to be modified in order to ease introducing any type of domain knowledge. \fromont{The main difficulty here is how one can introduce different types of priors to the network ? We suggest two possible ways to do that: either through the architecture by designing new models that take in consideration the prior such as in convolutional networks where the same feature detector is applied all over the image following the prior that the feature may appear in different positions. The other way is through learning, by using constraints such as in regularization.}
  
  Neural networks and deep learning are by far to be considered as intelligent models nor an instance of Artificial Intelligence. \fromont{We} do not think that such models are ready yet to be integrated to drive a car or take control over a robot. Such models are purely mechanical and have an extreme lack in reasoning. Up to now, we do not exactly know how a neural network takes a decision, and \fromont{it seems reasonable not to trust its decisions particularly in critical tasks.} \fromont{One can } qualify deep learning models by \emph{brute-force models}. \fromont{Unfortunately}, in such domain, the performance goes first, and the justification and the proof \fromont{do} not matter. It is an experimental-guided field.

  \item[2. Fundamental research: dictionary learning and deep learning]  \hfill \\
  During this thesis, \fromont{we found out} that representation learning \fromont{paradigm} is an important aspect in deep learning models that gives \fromont{them} a powerful capacity to learn complex \fromont{tasks}. While writing this thesis, \fromont{we} had to go back to the $40^{\prime}$s, the age of birth of neural networks. Years later, Minsky and Papert \cite{minsky69perceptrons, Minsky1988} showed the limitations of shallow neural networks, i.e., perceptrons. The main message of their criticism is that shallow networks can be able to solve complex tasks when they are able to \emph{represent} differently the input signal in hidden layers. Minsky and Papert proposed, at the time, to use perceptron-like layers to find such representations. As much as this proposition is interesting, \fromont{we} think that we can do better. There are many reasons to think so. We recall that back in the $40^{\prime}$s, the goal behind creating perceptrons and neural networks is to build machines, more precisely, intelligent machines. Therefore, such attentions had a large impact on most research directions including the architecture of such models where a clear attempt has appeared to mimic an electronic circuit to be easy to implement in real life. Hence, neural networks, old and modern version, have inherited most of their aspect from circuits. Seeing deep learning today's success, this certainly has advantages. However, \fromont{we} think that such aspect has carried with it many disadvantages as well which may be the reason behind most deep learning issues today. Using perceptron-like as an encoder in the hidden layer can have advantages in, at least, two cases:
  \begin{itemize*}
    \item Dealing with binary predicates as input: Using a perceptron-like in the hidden layer will allow to learn new predicates based on the input predicate. Such hidden representation can be built by combining different boolean inputs.
    \item Using handcrafted features: In this case, the input signal is composed of different pre-computed features. The most important aspect here is that each dimension \fromont{represents} the same feature. Therefore, combining different features does make sense in order to build a new feature.
  \end{itemize*}
  
  For long time, the research community has started using continues inputs, and raw data as input. Therefore, each component of the input signal changes from an example to another, taking as example an image as raw input. This makes perceptron-like hidden layer inefficient. 
  
  Another critical aspect in using perceptron-like as a hidden layer is information loss. A neuron is fundamentally used to take a decision. Aligning a \fromont{set} of a neurons to learn a new representation, based a continuous input, is not really optimal especially in \fromont{low} layers. It is clear that there will be a large information loss. \fromont{We} think that one of the reasons that today neural networks are deep is because there is a need to many layers to recover the lost information.
  
  In order to \fromont{alleviate} this issue, and in order to give neural networks \fromont{a more solid theoretical background}, \fromont{we} suggest to keep the idea of Minsky and Papert, i.e., to use hidden representations to represent in a different way the raw input signal. However, instead of using perceptron-like layers, \fromont{we suggest to use well studied, interpretable, and solid concepts based on approximation theory and data representation such as dictionary learning-like methods \cite{steffens2007history}}. The main idea consists in learning how to \fromont{well} represent data in the hidden layers \fromont{using flexible tools that we know and control what they do exactly}. Dictionary learning \cite{starckmurtaghfadili2015} provides a strong tool to represent a \fromont{signal} using a fixed number of atoms. In the case of a classification task, one can image a pool \fromont{of} a shared dictionary to represent all the samples independently from the class membership, followed by class-wise dictionaries, followed by a perceptron-like layer for decision step. Possible adaptation to the output can be done by \fromont{considering}  dictionary learning properties in \fromont{a} classification task. The main idea here is to find a new common space to represent all samples. This common space could be modeled using dictionaries, and could exploit their ability to be trained using unsupervised data through reconstruction. It is clear that one can build easily hierarchical representations following this scheme. However, the first issue of using dictionaries is security, since the model will explicitly memorize chunks of data.
  
  Following this approach, one may go further to exploit more solid theoretical frameworks for data representation such as tensor decomposition \cite{rabanser2017arxiv, kolda2009} in order to build more intelligent layers that are able to decompose a complex signal into elementary elements.

  \end{description}
  }

\end{generalconclusion}


\begin{spacing}{0.9}


\bibliographystyle{apalike}
\cleardoublepage
\bibliography{References/references} 

\begin{thebibliography}{}

\bibitem[Abadi et~al., 2016]{Abadi2016DLD}
Abadi, M., Chu, A., Goodfellow, I., McMahan, H.~B., Mironov, I., Talwar, K.,
  and Zhang, L. (2016).
\newblock Deep learning with differential privacy.
\newblock In {\em Proceedings of the 2016 ACM SIGSAC Conference on Computer and
  Communications Security}, CCS '16, pages 308--318, New York, NY, USA. ACM.

\bibitem[Abu{-}Mostafa, 1990]{abuMostafa90}
Abu{-}Mostafa, Y.~S. (1990).
\newblock Learning from hints in neural networks.
\newblock {\em Journal of Complexity}, 6(2):192--198.

\bibitem[Abu{-}Mostafa, 1992]{abuMostafab92}
Abu{-}Mostafa, Y.~S. (1992).
\newblock A method for learning from hints.
\newblock In {\em Advances in Neural Information Processing Systems 5, {[NIPS}
  Conference, Denver, Colorado, USA, November 30 - December 3, 1992]}, pages
  73--80.

\bibitem[Abu{-}Mostafa, 1993]{abuMostafa93}
Abu{-}Mostafa, Y.~S. (1993).
\newblock Hints and the vc dimension.
\newblock {\em Neural Computation}, 5(2):278--288.

\bibitem[Abu-Mostafa et~al., 2012]{AbuMostafa2012MLBook}
Abu-Mostafa, Y.~S., Magdon-Ismail, M., and Lin, H.-T. (2012).
\newblock {\em Learning From Data}.
\newblock AMLBook.

\bibitem[Ackley et~al., 1985]{AckleyCognScie1985}
Ackley, D.~H., Hinton, G.~E., and Sejnowski, T.~J. (1985).
\newblock A learning algorithm for {B}oltzmann machines.
\newblock {\em Cognitive Science}, 9:147--169.

\bibitem[Aggarwal, 2014]{Aggarwal2014DCA}
Aggarwal, C.~C. (2014).
\newblock {\em Data Classification: Algorithms and Applications}.
\newblock Chapman \& Hall/CRC, 1st edition.

\bibitem[Aizenberg et~al., 2000]{Aizenberg2000}
Aizenberg, I.~N., Aizenberg, N.~N., and Vandewalle, J.~P. (2000).
\newblock {\em Multi-Valued and Universal Binary Neurons: Theory, Learning and
  Applications}.
\newblock Kluwer Academic Publishers, Norwell, MA, USA.

\bibitem[Anastassiou, 2016]{Anastassiou2016}
Anastassiou, G.~A. (2016).
\newblock {\em Intelligent Systems II: Complete Approximation by Neural Network
  Operators}.
\newblock Springer Publishing Company, Incorporated, 1st edition.

\bibitem[Anastassiou and Duman, 2016]{anastassiou2016intelligent}
Anastassiou, G.~A. and Duman, O. (2016).
\newblock {\em Intelligent Mathematics II: Applied Mathematics and
  Approximation Theory}, volume 441.
\newblock Springer.

\bibitem[Anastassiou and Gal, 2002]{anastassiou2002approximation}
Anastassiou, G.~A. and Gal, S.~G. (2002).
\newblock Approximation theory. moduli of continuity and global smoothness
  preservation.

\bibitem[Andersen, 2018]{Andersen2018corr}
Andersen, P. (2018).
\newblock Deep reinforcement learning using capsules in advanced game
  environments.
\newblock {\em CoRR}, abs/1801.09597.

\bibitem[Argyriou et~al., 2006]{ArgyriouEP06nips}
Argyriou, A., Evgeniou, T., and Pontil, M. (2006).
\newblock Multi-task feature learning.
\newblock In {\em Advances in Neural Information Processing Systems 19,
  Proceedings of the Twentieth Annual Conference on Neural Information
  Processing Systems, Vancouver, British Columbia, Canada, December 4-7, 2006},
  pages 41--48.

\bibitem[Argyriou et~al., 2007]{ArgyriouMPY07nips}
Argyriou, A., Micchelli, C.~A., Pontil, M., and Ying, Y. (2007).
\newblock A spectral regularization framework for multi-task structure
  learning.
\newblock In {\em Advances in Neural Information Processing Systems 20,
  Proceedings of the Twenty-First Annual Conference on Neural Information
  Processing Systems, Vancouver, British Columbia, Canada, December 3-6, 2007},
  pages 25--32.

\bibitem[Arpit et~al., 2017]{ArpitJBKBKMFCBL17ICML}
Arpit, D., Jastrzebski, S.~K., Ballas, N., Krueger, D., Bengio, E., Kanwal,
  M.~S., Maharaj, T., Fischer, A., Courville, A.~C., Bengio, Y., and
  Lacoste{-}Julien, S. (2017).
\newblock A closer look at memorization in deep networks.
\newblock In {\em {ICML}}, volume~70 of {\em Proceedings of Machine Learning
  Research}, pages 233--242. {PMLR}.

\bibitem[Arulkumaran et~al., 2017]{Arulkumaran2017corr}
Arulkumaran, K., Deisenroth, M.~P., Brundage, M., and Bharath, A.~A. (2017).
\newblock Deep reinforcement learning: A brief survey.
\newblock {\em IEEE Signal Processing Magazine}, 34(6):26--38.

\bibitem[Atkins et~al., 2014]{Atkins2014}
Atkins, J.~L., Whincup, P.~H., Morris, R.~W., Lennon, L.~T., Papacosta, O., and
  Wannamethee, S.~G. (2014).
\newblock {Sarcopenic obesity and risk of cardiovascular disease and mortality:
  a population-based cohort study of older men.}
\newblock {\em Journal of the American Geriatrics Society}, 62(2):253--60.

\bibitem[Auli et~al., 2013]{auliGQZ13}
Auli, M., Galley, M., Quirk, C., and Zweig, G. (2013).
\newblock Joint language and translation modeling with recurrent neural
  networks.
\newblock In {\em Proceedings of the 2013 Conference on Empirical Methods in
  Natural Language Processing, {EMNLP} 2013, 18-21 October 2013, Grand Hyatt
  Seattle, Seattle, Washington, USA, {A} meeting of SIGDAT, a Special Interest
  Group of the {ACL}}, pages 1044--1054.

\bibitem[Ba and Caruana, 2014]{Ba2014NIPS}
Ba, L.~J. and Caruana, R. (2014).
\newblock Do deep nets really need to be deep?
\newblock In {\em Proceedings of the 27th International Conference on Neural
  Information Processing Systems - Volume 2}, NIPS'14, pages 2654--2662,
  Cambridge, MA, USA. MIT Press.

\bibitem[Bahdanau et~al., 2014]{BahdanauCB14CORR}
Bahdanau, D., Cho, K., and Bengio, Y. (2014).
\newblock Neural machine translation by jointly learning to align and
  translate.
\newblock {\em CoRR}, abs/1409.0473.

\bibitem[Baird, 1990]{Baird90}
Baird, H. (1990).
\newblock Document image defect models.
\newblock In {\em Proceddings, IAPR Workshop on Syntactic and Structural
  Pattern Recognition}, Murray Hill, NJ.

\bibitem[Baldi et~al., 1999]{baldi1999exploiting}
Baldi, P., Brunak, S., Frasconi, P., Soda, G., and Pollastri, G. (1999).
\newblock Exploiting the past and the future in protein secondary structure
  prediction.
\newblock {\em Bioinformatics}, 15(11):937--946.

\bibitem[Ballard, 1987]{ballard1987modular}
Ballard, D.~H. (1987).
\newblock Modular learning in neural networks.
\newblock In {\em Proc. AAAI}, pages 279--284.

\bibitem[Bar et~al., 2015]{chest-transfer}
Bar, Y., Diamant, I., Wolf, L., and Greenspan, H. (2015).
\newblock Deep learning with non-medical training used for chest pathology
  identification.
\newblock {\em Proc. SPIE, Medical Imaging: Computer-Aided Diagnosis},
  9414:94140V--7.

\bibitem[Barron, 1993]{barron93universalapproximation}
Barron, A.~R. (1993).
\newblock {Universal approximation bounds for superpositions of a sigmoidal
  function}.
\newblock {\em Information Theory, IEEE Transactions on}, 39(3):930--945.

\bibitem[Baxter, 2000]{Baxter2000MIB}
Baxter, J. (2000).
\newblock A model of inductive bias learning.
\newblock {\em J. Artif. Int. Res.}, 12(1):149--198.

\bibitem[Belanger and McCallum, 2016]{belangerM16}
Belanger, D. and McCallum, A. (2016).
\newblock Structured prediction energy networks.
\newblock In {\em Proceedings of the 33nd International Conference on Machine
  Learning, {ICML} 2016, New York City, NY, USA, June 19-24, 2016}, pages
  983--992.

\bibitem[Belharbi et~al., 2015a]{belharbiICMLWSDL2015}
Belharbi, S., Chatelain, C., Hérault, R., and Adam, S. (2015a).
\newblock Learning structured output dependencies using deep neural networks.
\newblock {\em Deep Learning Workshop in the 32nd International Conference on
  Machine Learning (ICML)}.

\bibitem[Belharbi et~al., 2015b]{belharbiCAP2015}
Belharbi, S., Chatelain, C., Hérault, R., and Adam, S. (2015b).
\newblock A unified neural based model for structured output problems.
\newblock {\em Conférence Francophone sur l'Apprentissage Automatique (CAP)}.

\bibitem[Belharbi et~al., 2017a]{belharbiIEEETNNLS2017}
Belharbi, S., Chatelain, C., Hérault, R., and Adam, S. (2017a).
\newblock Neural networks regularization through class-wise invariant
  representation learning.
\newblock {\em XXX}, XX(X):XXXX--XXXX.

\bibitem[Belharbi et~al., 2017b]{belharbi2017}
Belharbi, S., Chatelain, C., Hérault, R., Adam, S., Thureau, S., Chastan, M.,
  and Modzelewski, R. (2017b).
\newblock Spotting l3 slice in ct scans using deep convolutional network and
  transfer learning.
\newblock {\em Computers in Biology and Medicine}, 87:95 -- 103.

\bibitem[Belharbi et~al., 2017c]{BELHARBICBM2017}
Belharbi, S., Chatelain, C., Hérault, R., Adam, S., Thureau, S., Chastan, M.,
  and Modzelewski, R. (2017c).
\newblock Spotting l3 slice in ct scans using deep convolutional network and
  transfer learning.
\newblock {\em Computers in Biology and Medicine}, 87:95 -- 103.

\bibitem[Belharbi et~al., 2016a]{belharbiESANN2016}
Belharbi, S., Hérault, R., Chatelain, C., and Adam, S. (2016a).
\newblock Deep multi-task learning with evolving weights.
\newblock In {\em European Symposium on Artificial Neural Networks (ESANN)}.

\bibitem[Belharbi et~al., 2016b]{sbelharbiRFIA2016}
Belharbi, S., Hérault, R., Chatelain, C., and Adam, S. (2016b).
\newblock Pondération dynamique dans un cadre multi-tâche pour réseaux de
  neurones profonds.
\newblock {\em Reconnaissance des Formes et l'Intelligence Artificielle (RFIA)
  (Session spéciale "Apprentissage et vision")}.

\bibitem[Belharbi et~al., 2018]{belharbiNeurocomp2017}
Belharbi, S., Hérault, R., Chatelain, C., and Adam, S. (2018).
\newblock Deep neural networks regularization for structured output prediction.
\newblock {\em Neurocomputing}, 281C:169 -- 177.

\bibitem[Belharbi et~al., 2016c]{bel16}
Belharbi, S., R.Hérault, Chatelain, C., and Adam, S. (2016c).
\newblock Deep multi-task learning with evolving weights.
\newblock In {\em European Symposium on Artificial Neural Networks ({ESANN})}.

\bibitem[Belhumeur et~al., 2011]{belhumeur11}
Belhumeur, P.~N., Jacobs, D.~W., Kriegman, D.~J., and Kumar, N. (2011).
\newblock {Localizing parts of faces using a consensus of exemplars.}
\newblock In {\em {CVPR}}, pages 545--552. IEEE.

\bibitem[Bellman, 1957]{Bellman1957}
Bellman, R. (1957).
\newblock {\em Dynamic Programming}.
\newblock Princeton University Press, Princeton, NJ, USA, 1st edition.

\bibitem[Ben-David et~al., 2010]{BenDavid2010da}
Ben-David, S., Blitzer, J., Crammer, K., Kulesza, A., Pereira, F., and Vaughan,
  J.~W. (2010).
\newblock A theory of learning from different domains.
\newblock {\em Machine Learning}, 79(1):151--175.

\bibitem[Ben{-}David et~al., 2006]{BenDavidBCP06nips}
Ben{-}David, S., Blitzer, J., Crammer, K., and Pereira, F. (2006).
\newblock Analysis of representations for domain adaptation.
\newblock In {\em Advances in Neural Information Processing Systems 19,
  Proceedings of the Twentieth Annual Conference on Neural Information
  Processing Systems, Vancouver, British Columbia, Canada, December 4-7, 2006},
  pages 137--144.

\bibitem[Ben{-}David and Borbely, 2008]{BenDavidB08ml}
Ben{-}David, S. and Borbely, R.~S. (2008).
\newblock A notion of task relatedness yielding provable multiple-task learning
  guarantees.
\newblock {\em Machine Learning}, 73(3):273--287.

\bibitem[Ben{-}David et~al., 2002]{BenDavidGS02kdd}
Ben{-}David, S., Gehrke, J., and Schuller, R. (2002).
\newblock A theoretical framework for learning from a pool of disparate data
  sources.
\newblock In {\em {KDD}}, pages 443--449. {ACM}.

\bibitem[Ben-David and Schuller, 2003]{bendavid2003springer}
Ben-David, S. and Schuller, R. (2003).
\newblock Exploiting task relatedness for multiple task learning.
\newblock In Sch{\"o}lkopf, B. and Warmuth, M.~K., editors, {\em Learning
  Theory and Kernel Machines}, pages 567--580, Berlin, Heidelberg. Springer
  Berlin Heidelberg.

\bibitem[Bengio, 2009]{bengio09}
Bengio, Y. (2009).
\newblock {Learning Deep Architectures for AI}.
\newblock {\em Found. Trends Mach. Learn.}, 2(1):1--127.

\bibitem[Bengio, 2012]{bengio12}
Bengio, Y. (2012).
\newblock Practical recommendations for gradient-based training of deep
  architectures.
\newblock In Montavon, G., Orr, G.~B., and Müller, K.-R., editors, {\em Neural
  Networks: Tricks of the Trade (2nd ed.)}, volume 7700 of {\em Lecture Notes
  in Computer Science}, pages 437--478. Springer.

\bibitem[Bengio, 2013]{bengio2013corr}
Bengio, Y. (2013).
\newblock Deep learning of representations: Looking forward.
\newblock {\em CoRR}, abs/1305.0445.

\bibitem[Bengio et~al., 2013a]{Bengio2013RLR}
Bengio, Y., Courville, A., and Vincent, P. (2013a).
\newblock Representation learning: A review and new perspectives.
\newblock {\em IEEE Trans. Pattern Anal. Mach. Intell.}, 35(8):1798--1828.

\bibitem[Bengio et~al., 2013b]{bengio13}
Bengio, Y., Courville, A.~C., and Vincent, P. (2013b).
\newblock {Representation Learning: {A} Review and New Perspectives}.
\newblock {\em {IEEE} PAMI}, 35(8):1798--1828.

\bibitem[Bengio et~al., 2006]{bengio06NIPS}
Bengio, Y., Lamblin, P., Popovici, D., and Larochelle, H. (2006).
\newblock Greedy layer-wise training of deep networks.
\newblock In {\em Advances in Neural information Processing Systems 19, NIPS
  2006}, pages 153--160.

\bibitem[Bengio et~al., 2007]{bengio07}
Bengio, Y., Lamblin, P., Popovici, D., and Larochelle, H. (2007).
\newblock {Greedy Layer-Wise Training of Deep Networks}.
\newblock In Sch{\"o}lkopf, B., Platt, J., and Hoffman, T., editors, {\em
  {NIPS}}, pages 153--160.

\bibitem[Bengio and Lecun, 2007]{bengioetlecun2007}
Bengio, Y. and Lecun, Y. (2007).
\newblock {\em Scaling learning algorithms towards AI}.
\newblock MIT Press.

\bibitem[Bengio et~al., 1994]{Bengio1994RNNLTD}
Bengio, Y., Simard, P., and Frasconi, P. (1994).
\newblock Learning long-term dependencies with gradient descent is difficult.
\newblock {\em Trans. Neur. Netw.}, 5(2):157--166.

\bibitem[Bermingham et~al., 2015]{bermingham2015application}
Bermingham, M.~L., Pong-Wong, R., Spiliopoulou, A., Hayward, C., Rudan, I.,
  Campbell, H., Wright, A.~F., Wilson, J.~F., Agakov, F., Navarro, P., et~al.
  (2015).
\newblock Application of high-dimensional feature selection: evaluation for
  genomic prediction in man.
\newblock {\em Scientific reports}, 5:10312.

\bibitem[Bikel et~al., 1999]{bikel99}
Bikel, D.~M., Schwartz, R., and Weischedel, R.~M. (1999).
\newblock {An algorithm that learns what's in a name}.
\newblock {\em Machine learning}, 34(1-3):211--231.

\bibitem[Bishop, 1995]{bishop1995ICANN}
Bishop, C. (1995).
\newblock Regularization and complexity control in feed-forward networks.
\newblock In {\em Proceedings International Conference on Artificial Neural
  Networks ICANN'95}, volume~1, page 141–148. EC2 et Cie.

\bibitem[Blitzer et~al., 2007]{BlitzerCKPW07nips}
Blitzer, J., Crammer, K., Kulesza, A., Pereira, F., and Wortman, J. (2007).
\newblock Learning bounds for domain adaptation.
\newblock In {\em Advances in Neural Information Processing Systems 20,
  Proceedings of the Twenty-First Annual Conference on Neural Information
  Processing Systems, Vancouver, British Columbia, Canada, December 3-6, 2007},
  pages 129--136.

\bibitem[Bonilla et~al., 2007]{BonillaCW07nips}
Bonilla, E.~V., Chai, K. M.~A., and Williams, C. K.~I. (2007).
\newblock Multi-task gaussian process prediction.
\newblock In {\em Advances in Neural Information Processing Systems 20,
  Proceedings of the Twenty-First Annual Conference on Neural Information
  Processing Systems, Vancouver, British Columbia, Canada, December 3-6, 2007},
  pages 153--160.

\bibitem[Bousmalis et~al., 2016]{BousmalisTSKE16nips}
Bousmalis, K., Trigeorgis, G., Silberman, N., Krishnan, D., and Erhan, D.
  (2016).
\newblock Domain separation networks.
\newblock In {\em {NIPS}}, pages 343--351.

\bibitem[Boyd and Vandenberghe, 2004]{Boyd2004}
Boyd, S. and Vandenberghe, L. (2004).
\newblock {\em Convex Optimization}.
\newblock Cambridge University Press, New York, NY, USA.

\bibitem[Breiman, 1996]{Breiman1996}
Breiman, L. (1996).
\newblock Bagging predictors.
\newblock {\em Machine Learning}, 24(2):123--140.

\bibitem[Breiman, 2001]{breiman2001}
Breiman, L. (2001).
\newblock Random forests.
\newblock {\em Mach. Learn.}, 45(1):5--32.

\bibitem[Bridle and Cox, 1990]{BridleC90nips}
Bridle, J.~S. and Cox, S. (1990).
\newblock Recnorm: Simultaneous normalisation and classification applied to
  speech recognition.
\newblock In {\em {NIPS}}, pages 234--240. Morgan Kaufmann.

\bibitem[Bromley et~al., 1994]{bromley1993}
Bromley, J., Guyon, I., LeCun, Y., S\"{a}ckinger, E., and Shah, R. (1994).
\newblock Signature verification using a "siamese" time delay neural network.
\newblock In Cowan, J.~D., Tesauro, G., and Alspector, J., editors, {\em
  Advances in Neural Information Processing Systems 6}, pages 737--744.
  Morgan-Kaufmann.

\bibitem[Bryson and Ho, 1969]{bryson1969applied}
Bryson, A. and Ho, Y. (1969).
\newblock {\em Applied optimal control: optimization, estimation, and control}.
\newblock Blaisdell Pub. Co.

\bibitem[Bryson, 1961]{bryson1961}
Bryson, A.~E. (1961).
\newblock A gradient method for optimizing multi-stage allocation processes.
\newblock In {\em Proc. Harvard Univ. Symposium on digital computers and their
  applications}.

\bibitem[Bryson and Denham, 1961]{BRYSON-DENHAM-61A}
Bryson, Jr., A.~E. and Denham, W.~F. (1961).
\newblock A steepest-ascent method for solving optimum programming problems.
\newblock Technical Report BR-1303, Raytheon Company, Missle and Space
  Division.

\bibitem[Candes and Tao, 2005]{Candes2005}
Candes, E.~J. and Tao, T. (2005).
\newblock Decoding by linear programming.
\newblock {\em IEEE Trans. Inf. Theor.}, 51(12):4203--4215.

\bibitem[Carlini et~al., 2018]{carlini2018memorization}
Carlini, N., Liu, C., Kos, J., Erlingsson, {\'{U}}., and Song, D. (2018).
\newblock The secret sharer: Measuring unintended neural network memorization
  {\&} extracting secrets.
\newblock {\em CoRR}, abs/1802.08232.

\bibitem[Caruana, 1993]{Caruana93icml}
Caruana, R. (1993).
\newblock Multitask learning: {A} knowledge-based source of inductive bias.
\newblock In {\em {ICML}}, pages 41--48. Morgan Kaufmann.

\bibitem[Caruana, 1997]{caruana97ML}
Caruana, R. (1997).
\newblock Multitask learning.
\newblock {\em Machine Learning}, 28(1):41--75.

\bibitem[Chapelle et~al., 2006]{chapelle06MIT}
Chapelle, O., Sch{\"o}lkopf, B., and Zien, A. (2006).
\newblock {\em Semi-supervised learning}.
\newblock Adaptive computation and machine learning. MIT Press.

\bibitem[Chartrand, 2007]{chartrand2007}
Chartrand, R. (2007).
\newblock Exact reconstruction of sparse signals via nonconvex minimization.
\newblock {\em IEEE Signal Processing Letters}, 14(10):707--710.

\bibitem[Chartrand, 2009]{chartrand2009}
Chartrand, R. (2009).
\newblock Fast algorithms for nonconvex compressive sensing: Mri reconstruction
  from very few data.
\newblock In {\em 2009 IEEE International Symposium on Biomedical Imaging: From
  Nano to Macro}, pages 262--265.

\bibitem[Checco and Corinto, 2006]{drusen-cnn}
Checco, P. and Corinto, F. (2006).
\newblock Cnn-based algorithm for drusen identification.
\newblock In {\em International Symposium on Circuits and Systems}.

\bibitem[Chen and Chaudhari, 2004]{ChenC04aISNN}
Chen, J. and Chaudhari, N.~S. (2004).
\newblock Capturing long-term dependencies for protein secondary structure
  prediction.
\newblock In {\em Advances in Neural Networks - {ISNN} 2004, International
  Symposium on Neural Networks, Dalian, China, August 19-21, 2004, Proceedings,
  Part {II}}, pages 494--500.

\bibitem[Chen et~al., 2012]{ChenXWS12icml}
Chen, M., Xu, Z.~E., Weinberger, K.~Q., and Sha, F. (2012).
\newblock Marginalized denoising autoencoders for domain adaptation.
\newblock In {\em {ICML}}. icml.cc / Omnipress.

\bibitem[Chen et~al., 2009]{Chen-tech-report2009}
Chen, X., Xu, F., and Ye, Y. (2009).
\newblock {Lower Bound Theory of Nonzero Entries in Solutions of l2-lp
  Minimization}.
\newblock Technical report, The Hong Kong Polytechnic University.

\bibitem[Chi and Mu, 2017]{chi2017corrdeeptesla}
Chi, L. and Mu, Y. (2017).
\newblock Deep steering: Learning end-to-end driving model from spatial and
  temporal visual cues.
\newblock {\em CoRR}, abs/1708.03798.

\bibitem[Chicco et~al., 2014]{Chicco2014DAN}
Chicco, D., Sadowski, P., and Baldi, P. (2014).
\newblock Deep autoencoder neural networks for gene ontology annotation
  predictions.
\newblock In {\em Proceedings of the 5th ACM Conference on Bioinformatics,
  Computational Biology, and Health Informatics}, BCB '14, pages 533--540, New
  York, NY, USA. ACM.

\bibitem[Cho et~al., 2014]{ChoMBB14SSST}
Cho, K., van Merrienboer, B., Bahdanau, D., and Bengio, Y. (2014).
\newblock On the properties of neural machine translation: Encoder-decoder
  approaches.
\newblock In {\em Proceedings of SSST@EMNLP 2014, Eighth Workshop on Syntax,
  Semantics and Structure in Statistical Translation, Doha, Qatar, 25 October
  2014}, pages 103--111.

\bibitem[Chollet, 2015]{chollet2015keras}
Chollet, F. (2015).
\newblock Keras.
\newblock \url{https://github.com/fchollet/keras}.

\bibitem[Chopra et~al., 2005]{chopraHL05}
Chopra, S., Hadsell, R., and LeCun, Y. (2005).
\newblock Learning a similarity metric discriminatively, with application to
  face verification.
\newblock In {\em 2005 {IEEE} Computer Society Conference on Computer Vision
  and Pattern Recognition {(CVPR} 2005), 20-26 June 2005, San Diego, CA,
  {USA}}, pages 539--546.

\bibitem[Chung et~al., 2009]{Chung2009}
Chung, H., Cobzas, D., Birdsell, L., Lieffers, J., and Baracos, V. (2009).
\newblock {Automated segmentation of muscle and adipose tissue on CT images for
  human body composition analysis}.
\newblock {\em Proceedings of SPIE}, 7261:72610K--72610K--8.

\bibitem[Chung et~al., 2014]{Chung2014DLWNIPS}
Chung, J., G{\"{u}}lçehre, {\c{C}}., Cho, K., and Bengio, Y. (2014).
\newblock Empirical evaluation of gated recurrent neural networks on sequence
  modeling.
\newblock {\em arXiv e-prints}, abs/1412.3555.
\newblock Presented at the Deep Learning workshop at NIPS2014.

\bibitem[Chung et~al., 2015]{ChungGCB15ICML}
Chung, J., Çaglar G{\"{u}}l{\c{c}}ehre, Cho, K., and Bengio, Y. (2015).
\newblock Gated feedback recurrent neural networks.
\newblock In {\em Proceedings of the 32nd International Conference on Machine
  Learning, {ICML} 2015, Lille, France, 6-11 July 2015}, pages 2067--2075.

\bibitem[Cire\c{s}an et~al., 2010a]{Ciresan2010DBS}
Cire\c{s}an, D.~C., Meier, U., Gambardella, L.~M., and Schmidhuber, J. (2010a).
\newblock Deep, big, simple neural nets for handwritten digit recognition.
\newblock {\em Neural Comput.}, 22(12):3207--3220.

\bibitem[Cire\c{s}an et~al., 2010b]{Ciresan2010}
Cire\c{s}an, D.~C., Meier, U., Gambardella, L.~M., and Schmidhuber, J. (2010b).
\newblock Deep, big, simple neural nets for handwritten digit recognition.
\newblock {\em Neural Computation}, 22(12):3207--3220.

\bibitem[Ciresan et~al., 2012a]{ciresan12}
Ciresan, D., Meier, U., and Schmidhuber, J. (2012a).
\newblock {Multi-column deep neural networks for image classification}.
\newblock In {\em {IN PROCEEDINGS OF THE 25TH IEEE CONFERENCE ON COMPUTER
  VISION AND PATTERN RECOGNITION (CVPR 2012}}, pages 3642--3649.

\bibitem[Ciresan et~al., 2012b]{ciresanGGS12}
Ciresan, D.~C., Giusti, A., Gambardella, L.~M., and Schmidhuber, J. (2012b).
\newblock Deep neural networks segment neuronal membranes in electron
  microscopy images.
\newblock In {\em Advances in Neural Information Processing Systems 25: 26th
  Annual Conference on Neural Information Processing Systems 2012. Proceedings
  of a meeting held December 3-6, 2012, Lake Tahoe, Nevada, United States.},
  pages 2852--2860.

\bibitem[Cireşan et~al., 2012]{Ciresan2012}
Cireşan, D.~C., Meier, U., and Schmidhuber, J. (2012).
\newblock Transfer learning for latin and chinese characters with deep neural
  networks.
\newblock In {\em International Joint Conference on Neural Networks}, pages
  1--6.

\bibitem[Collobert and Weston, 2008a]{collobert08ICML}
Collobert, R. and Weston, J. (2008a).
\newblock A unified architecture for natural language processing: deep neural
  networks with multitask learning.
\newblock In {\em Machine Learning, Proceedings of he 25th International
  Conference, ICML 2008}, pages 160--167.

\bibitem[Collobert and Weston, 2008b]{CollobertWeston2008}
Collobert, R. and Weston, J. (2008b).
\newblock A unified architecture for natural language processing: deep neural
  networks with multitask learning.
\newblock In {\em Machine Learning, Proceedings of the Twenty-Fifth
  International Conference {(ICML} 2008), Helsinki, Finland, June 5-9, 2008},
  pages 160--167.

\bibitem[Cortes and Vapnik, 1995]{Cortes1995svm}
Cortes, C. and Vapnik, V. (1995).
\newblock Support-vector networks.
\newblock {\em Machine Learning}, 20(3):273--297.

\bibitem[Crammer et~al., 2008]{CrammerKW08jmlr}
Crammer, K., Kearns, M.~J., and Wortman, J. (2008).
\newblock Learning from multiple sources.
\newblock {\em Journal of Machine Learning Research}, 9:1757--1774.

\bibitem[Cristinacce and Cootes, 2006]{cristinacce06}
Cristinacce, D. and Cootes, T. (2006).
\newblock {Feature Detection and Tracking with Constrained Local Models}.
\newblock In {\em {BMVC}}, pages 95.1--95.10.

\bibitem[Cs{\'a}ji, 2001]{csaji2001mscthesis}
Cs{\'a}ji, B.~C. (2001).
\newblock Approximation with artificial neural networks.
\newblock Master's thesis, Faculty of Sciences, Etvs Lornd University, Hungary,
  Hungary.

\bibitem[Cunliffe et~al., 2015]{Cunliffe2015}
Cunliffe, A., White, B., Justusson, J., Straus, C., Malik, R., Hallaq, A.-H.,
  and Armato, S. (2015).
\newblock {Comparison of Two Deformable Registration Algorithms in the Presence
  of Radiologic Change Between Serial Lung CT Scans}.
\newblock {\em Journal of Digital Imaging}, 28(6):755--760.

\bibitem[Cybenko, 1989]{Cybenko1989}
Cybenko, G. (1989).
\newblock Approximation by superpositions of a sigmoidal function.
\newblock {\em Mathematics of Control, Signals and Systems}, 2(4):303--314.

\bibitem[Dai et~al., 2007]{DaiYXY07icml}
Dai, W., Yang, Q., Xue, G., and Yu, Y. (2007).
\newblock Boosting for transfer learning.
\newblock In {\em Machine Learning, Proceedings of the Twenty-Fourth
  International Conference {(ICML} 2007), Corvallis, Oregon, USA, June 20-24,
  2007}, pages 193--200.

\bibitem[De~Vries et~al., 2016]{devriesESANN2016}
De~Vries, H., Memisevic, R., and Courville, A. (2016).
\newblock Deep learning vector quantization.
\newblock In {\em European Symposium on Artificial Neural Networks (ESANN)}.

\bibitem[Dean et~al., 2012]{Dean2012xx}
Dean, J., Corrado, G.~S., Monga, R., Chen, K., Devin, M., Le, Q.~V., Mao,
  M.~Z., Ranzato, M., Senior, A., Tucker, P., Yang, K., and Ng, A.~Y. (2012).
\newblock Large scale distributed deep networks.
\newblock In {\em Proceedings of the 25th International Conference on Neural
  Information Processing Systems - Volume 1}, NIPS'12, pages 1223--1231, USA.
  Curran Associates Inc.

\bibitem[Dechter, 1986]{Dechter86}
Dechter, R. (1986).
\newblock Learning while searching in constraint-satisfaction-problems.
\newblock In Kehler, T., editor, {\em AAAI}, pages 178--185. Morgan Kaufmann.

\bibitem[Demyanov, 2015]{Demyanovtehsis2015}
Demyanov, S. (2015).
\newblock {\em Regularization methods for neural networks and related models}.
\newblock PhD thesis, The University of Melbourne, Department of Computing and
  Information Systems.

\bibitem[Deng et~al., 2009a]{imagenet09}
Deng, J., Dong, W., Socher, R., Li, L.-J., Li, K., and Fei-Fei, L. (2009a).
\newblock {ImageNet: A Large-Scale Hierarchical Image Database}.
\newblock In {\em CVPR09}.

\bibitem[Deng et~al., 2009b]{imagenetcvpr09}
Deng, J., Dong, W., Socher, R., Li, L.-J., Li, K., and Li, F.-F. (2009b).
\newblock Imagenet: A large-scale hierarchical image database.
\newblock In {\em CVPR}, pages 248--255.

\bibitem[Deng and Yu, 2014]{Deng2014DLM}
Deng, L. and Yu, D. (2014).
\newblock Deep learning: Methods and applications.
\newblock {\em Found. Trends Signal Process.}, 7(3\&\#8211;4):197--387.

\bibitem[Desjardins et~al., 2015]{DesjardinsNIPS2015}
Desjardins, G., Simonyan, K., Pascanu, R., and kavukcuoglu, k. (2015).
\newblock Natural neural networks.
\newblock In Cortes, C., Lawrence, N.~D., Lee, D.~D., Sugiyama, M., and
  Garnett, R., editors, {\em Advances in Neural Information Processing Systems
  28}, pages 2071--2079. Curran Associates, Inc.

\bibitem[Doersch, 2016]{Doersch16CORR}
Doersch, C. (2016).
\newblock Tutorial on variational autoencoders.
\newblock {\em CoRR}, abs/1606.05908.

\bibitem[Donoho, 2004]{donoho2004}
Donoho, D. (2004).
\newblock {For most large underdetermined systems of linear equations the
  minimal l1 -norm solution is also the sparsest Solution}.
\newblock Technical report, Stanford University.

\bibitem[Dosovitskiy et~al., 2017]{DosovitskiySTB17PAMI}
Dosovitskiy, A., Springenberg, J.~T., Tatarchenko, M., and Brox, T. (2017).
\newblock Learning to generate chairs, tables and cars with convolutional
  networks.
\newblock {\em {IEEE} Trans. Pattern Anal. Mach. Intell.}, 39(4):692--705.

\bibitem[Dozat, 2016]{dozat2016incorporating}
Dozat, T. (2016).
\newblock Incorporating nesterov momentum into adam.

\bibitem[Dreyfus, 1962]{dreyfus1962}
Dreyfus, S.~E. (1962).
\newblock The numerical solution of variational problems.
\newblock {\em Journal of Mathematical Analysis and Applications}, 5(1):30--45.

\bibitem[Dubey et~al., 2018]{dubey2018investigating}
Dubey, R., Agrawal, P., Pathak, D., Griffiths, T.~L., and Efros, A.~A. (2018).
\newblock Investigating human priors for playing video games.

\bibitem[Duchi et~al., 2010]{duchi1}
Duchi, J., Hazan, E., and Singer, Y. (2010).
\newblock {Adaptive Subgradient Methods for Online Learning and Stochastic
  Optimization.}
\newblock In {\em {COLT}}, pages 257--269.

\bibitem[Duong et~al., 2015]{DuongCBC15acl}
Duong, L., Cohn, T., Bird, S., and Cook, P. (2015).
\newblock Low resource dependency parsing: Cross-lingual parameter sharing in a
  neural network parser.
\newblock In {\em {ACL} {(2)}}, pages 845--850. The Association for Computer
  Linguistics.

\bibitem[Dwork, 2011]{Dwork11cacm}
Dwork, C. (2011).
\newblock A firm foundation for private data analysis.
\newblock {\em Commun. {ACM}}, 54(1):86--95.

\bibitem[Dwork et~al., 2006]{DworkMNS06tcc}
Dwork, C., McSherry, F., Nissim, K., and Smith, A.~D. (2006).
\newblock Calibrating noise to sensitivity in private data analysis.
\newblock In {\em {TCC}}, volume 3876 of {\em Lecture Notes in Computer
  Science}, pages 265--284. Springer.

\bibitem[Dwork and Roth, 2014]{DworkR14fttcs}
Dwork, C. and Roth, A. (2014).
\newblock The algorithmic foundations of differential privacy.
\newblock {\em Foundations and Trends in Theoretical Computer Science},
  9(3-4):211--407.

\bibitem[El-Yacoubi et~al., 2002]{elyacoubi02}
El-Yacoubi, M., Gilloux, M., and Bertille, J.-M. (2002).
\newblock {A statistical approach for phrase location and recognition within a
  text line: An application to street name recognition}.
\newblock {\em IEEE PAMI}, 24(2):172--188.

\bibitem[Ellis, 1965]{ellis1965transfer}
Ellis, H.~C. (1965).
\newblock {\em Invariant {Subspaces}}.
\newblock Macmillan, New York.

\bibitem[Erhan et~al., 2010]{erhan10whydoes}
Erhan, D., Bengio, Y., Courville, A., Manzagol, P.-A., Vincent, P., and Bengio,
  S. (2010).
\newblock Why does unsupervised pre-training help deep learning?
\newblock {\em J. Mach. Learn. Res.}, 11:625{\textendash}660.

\bibitem[Erhan et~al., 2014]{Erhan14}
Erhan, D., Szegedy, C., Toshev, A., and Anguelov, D. (2014).
\newblock Scalable object detection using deep neural networks.
\newblock In {\em CVPR}, pages 2155--2162.

\bibitem[Evgeniou and Pontil, 2004]{EvgeniouP04kdd}
Evgeniou, T. and Pontil, M. (2004).
\newblock Regularized multi--task learning.
\newblock In {\em Proceedings of the Tenth {ACM} {SIGKDD} International
  Conference on Knowledge Discovery and Data Mining, Seattle, Washington, USA,
  August 22-25, 2004}, pages 109--117.

\bibitem[Fahlman et~al., 1983]{FahlmanHS83AAAI}
Fahlman, S.~E., Hinton, G.~E., and Sejnowski, T.~J. (1983).
\newblock Massively parallel architectures for {AI:} netl, thistle, and
  boltzmann machines.
\newblock In {\em Proceedings of the National Conference on Artificial
  Intelligence. Washington, D.C., August 22-26, 1983.}, pages 109--113.

\bibitem[Fan and R., 2001]{fan2001norm}
Fan, J. and R., L. (2001).
\newblock Variable selection via nonconcave penalized likelihood and its oracle
  properties.
\newblock {\em Journal of the American Statistical Association}, 96:1348--1360.

\bibitem[Farabet et~al., 2013]{farabet13}
Farabet, C., Couprie, C., Najman, L., and LeCun, Y. (2013).
\newblock {Learning Hierarchical Features for Scene Labeling}.
\newblock {\em {IEEE} PAMI}, 35(8):1915--1929.

\bibitem[Flamary et~al., 2018]{flamary17}
Flamary, R., Cuturi, M., Courty, N., and Rakotomamonjy, A. (2018).
\newblock Wasserstein discriminant analysis.
\newblock {\em Machine learning}.

\bibitem[Fridman et~al., 2018]{Fridman2018corr}
Fridman, L., Jenik, B., and Terwilliger, J. (2018).
\newblock Deeptraffic: Driving fast through dense traffic with deep
  reinforcement learning.
\newblock {\em CoRR}, abs/1801.02805.

\bibitem[Fridman, 1993]{fridman93}
Fridman, M. (1993).
\newblock {\em {Hidden markov model regression}}.
\newblock PhD thesis, Graduate School of Arts and Sciences, University of
  Pennsylvania.

\bibitem[Fukada et~al., 1999]{fukada99boundary}
Fukada, T., Schuster, M., and Sagisaka, Y. (1999).
\newblock Phoneme boundary estimation using bidirectional recurrent neural
  networks and its applications.
\newblock {\em Systems and Computers in Japan}, 30(4):20--30.

\bibitem[Fukushima, 1979]{Fukushima1979neocognitron}
Fukushima, K. (1979).
\newblock Neural network model for a mechanism of pattern recognition
  unaffected by shift in position - {Neocognitron}.
\newblock {\em Trans. IECE}, J62-A(10):658--665.

\bibitem[Fukushima, 1980]{fukushima1980}
Fukushima, K. (1980).
\newblock Neocognitron: A self-organizing neural network for a mechanism of
  pattern recognition unaffected by shift in position.
\newblock {\em Biological Cybernetics}, 36(4):193--202.

\bibitem[Fukushima, 2011]{Fukushima2011}
Fukushima, K. (2011).
\newblock {Increasing robustness against background noise: visual pattern
  recognition by a Neocognitron}.
\newblock {\em Neural Networks}, 24(7):767--778.

\bibitem[Fukushima, 2013]{Fukushima2013b}
Fukushima, K. (2013).
\newblock {Training multi-layered neural network Neocognitron}.
\newblock {\em Neural Networks}, 40:18--31.

\bibitem[Fukushima and Miyake, 1982]{fukushimaM82}
Fukushima, K. and Miyake, S. (1982).
\newblock Neocognitron: {A} new algorithm for pattern recognition tolerant of
  deformations and shifts in position.
\newblock {\em Pattern Recognition}, 15(6):455--469.

\bibitem[Funahashi, 1989]{Funahashi1989}
Funahashi, K. (1989).
\newblock On the approximate realization of continuous mappings by neural
  networks.
\newblock {\em Neural Networks}, 2(3):183--192.

\bibitem[Gamba et~al., 1961]{Gamba1961}
Gamba, A., Gamberini, L., Palmieri, G., and Sanna, R. (1961).
\newblock Further experiments with papa.
\newblock {\em Il Nuovo Cimento (1955-1965)}, 20(2):112--115.

\bibitem[Ganin et~al., 2016]{GaninJMLRv172016}
Ganin, Y., Ustinova, E., Ajakan, H., Germain, P., Larochelle, H., Laviolette,
  F., Marchand, M., and Lempitsky, V. (2016).
\newblock Domain-adversarial training of neural networks.
\newblock {\em Journal of Machine Learning Research}, 17(59):1--35.

\bibitem[Gao et~al., 2008]{GaoFJH08kdd}
Gao, J., Fan, W., Jiang, J., and Han, J. (2008).
\newblock Knowledge transfer via multiple model local structure mapping.
\newblock In {\em Proceedings of the 14th {ACM} {SIGKDD} International
  Conference on Knowledge Discovery and Data Mining, Las Vegas, Nevada, USA,
  August 24-27, 2008}, pages 283--291.

\bibitem[Ge et~al., 2011]{Ge2011}
Ge, D., Jiang, X., and Ye, Y. (2011).
\newblock A note on the complexity of lp minimization.
\newblock {\em Mathematical Programming}, 129(2):285--299.

\bibitem[Geman et~al., 1992]{Geman1992}
Geman, S., Bienenstock, E., and Doursat, R. (1992).
\newblock Neural networks and the bias/variance dilemma.
\newblock {\em Neural Comput.}, 4(1):1--58.

\bibitem[Germain et~al., 2017]{germain2017pac}
Germain, P., Habrard, A., Laviolette, F., and Morvant, E. (2017).
\newblock Pac-bayes and domain adaptation.
\newblock {\em arXiv}.

\bibitem[Ghosh et~al., 2011]{Ghosh2011}
Ghosh, S., Alomari, R.~S., Chaudhary, V., and Dhillon, G. (2011).
\newblock {Automatic lumbar vertebra segmentation from clinical CT for wedge
  compression fracture diagnosis}.
\newblock {\em Proceedings of the SPIE}, 3:796303--9.

\bibitem[Girosi and Poggio, 1989]{Girosi1989}
Girosi, F. and Poggio, T. (1989).
\newblock Representation properties of networks: Kolmogorov's theorem is
  irrelevant.
\newblock {\em Neural Computation}, 1(4):465--469.

\bibitem[Glocker et~al., 2013]{Glocker2013}
Glocker, B., D.Zikic, E.Konukoglu, Haynor, D., and Criminisi, A. (2013).
\newblock {Vertebrae localization in pathological spine CT via dense
  classification from sparse annotations.}
\newblock {\em MICCAI}, 16(Pt 2):262--70.

\bibitem[Glocker et~al., 2012]{Glocker2012}
Glocker, B., Feulner, J., Criminisi, A., Haynor, D.~R., and Konukoglu, E.
  (2012).
\newblock {\em Automatic Localization and Identification of Vertebrae in
  Arbitrary Field-of-View CT Scans}, pages 590--598.
\newblock Springer Berlin Heidelberg, Berlin, Heidelberg.

\bibitem[Glocker et~al., 2014]{Glocker2014}
Glocker, B., Zikic, D., and Haynor, D.~R. (2014).
\newblock {\em Robust Registration of Longitudinal Spine CT}, pages 251--258.
\newblock Springer International Publishing.

\bibitem[Glorot and Bengio, 2010]{glorot10}
Glorot, X. and Bengio, Y. (2010).
\newblock {Understanding the difficulty of training deep feedforward neural
  networks}.
\newblock In {\em {International conference on artificial intelligence and
  statistics}}, pages 249--256.

\bibitem[Glorot et~al., 2011a]{AISTATS2011GlorotBB11}
Glorot, X., Bordes, A., and Bengio, Y. (2011a).
\newblock Deep sparse rectifier neural networks.
\newblock In Gordon, G.~J. and Dunson, D.~B., editors, {\em Proceedings of the
  Fourteenth International Conference on Artificial Intelligence and Statistics
  (AISTATS-11)}, volume~15, pages 315--323. Journal of Machine Learning
  Research - Workshop and Conference Proceedings.

\bibitem[Glorot et~al., 2011b]{GlorotBB11icml}
Glorot, X., Bordes, A., and Bengio, Y. (2011b).
\newblock Domain adaptation for large-scale sentiment classification: {A} deep
  learning approach.
\newblock In {\em Proceedings of the 28th International Conference on Machine
  Learning, {ICML} 2011, Bellevue, Washington, USA, June 28 - July 2, 2011},
  pages 513--520.

\bibitem[Goertzel, 2015]{Geortzel2015}
Goertzel, B. (2015).
\newblock Are there deep reasons underlying the pathologies of today's deep
  learning algorithms?
\newblock In Bieger, J., Goertzel, B., and Potapov, A., editors, {\em
  Artificial General Intelligence}, pages 70--79, Cham. Springer International
  Publishing.

\bibitem[Golodetz et~al., 2009]{Golodetz2009}
Golodetz, S., Voiculescu, I., and Cameron, S. (2009).
\newblock {Automatic spine identification in abdominal CT slices using image
  partition forests}.
\newblock {\em International Symposium on Image and Signal Processing and
  Analysis}.

\bibitem[Gomez and Schmidhuber, 2005]{Gomez2005CRN}
Gomez, F.~J. and Schmidhuber, J. (2005).
\newblock Co-evolving recurrent neurons learn deep memory pomdps.
\newblock In {\em Proceedings of the 7th Annual Conference on Genetic and
  Evolutionary Computation}, GECCO '05, pages 491--498, New York, NY, USA. ACM.

\bibitem[Goodfellow et~al., 2016]{Goodfellowbook2016}
Goodfellow, I., Bengio, Y., and Courville, A. (2016).
\newblock {\em Deep Learning}.
\newblock MIT Press.
\newblock \url{http://www.deeplearningbook.org}.

\bibitem[Goodfellow et~al., 2014a]{goodfellow2014iclr}
Goodfellow, I., Bulatov, Y., Ibarz, J., Arnoud, S., and Shet, V. (2014a).
\newblock Multi-digit number recognition from street view imagery using deep
  convolutional neural networks.
\newblock In {\em International Conference on Learning Representations
  (ICLR2014)}.

\bibitem[Goodfellow et~al., 2014b]{GoodfellowNIPS2014}
Goodfellow, I., Pouget-Abadie, J., Mirza, M., Xu, B., Warde-Farley, D., Ozair,
  S., Courville, A., and Bengio, Y. (2014b).
\newblock Generative adversarial nets.
\newblock In Ghahramani, Z., Welling, M., Cortes, C., Lawrence, N.~D., and
  Weinberger, K.~Q., editors, {\em Advances in Neural Information Processing
  Systems 27}, pages 2672--2680. Curran Associates, Inc.

\bibitem[Goodfellow et~al., 2014c]{GoodfellowSS14CORR}
Goodfellow, I.~J., Shlens, J., and Szegedy, C. (2014c).
\newblock Explaining and harnessing adversarial examples.
\newblock {\em CoRR}, abs/1412.6572.

\bibitem[Goodfellow et~al., 2013]{GoodfellowWMCB13ICML}
Goodfellow, I.~J., Warde{-}Farley, D., Mirza, M., Courville, A.~C., and Bengio,
  Y. (2013).
\newblock Maxout networks.
\newblock In {\em Proceedings of the 30th International Conference on Machine
  Learning, {ICML} 2013, Atlanta, GA, USA, 16-21 June 2013}, pages 1319--1327.

\bibitem[Gordon and Desjardins, 1995]{Gordon1995}
Gordon, D.~F. and Desjardins, M. (1995).
\newblock Evaluation and selection of biases in machine learning.
\newblock {\em Machine Learning}, 20(1):5--22.

\bibitem[Gou{\'e}rant et~al., 2013]{Gouerant2013}
Gou{\'e}rant, S., Leheurteur, M., Chaker, M., Modzelewski, R., Rigal, O.,
  Veyret, C., Lauridant, G., and Clatot, F. (2013).
\newblock A higher body mass index and fat mass are factors predictive of
  docetaxel dose intensity.
\newblock {\em Anticancer research}, 33(12):5655.

\bibitem[Graves, 2012]{Graves2012CSI}
Graves, A. (2012).
\newblock {\em Supervised Sequence Labelling with Recurrent Neural Networks},
  volume 385 of {\em Studies in Computational Intelligence}.
\newblock Springer.

\bibitem[Graves, 2013a]{Graves13}
Graves, A. (2013a).
\newblock Generating sequences with recurrent neural networks.
\newblock {\em CoRR}, abs/1308.0850.

\bibitem[Graves, 2013b]{Graves13ARXIVCORR}
Graves, A. (2013b).
\newblock Generating sequences with recurrent neural networks.
\newblock {\em CoRR}, abs/1308.0850.

\bibitem[Graves and Jaitly, 2014a]{gravesJ14}
Graves, A. and Jaitly, N. (2014a).
\newblock Towards end-to-end speech recognition with recurrent neural networks.
\newblock In {\em Proceedings of the 31th International Conference on Machine
  Learning, {ICML} 2014, Beijing, China, 21-26 June 2014}, pages 1764--1772.

\bibitem[Graves and Jaitly, 2014b]{Graves2014ICML}
Graves, A. and Jaitly, N. (2014b).
\newblock Towards end-to-end speech recognition with recurrent neural networks.
\newblock In {\em Proceedings of the 31st International Conference on
  International Conference on Machine Learning - Volume 32}, ICML'14, pages
  II--1764--II--1772. JMLR.org.

\bibitem[Graves et~al., 2009]{graves2009novel}
Graves, A., Liwicki, M., Fern{\'a}ndez, S., Bertolami, R., Bunke, H., and
  Schmidhuber, J. (2009).
\newblock A novel connectionist system for unconstrained handwriting
  recognition.
\newblock {\em IEEE transactions on pattern analysis and machine intelligence},
  31(5):855--868.

\bibitem[Graves et~al., 2013]{GravesMH13ICASSP}
Graves, A., Mohamed, A., and Hinton, G.~E. (2013).
\newblock Speech recognition with deep recurrent neural networks.
\newblock In {\em {IEEE} International Conference on Acoustics, Speech and
  Signal Processing, {ICASSP} 2013, Vancouver, BC, Canada, May 26-31, 2013},
  pages 6645--6649.

\bibitem[Graves and Schmidhuber, 2009]{GravesNIPS2008}
Graves, A. and Schmidhuber, J. (2009).
\newblock Offline handwriting recognition with multidimensional recurrent
  neural networks.
\newblock In Koller, D., Schuurmans, D., Bengio, Y., and Bottou, L., editors,
  {\em Advances in Neural Information Processing Systems 21}, pages 545--552.
  Curran Associates, Inc.

\bibitem[Graves et~al., 2014]{graves2014neuralturingmachine}
Graves, A., Wayne, G., and Danihelka, I. (2014).
\newblock Neural turing machines.
\newblock {\em arXiv preprint arXiv:1410.5401}.

\bibitem[Graves et~al., 2016]{GravesNature2016}
Graves, A., Wayne, G., Reynolds, M., Harley, T., Danihelka, I.,
  Grabska-Barwi\'{n}ska, A., Colmenarejo, S.~G., Grefenstette, E., Ramalho, T.,
  Agapiou, J., Badia, A.~P., Hermann, K.~M., Zwols, Y., Ostrovski, G., Cain,
  A., King, H., Summerfield, C., Blunsom, P., Kavukcuoglu, K., and Hassabis, D.
  (2016).
\newblock {Hybrid computing using a neural network with dynamic external
  memory}.
\newblock {\em Nature}, 538(7626):471--476.

\bibitem[Griewank, 2012]{Griewank2012}
Griewank, A. (2012).
\newblock {\em Documenta Mathematica - Extra Volume ISMP}, pages 389--400.

\bibitem[Grossberg, 1973]{grossberg1973c}
Grossberg, S. (1973).
\newblock {Contour enhancement, short term memory, and constancies in
  reverberating neural networks}.
\newblock {\em Studies in Applied Mathematics}, 52(3):213--257.

\bibitem[Grossberg, 1982]{Grossberg1982}
Grossberg, S. (1982).
\newblock {\em Contour Enhancement, Short Term Memory, and Constancies in
  Reverberating Neural Networks}, pages 332--378.
\newblock Springer Netherlands, Dordrecht.

\bibitem[Gr\"{u}nwald, 2007]{Grunwald2007}
Gr\"{u}nwald, P.~D. (2007).
\newblock {\em The Minimum Description Length Principle (Adaptive Computation
  and Machine Learning)}.
\newblock The MIT Press.

\bibitem[Grünwald, 2005]{Grunwald05atutorial}
Grünwald, P. (2005).
\newblock A tutorial introduction to the minimum description length principle.
\newblock In {\em Advances in Minimum Description Length: Theory and
  Applications}. MIT Press.

\bibitem[Gu and Rigazio, 2014]{shixiang2014}
Gu, S. and Rigazio, L. (2014).
\newblock Towards deep neural network architectures robust to adversarial
  examples.
\newblock {\em CoRR}, abs/1412.5068.

\bibitem[Hadamard, 1908]{hadamard1908memoire}
Hadamard, J. (1908).
\newblock {\em M{\'e}moire sur le probl{\`e}me d'analyse relatif {\`a}
  l'{\'e}quilibre des plaques {\'e}lastiques encastr{\'e}es}, volume~33.
\newblock Imprimerie nationale.

\bibitem[Hadsell et~al., 2006]{hadsellCL06}
Hadsell, R., Chopra, S., and LeCun, Y. (2006).
\newblock Dimensionality reduction by learning an invariant mapping.
\newblock In {\em 2006 {IEEE} Computer Society Conference on Computer Vision
  and Pattern Recognition {(CVPR} 2006), 17-22 June 2006, New York, NY, {USA}},
  pages 1735--1742.

\bibitem[Hafemann et~al., 2016]{Hafemann2016}
Hafemann, L.~G., Sabourin, R., and Oliveira, L.~S. (2016).
\newblock Writer-independent feature learning for offline signature
  verification using deep convolutional neural networks.
\newblock {\em CoRR}, abs/1604.00974.

\bibitem[Hammer, 1998]{Hammer98onthe}
Hammer, B. (1998).
\newblock On the approximation capability of recurrent neural networks.
\newblock In {\em In International Symposium on Neural Computation}, pages
  12--4.

\bibitem[Hansel et~al., 1992]{hansel1992memorization}
Hansel, D., Mato, G., and Meunier, C. (1992).
\newblock Memorization without generalization in a multilayered neural network.
\newblock {\em EPL (Europhysics Letters)}, 20(5):471.

\bibitem[Hashimoto et~al., 2017]{HashimotoXTS17emnlp}
Hashimoto, K., Xiong, C., Tsuruoka, Y., and Socher, R. (2017).
\newblock A joint many-task model: Growing a neural network for multiple {NLP}
  tasks.
\newblock In {\em {EMNLP}}, pages 1923--1933. Association for Computational
  Linguistics.

\bibitem[Hassoun, 1995]{Hassoun1995}
Hassoun, M.~H. (1995).
\newblock {\em Fundamentals of Artificial Neural Networks}.
\newblock MIT Press, Cambridge, MA, USA, 1st edition.

\bibitem[Hastie et~al., 2015]{Hastie2015book}
Hastie, T., Tibshirani, R., and Wainwright, M. (2015).
\newblock {\em Statistical Learning with Sparsity: The Lasso and
  Generalizations}.
\newblock Chapman \& Hall/CRC.

\bibitem[Havaei et~al., 2015]{Havaei15}
Havaei, M., Davy, A., Warde{-}Farley, D., Biard, A., Courville, A., Bengio, Y.,
  Pal, C., Jodoin, P., and Larochelle, H. (2015).
\newblock Brain tumor segmentation with deep neural networks.
\newblock {\em CoRR}, abs/1505.03540.

\bibitem[He et~al., 2015]{HeZRS15iccv}
He, K., Zhang, X., Ren, S., and Sun, J. (2015).
\newblock Delving deep into rectifiers: Surpassing human-level performance on
  imagenet classification.
\newblock In {\em ICCV 2015}, pages 1026--1034.

\bibitem[He et~al., 2016a]{heZRS16}
He, K., Zhang, X., Ren, S., and Sun, J. (2016a).
\newblock Deep residual learning for image recognition.
\newblock In {\em 2016 IEEE Conference on Computer Vision and Pattern
  Recognition (CVPR)}, pages 770--778.

\bibitem[He et~al., 2016b]{HeZRS16ECCV}
He, K., Zhang, X., Ren, S., and Sun, J. (2016b).
\newblock Identity mappings in deep residual networks.
\newblock In {\em Computer Vision - {ECCV} 2016 - 14th European Conference,
  Amsterdam, The Netherlands, October 11-14, 2016, Proceedings, Part {IV}},
  pages 630--645.

\bibitem[Hebb, 1949]{hebb1949}
Hebb, D.~O. (1949).
\newblock {\em The organization of behavior: {A} neuropsychological theory}.
\newblock Wiley, New York.

\bibitem[Hecht-Nielsen, 1989a]{HechtNielsen1989}
Hecht-Nielsen, R. (1989a).
\newblock {\em Neurocomputing}.
\newblock Addison-Wesley Longman Publishing Co., Inc., Boston, MA, USA.

\bibitem[Hecht-Nielsen, 1989b]{hecht1989}
Hecht-Nielsen, R. (1989b).
\newblock Theory of the backpropagation neural network.
\newblock In {\em International Joint Conference on Neural Networks (IJCNN)},
  pages 593--605. IEEE.

\bibitem[Heckerman et~al., 1995]{heckerman1995}
Heckerman, D., Geiger, D., and Chickering, D.~M. (1995).
\newblock Learning bayesian networks: The combination of knowledge and
  statistical data.
\newblock {\em Machine Learning}, 20(3):197--243.

\bibitem[Hinton, 2002]{Hinton2002CD}
Hinton, G.~E. (2002).
\newblock Training products of experts by minimizing contrastive divergence.
\newblock {\em Neural Comput.}, 14(8):1771--1800.

\bibitem[Hinton, 2007a]{Hinton2007}
Hinton, G.~E. (2007a).
\newblock Learning multiple layers of representation.
\newblock {\em Trends in Cognitive Sciences}, 11:428--434.

\bibitem[Hinton, 2007b]{Hinton2007Trends}
Hinton, G.~E. (2007b).
\newblock Learning multiple layers of representation.
\newblock {\em Trends in Cognitive Sciences}, 11:428--434.

\bibitem[Hinton, 2012]{Hinton12CD}
Hinton, G.~E. (2012).
\newblock A practical guide to training restricted boltzmann machines.
\newblock In Montavon, G., Orr, G.~B., and Müller, K.-R., editors, {\em Neural
  Networks: Tricks of the Trade (2nd ed.)}, volume 7700 of {\em Lecture Notes
  in Computer Science}, pages 599--619. Springer.

\bibitem[Hinton et~al., 1986]{Hinton1986PRALLEL}
Hinton, G.~E., McClelland, J.~L., and Rumelhart, D.~E. (1986).
\newblock Parallel distributed processing: Explorations in the microstructure
  of cognition, vol. 1.
\newblock chapter Distributed Representations, pages 77--109. MIT Press,
  Cambridge, MA, USA.

\bibitem[Hinton et~al., 2006a]{hinton06}
Hinton, G.~E., Osindero, S., and Teh, Y.-W. (2006a).
\newblock {A fast learning algorithm for deep belief nets}.
\newblock {\em Neural Comput.}, 18(7):1527--1554.

\bibitem[Hinton et~al., 2006b]{hinton06NC}
Hinton, G.~E., Osindero, S., and Teh, Y.~W. (2006b).
\newblock A fast learning algorithm for deep belief nets.
\newblock {\em Neural Computation}, 18(7):1527--1554.

\bibitem[Hinton et~al., 2018]{capsulese2018matrix}
Hinton, G.~E., Sabour, S., and Frosst, N. (2018).
\newblock Matrix capsules with {EM} routing.
\newblock In {\em International Conference on Learning Representations}.

\bibitem[Hinton and Salakhutdinov, 2006]{hinton06SCI}
Hinton, G.~E. and Salakhutdinov, R. (2006).
\newblock Reducing the dimensionality of data with neural networks.
\newblock {\em Science}, 313(5786):504--507.

\bibitem[Hinton et~al., 1984]{hintontechrepo1984}
Hinton, G.~E., Sejnowski, T.~J., and Ackley, D.~H. (1984).
\newblock {B}oltzmann machines: {C}onstraint satisfaction networks that learn.
\newblock Technical Report CMU-CS-84-119, Computer Science Department, Carnegie
  Mellon University, Pittsburgh, PA.

\bibitem[Ho, 1995]{ho1995}
Ho, T.~K. (1995).
\newblock Random decision forests.
\newblock In {\em Proceedings of the Third International Conference on Document
  Analysis and Recognition (Volume 1) - Volume 1}, ICDAR '95, pages 278--,
  Washington, DC, USA. IEEE Computer Society.

\bibitem[Hochreiter, 1991]{Hochreiter91}
Hochreiter, S. (1991).
\newblock {Untersuchungen zu dynamischen neuronalen Netzen. Diploma thesis,
  Institut f\"{u}r Informatik, Lehrstuhl Prof. Brauer, Technische
  Universit\"{a}t M\"{u}nchen}.
\newblock Advisor: J. Schmidhuber.

\bibitem[Hochreiter et~al., 2001]{Hochreiter2001}
Hochreiter, S., Bengio, Y., Frasconi, P., and Schmidhuber, J. (2001).
\newblock Gradient flow in recurrent nets: the difficulty of learning long-term
  dependencies.
\newblock In Kremer and Kolen, editors, {\em A Field Guide to Dynamical
  Recurrent Neural Networks}. IEEE Press.

\bibitem[Hochreiter and Schmidhuber, 1997]{Hochreiter1997LSM}
Hochreiter, S. and Schmidhuber, J. (1997).
\newblock Long short-term memory.
\newblock {\em Neural Comput.}, 9(8):1735--1780.

\bibitem[Hoffman et~al., 2016]{HoffmanWYD16}
Hoffman, J., Wang, D., Yu, F., and Darrell, T. (2016).
\newblock Fcns in the wild: Pixel-level adversarial and constraint-based
  adaptation.
\newblock {\em CoRR}, abs/1612.02649.

\bibitem[Hornik et~al., 1989]{hornik1989}
Hornik, K., Stinchcombe, M., and White, H. (1989).
\newblock Multilayer feedforward networks are universal approximators.
\newblock {\em Neural Networks}, 2(5):359--366.

\bibitem[Hornik et~al., 1990]{Hornik1990}
Hornik, K., Stinchcombe, M., and White, H. (1990).
\newblock Universal approximation of an unknown mapping and its derivatives
  using multilayer feedforward networks.
\newblock {\em Neural Netw.}, 3(5):551--560.

\bibitem[Hosu and Rebedea, 2016]{HosuR16corr}
Hosu, I. and Rebedea, T. (2016).
\newblock Playing atari games with deep reinforcement learning and human
  checkpoint replay.
\newblock {\em CoRR}, abs/1607.05077.

\bibitem[Huang and Jain, 2013]{Huang13}
Huang, G.~B. and Jain, V. (2013).
\newblock Deep and wide multiscale recursive networks for robust image
  labeling.
\newblock {\em CoRR}, abs/1310.0354.

\bibitem[Huang et~al., 2009]{Huang2009}
Huang, S.~H., Chu, Y.~H., Lai, S.~H., and Novak, C.~L. (2009).
\newblock {Learning-Based Vertebra Detection and Iterative Normalized-Cut
  Segmentation for Spinal MRI}.
\newblock {\em IEEE Transactions on Medical Imaging}, 28(10):1595--1605.

\bibitem[Hubel and Wiesel, 1962]{Hubel62}
Hubel, D.~H. and Wiesel, T. (1962).
\newblock Receptive fields, binocular interaction, and functional architecture
  in the cat's visual cortex.
\newblock {\em Journal of Physiology (London)}, 160:106--154.

\bibitem[III, 2007]{daumeiii2007ACLMain}
III, H.~D. (2007).
\newblock Frustratingly easy domain adaptation.
\newblock In {\em Proceedings of the 45th Annual Meeting of the Association of
  Computational Linguistics}, pages 256--263, Prague, Czech Republic.
  Association for Computational Linguistics.

\bibitem[(III.), 1962]{Ridgway1962}
(III.), W. C.~R. (1962).
\newblock {\em An Adaptive Logic System with Generalizing Properties}.
\newblock PhD thesis, Stanford University, Stanford Electronics Labs.

\bibitem[Ioffe and Szegedy, 2015]{IoffeICML15}
Ioffe, S. and Szegedy, C. (2015).
\newblock Batch normalization: Accelerating deep network training by reducing
  internal covariate shift.
\newblock In {\em Proceedings of the 32nd International Conference on Machine
  Learning, {ICML} 2015, Lille, France, 6-11 July 2015}, pages 448--456.

\bibitem[Irie and Miyake, 1988]{irie1988capabilities}
Irie, B. and Miyake, S. (1988).
\newblock Capabilities of three-layered perceptrons.
\newblock In {\em IEEE International Conference on Neural Networks}, volume~1,
  page 218.

\bibitem[Ivakhnenko, 1971]{ivakhnenko1971}
Ivakhnenko, A.~G. (1971).
\newblock Polynomial theory of complex systems.
\newblock {\em IEEE Transactions on Systems, Man and Cybernetics},
  (4):364--378.

\bibitem[Ivakhnenko and Lapa, 1965]{ivakhnenko1965}
Ivakhnenko, A.~G. and Lapa, V.~G. (1965).
\newblock {\em Cybernetic Predicting Devices}.
\newblock CCM Information Corporation.

\bibitem[Ivakhnenko et~al., 1967]{ivakhnenko1967}
Ivakhnenko, A.~G., Lapa, V.~G., and McDonough, R.~N. (1967).
\newblock {\em Cybernetics and forecasting techniques}.
\newblock American Elsevier, NY.

\bibitem[Jaderberg et~al., 2014]{JaderbergSVZ14b2014}
Jaderberg, M., Simonyan, K., Vedaldi, A., and Zisserman, A. (2014).
\newblock Deep structured output learning for unconstrained text recognition.
\newblock {\em CoRR}, abs/1412.5903.

\bibitem[James et~al., 2014]{James2014book}
James, G., Witten, D., Hastie, T., and Tibshirani, R. (2014).
\newblock {\em An Introduction to Statistical Learning: With Applications in
  R}.
\newblock Springer Publishing Company, Incorporated.

\bibitem[Jarrett et~al., 2009]{JarrettKRL09ICCV}
Jarrett, K., Kavukcuoglu, K., Ranzato, M., and LeCun, Y. (2009).
\newblock What is the best multi-stage architecture for object recognition?
\newblock In {\em ICCV 2009}, pages 2146--2153.

\bibitem[Jiang and Zhai, 2007]{JiangZ07acl}
Jiang, J. and Zhai, C. (2007).
\newblock Instance weighting for domain adaptation in {NLP}.
\newblock In {\em {ACL} 2007, Proceedings of the 45th Annual Meeting of the
  Association for Computational Linguistics, June 23-30, 2007, Prague, Czech
  Republic}.

\bibitem[Jiang, 2015]{Jiang2015}
Jiang, X. (2015).
\newblock Representational transfer in deep belief networks.
\newblock In {\em 28th Canadian Conference on Artificial Intelligence}, pages
  338--342.

\bibitem[Jones, 1999]{jones99}
Jones, D.~T. (1999).
\newblock {Protein secondary structure prediction based on position-specific
  scoring matrices}.
\newblock {\em Journal of Molecular Biology}, 292(2):195--202.

\bibitem[Joulin and Mikolov, 2015]{JoulinM15NIPS}
Joulin, A. and Mikolov, T. (2015).
\newblock Inferring algorithmic patterns with stack-augmented recurrent nets.
\newblock In {\em Advances in Neural Information Processing Systems 28: Annual
  Conference on Neural Information Processing Systems 2015, December 7-12,
  2015, Montreal, Quebec, Canada}, pages 190--198.

\bibitem[J{\'{o}}zefowicz et~al., 2016]{JozefowiczVSSW16corr}
J{\'{o}}zefowicz, R., Vinyals, O., Schuster, M., Shazeer, N., and Wu, Y.
  (2016).
\newblock Exploring the limits of language modeling.
\newblock {\em CoRR}, abs/1602.02410.

\bibitem[Kadoury et~al., 2011]{Kadoury2011}
Kadoury, S., Labelle, H., and Paragios, N. (2011).
\newblock Automatic inference of articulated spine models in {CT} images using
  high-order markov random fields.
\newblock {\em Medical Image Analysis}, 15(4):426--437.

\bibitem[Kaido et~al., 2013]{Kaido2013}
Kaido, T., Ogawa, K., Fujimoto, Y., Ogura, Y., Hata, K., Ito, T., Tomiyama, K.,
  Yagi, S., Mori, A., and Uemoto, S. (2013).
\newblock Impact of sarcopenia on survival in patients undergoing living donor
  liver transplantation.
\newblock {\em American Journal of Transplantation}, 13(6):1549--1556.

\bibitem[Karpathy and Li, 2015]{karpathyL15}
Karpathy, A. and Li, F. (2015).
\newblock Deep visual-semantic alignments for generating image descriptions.
\newblock In {\em {IEEE} Conference on Computer Vision and Pattern Recognition,
  {CVPR} 2015, Boston, MA, USA, June 7-12, 2015}, pages 3128--3137.

\bibitem[Kearns and Vazirani, 1994]{Kearns1994book}
Kearns, M.~J. and Vazirani, U.~V. (1994).
\newblock {\em An Introduction to Computational Learning Theory}.
\newblock MIT Press, Cambridge, MA, USA.

\bibitem[Kelley, 1960]{kelley1960gradient}
Kelley, H.~J. (1960).
\newblock Gradient theory of optimal flight paths.
\newblock {\em Ars Journal}, 30(10):947--954.

\bibitem[Kendall et~al., 2017]{KendallGC17corr}
Kendall, A., Gal, Y., and Cipolla, R. (2017).
\newblock Multi-task learning using uncertainty to weigh losses for scene
  geometry and semantics.
\newblock {\em CoRR}, abs/1705.07115.

\bibitem[Kim, 2014a]{Kim14fcorr}
Kim, Y. (2014a).
\newblock Convolutional neural networks for sentence classification.
\newblock {\em CoRR}, abs/1408.5882.

\bibitem[Kim, 2014b]{Kim14}
Kim, Y. (2014b).
\newblock Convolutional neural networks for sentence classification.
\newblock In {\em Proceedings of the 2014 Conference on Empirical Methods in
  Natural Language Processing, {EMNLP} 2014, October 25-29, 2014, Doha, Qatar,
  {A} meeting of SIGDAT, a Special Interest Group of the {ACL}}, pages
  1746--1751.

\bibitem[Kingma and Ba, 2014]{KingmaB14}
Kingma, D.~P. and Ba, J. (2014).
\newblock Adam: A method for stochastic optimization.
\newblock {\em CoRR}, abs/1412.6980.

\bibitem[Kingma and Welling, 2013]{KingmaW13}
Kingma, D.~P. and Welling, M. (2013).
\newblock Auto-encoding variational bayes.
\newblock {\em CoRR}, abs/1312.6114.

\bibitem[Kiros et~al., 2014]{KirosSZ14CORRARXIV}
Kiros, R., Salakhutdinov, R., and Zemel, R.~S. (2014).
\newblock Unifying visual-semantic embeddings with multimodal neural language
  models.
\newblock {\em CoRR}, abs/1411.2539.

\bibitem[Kolda and Bader, 2009]{kolda2009}
Kolda, T.~G. and Bader, B.~W. (2009).
\newblock Tensor decompositions and applications.
\newblock {\em SIAM Rev.}, 51(3):455--500.

\bibitem[Kolmogorov, 1957]{kolmogorov1957}
Kolmogorov, A.~K. (1957).
\newblock On the representation of continuous functions of several variables by
  superposition of continuous functions of one variable and addition.
\newblock {\em Doklady Akademii Nauk SSSR}, 114:369--373.

\bibitem[Kolmogorov, 1965]{Kolmogorov65}
Kolmogorov, A.~N. (1965).
\newblock On the representation of continuous functions of several variables by
  superposition of continuous functions of one variable and addition.
\newblock {\em Doklady Akademii. Nauk USSR,}, 114:679--681.

\bibitem[Krizhevsky, 2009]{krizhevsky09learningmultiple}
Krizhevsky, A. (2009).
\newblock Learning multiple layers of features from tiny images.
\newblock Technical report.

\bibitem[Krizhevsky et~al., 2012]{krizhevsky12}
Krizhevsky, A., Sutskever, I., and Hinton, G.~E. (2012).
\newblock {ImageNet Classification with Deep Convolutional Neural Networks}.
\newblock In Pereira, F., Burges, C., Bottou, L., and Weinberger, K., editors,
  {\em {Advances in Neural Information Processing Systems 25}}, pages
  1097--1105. Curran Associates, Inc.

\bibitem[Krueger et~al., 2017]{krueger2017deep}
Krueger, D., Ballas, N., Jastrzebski, S., Arpit, D., Kanwal, M.~S., Maharaj,
  T., Bengio, E., Fischer, A., and Courville, A. (2017).
\newblock Deep nets don't learn via memorization.

\bibitem[Krupka and Tishby, 2007]{krupka2007}
Krupka, E. and Tishby, N. (2007).
\newblock Incorporating prior knowledge on features into learning.
\newblock In {\em Proceedings of the Eleventh International Conference on
  Artificial Intelligence and Statistics, {AISTATS} 2007, San Juan, Puerto
  Rico, March 21-24, 2007}, pages 227--234.

\bibitem[Kumar et~al., 2016]{kumar16pmlr}
Kumar, A., Irsoy, O., Ondruska, P., Iyyer, M., Bradbury, J., Gulrajani, I.,
  Zhong, V., Paulus, R., and Socher, R. (2016).
\newblock Ask me anything: Dynamic memory networks for natural language
  processing.
\newblock In Balcan, M.~F. and Weinberger, K.~Q., editors, {\em Proceedings of
  The 33rd International Conference on Machine Learning}, volume~48 of {\em
  Proceedings of Machine Learning Research}, pages 1378--1387, New York, New
  York, USA. PMLR.

\bibitem[Lafferty et~al., 2001]{lafferty01}
Lafferty, J.~D., McCallum, A., and Pereira, F. C.~N. (2001).
\newblock {Conditional Random Fields: Probabilistic Models for Segmenting and
  Labeling Sequence Data}.
\newblock In {\em { ICML}}, pages 282--289.

\bibitem[Lai, 2015]{Lai2015}
Lai, M. (2015).
\newblock Deep learning for medical image segmentation.
\newblock {\em CoRR}, abs/1505.02000.

\bibitem[Lanic et~al., 2014]{Lanic2014}
Lanic, H., Kraut-Tauzia, J., Modzelewski, R., Clatot, F., Mareschal, S.,
  Picquenot, J.~M., Stamatoullas, A., Leprêtre, S., Tilly, H., and Jardin, F.
  (2014).
\newblock Sarcopenia is an independent prognostic factor in elderly patients
  with diffuse large b-cell lymphoma treated with immunochemotherapy.
\newblock {\em Leukemia \& Lymphoma}, 55(4):817--823.

\bibitem[Lawrence and Platt, 2004]{LawrenceP04icml}
Lawrence, N.~D. and Platt, J.~C. (2004).
\newblock Learning to learn with the informative vector machine.
\newblock In {\em Machine Learning, Proceedings of the Twenty-first
  International Conference {(ICML} 2004), Banff, Alberta, Canada, July 4-8,
  2004}.

\bibitem[Le et~al., 2012]{le12}
Le, V., Brandt, J., Lin, Z., Bourdev, L.~D., and Huang, T.~S. (2012).
\newblock {Interactive Facial Feature Localization}.
\newblock In {\em {ECCV, 2012, Proceedings, Part {III}}}, pages 679--692.

\bibitem[LeCun, 1985]{LeCun85}
LeCun, Y. (1985).
\newblock Une proc\'{e}dure d'apprentissage pour r\'{e}seau \`{a} seuil
  asym\'{e}trique.
\newblock {\em Proceedings of Cognitiva 85, Paris}, pages 599--604.

\bibitem[LeCun et~al., 1989a]{LeCun89}
LeCun, Y., Boser, B., Denker, J.~S., Henderson, D., Howard, R.~E., Hubbard, W.,
  and Jackel, L.~D. (1989a).
\newblock Back-propagation applied to handwritten zip code recognition.
\newblock {\em Neural Computation}, 1(4):541--551.

\bibitem[LeCun et~al., 1989b]{leCun1989}
LeCun, Y., Boser, B., Denker, J.~S., Henderson, D., Howard, R.~E., Hubbard, W.,
  and Jackel, L.~D. (1989b).
\newblock Backpropagation applied to handwritten zip code recognition.
\newblock {\em Neural Comput.}, 1(4):541--551.

\bibitem[LeCun et~al., 1990]{LeCun90}
LeCun, Y., Boser, B., Denker, J.~S., Henderson, D., Howard, R.~E., Hubbard, W.,
  and Jackel, L.~D. (1990).
\newblock Handwritten digit recognition with a back-propagation network.
\newblock In Touretzky, D.~S., editor, {\em Advances in Neural Information
  Processing Systems 2}, pages 396--404. Morgan Kaufmann.

\bibitem[Lecun et~al., 1998]{lecun98gradient}
Lecun, Y., Bottou, L., Bengio, Y., and Haffner, P. (1998).
\newblock Gradient-based learning applied to document recognition.
\newblock In {\em Proceedings of the IEEE}, pages 2278--2324.

\bibitem[LeCun et~al., 1998]{LeCun1998backprop}
LeCun, Y., Bottou, L., Orr, G.~B., and M\"{u}ller, K.-R. (1998).
\newblock Efficient backprop.
\newblock In {\em Neural Networks: Tricks of the Trade, This Book is an
  Outgrowth of a 1996 NIPS Workshop}, pages 9--50, London, UK, UK.
  Springer-Verlag.

\bibitem[Lee et~al., 2009]{lee2009}
Lee, H., Grosse, R., Ranganath, R., and Ng, A.~Y. (2009).
\newblock Convolutional deep belief networks for scalable unsupervised learning
  of hierarchical representations.
\newblock In {\em Proceedings of the 26th Annual International Conference on
  Machine Learning}, ICML '09, pages 609--616, New York, NY, USA. ACM.

\bibitem[Leibniz, 1676]{leibniz1676}
Leibniz, G.~W. (1676).
\newblock Memoir using the chain rule (cited in {TMME} 7:2\&3 p 321-332, 2010).

\bibitem[Lerouge et~al., 2015]{lerouge15}
Lerouge, J., Herault, R., Chatelain, C., Jardin, F., and Modzelewski, R.
  (2015).
\newblock {IODA : An input / output deep architecture for image labeling}.
\newblock {\em Pattern Recognition}, 48(9):2847--2858.

\bibitem[Leshno et~al., 1993]{LeshnoLPS93NN}
Leshno, M., Lin, V.~Y., Pinkus, A., and Schocken, S. (1993).
\newblock Multilayer feedforward networks with a nonpolynomial activation
  function can approximate any function.
\newblock {\em Neural Networks}, 6(6):861--867.

\bibitem[L'H\^{o}pital, 1696]{de1716analyse}
L'H\^{o}pital, G. F.~A. (1696).
\newblock {\em Analyse des infiniment petits, pour l'intelligence des lignes
  courbes}.
\newblock Paris: L'Imprimerie Royale.

\bibitem[Li et~al., 2016a]{li2016pruning}
Li, H., Kadav, A., Durdanovic, I., Samet, H., and Graf, H.~P. (2016a).
\newblock Pruning filters for efficient convnets.
\newblock {\em arXiv preprint arXiv:1608.08710}.

\bibitem[Li et~al., 2016b]{LiUBBSB16}
Li, X., Uricchio, T., Ballan, L., Bertini, M., Snoek, C. G.~M., and Bimbo,
  A.~D. (2016b).
\newblock Socializing the semantic gap: {A} comparative survey on image tag
  assignment, refinement, and retrieval.
\newblock {\em {ACM} Comput. Surv.}, 49(1):14:1--14:39.

\bibitem[Light, 1992]{light1992ridge}
Light, W. (1992).
\newblock Ridge functions, sigmoidal functions and neural networks.
\newblock {\em Approximation theory VII}, pages 163--206.

\bibitem[Linnainmaa, 1970]{Linnainmaa1970}
Linnainmaa, S. (1970).
\newblock The representation of the cumulative rounding error of an algorithm
  as a {Taylor} expansion of the local rounding errors.
\newblock Master's thesis, Univ. Helsinki.

\bibitem[Linnainmaa, 1976]{Linnainmaa1976}
Linnainmaa, S. (1976).
\newblock Taylor expansion of the accumulated rounding error.
\newblock {\em BIT Numerical Mathematics}, 16(2):146--160.

\bibitem[Lippmann, 1988]{Lippmann1988}
Lippmann, R.~P. (1988).
\newblock An introduction to computing with neural nets.
\newblock {\em SIGARCH Computer Architecture News}, 16(1):7--25.

\bibitem[Liu and Nocedal, 1989]{Liu1989}
Liu, D.~C. and Nocedal, J. (1989).
\newblock On the limited memory bfgs method for large scale optimization.
\newblock {\em Math. Program.}, 45(3):503--528.

\bibitem[Liu et~al., 2014]{liu2014}
Liu, S., Yang, N., Li, M., and Zhou, M. (2014).
\newblock A recursive recurrent neural network for statistical machine
  translation.
\newblock In {\em Proceedings of the 52nd Annual Meeting of the Association for
  Computational Linguistics (Volume 1: Long Papers)}, pages 1491--1500,
  Baltimore, Maryland. Association for Computational Linguistics.

\bibitem[Long et~al., 2015]{longSD15}
Long, J., Shelhamer, E., and Darrell, T. (2015).
\newblock Fully convolutional networks for semantic segmentation.
\newblock In {\em {IEEE} Conference on Computer Vision and Pattern Recognition,
  {CVPR} 2015, Boston, MA, USA, June 7-12, 2015}, pages 3431--3440.

\bibitem[Long and Wang, 2015]{Long015acorr}
Long, M. and Wang, J. (2015).
\newblock Learning multiple tasks with deep relationship networks.
\newblock {\em CoRR}, abs/1506.02117.

\bibitem[Lu et~al., 2017]{LuKZCJF17cvpr}
Lu, Y., Kumar, A., Zhai, S., Cheng, Y., Javidi, T., and Feris, R.~S. (2017).
\newblock Fully-adaptive feature sharing in multi-task networks with
  applications in person attribute classification.
\newblock In {\em {CVPR}}, pages 1131--1140. {IEEE} Computer Society.

\bibitem[Luenberger, 1969]{Luenberger1969}
Luenberger, D.~G. (1969).
\newblock {\em {Optimization by vector space methods}}.
\newblock Decision and control. Wiley, New York, NY.

\bibitem[Ma and Lu, 2013]{Ma2013}
Ma, J. and Lu, L. (2013).
\newblock {Hierarchical segmentation and identification of thoracic vertebra
  using learning-based edge detection and coarse-to-fine deformable model}.
\newblock {\em Computer Vision and Image Understanding}, 117(9):1072--1083.

\bibitem[Maas et~al., 2013]{Maas13rectifiernonlinearities}
Maas, A.~L., Hannun, A.~Y., and Ng, A.~Y. (2013).
\newblock Rectifier nonlinearities improve neural network acoustic models.
\newblock In {\em in ICML Workshop on Deep Learning for Audio, Speech and
  Language Processing}.

\bibitem[Mahmud and Ray, 2007]{MahmudR07nips}
Mahmud, M.~M. and Ray, S.~R. (2007).
\newblock Transfer learning using kolmogorov complexity: Basic theory and
  empirical evaluations.
\newblock In {\em Advances in Neural Information Processing Systems 20,
  Proceedings of the Twenty-First Annual Conference on Neural Information
  Processing Systems, Vancouver, British Columbia, Canada, December 3-6, 2007},
  pages 985--992.

\bibitem[Major et~al., 2013]{Major13}
Major, D., Hladůvka, J., Schulze, F., and B\"{u}hler, K. (2013).
\newblock {Automated landmarking and labeling of fully and partially scanned
  spinal columns in CT images}.
\newblock {\em Medical Image Analysis}, 17(8):1151--1163.

\bibitem[Makhzani et~al., 2015]{MakhzaniSJG15CORR}
Makhzani, A., Shlens, J., Jaitly, N., and Goodfellow, I.~J. (2015).
\newblock Adversarial autoencoders.
\newblock {\em CoRR}, abs/1511.05644.

\bibitem[Malon et~al., 2008]{histo-cnn}
Malon, C., Miller, M., Burger, H.~C., Cosatto, E., and Graf, H.~P. (2008).
\newblock Identifying histological elements with convolutional neural networks.
\newblock In {\em Int. Conf. on Soft Computing As Transdisciplinary Science and
  Technology}, pages 450--456.

\bibitem[Marcus, 2018]{marcus2018corr}
Marcus, G. (2018).
\newblock Deep learning: {A} critical appraisal.
\newblock {\em CoRR}, abs/1801.00631.

\bibitem[Marcus, 2003]{marcus2003algebraic}
Marcus, G.~F. (2003).
\newblock {\em The algebraic mind: Integrating connectionism and cognitive
  science}.
\newblock MIT press.

\bibitem[Martens, 2010]{Martens10}
Martens, J. (2010).
\newblock Deep learning via hessian-free optimization.
\newblock In {\em Proceedings of the 27th International Conference on Machine
  Learning (ICML-10), June 21-24, 2010, Haifa, Israel}, pages 735--742.

\bibitem[Martens and Sutskever, 2011]{Martens2011hessfreeICML}
Martens, J. and Sutskever, I. (2011).
\newblock Learning recurrent neural networks with {Hessian}-free optimization.
\newblock In {\em ICML 2011}, pages 1033--1040.

\bibitem[Martin et~al., 2013]{Martin2013}
Martin, L., Birdsell, L., MacDonald, N., Reiman, T., Clandinin, M.~T.,
  McCargar, L.~J., Murphy, R., Ghosh, S., Sawyer, M.~B., and Baracos, V.~E.
  (2013).
\newblock {Cancer cachexia in the age of obesity: Skeletal muscle depletion is
  a powerful prognostic factor, independent of body mass index}.
\newblock {\em Journal of Clinical Oncology}, 31(12):1539--1547.

\bibitem[Masci et~al., 2011]{Masci2011}
Masci, J., Meier, U., Cire{\c{s}}an, D., and Schmidhuber, J. (2011).
\newblock {\em Stacked Convolutional Auto-Encoders for Hierarchical Feature
  Extraction}, pages 52--59.
\newblock Springer Berlin Heidelberg, Berlin, Heidelberg.

\bibitem[McCulloch and Pitts, 1943]{McCulloch1943}
McCulloch, W.~S. and Pitts, W. (1943).
\newblock A logical calculus of the ideas immanent in nervous activity.
\newblock {\em The bulletin of mathematical biophysics}, 5(4):115--133.

\bibitem[McInerney and Terzopoulos, 1996]{mcinerney1996deformable}
McInerney, T. and Terzopoulos, D. (1996).
\newblock Deformable models in medical image analysis: a survey.
\newblock {\em Medical image analysis}, 1(2):91--108.

\bibitem[{Michael Kelm} et~al., 2013]{MichaelKelm2013}
{Michael Kelm}, B., Wels, M., {Kevin Zhou}, S., Seifert, S., Suehling, M.,
  Zheng, Y., and Comaniciu, D. (2013).
\newblock {Spine detection in CT and MR using iterated marginal space
  learning}.
\newblock {\em Medical Image Analysis}, 17(8):1283--1292.

\bibitem[Mihalkova et~al., 2007]{MihalkovaHM07aaai}
Mihalkova, L., Huynh, T.~N., and Mooney, R.~J. (2007).
\newblock Mapping and revising markov logic networks for transfer learning.
\newblock In {\em Proceedings of the Twenty-Second {AAAI} Conference on
  Artificial Intelligence, July 22-26, 2007, Vancouver, British Columbia,
  Canada}, pages 608--614.

\bibitem[Mihalkova and Mooney, 2006]{mihalkova2006transfer}
Mihalkova, L. and Mooney, R. (2006).
\newblock Transfer learning with markov logic networks.
\newblock In {\em ICML workshop on structural knowledge transfer for machine
  learning}.

\bibitem[Mihalkova and Mooney, 2008]{mihalkova2008transfer}
Mihalkova, L. and Mooney, R.~J. (2008).
\newblock Transfer learning by mapping with minimal target data.
\newblock In {\em Proceedings of the AAAI-08 workshop on transfer learning for
  complex tasks}.

\bibitem[Mikolov, 2012]{mikolov2012statistical}
Mikolov, T. (2012).
\newblock {\em Statistical language models based on neural networks}.
\newblock PhD thesis, Brno University of Technology.

\bibitem[Mikolov et~al., 2013]{MikolovSCCD13nips}
Mikolov, T., Sutskever, I., Chen, K., Corrado, G.~S., and Dean, J. (2013).
\newblock Distributed representations of words and phrases and their
  compositionality.
\newblock In {\em {NIPS}}, pages 3111--3119.

\bibitem[Minsky and Papert, 1969]{minsky69perceptrons}
Minsky, M. and Papert, S. (1969).
\newblock {\em Perceptrons: An Introduction to Computational Geometry}.
\newblock MIT Press, Cambridge, MA, USA.

\bibitem[Minsky and Papert, 1988]{Minsky1988}
Minsky, M.~L. and Papert, S.~A. (1988).
\newblock {\em Perceptrons: Expanded Edition}.
\newblock MIT Press, Cambridge, MA, USA.

\bibitem[Misra et~al., 2016]{MisraSGH16cvpr}
Misra, I., Shrivastava, A., Gupta, A., and Hebert, M. (2016).
\newblock Cross-stitch networks for multi-task learning.
\newblock In {\em {CVPR}}, pages 3994--4003. {IEEE} Computer Society.

\bibitem[Mitchell, 1980]{Mitchell80indbias}
Mitchell, T.~M. (1980).
\newblock The need for biases in learning generalizations.
\newblock Technical report, Rutgers University, New Brunswick, NJ.

\bibitem[Mitchell, 1997]{mitchell1997}
Mitchell, T.~M. (1997).
\newblock {\em Machine Learning}.
\newblock McGraw-Hill, Inc., New York, NY, USA, 1 edition.

\bibitem[Mitsiopoulos et~al., 1998]{Mitsiopoulos1998}
Mitsiopoulos, N., Baumgartner, R.~N., Heymsfield, S.~B., Lyons, W., Gallagher,
  D., and Ross, R. (1998).
\newblock {Cadaver validation of skeletal muscle measurement by magnetic
  resonance imaging and computerized tomography.}
\newblock {\em Journal of applied physiology}, 85(1):115--122.

\bibitem[Mnih et~al., 2013]{mnihatari2013}
Mnih, V., Kavukcuoglu, K., Silver, D., Graves, A., Antonoglou, I., Wierstra,
  D., and Riedmiller, M. (2013).
\newblock Playing atari with deep reinforcement learning.
\newblock In {\em NIPS Deep Learning Workshop}.

\bibitem[Mnih et~al., 2015]{MnihKSRVBGRFOPB15Nature}
Mnih, V., Kavukcuoglu, K., Silver, D., Rusu, A.~A., Veness, J., Bellemare,
  M.~G., Graves, A., Riedmiller, M.~A., Fidjeland, A., Ostrovski, G., Petersen,
  S., Beattie, C., Sadik, A., Antonoglou, I., King, H., Kumaran, D., Wierstra,
  D., Legg, S., and Hassabis, D. (2015).
\newblock Human-level control through deep reinforcement learning.
\newblock {\em Nature}, 518(7540):529--533.

\bibitem[Mnih et~al., 2011]{mnihLH11}
Mnih, V., Larochelle, H., and Hinton, G.~E. (2011).
\newblock Conditional restricted boltzmann machines for structured output
  prediction.
\newblock In {\em {UAI} 2011, Proceedings of the Twenty-Seventh Conference on
  Uncertainty in Artificial Intelligence, Barcelona, Spain, July 14-17, 2011},
  pages 514--522.

\bibitem[Mohri et~al., 2012]{Mohri2012bookML}
Mohri, M., Rostamizadeh, A., and Talwalkar, A. (2012).
\newblock {\em Foundations of Machine Learning}.
\newblock The MIT Press.

\bibitem[Molchanov et~al., 2016]{molchanov2016pruning}
Molchanov, P., Tyree, S., Karras, T., Aila, T., and Kautz, J. (2016).
\newblock Pruning convolutional neural networks for resource efficient transfer
  learning.
\newblock {\em arXiv preprint arXiv:1611.06440}.

\bibitem[Moller, 1993]{Moller93}
Moller, M.~F. (1993).
\newblock Exact calculation of the product of the {H}essian matrix of
  feed-forward network error functions and a vector in {O}({N}) time.
\newblock Technical Report PB-432, Computer Science Department, Aarhus
  University, Denmark.

\bibitem[Mont\'{u}far et~al., 2014]{Montufar2014NIPS}
Mont\'{u}far, G., Pascanu, R., Cho, K., and Bengio, Y. (2014).
\newblock On the number of linear regions of deep neural networks.
\newblock In {\em Proceedings of the 27th International Conference on Neural
  Information Processing Systems - Volume 2}, NIPS'14, pages 2924--2932,
  Cambridge, MA, USA. MIT Press.

\bibitem[Mourad and Reilly, 2010]{MouradR2010}
Mourad, N. and Reilly, J.~P. (2010).
\newblock Minimizing nonconvex functions for sparse vector reconstruction.
\newblock {\em {IEEE} Trans. Signal Processing}, 58(7):3485--3496.

\bibitem[Nair and Hinton, 2010]{NairHicml10}
Nair, V. and Hinton, G.~E. (2010).
\newblock Rectified linear units improve restricted boltzmann machines.
\newblock In Fürnkranz, J. and Joachims, T., editors, {\em Proceedings of the
  27th International Conference on Machine Learning (ICML-10)}, pages 807--814.
  Omnipress.

\bibitem[Nanni et~al., 2010]{nanni2010}
Nanni, L., Lumini, A., and Brahnam, S. (2010).
\newblock Local binary patterns variants as texture descriptors for medical
  image analysis.
\newblock {\em Artificial Intelligence in Medicine}, 49(2):117 -- 125.

\bibitem[Natarajan, 1995]{Natarajan1995}
Natarajan, B.~K. (1995).
\newblock Sparse approximate solutions to linear systems.
\newblock {\em SIAM J. Comput.}, 24(2):227--234.

\bibitem[Nesterov, 1983]{Nesterov1983wy}
Nesterov, Y. (1983).
\newblock {A method of solving a convex programming problem with convergence
  rate O(1/sqr(k))}.
\newblock {\em Soviet Mathematics Doklady}, 27:372--376.

\bibitem[Nguyen et~al., 2015]{NguyenYC15cvpr}
Nguyen, A.~M., Yosinski, J., and Clune, J. (2015).
\newblock Deep neural networks are easily fooled: High confidence predictions
  for unrecognizable images.
\newblock In {\em {CVPR}}, pages 427--436. {IEEE} Computer Society.

\bibitem[Nguyen et~al., 2017]{NguyenAOI17corrfiltering}
Nguyen, D.~T., Alam, F., Ofli, F., and Imran, M. (2017).
\newblock Automatic image filtering on social networks using deep learning and
  perceptual hashing during crises.
\newblock {\em CoRR}, abs/1704.02602.

\bibitem[Nicolas et~al., 2006]{nicolas06}
Nicolas, S., Paquet, T., and Heutte, L. (2006).
\newblock {A Markovian Approach for Handwritten Document Segmentation.}
\newblock In {\em {ICPR (3)}}, pages 292--295.

\bibitem[Nielsen, 1987]{hechtnielsenkolmogorov1987}
Nielsen, R.~H. (1987).
\newblock {K}olmogorov's mapping neural network existence theorem.
\newblock In {\em {P}roceedings of the IEEE First International Conference on
  Neural Networks {\rm ({S}an {D}iego, {CA})}}, volume III, pages 11--13.
  Piscataway, NJ: IEEE.

\bibitem[Ning et~al., 2005]{ningDLPBB05}
Ning, F., Delhomme, D., LeCun, Y., Piano, F., Bottou, L., and Barbano, P.~E.
  (2005).
\newblock Toward automatic phenotyping of developing embryos from videos.
\newblock {\em {IEEE} Trans. Image Processing}, 14(9):1360--1371.

\bibitem[Niyogi et~al., 1998]{niyogi98}
Niyogi, P., Girosi, F., and Poggio, T. (1998).
\newblock Incorporating prior information in machine learning by creating
  virtual examples.
\newblock {\em Proceedings of the IEEE}, 86(11):2196--2209.

\bibitem[Noh et~al., 2015]{nohHH15}
Noh, H., Hong, S., and Han, B. (2015).
\newblock Learning deconvolution network for semantic segmentation.
\newblock In {\em 2015 {IEEE} International Conference on Computer Vision,
  {ICCV} 2015, Santiago, Chile, December 7-13, 2015}, pages 1520--1528.

\bibitem[Noto and Craven, 2012]{noto12}
Noto, K. and Craven, M. (2012).
\newblock {Learning Hidden Markov Models for Regression using Path
  Aggregation}.
\newblock {\em CoRR}, abs/1206.3275.

\bibitem[Novikoff, 1962]{novikoff1962}
Novikoff, A.~B. (1962).
\newblock On convergence proofs on perceptrons.
\newblock In {\em Proceedings of the Symposium on the Mathematical Theory of
  Automata}.

\bibitem[Och, 2003]{och03}
Och, F.~J. (2003).
\newblock {Minimum error rate training in statistical machine translation}.
\newblock In {\em {Proceedings of the ACL}}, volume~1.

\bibitem[Ojala et~al., 2002]{ojala2002}
Ojala, T., Pietikainen, M., and Maenpaa, T. (2002).
\newblock Multiresolution gray-scale and rotation invariant texture
  classification with local binary patterns.
\newblock {\em IEEE Transactions on Pattern Analysis and Machine Intelligence},
  24(7):971--987.

\bibitem[Oktay and Akgul, 2011]{Oktay2011}
Oktay, A.~B. and Akgul, Y.~S. (2011).
\newblock {Localization of the lumbar discs using machine learning and exact
  probabilistic inference}.
\newblock In {\em Lecture Notes in Artificial Intelligence and Lecture Notes in
  Bioinformatics}, pages 158--165.

\bibitem[Olazaran, 1996]{Olazaran1996}
Olazaran, M. (1996).
\newblock A sociological study of the official history of the perceptrons
  controversy.
\newblock {\em Social Studies of Science}, 26(3):611--659.

\bibitem[Olivas et~al., 2009]{Olivas2009HRM}
Olivas, E.~S., Guerrero, J. D.~M., Sober, M.~M., Benedito, J. R.~M., and Lopez,
  A. J.~S. (2009).
\newblock {\em Handbook Of Research On Machine Learning Applications and
  Trends: Algorithms, Methods and Techniques - 2 Volumes}.
\newblock Information Science Reference - Imprint of: IGI Publishing, Hershey,
  PA.

\bibitem[Oster et~al., 2009]{OsterDL09neco}
Oster, M., Douglas, R.~J., and Liu, S. (2009).
\newblock Computation with spikes in a winner-take-all network.
\newblock {\em Neural Computation}, 21(9):2437--2465.

\bibitem[Pan and Yang, 2010]{Pan2010STL}
Pan, S.~J. and Yang, Q. (2010).
\newblock A survey on transfer learning.
\newblock {\em IEEE Trans. on Knowl. and Data Eng.}, 22(10):1345--1359.

\bibitem[Pandey and Dukkipati, 2014]{pmlrpandey14}
Pandey, G. and Dukkipati, A. (2014).
\newblock {To go deep or wide in learning?}
\newblock In Kaski, S. and Corander, J., editors, {\em Proceedings of the
  Seventeenth International Conference on Artificial Intelligence and
  Statistics}, volume~33 of {\em Proceedings of Machine Learning Research},
  pages 724--732, Reykjavik, Iceland. PMLR.

\bibitem[Papernot et~al., 2016]{PapernotAEGT16corr}
Papernot, N., Abadi, M., Erlingsson, {\'{U}}., Goodfellow, I.~J., and Talwar,
  K. (2016).
\newblock Semi-supervised knowledge transfer for deep learning from private
  training data.
\newblock {\em CoRR}, abs/1610.05755.

\bibitem[Paredes et~al., 2012]{pmlrv22romera12}
Paredes, B.~R., Argyriou, A., Berthouze, N., and Pontil, M. (2012).
\newblock Exploiting unrelated tasks in multi-task learning.
\newblock In Lawrence, N.~D. and Girolami, M., editors, {\em Proceedings of the
  Fifteenth International Conference on Artificial Intelligence and
  Statistics}, volume~22 of {\em Proceedings of Machine Learning Research},
  pages 951--959, La Palma, Canary Islands. PMLR.

\bibitem[Park et~al., 2016]{park2016faster}
Park, J., Li, S., Wen, W., Tang, P. T.~P., Li, H., Chen, Y., and Dubey, P.
  (2016).
\newblock Faster cnns with direct sparse convolutions and guided pruning.

\bibitem[Parker, 1985]{Parker85}
Parker, D.~B. (1985).
\newblock Learning-logic.
\newblock Technical Report TR-47, Center for Comp. Research in Economics and
  Management Sci., MIT.

\bibitem[Pascanu et~al., 2012]{Pascanu2012arxiv}
Pascanu, R., Mikolov, T., and Bengio, Y. (2012).
\newblock Understanding the exploding gradient problem.
\newblock {\em CoRR}, abs/1211.5063.

\bibitem[Pascanu et~al., 2013a]{PascanuMB13}
Pascanu, R., Mikolov, T., and Bengio, Y. (2013a).
\newblock On the difficulty of training recurrent neural networks.
\newblock In {\em Proceedings of the 30th International Conference on Machine
  Learning, {ICML} 2013, Atlanta, GA, USA, 16-21 June 2013}, pages 1310--1318.

\bibitem[Pascanu et~al., 2013b]{Pascanu2013}
Pascanu, R., Mikolov, T., and Bengio, Y. (2013b).
\newblock On the difficulty of training recurrent neural networks.
\newblock In {\em Proceedings of the 30th International Conference on
  International Conference on Machine Learning - Volume 28}, ICML'13, pages
  III--1310--III--1318. JMLR.org.

\bibitem[Pearlmutter, 1994]{Pearlmutter93}
Pearlmutter, B.~A. (1994).
\newblock Fast exact multiplication by the {Hessian}.
\newblock {\em Neural Computation}, 6(1):147--160.

\bibitem[Peng et~al., 2011]{Peng2011}
Peng, P., {Van Vledder}, M., Tsai, S., {De Jong}, M., Makary, M., Ng, J., Edil,
  B., Wolfgang, C., Schulick, R., Choti, M., Kamel, I., and Pawlik, T. (2011).
\newblock Sarcopenia negatively impacts short-term outcomes in patients
  undergoing hepatic resection for colorectal liver metastasis.
\newblock {\em HPB}, 13(7):439--446.

\bibitem[Peyr{\'e} et~al., 2017]{peyre17}
Peyr{\'e}, G., Cuturi, M., et~al. (2017).
\newblock Computational optimal transport.
\newblock Technical report.

\bibitem[Pham et~al., 2000]{pham2000current}
Pham, D.~L., Xu, C., and Prince, J.~L. (2000).
\newblock Current methods in medical image segmentation 1.
\newblock {\em Annual review of biomedical engineering}, 2(1):315--337.

\bibitem[Polyak, 1964]{polyak1964some}
Polyak, B.~T. (1964).
\newblock Some methods of speeding up the convergence of iteration methods.
\newblock {\em USSR Computational Mathematics and Mathematical Physics},
  4(5):1--17.

\bibitem[Poole et~al., 2014]{PooleSG14CORR}
Poole, B., Sohl{-}Dickstein, J., and Ganguli, S. (2014).
\newblock Analyzing noise in autoencoders and deep networks.
\newblock {\em CoRR}, abs/1406.1831.

\bibitem[Povey et~al., 2014]{PoveyZK14CORR}
Povey, D., Zhang, X., and Khudanpur, S. (2014).
\newblock Parallel training of deep neural networks with natural gradient and
  parameter averaging.
\newblock {\em CoRR}, abs/1410.7455.

\bibitem[Rabanser et~al., 2017]{rabanser2017arxiv}
Rabanser, S., Shchur, O., and G{\"{u}}nnemann, S. (2017).
\newblock Introduction to tensor decompositions and their applications in
  machine learning.
\newblock {\em CoRR}, abs/1711.10781.

\bibitem[Rabiner, 1989]{rabiner89}
Rabiner, L. (1989).
\newblock {A tutorial on hidden Markov models and selected applications in
  speech recognition}.
\newblock {\em Proceedings of the IEEE}, 77(2):257--286.

\bibitem[Radford et~al., 2015]{RadfordMC15corr}
Radford, A., Metz, L., and Chintala, S. (2015).
\newblock Unsupervised representation learning with deep convolutional
  generative adversarial networks.
\newblock {\em CoRR}, abs/1511.06434.

\bibitem[Raiko et~al., 2012]{raikoPMLR2012}
Raiko, T., Valpola, H., and Lecun, Y. (2012).
\newblock Deep learning made easier by linear transformations in perceptrons.
\newblock In Lawrence, N.~D. and Girolami, M., editors, {\em Proceedings of the
  Fifteenth International Conference on Artificial Intelligence and
  Statistics}, volume~22 of {\em Proceedings of Machine Learning Research},
  pages 924--932, La Palma, Canary Islands. PMLR.

\bibitem[Raina et~al., 2007]{RainaBLPN07icml}
Raina, R., Battle, A., Lee, H., Packer, B., and Ng, A.~Y. (2007).
\newblock Self-taught learning: transfer learning from unlabeled data.
\newblock In {\em Machine Learning, Proceedings of the Twenty-Fourth
  International Conference {(ICML} 2007), Corvallis, Oregon, USA, June 20-24,
  2007}, pages 759--766.

\bibitem[Raina et~al., 2009]{Raina2009LDU}
Raina, R., Madhavan, A., and Ng, A.~Y. (2009).
\newblock Large-scale deep unsupervised learning using graphics processors.
\newblock In {\em Proceedings of the 26th Annual International Conference on
  Machine Learning}, ICML '09, pages 873--880, New York, NY, USA. ACM.

\bibitem[Ramachandran et~al., 2017]{RamachandranLL17emnlp}
Ramachandran, P., Liu, P.~J., and Le, Q.~V. (2017).
\newblock Unsupervised pretraining for sequence to sequence learning.
\newblock In {\em {EMNLP}}, pages 383--391. Association for Computational
  Linguistics.

\bibitem[Ramsundar et~al., 2015]{RamsundarKRWKP15corr}
Ramsundar, B., Kearnes, S.~M., Riley, P., Webster, D., Konerding, D.~E., and
  Pande, V.~S. (2015).
\newblock Massively multitask networks for drug discovery.
\newblock {\em CoRR}, abs/1502.02072.

\bibitem[Ranzato et~al., 2007a]{aurelio07}
Ranzato, A., Poultney, C., Chopra, S., and Lecun, Y. (2007a).
\newblock {Efficient Learning of Sparse Representations with an Energy-Based
  Model}.
\newblock In {\em {NIPS}}, pages 1137--1144.

\bibitem[Ranzato et~al., 2007b]{RanzatoBL2007}
Ranzato, M., Boureau, Y., and LeCun, Y. (2007b).
\newblock Sparse feature learning for deep belief networks.
\newblock In {\em Advances in Neural Information Processing Systems 20,
  Proceedings of the Twenty-First Annual Conference on Neural Information
  Processing Systems, Vancouver, British Columbia, Canada, December 3-6, 2007},
  pages 1185--1192.

\bibitem[Ranzato et~al., 2007c]{ranzato-cvpr-07}
Ranzato, M., Huang, F., Boureau, Y., and LeCun, Y. (2007c).
\newblock Unsupervised learning of invariant feature hierarchies with
  applications to object recognition.
\newblock In {\em Proc. Computer Vision and Pattern Recognition Conference
  (CVPR'07)}. IEEE Press.

\bibitem[Ranzato et~al., 2006]{RanzatoPCL2006}
Ranzato, M., Poultney, C.~S., Chopra, S., and LeCun, Y. (2006).
\newblock Efficient learning of sparse representations with an energy-based
  model.
\newblock In {\em Advances in Neural Information Processing Systems 19,
  Proceedings of the Twentieth Annual Conference on Neural Information
  Processing Systems, Vancouver, British Columbia, Canada, December 4-7, 2006},
  pages 1137--1144.

\bibitem[Razavian et~al., 2014]{RazavianASC14cvpr}
Razavian, A.~S., Azizpour, H., Sullivan, J., and Carlsson, S. (2014).
\newblock {CNN} features off-the-shelf: An astounding baseline for recognition.
\newblock In {\em {CVPR} Workshops}, pages 512--519. {IEEE} Computer Society.

\bibitem[Ren et~al., 2015]{Shaoqing15}
Ren, S., He, K., Girshick, R., and Sun, J. (2015).
\newblock Faster r-cnn: Towards real-time object detection with region proposal
  networks.
\newblock {\em NIPS 28}, pages 91--99.

\bibitem[Riesenhuber and Poggio, 1999]{riesenhuber99}
Riesenhuber, M. and Poggio, T. (1999).
\newblock Hierarchical models of object recognition in cortex.
\newblock {\em Nature Neuroscience}, 2(11).

\bibitem[Rifai et~al., 2011a]{salah2011}
Rifai, S., Mesnil, G., Vincent, P., Muller, X., Bengio, Y., Dauphin, Y., and
  Glorot, X. (2011a).
\newblock Higher order contractive auto-encoder.
\newblock In {\em European Conference on Machine Learning and Principles and
  Practice of Knowledge Discovery in Databases (ECML PKDD)}.

\bibitem[Rifai et~al., 2011b]{rifaiVMGB11}
Rifai, S., Vincent, P., Muller, X., Glorot, X., and Bengio, Y. (2011b).
\newblock Contractive auto-encoders: Explicit invariance during feature
  extraction.
\newblock In {\em Proceedings of the 28th International Conference on Machine
  Learning, {ICML} 2011, Bellevue, Washington, USA, June 28 - July 2, 2011},
  pages 833--840.

\bibitem[Rissanen, 1978]{Rissanen1978}
Rissanen, J. (1978).
\newblock Modeling by shortest data description.
\newblock {\em Automatica}, 14(5):465--471.

\bibitem[Ronneberger et~al., 2015]{ronnebergerFB15}
Ronneberger, O., Fischer, P., and Brox, T. (2015).
\newblock U-net: Convolutional networks for biomedical image segmentation.
\newblock In {\em Medical Image Computing and Computer-Assisted Intervention -
  {MICCAI} 2015 - 18th International Conference Munich, Germany, October 5 - 9,
  2015, Proceedings, Part {III}}, pages 234--241.

\bibitem[Rosenblatt, 1958]{Rosenblatt58}
Rosenblatt, F. (1958).
\newblock The perceptron: A probabilistic model for information storage and
  organization in the brain.
\newblock {\em Psychological Review}, pages 65--386.

\bibitem[Rosenblatt, 1962]{Rosenblatt1962}
Rosenblatt, F. (1962).
\newblock {\em Principles of Neurodynamics: Perceptrons and the Theory of Brain
  Mechanisms}.
\newblock Spartan Books, Washington.

\bibitem[Roth et~al., 2014]{Roth14}
Roth, H.~R., Yao, J., Lu, L., Stieger, J., Burns, J.~E., and Summers, R.~M.
  (2014).
\newblock Detection of sclerotic spine metastases via random aggregation of
  deep convolutional neural network classifications.
\newblock {\em CoRR}, abs/1407.5976.

\bibitem[Ruder et~al., 2017]{RuderBAS17corr}
Ruder, S., Bingel, J., Augenstein, I., and S{\o}gaard, A. (2017).
\newblock Sluice networks: Learning what to share between loosely related
  tasks.
\newblock {\em CoRR}, abs/1705.08142.

\bibitem[Rudin, 1964]{rudin1964principles}
Rudin, W. (1964).
\newblock {\em Principles of mathematical analysis}.
\newblock McGraw-hill New York.

\bibitem[Rumelhart et~al., 1986]{Rumelhart86}
Rumelhart, D.~E., Hinton, G.~E., and Williams, R.~J. (1986).
\newblock Learning internal representations by error propagation.
\newblock In Rumelhart, D.~E. and McClelland, J.~L., editors, {\em Parallel
  Distributed Processing}, volume~1, pages 318--362. MIT Press.

\bibitem[Sabour et~al., 2017]{SabourFH17}
Sabour, S., Frosst, N., and Hinton, G.~E. (2017).
\newblock Dynamic routing between capsules.
\newblock In {\em Advances in Neural Information Processing Systems 30: Annual
  Conference on Neural Information Processing Systems 2017, 4-9 December 2017,
  Long Beach, CA, {USA}}, pages 3859--3869.

\bibitem[Sagonas et~al., 2013]{sagonas13}
Sagonas, C., Tzimiropoulos, G., Zafeiriou, S., and Pantic, M. (2013).
\newblock A semi-automatic methodology for facial landmark annotation.
\newblock In {\em CVPR Workshops}, pages 896--903.

\bibitem[Salakhutdinov and Hinton, 2009]{SalHinton07DBM}
Salakhutdinov, R. and Hinton, G. (2009).
\newblock Deep {B}oltzmann machines.
\newblock In {\em Proceedings of the International Conference on Artificial
  Intelligence and Statistics}, volume~5, pages 448--455.

\bibitem[Salimans et~al., 2015]{SalimansKW15}
Salimans, T., Kingma, D.~P., and Welling, M. (2015).
\newblock Markov chain monte carlo and variational inference: Bridging the gap.
\newblock In {\em Proceedings of the 32nd International Conference on Machine
  Learning, {ICML} 2015, Lille, France, 6-11 July 2015}, pages 1218--1226.

\bibitem[Sathyanarayana et~al., 2016]{Sathyanarayana2016JMIR}
Sathyanarayana, A., Joty, S., Fernandez-Luque, L., Ofli, F., Srivastava, J.,
  Elmagarmid, A., Arora, T., and Taheri, S. (2016).
\newblock Sleep quality prediction from wearable data using deep learning.
\newblock {\em JMIR mHealth and uHealth}, 4(4).

\bibitem[Savva et~al., 2016]{Savva2016}
Savva, A.~D., Economopoulos, T.~L., and Matsopoulos, G.~K. (2016).
\newblock {Geometry-based vs. intensity-based medical image registration: A
  comparative study on 3D CT data}.
\newblock {\em Computers in Biology and Medicine}, 69:120--133.

\bibitem[Schmid, 1994]{schmid94}
Schmid, H. (1994).
\newblock {Part-of-speech tagging with neural networks}.
\newblock {\em conference on Computational linguistics}, 12:44--49.

\bibitem[Schmidhuber, 1989]{Schmidhuber89cs}
Schmidhuber, J. (1989).
\newblock A local learning algorithm for dynamic feedforward and recurrent
  networks.
\newblock {\em Connection Science}, 1(4):403--412.

\bibitem[Schmidhuber, 1992]{schmidhuber1992}
Schmidhuber, J. (1992).
\newblock Learning complex, extended sequences using the principle of history
  compression.
\newblock {\em Neural Computation}, 4(2):234--242.

\bibitem[Schmidhuber, 2013]{mydeep2013}
Schmidhuber, J. (2013).
\newblock My first {Deep Learning} system of 1991 $+$ {Deep Learning} timeline
  1962-2013.
\newblock Technical Report arXiv:1312.5548v1 [cs.NE], The Swiss AI Lab IDSIA.

\bibitem[Schmidhuber, 2015]{SchmidhuberNeuralnets2015}
Schmidhuber, J. (2015).
\newblock Deep learning in neural networks: An overview.
\newblock {\em Neural Networks}, 61:85--117.
\newblock Published online 2014; based on TR arXiv:1404.7828 [cs.NE].

\bibitem[Scholkopf and Smola, 2001]{scholkopf2001}
Scholkopf, B. and Smola, A.~J. (2001).
\newblock {\em Learning with Kernels: Support Vector Machines, Regularization,
  Optimization, and Beyond}.
\newblock MIT Press, Cambridge, MA, USA.

\bibitem[Schuster, 1999]{schuster1999supervisedtehsis}
Schuster, M. (1999).
\newblock {\em On supervised learning from sequential data with applications
  for speech recognition}.
\newblock PhD thesis, Daktaro disertacija, Nara Institute of Science and
  Technology.

\bibitem[Schuster and Paliwal, 1997]{Schuster1997BRNN}
Schuster, M. and Paliwal, K. (1997).
\newblock Bidirectional recurrent neural networks.
\newblock {\em Trans. Sig. Proc.}, 45(11):2673--2681.

\bibitem[Sermanet et~al., 2013]{Sermanet2013CVPR}
Sermanet, P., Kavukcuoglu, K., Chintala, S., and Lecun, Y. (2013).
\newblock Pedestrian detection with unsupervised multi-stage feature learning.
\newblock In {\em Proceedings of the 2013 IEEE Conference on Computer Vision
  and Pattern Recognition}, CVPR '13, pages 3626--3633, Washington, DC, USA.
  IEEE Computer Society.

\bibitem[Shalev-Shwartz and Ben-David, 2014]{Shalev2014bookML}
Shalev-Shwartz, S. and Ben-David, S. (2014).
\newblock {\em Understanding Machine Learning: From Theory to Algorithms}.
\newblock Cambridge University Press, New York, NY, USA.

\bibitem[Shen et~al., 2004]{Shen2004}
Shen, W., Punyanitya, M., Wang, Z., Gallagher, D., St-Onge, M.-P., Albu, J.,
  Heymsfield, S.~B., and Heshka, S. (2004).
\newblock {Total body skeletal muscle and adipose tissue volumes: estimation
  from a single abdominal cross-sectional image.}
\newblock {\em Journal of applied physiology}, 97(6):2333--2338.

\bibitem[Shen et~al., 2014]{ShenHGDM14cikm}
Shen, Y., He, X., Gao, J., Deng, L., and Mesnil, G. (2014).
\newblock A latent semantic model with convolutional-pooling structure for
  information retrieval.
\newblock In {\em {CIKM}}, pages 101--110. {ACM}.

\bibitem[Shimodaira, 2000]{Shimodaira2000}
Shimodaira, H. (2000).
\newblock Improving predictive inference under covariate shift by weighting the
  log-likelihood function.
\newblock {\em Journal of Statistical Planning and Inference}, 90(2):227--244.

\bibitem[Shin et~al., 2016]{lung-transfer}
Shin, H.~C., Roth, H.~R., Gao, M., Lu, L., Xu, Z., Nogues, I., Yao, J.,
  Mollura, D., and Summers, R.~M. (2016).
\newblock Deep convolutional neural networks for computer-aided detection: Cnn
  architectures, dataset characteristics and transfer learning.
\newblock {\em IEEE Transactions on Medical Imaging}, 35(5):1285--1298.

\bibitem[Sietsma and Dow, 1991]{SIETSMA1991}
Sietsma, J. and Dow, R.~J. (1991).
\newblock Creating artificial neural networks that generalize.
\newblock {\em Neural Networks}, 4(1):67 -- 79.

\bibitem[Silver et~al., 2016]{SilverHMGSDSAPL16}
Silver, D., Huang, A., Maddison, C.~J., Guez, A., Sifre, L., van~den Driessche,
  G., Schrittwieser, J., Antonoglou, I., Panneershelvam, V., Lanctot, M.,
  Dieleman, S., Grewe, D., Nham, J., Kalchbrenner, N., Sutskever, I.,
  Lillicrap, T.~P., Leach, M., Kavukcuoglu, K., Graepel, T., and Hassabis, D.
  (2016).
\newblock Mastering the game of go with deep neural networks and tree search.
\newblock {\em Nature}, 529(7587):484--489.

\bibitem[Simard et~al., 1993]{simard1993}
Simard, P., Le~Cun, Y., and Denker, J. (1993).
\newblock {Efficient Pattern Recognition Using a New Transformation Distance}.
\newblock In {\em Advances in Neural Information Processing Systems}, volume~5,
  pages 50--58.

\bibitem[Simard et~al., 1992]{simVicLeCDen92}
Simard, P., Victorri, B., Lecun, Y., and Denker, J. (1992).
\newblock {Tangent Prop --- a formalism for specifying selected invariances in
  an adaptive network}.
\newblock In Moody, J.~E., Hanson, S.~J., and Lippmann, R.~P., editors, {\em
  Advances in Neural Information Processing Systems 4}, pages 895--903, San
  Mateo, CA. Morgan Kaufmann.

\bibitem[Simard et~al., 2003]{Simard2003}
Simard, P.~Y., Steinkraus, D., and Platt, J.~C. (2003).
\newblock Best practices for convolutional neural networks applied to visual
  document analysis.
\newblock In {\em Proceedings of the Seventh International Conference on
  Document Analysis and Recognition - Volume 2}, ICDAR '03, pages 958--,
  Washington, DC, USA. IEEE Computer Society.

\bibitem[Simonyan et~al., 2013]{SimonyanVZ13}
Simonyan, K., Vedaldi, A., and Zisserman, A. (2013).
\newblock Deep inside convolutional networks: Visualising image classification
  models and saliency maps.
\newblock {\em CoRR}, abs/1312.6034.

\bibitem[Simonyan and Zisserman, 2014]{SimonyanZ14aCORR}
Simonyan, K. and Zisserman, A. (2014).
\newblock Very deep convolutional networks for large-scale image recognition.
\newblock {\em CoRR}, abs/1409.1556.

\bibitem[Sjoberg et~al., 1995]{sjoberg1995}
Sjoberg, J., Sjoeberg, J., Sjöberg, J., and Ljung, L. (1995).
\newblock Overtraining, regularization and searching for a minimum, with
  application to neural networks.
\newblock {\em International Journal of Control}, 62:1391--1407.

\bibitem[Sleator and Temperley, 1993]{sleator95}
Sleator, D.~D. and Temperley, D. (1993).
\newblock Parsing {English} with a link grammar.
\newblock In {\em Proc. Third International Workshop on Parsing Technologies},
  pages 277--292.

\bibitem[Smolensky, 1986]{smolensky86}
Smolensky, P. (1986).
\newblock Parallel distributed processing: Explorations in the microstructure
  of cognition, vol. 1.
\newblock chapter Information Processing in Dynamical Systems: Foundations of
  Harmony Theory, pages 194--281. MIT Press, Cambridge, MA, USA.

\bibitem[Socher et~al., 2013]{Socheretal2013}
Socher, R., Perelygin, A., Wu, J., Chuang, J., Manning, C.~D., Ng, A.~Y., and
  Potts, C. (2013).
\newblock Recursive deep models for semantic compositionality over a sentiment
  treebank.
\newblock In {\em Proceedings of the 2013 Conference on {E}mpirical {M}ethods
  in {N}atural {L}anguage {P}rocessing}, pages 1631--1642, Stroudsburg, PA.
  Association for Computational Linguistics.

\bibitem[S{\o}gaard and Goldberg, 2016]{SogaardG16acl}
S{\o}gaard, A. and Goldberg, Y. (2016).
\newblock Deep multi-task learning with low level tasks supervised at lower
  layers.
\newblock In {\em {ACL} {(2)}}. The Association for Computer Linguistics.

\bibitem[Sohn et~al., 2015]{SohnLY15}
Sohn, K., Lee, H., and Yan, X. (2015).
\newblock Learning structured output representation using deep conditional
  generative models.
\newblock In {\em NIPS 2015}, pages 3483--3491.

\bibitem[Srivastava, 2013]{SrivastavaDropout2013}
Srivastava, N. (2013).
\newblock {Improving Neural Networks with Dropout}.
\newblock Master's thesis, University of Toronto, Toronto, Canada.

\bibitem[Srivastava et~al., 2014]{srivastava14a}
Srivastava, N., Hinton, G., Krizhevsky, A., Sutskever, I., and Salakhutdinov,
  R. (2014).
\newblock Dropout: A simple way to prevent neural networks from overfitting.
\newblock {\em Journal of Machine Learning Research}, 15:1929--1958.

\bibitem[Srivastava et~al., 2016]{Srivastava2016arxiv}
Srivastava, R.~K., Greff, K., and Schmidhuber, J. (2016).
\newblock {Highway Networks}.

\bibitem[Starck et~al., 2015]{starckmurtaghfadili2015}
Starck, J.-L., Murtagh, F., and Fadili, J. (2015).
\newblock {\em Dictionary Learning}, page 263–274.
\newblock Cambridge University Press, 2 edition.

\bibitem[Steffens, 2007]{steffens2007history}
Steffens, K.-G. (2007).
\newblock {\em The history of approximation theory: from Euler to Bernstein}.
\newblock Springer Science \& Business Media.

\bibitem[Steinkrau et~al., 2005]{Steinkrau2005UGM}
Steinkrau, D., Simard, P.~Y., and Buck, I. (2005).
\newblock Using gpus for machine learning algorithms.
\newblock In {\em Proceedings of the Eighth International Conference on
  Document Analysis and Recognition}, ICDAR '05, pages 1115--1119, Washington,
  DC, USA. IEEE Computer Society.

\bibitem[Stuner et~al., 2016]{Stuner16}
Stuner, B., Chatelain, C., and Paquet, T. (2016).
\newblock Cohort of {LSTM} and lexicon verification for handwriting recognition
  with gigantic lexicon.
\newblock {\em CoRR}, abs/1612.07528.

\bibitem[Suddarth and Kergosien, 1990]{suddarth90NN}
Suddarth, S.~C. and Kergosien, Y.~L. (1990).
\newblock Rule-injection hints as a means of improving network performance and
  learning time.
\newblock In {\em Neural Networks, {EURASIP} workshop 1990}, pages 120--129.

\bibitem[Sugiyama, 2007]{Sugiyama2007}
Sugiyama, M. (2007).
\newblock Dimensionality reduction of multimodal labeled data by local fisher
  discriminant analysis.
\newblock {\em J. Mach. Learn. Res.}, 8:1027--1061.

\bibitem[Sukhbaatar et~al., 2015]{SukhbaatarSWF15CORR}
Sukhbaatar, S., Szlam, A., Weston, J., and Fergus, R. (2015).
\newblock Weakly supervised memory networks.
\newblock {\em CoRR}, abs/1503.08895.

\bibitem[Sun and Cheney, 1992]{Sun1992}
Sun, X. and Cheney, E.~W. (1992).
\newblock The fundamentality of sets of ridge functions.
\newblock {\em aequationes mathematicae}, 44(2):226--235.

\bibitem[Sutskever et~al., 2013]{sutskever1}
Sutskever, I., Martens, J., Dahl, G., and Hinton, G. (2013).
\newblock {On the importance of initialization and momentum in deep learning}.
\newblock In {\em {ICML}}, volume~28, pages 1139--1147.

\bibitem[Sutskever et~al., 2014a]{sutskeverVL14}
Sutskever, I., Vinyals, O., and Le, Q.~V. (2014a).
\newblock Sequence to sequence learning with neural networks.
\newblock In {\em Advances in Neural Information Processing Systems 27: Annual
  Conference on Neural Information Processing Systems 2014, December 8-13 2014,
  Montreal, Quebec, Canada}, pages 3104--3112.

\bibitem[Sutskever et~al., 2014b]{Sutskever2014SSL}
Sutskever, I., Vinyals, O., and Le, Q.~V. (2014b).
\newblock Sequence to sequence learning with neural networks.
\newblock In {\em Proceedings of the 27th International Conference on Neural
  Information Processing Systems - Volume 2}, NIPS'14, pages 3104--3112,
  Cambridge, MA, USA. MIT Press.

\bibitem[Syed and Yona, 2009]{syed09}
Syed, U. and Yona, G. (2009).
\newblock Enzyme function prediction with interpretable models.
\newblock {\em Computational Systems Biology. Humana press}, pages 373--420.

\bibitem[Sze et~al., 2017]{szegpus2017}
Sze, V., Chen, Y.~H., Yang, T.~J., and Emer, J.~S. (2017).
\newblock Efficient processing of deep neural networks: A tutorial and survey.
\newblock {\em Proceedings of the IEEE}, 105(12):2295--2329.

\bibitem[{Szegedy} et~al., 2014]{szegedy15}
{Szegedy}, C., {Liu}, W., {Jia}, Y., {Sermanet}, P., {Reed}, S., {Anguelov},
  D., {Erhan}, D., {Vanhoucke}, V., and {Rabinovich}, A. (2014).
\newblock {Going Deeper with Convolutions}.
\newblock {\em CoRR}, abs/1409.4842.

\bibitem[Szegedy et~al., 2014]{szegedyLJSRAEVR14}
Szegedy, C., Liu, W., Jia, Y., Sermanet, P., Reed, S.~E., Anguelov, D., Erhan,
  D., Vanhoucke, V., and Rabinovich, A. (2014).
\newblock Going deeper with convolutions.
\newblock {\em CoRR}, abs/1409.4842.

\bibitem[Szegedy et~al., 2013a]{Szegedy13}
Szegedy, C., Toshev, A., and Erhan, D. (2013a).
\newblock Deep neural networks for object detection.
\newblock In Burges, C. J.~C., Bottou, L., Welling, M., Ghahramani, Z., and
  Weinberger, K.~Q., editors, {\em NIPS 26}, pages 2553--2561.

\bibitem[Szegedy et~al., 2013b]{SzegedyZSBEGF13CORR}
Szegedy, C., Zaremba, W., Sutskever, I., Bruna, J., Erhan, D., Goodfellow,
  I.~J., and Fergus, R. (2013b).
\newblock Intriguing properties of neural networks.
\newblock {\em CoRR}, abs/1312.6199.

\bibitem[Szummer and Qi, 2004]{szummer04}
Szummer, M. and Qi, Y. (2004).
\newblock {Contextual Recognition of Hand-drawn Diagrams with Conditional
  Random Fields}.
\newblock In {\em {IWFHR}}, pages 32--37.

\bibitem[Tang and Eliasmith, 2010]{TangE10ICML}
Tang, Y. and Eliasmith, C. (2010).
\newblock Deep networks for robust visual recognition.
\newblock In {\em Proceedings of the 27th International Conference on Machine
  Learning (ICML-10), June 21-24, 2010, Haifa, Israel}, pages 1055--1062.

\bibitem[Terlaky, 1985]{terlaky1985}
Terlaky, T. (1985).
\newblock On lp programming.
\newblock {\em European Journal of Operational Research}, 22(1):70--100.

\bibitem[Thrun, 1996]{Thrun96islearning}
Thrun, S. (1996).
\newblock Is learning the n-th thing any easier than learning the first?
\newblock In {\em Advances in Neural Information Processing Systems}, pages
  640--646. The MIT Press.

\bibitem[Thrun and Mitchell, 1995]{ThrunM95ijcai}
Thrun, S. and Mitchell, T.~M. (1995).
\newblock Learning one more thing.
\newblock In {\em Proceedings of the Fourteenth International Joint Conference
  on Artificial Intelligence, {IJCAI} 95, Montr{\'{e}}al Qu{\'{e}}bec, Canada,
  August 20-25 1995, 2 Volumes}, pages 1217--1225.

\bibitem[Tibshirani, 1994]{Tibshirani94regressionshrinkage}
Tibshirani, R. (1994).
\newblock Regression shrinkage and selection via the lasso.
\newblock {\em Journal of the Royal Statistical Society, Series B},
  58:267--288.

\bibitem[Tikhonov, 1963]{tikhonov1963}
Tikhonov, A.~N. (1963).
\newblock On solving ill-posed problem and method of regularization.
\newblock {\em Doklady Akademii Nauk USSR}, 153:501--504.

\bibitem[Tikhonov and Arsenin, 1977]{tikhonov77}
Tikhonov, A.~N. and Arsenin, V.~Y. (1977).
\newblock {\em Solutions of Ill-posed problems}.
\newblock W.H.~Winston.

\bibitem[Tsechpenakis et~al., 2007]{tsechpenakis07}
Tsechpenakis, G., Wang, J., Mayer, B., and Metaxas, D. (2007).
\newblock {Coupling CRFs and Deformable Models for 3D Medical Image
  Segmentation}.
\newblock In {\em {ICCV}}, pages 1--8.

\bibitem[Tzeng et~al., 2017]{TzengCVPR2017}
Tzeng, E., Hoffman, J., Darrell, T., and Saenko, K. (2017).
\newblock Adversarial discriminative domain adaptation.
\newblock In {\em CVPR}.

\bibitem[Ulyanov et~al., 2017]{UlyanovVL17}
Ulyanov, D., Vedaldi, A., and Lempitsky, V. (2017).
\newblock Deep image prior.
\newblock {\em arXiv:1711.10925}.

\bibitem[Urban et~al., 2014]{Urban14}
Urban, G., Bendszus, M., Hamprecht, F.~A., and Kleesiek, J. (2014).
\newblock Multi-modal brain tumor segmentation using deep convolutional neural
  networks.
\newblock In {\em MICCAI BraTS Challenge Proceedings}, pages 31--35.

\bibitem[Utgoff, 1986]{Utgoff1986}
Utgoff, P.~E. (1986).
\newblock {\em Machine Learning of Inductive Bias}.
\newblock Kluwer, B.V., Deventer, The Netherlands, The Netherlands.

\bibitem[Valiant, 1984]{Valiant1984}
Valiant, L.~G. (1984).
\newblock A theory of the learnable.
\newblock {\em Commun. ACM}, 27(11):1134--1142.

\bibitem[van~den Oord et~al., 2013]{OordDS13nips}
van~den Oord, A., Dieleman, S., and Schrauwen, B. (2013).
\newblock Deep content-based music recommendation.
\newblock In {\em {NIPS}}, pages 2643--2651.

\bibitem[Vapnik and Chervonenkis, 1971]{vapnik1971}
Vapnik, V.~N. and Chervonenkis, A.~Y. (1971).
\newblock On the uniform convergence of relative frequencies of events to their
  probabilities.
\newblock {\em Theory of Probability and its Applications}, 16(2):264--280.

\bibitem[Vincent et~al., 2008]{vincent2008}
Vincent, P., Hugo, L., Bengio, Y., and Manzagol, P.-A. (2008).
\newblock Extracting and composing robust features with denoising autoencoders.
\newblock In {\em Proceedings of the 25th international conference on Machine
  learning}, ICML '08, pages 1096--1103, New York, NY, USA. ACM.

\bibitem[Vincent et~al., 2010]{vincent10}
Vincent, P., Larochelle, H., Lajoie, I., Bengio, Y., and Manzagol, P. (2010).
\newblock {Stacked Denoising Autoencoders: Learning Useful Representations in a
  Deep Network with a Local Denoising Criterion}.
\newblock {\em JMLR}, 11:3371--3408.

\bibitem[Vinyals et~al., 2015a]{VinyalsKKPSH15NIPS}
Vinyals, O., Kaiser, L., Koo, T., Petrov, S., Sutskever, I., and Hinton, G.~E.
  (2015a).
\newblock Grammar as a foreign language.
\newblock In {\em Advances in Neural Information Processing Systems 28: Annual
  Conference on Neural Information Processing Systems 2015, December 7-12,
  2015, Montreal, Quebec, Canada}, pages 2773--2781.

\bibitem[Vinyals et~al., 2015b]{VinyalsTBE15CVPR}
Vinyals, O., Toshev, A., Bengio, S., and Erhan, D. (2015b).
\newblock Show and tell: {A} neural image caption generator.
\newblock In {\em {IEEE} Conference on Computer Vision and Pattern Recognition,
  {CVPR} 2015, Boston, MA, USA, June 7-12, 2015}, pages 3156--3164.

\bibitem[Wager et~al., 2013]{wagerNIPS2013}
Wager, S., Wang, S., and Liang, P.~S. (2013).
\newblock Dropout training as adaptive regularization.
\newblock In Burges, C. J.~C., Bottou, L., Welling, M., Ghahramani, Z., and
  Weinberger, K.~Q., editors, {\em Advances in Neural Information Processing
  Systems 26}, pages 351--359. Curran Associates, Inc.

\bibitem[Wallis and Miller, 1991]{Wallis91}
Wallis, J. and Miller, T. (1991).
\newblock Three-dimensional display in nuclear medicine and radiology.
\newblock {\em Journal of nuclear medicine : official publication, Society of
  Nuclear Medicine}, 32(3):534--546.

\bibitem[Wallis, 1992]{wallis92}
Wallis, J.~W. (1992).
\newblock {\em Cardiovascular Nuclear Medicine and MRI: Quantitation and
  Clinical Applications}, pages 89--100.
\newblock Springer Netherlands.

\bibitem[Wallis et~al., 1989]{wallis89}
Wallis, J.~W., Miller, T.~R., Lerner, C.~A., and Kleerup, E.~C. (1989).
\newblock Three-dimensional display in nuclear medicine.
\newblock {\em IEEE Trans. on Medical Imaging}, 8(4):297--230.

\bibitem[Wan et~al., 2013]{WanZZLFICML13}
Wan, L., Zeiler, M.~D., Zhang, S., LeCun, Y., and Fergus, R. (2013).
\newblock Regularization of neural networks using dropconnect.
\newblock In {\em ICML (3)}, volume~28 of {\em JMLR Proceedings}, pages
  1058--1066. JMLR.org.

\bibitem[Wang and Mahadevan, 2008]{WangM08icml}
Wang, C. and Mahadevan, S. (2008).
\newblock Manifold alignment using procrustes analysis.
\newblock In {\em Machine Learning, Proceedings of the Twenty-Fifth
  International Conference {(ICML} 2008), Helsinki, Finland, June 5-9, 2008},
  pages 1120--1127.

\bibitem[Wang et~al., 2005]{wang2005self}
Wang, X., Li, L., Lockington, D., Pullar, D., and Jeng, D. (2005).
\newblock Self-organizing polynomial neural network for modelling complex
  hydrological processes.
\newblock {\em Research Report No R861, Department of Civil Engineering}.

\bibitem[Wang and Wang, 2014]{WangW14acmm}
Wang, X. and Wang, Y. (2014).
\newblock Improving content-based and hybrid music recommendation using deep
  learning.
\newblock In {\em {ACM} Multimedia}, pages 627--636. {ACM}.

\bibitem[Warde{-}Farley et~al., 2014]{WardeFarleyGCB2014ICLR}
Warde{-}Farley, D., Goodfellow, I.~J., Courville, A.~C., and Bengio, Y. (2014).
\newblock An empirical analysis of dropout in piecewise linear networks.
\newblock In {\em International Conference on Learning Representations
  (ICLR2014)}.

\bibitem[Weng et~al., 1992]{weng1992}
Weng, J., Ahuja, N., and Huang, T.~S. (1992).
\newblock Cresceptron: a self-organizing neural network which grows adaptively.
\newblock In {\em International Joint Conference on Neural Networks (IJCNN)},
  volume~1, pages 576--581. IEEE.

\bibitem[Weng et~al., 1997]{weng1997}
Weng, J.~J., Ahuja, N., and Huang, T.~S. (1997).
\newblock Learning recognition and segmentation using the cresceptron.
\newblock {\em International Journal of Computer Vision}, 25(2):109--143.

\bibitem[Werbos, 1974]{Werbos74}
Werbos, P.~J. (1974).
\newblock {\em Beyond Regression: New Tools for Prediction and Analysis in the
  Behavioral Sciences}.
\newblock PhD thesis, Harvard University.

\bibitem[Werbos, 1981]{Werbos81sensitivity}
Werbos, P.~J. (1981).
\newblock Applications of advances in nonlinear sensitivity analysis.
\newblock In {\em Proceedings of the 10th IFIP Conference, 31.8 - 4.9, NYC},
  pages 762--770.

\bibitem[Weston et~al., 2014]{weston2014memory}
Weston, J., Chopra, S., and Bordes, A. (2014).
\newblock Memory networks.
\newblock {\em arXiv preprint arXiv:1410.3916}.

\bibitem[Weston et~al., 2008]{WestonRC2008}
Weston, J., Ratle, F., and Collobert, R. (2008).
\newblock Deep learning via semi-supervised embedding.
\newblock In {\em Machine Learning, Proceedings of the Twenty-Fifth
  International Conference {(ICML} 2008), Helsinki, Finland, June 5-9, 2008},
  pages 1168--1175.

\bibitem[Weston et~al., 2012]{weston2012}
Weston, J., Ratle, F., Mobahi, H., and Collobert, R. (2012).
\newblock Deep learning via semi-supervised embedding.
\newblock In Montavon, G., Orr, G., and Muller, K.-R., editors, {\em Neural
  Networks: Tricks of the Trade}. Springer.

\bibitem[Widrow, 1960]{Widrow1960}
Widrow, B. (1960).
\newblock {An Adaptive "ADALINE" Neuron Using Chemical "Memistors"}.
\newblock Technical report, Solid-State Electronics Laboratory, Stanford
  Electronics Laboratories, Stanford University, Stanford, California.

\bibitem[Wiesel and Hubel, 1959]{wiesel1959}
Wiesel, D.~H. and Hubel, T.~N. (1959).
\newblock Receptive fields of single neurones in the cat's striate cortex.
\newblock {\em J. Physiol.}, 148:574--591.

\bibitem[Wiesler et~al., 2014]{Wiesler2014ICASSP}
Wiesler, S., Richard, A., Schlüter, R., and Ney, H. (2014).
\newblock {M}ean-normalized stochastic gradient for large-scale deep learning.
\newblock In {\em IEEE International Conference on Acoustics, Speech and Signal
  Processing (ICASSP), 2014 : 4 - 9 May 2014, Florence, Italy}, International
  Conference on Acoustics Speech and Signal Processing ICASSP, pages 180--184,
  Piscataway, NJ. IEEE International Conference on Acoustics, Speech and Signal
  Processing, Florence (Italy), 4 May 2014 - 9 May 2014, IEEE.

\bibitem[Wiesler et~al., 2011]{Wiesler2011}
Wiesler, S., Schlüter, R., and Ney, H. (2011).
\newblock A convergence analysis of log-linear training and its application to
  speech recognition.
\newblock In {\em 2011 IEEE Workshop on Automatic Speech Recognition
  Understanding}, pages 1--6.

\bibitem[Winter and Widrow, 1988]{Winter1988}
Winter, R. and Widrow, B. (1988).
\newblock Madaline rule ii: a training algorithm for neural networks.
\newblock In {\em IEEE 1988 International Conference on Neural Networks}, pages
  401--408 vol.1.

\bibitem[Wolpert, 1996]{Wolpert1996}
Wolpert, D.~H. (1996).
\newblock The lack of a priori distinctions between learning algorithms.
\newblock {\em Neural Comput.}, 8(7):1341--1390.

\bibitem[Woodworth and Thorndike, 1901]{woodworth1901influence}
Woodworth, R.~S. and Thorndike, E. (1901).
\newblock The influence of improvement in one mental function upon the
  efficiency of other functions.(i).
\newblock {\em Psychological review}, 8(3):247.

\bibitem[Wu and Srihari, 2004]{wu2004}
Wu, X. and Srihari, R. (2004).
\newblock Incorporating prior knowledge with weighted margin support vector
  machines.
\newblock In {\em Proceedings of the Tenth ACM SIGKDD International Conference
  on Knowledge Discovery and Data Mining}, KDD '04, pages 326--333, New York,
  NY, USA. ACM.

\bibitem[Xu et~al., 2015]{XuBKCCSZB15ICML}
Xu, K., Ba, J., Kiros, R., Cho, K., Courville, A.~C., Salakhutdinov, R., Zemel,
  R.~S., and Bengio, Y. (2015).
\newblock Show, attend and tell: Neural image caption generation with visual
  attention.
\newblock In {\em Proceedings of the 32nd International Conference on Machine
  Learning, {ICML} 2015, Lille, France, 6-11 July 2015}, pages 2048--2057.

\bibitem[Xue and Ye, 2000]{xue2000efficient}
Xue, G. and Ye, Y. (2000).
\newblock An efficient algorithm for minimizing a sum of p-norms.
\newblock {\em SIAM Journal on Optimization}, 10(2):551--579.

\bibitem[Xue et~al., 2007]{XueLCK07jmlr}
Xue, Y., Liao, X., Carin, L., and Krishnapuram, B. (2007).
\newblock Multi-task learning for classification with dirichlet process priors.
\newblock {\em Journal of Machine Learning Research}, 8:35--63.

\bibitem[Yang and Hospedales, 2016a]{YangH16corr}
Yang, Y. and Hospedales, T.~M. (2016a).
\newblock Deep multi-task representation learning: {A} tensor factorisation
  approach.
\newblock {\em CoRR}, abs/1605.06391.

\bibitem[Yang and Hospedales, 2016b]{YangH16acorr}
Yang, Y. and Hospedales, T.~M. (2016b).
\newblock Trace norm regularised deep multi-task learning.
\newblock {\em CoRR}, abs/1606.04038.

\bibitem[Yip et~al., 2015]{Yip2015}
Yip, C., Dinkel, C., Mahajan, A., Siddique, M., Cook, G., and Goh, V. (2015).
\newblock {Imaging body composition in cancer patients: visceral obesity,
  sarcopenia and sarcopenic obesity may impact on clinical outcome}.
\newblock {\em Insights into Imaging}, pages 489--497.

\bibitem[Yosinski et~al., 2014]{YosinskiCBL14nips}
Yosinski, J., Clune, J., Bengio, Y., and Lipson, H. (2014).
\newblock How transferable are features in deep neural networks?
\newblock In {\em {NIPS}}, pages 3320--3328.

\bibitem[Yu et~al., 2016]{yu2016deep}
Yu, A., Palefsky-Smith, R., and Bedi, R. (2016).
\newblock Deep reinforcement learning for simulated autonomous vehicle control.
\newblock {\em Course Project Reports: Winter}, pages 1--7.

\bibitem[Yu et~al., 2007]{yu2007}
Yu, T., Jan, T., Simoff, S., and Debenham, J. (2007).
\newblock Incorporating prior domain knowledge into inductive machine learning.
\newblock {\em Unpublished doctoral dissertation Computer Sciences}.

\bibitem[Yu et~al., 2010]{YuSJ10Neurocomp}
Yu, T., Simoff, S., and Jan, T. (2010).
\newblock {VQSVM:} {A} case study for incorporating prior domain knowledge into
  inductive machine learning.
\newblock {\em Neurocomputing}, 73(13-15):2614--2623.

\bibitem[Zeiler, 2012a]{zeiler1}
Zeiler, M. (2012a).
\newblock {ADADELTA: An Adaptive Learning Rate Method}.
\newblock {\em CoRR}, abs/1212.5701.

\bibitem[Zeiler, 2012b]{zeiler12}
Zeiler, M.~D. (2012b).
\newblock {ADADELTA:} an adaptive learning rate method.
\newblock {\em CoRR}, abs/1212.5701.

\bibitem[Zeiler and Fergus, 2013]{matthew2013ICLR}
Zeiler, M.~D. and Fergus, R. (2013).
\newblock Stochastic pooling for regularization of deep convolutional neural
  networks.
\newblock In {\em International Conference on Learning Representations
  (ICLR2013)}.

\bibitem[Zeiler and Fergus, 2014]{ZeilerF14eccv}
Zeiler, M.~D. and Fergus, R. (2014).
\newblock Visualizing and understanding convolutional networks.
\newblock In {\em {ECCV} {(1)}}, volume 8689 of {\em Lecture Notes in Computer
  Science}, pages 818--833. Springer.

\bibitem[Zen et~al., 2009]{zen09}
Zen, H., Tokuda, K., and Black, A. (2009).
\newblock {Statistical parametric speech synthesis}.
\newblock {\em Speech Communication}, 51(11):1039--1064.

\bibitem[Zhang et~al., 2016]{zhangBHRV16}
Zhang, C., Bengio, S., Hardt, M., Recht, B., and Vinyals, O. (2016).
\newblock Understanding deep learning requires rethinking generalization.
\newblock {\em CoRR}, abs/1611.03530.

\bibitem[Zhang et~al., 2014a]{zhang14}
Zhang, J., Shan, S., Kan, M., and Chen, X. (2014a).
\newblock {Coarse-to-Fine Auto-Encoder Networks {(CFAN)} for Real-Time Face
  Alignment}.
\newblock In {\em {ECCV, Part {II}}}, pages 1--16.

\bibitem[Zhang et~al., 2017a]{ZhangZGZ17cvpr}
Zhang, K., Zuo, W., Gu, S., and Zhang, L. (2017a).
\newblock Learning deep {CNN} denoiser prior for image restoration.
\newblock In {\em {CVPR}}, pages 2808--2817. {IEEE} Computer Society.

\bibitem[Zhang et~al., 2017b]{ZhangDNK17CORR}
Zhang, X., Das, S., Neopane, O., and Kreutz{-}Delgado, K. (2017b).
\newblock A design methodology for efficient implementation of deconvolutional
  neural networks on an {FPGA}.
\newblock {\em CoRR}, abs/1705.02583.

\bibitem[Zhang et~al., 2017c]{ZhangDG17}
Zhang, Y., David, P., and Gong, B. (2017c).
\newblock Curriculum domain adaptation for semantic segmentation of urban
  scenes.
\newblock {\em CoRR}, abs/1707.09465.

\bibitem[Zhang and Yang, 2017]{ZhangY17aa}
Zhang, Y. and Yang, Q. (2017).
\newblock A survey on multi-task learning.
\newblock {\em CoRR}, abs/1707.08114.

\bibitem[Zhang et~al., 2014b]{zhang14ECCV}
Zhang, Z., Luo, P., Loy, C.~C., and Tang, X. (2014b).
\newblock Facial landmark detection by deep multi-task learning.
\newblock In {\em Computer Vision, ECCV 2014, 13th European Conference}, pages
  94--108.

\bibitem[Zhuang et~al., 2015]{ZhuangCLPH15ijcai}
Zhuang, F., Cheng, X., Luo, P., Pan, S.~J., and He, Q. (2015).
\newblock Supervised representation learning: Transfer learning with deep
  autoencoders.
\newblock In {\em {IJCAI}}, pages 4119--4125. {AAAI} Press.

\end{thebibliography}



\end{spacing}

\begin{appendices} 

\chapter{Definitions and Technical Details}  
\label{chap:chapterapp4}

\ifpdf
    \graphicspath{{Appendix4/Figs/Raster/}{Appendix4/Figs/PDF/}{Appendix4/Figs/}}
\else
    \graphicspath{{Appendix4/Figs/Vector/}{Appendix4/Figs/}}
\fi

\makeatletter
\def\input@path{{Appendix4/}}
\makeatother

Due to the space limitation in the introductory chapter of this thesis (\autoref{chap:chapter0}), and in order to ease its reading, we decide to provide a separate appendix that covers some basic definitions in machine learning (Sec.\ref{sec:appendix4machinelearningdefintions}), and some technical details (Sec.\ref{sec:appendix4technicaldetails}) which are not mandatory in order to go through the rest of the thesis. However, they cover some important aspects including:
\begin{enumerate}
  \item bias-variance tradeoff (Sec.\ref{sub:appendix4biasvariance}),
  \item feedforward networks (Sec.\ref{sub:appendix4fnn}), including 
  \begin{enumerate*}
    \item backpropagation, derivatives computation, and issues (Sec.\ref{subsub:appendix4backprop}),
    \item nonlinear activation functions (Sec.\ref{subsub:appendix4nonlinearfunctions}, \ref{subsub:appendix4perceptronforclassification}),
    \item universal approximation theorem (Sec.\ref{subsub:appendix4univapproxtheoremanddepth}),
    \item and other neural architectures (Sec.\ref{subsub:appendix4otherneuralarchitectures}).
  \end{enumerate*}
  \item and the impact of some regularization approaches on the obtained solution  (Sec.\ref{sub:appendix4regularization}) including  $L_p$ norm regularization (Sec.\ref{subsub:appendix4lpnorm}), and early stopping (Sec.\ref{subsub:appendix4earlystopping}).
\end{enumerate}

\section[Machine Learning Definitions]{Machine Learning Definitions}
\label{sec:appendix4machinelearningdefintions}

We discuss in this section further details on machine learning (Sec.\ref{sec:machinel0}). We illustrate its use throughout different applications in real life (Sec.\ref{sub:appendix4applicationsandproblems}), while we provide some basic definitions (Sec.\ref{sub:appendix4definitionsandterminology}) and different learning scenarios (Sec.\ref{sub:appendix4learningscenarios}).

\subsection{Applications}
\label{sub:appendix4applicationsandproblems}
Machine learning algorithms have been successfully deployed in a variety of applications, including
\begin{itemize*}
    \item Text or document classification, e.g., spam detection;
    \item Natural language processing, e.g., part-of-speech tagging, statistical parsing, name-entity recognition;
    \item Speech recognition, speech synthesis, speaker verification;
    \item Computational biology applications, e.g., protein function or structural prediction;
    \item Computer vision tasks, e.g., image recognition, face detection;
    \item Fraud detection (credit card, telephone), and network intrusion;
    \item Games, e.g., chess, backgammon, go;
    \item Unassisted vehicle control (robots, navigation);
    \item Medical diagnosis;
    \item Recommendation systems, search engines, information extraction systems.
\end{itemize*}

This list is by no means comprehensive, and learning algorithms are applied to new applications every day. Moreover, such applications correspond to a wide variety of learning problems. Some major classes of learning problems are:
\begin{itemize}
    \item \emph{Classification}: Assign a category to each item. For example, document classification may assign items with categories such as \emph{politics}, \emph{business}, \emph{sports}, or \emph{weather}.
    \item \emph{Regression}: Predict a real value for each item. Examples of regression include prediction of stock values or variations of economic variables. In this problem, the penalty for an incorrect prediction depends on the magnitude of difference between the true and predicted values.
    \item \emph{Ranking}: Order items according to some criterion. Web search, e.g., returning web pages relevant to a search query, is the canonical ranking example.
    \item \emph{Clustering}: Partition items into homogeneous regions. Clustering is often performed to analyze very large data sets. For example, in the context of social network analysis, clustering algorithms attempt to identify \quotes{communities} within large groups of people.
    \item \emph{Dimensionality reduction or manifold learning}: Transform an initial representation of items into a lower-dimensional representation of the items while preserving some properties of the initial representation. A common example involves preprocessing digital images in computer vision tasks.
\end{itemize}

In the next section, we provide basic definitions and terminology that are used in machine learning.

\subsection{Terminology}
\label{sub:appendix4definitionsandterminology}
We use the canonical problem of spam detection as a running example to illustrate some basic definitions and to describe the use and evaluation of machine learning algorithms in practice \cite{Mohri2012bookML}. Spam detection is the problem of learning to automatically classify email messages as either \texttt{SPAM} or \texttt{not-SPAM}.

\begin{itemize}
    \item \emph{Examples}: Items or instances of data used for learning or evaluation. In our spam problem, these examples correspond to the collection of email messages we will use for learning and testing.
    \item \emph{Features}: The set of attributes, often represented as a vector, associated to an example. In the case of email messages, some relevant features may include the length of the message, the name of the sender, various characteristics of the header, the presence of certain keywords in the body of the message, and so on.
    \item \emph{Labels}: Values or categories assigned to examples. In classification problems, examples are assigned specific categories, for instance, the \texttt{SPAM} and \texttt{not-SPAM} categories in our binary classification problem. In regression, items are assigned real-valued labels.
    \item \emph{Training samples}: Examples used to train a model. In our spam problem, the training samples consist of a set of email examples along with their associated labels.
    \item \emph{Validation samples}: Examples used to tune the parameters of a model when working with labeled data. Models typically have one or more free parameters, and the validation samples are used to select appropriate values for such free parameters.
    \item \emph{Test samples}: Examples used to evaluate the performance of a learned model. The test samples are separate from the training and validation data and is not made available in the learning stage. In the spam problem, the test samples consist of a collection of email examples for which the learned model must predict labels based on features. These predictions are then compared with the labels of the test samples to measure the performance of the model.
    \item \emph{Loss function}: A function that measures the difference, or loss, between a predicted label and a true label. Denoting the set of all labels as $\Y$ and the set of possible predictions as $\Y^{\prime}$, a loss function ${\ell: \Y^{\prime} \times \Y \to \R_{+}}$. In most cases, $\Y^{\prime} = \Y$ and the loss function is bounded, but these conditions do not always hold. Common examples of loss functions include the 0-1 (or misclassification) loss defined over $\{-1, +1\} \times \{-1, +1\}$ by $\ell(y^{\prime}, y) = \ind_{y^{\prime} \neq y}$ and the squared loss defined over $I \times I$ by ${\ell(y^{\prime}, y) = (y^{\prime} - y)^2}$, where $I \subseteq \R$ is typically a bounded interval.
    \item \emph{Hypothesis set}: A set of functions mapping features (feature vectors) to the set of label $\Y$. In our example, these may be a set of functions mapping email features to $\Y = \{\text{\texttt{SPAM }}, \text{\texttt{not-SPAM}}\}$. More generally, hypotheses may be functions mapping features to a different set $\Y^{\prime}$. They could be linear functions mapping email feature vector to real numbers interpreted as scores ($\Y = \R$), with higher score values more indicative of $\texttt{SPAM}$ than lower ones.
\end{itemize}

We now define the learning stages of our spam problem. We start with a given collection of labeled examples. We first randomly partition the data into  training samples, validation samples, and  test samples. The size of each of these samples depends on a number of different considerations. For example, the amount of data reserved for validation depends on the number of free parameters of the model. Also, when labeled samples are relatively small, the amount of training data is often chosen to be larger than that of test data since the learning performance directly depends on the number of training samples.

Next, we associate relevant features to the examples. This is a critical step in the design of machine learning solutions. Useful features can effectively guide the learning of the model, while poor or uninformative ones can be misleading. Although it is critical, to a large extent, the choice of the features is left to the user. This choice reflects the user's prior knowledge about the learning task which in practice can have a dramatic effect on the performance results.

Now, we use the features selected to train our model by fixing different values of its free parameters. For each value of these parameters, the learning algorithm selects a different hypothesis, i.e., model, out of the hypothesis set. We choose among them the model resulting in the best performance on the validation samples. Finally, using that model, we predict the labels of the examples in the test samples. The performance of the model is evaluated by using the loss function associated to the task, e.g., the 0-1 loss in our spam detection task, to compare the predicted and true value.

\begin{figure}[!htbp]
  \begin{center}
		\begin{tikzpicture}
\coordinate (b1) at (0, 0);
\coordinate (b2) at (1.5, -0.2);
\coordinate (b3) at (-1, 1.5);
\coordinate (b4) at (1, 0.3);
\coordinate (b5) at (2, 0.3);
\coordinate (b6) at (4, 0.7);
\coordinate (b7) at (0.2, 0.8);
\coordinate (r1) at (2.6, 0.8);
\coordinate (b8) at (2, 1.5);
\coordinate (r2) at (3.6, 1.5);
\coordinate (r3) at (1.1, 1.8);
\coordinate (b9) at (0.33, 2.1);
\coordinate (r4) at (2.6, 2.5);
\coordinate (r5) at (1.5, 2.8);
\coordinate (r6) at (0.28, 3);
\coordinate (r7) at (3, 3);
\coordinate (r8) at (1.8, 3.8);
\coordinate (s1) at (-2, 2.3);
\coordinate (s2) at (0.8, 2.5);
\coordinate (s3) at (0.7, 1);
\coordinate (s4) at (2.4, 2.2);
\coordinate (s5) at (2.25, 0.5);
\coordinate (s6) at (3.3, 0.3);
\coordinate (s7) at (4.3, 1.8);
\coordinate (s8) at (4.8, 0.3);


\coordinate (b1r) at (7, 0);
\coordinate (b2r) at (8.5, -0.2);
\coordinate (b3r) at (6, 1.5);
\coordinate (b4r) at (8, 0.3);
\coordinate (b5r) at (9, 0.3);
\coordinate (b6r) at (11, 0.7);
\coordinate (b7r) at (7.2, 0.8);
\coordinate (r1r) at (9.6, 0.8);
\coordinate (b8r) at (9, 1.5);
\coordinate (r2r) at (10.6, 1.5);
\coordinate (r3r) at (8.1, 1.8);
\coordinate (b9r) at (7.33, 2.1);
\coordinate (r4r) at (9.6, 2.5);
\coordinate (r5r) at (8.5, 2.8);
\coordinate (r6r) at (7.28, 3);
\coordinate (r7r) at (10, 3);
\coordinate (r8r) at (8.8, 3.8);

\coordinate (s1r) at (5, 2.3);
\coordinate (s2r) at (11.5, -0.7);

\draw (b1) node[circle, draw=black!80, fill=blue, inner sep=0pt, minimum size=12pt, line width=0.5mm] (cb1) {};
\draw (b2) node[circle, draw=black!80, fill=blue, inner sep=0pt, minimum size=12pt, line width=0.5mm] (cb2) {};
\draw (b3) node[circle, draw=black!80, fill=blue, inner sep=0pt, minimum size=12pt, line width=0.5mm] (cb3) {};
\draw (b4) node[circle, draw=black!80, fill=blue, inner sep=0pt, minimum size=12pt, line width=0.5mm] (cb4) {};
\draw (b5) node[circle, draw=black!80, fill=blue, inner sep=0pt, minimum size=12pt, line width=0.5mm] (cb5) {};
\draw (b6) node[circle, draw=black!80, fill=blue, inner sep=0pt, minimum size=12pt, line width=0.5mm] (cb6) {};
\draw (b7) node[circle, draw=black!80, fill=blue, inner sep=0pt, minimum size=12pt, line width=0.5mm] (cb7) {};
\draw (r1) node[circle, draw=black!80, fill=red, inner sep=0pt, minimum size=12pt, line width=0.5mm] (cr1) {};
\draw (b8) node[circle, draw=black!80, fill=blue, inner sep=0pt, minimum size=12pt, line width=0.5mm] (cb8) {};
\draw (r2) node[circle, draw=black!80, fill=red, inner sep=0pt, minimum size=12pt, line width=0.5mm] (cr2) {};
\draw (r3) node[circle, draw=black!80, fill=red, inner sep=0pt, minimum size=12pt, line width=0.5mm] (cr3) {};
\draw (b9) node[circle, draw=black!80, fill=blue, inner sep=0pt, minimum size=12pt, line width=0.5mm] (cb9) {};
\draw (r4) node[circle, draw=black!80, fill=red, inner sep=0pt, minimum size=12pt, line width=0.5mm] (cr4) {};
\draw (r5) node[circle, draw=black!80, fill=red, inner sep=0pt, minimum size=12pt, line width=0.5mm] (cr5) {};
\draw (r6) node[circle, draw=black!80, fill=red, inner sep=0pt, minimum size=12pt, line width=0.5mm] (cr6) {};
\draw (r7) node[circle, draw=black!80, fill=red, inner sep=0pt, minimum size=12pt, line width=0.5mm] (cr7) {};
\draw (r8) node[circle, draw=black!80, fill=red, inner sep=0pt, minimum size=12pt, line width=0.5mm] (cr8) {};

\draw[-, thick, line width=0.8mm] (s1) -- (s2) node {};
\draw[-, thick, line width=0.8mm] (s2) -- (s3) node {};
\draw[-, thick, line width=0.8mm] (s3) -- (s4) node {};
\draw[-, thick, line width=0.8mm] (s4) -- (s5) node {};
\draw[-, thick, line width=0.8mm] (s5) -- (s6) node {};
\draw[-, thick, line width=0.8mm] (s6) -- (s7) node {};
\draw[-, thick, line width=0.8mm] (s7) -- (s8) node {};

\draw (b1r) node[circle, draw=black!80, fill=blue, inner sep=0pt, minimum size=12pt, line width=0.5mm] (cb1r) {};
\draw (b2r) node[circle, draw=black!80, fill=blue, inner sep=0pt, minimum size=12pt, line width=0.5mm] (cb2r) {};
\draw (b3r) node[circle, draw=black!80, fill=blue, inner sep=0pt, minimum size=12pt, line width=0.5mm] (cb3r) {};
\draw (b4r) node[circle, draw=black!80, fill=blue, inner sep=0pt, minimum size=12pt, line width=0.5mm] (cb4r) {};
\draw (b5r) node[circle, draw=black!80, fill=blue, inner sep=0pt, minimum size=12pt, line width=0.5mm] (cb5r) {};
\draw (b6r) node[circle, draw=black!80, fill=blue, inner sep=0pt, minimum size=12pt, line width=0.5mm] (cb6r) {};
\draw (b7r) node[circle, draw=black!80, fill=blue, inner sep=0pt, minimum size=12pt, line width=0.5mm] (cb7r) {};
\draw (r1r) node[circle, draw=black!80, fill=red, inner sep=0pt, minimum size=12pt, line width=0.5mm] (cr1r) {};
\draw (b8r) node[circle, draw=black!80, fill=blue, inner sep=0pt, minimum size=12pt, line width=0.5mm] (cb8r) {};
\draw (r2r) node[circle, draw=black!80, fill=red, inner sep=0pt, minimum size=12pt, line width=0.5mm] (cr2r) {};
\draw (r3r) node[circle, draw=black!80, fill=red, inner sep=0pt, minimum size=12pt, line width=0.5mm] (cr3r) {};
\draw (b9r) node[circle, draw=black!80, fill=blue, inner sep=0pt, minimum size=12pt, line width=0.5mm] (cb9r) {};
\draw (r4r) node[circle, draw=black!80, fill=red, inner sep=0pt, minimum size=12pt, line width=0.5mm] (cr4r) {};
\draw (r5r) node[circle, draw=black!80, fill=red, inner sep=0pt, minimum size=12pt, line width=0.5mm] (cr5r) {};
\draw (r6r) node[circle, draw=black!80, fill=red, inner sep=0pt, minimum size=12pt, line width=0.5mm] (cr6r) {};
\draw (r7r) node[circle, draw=black!80, fill=red, inner sep=0pt, minimum size=12pt, line width=0.5mm] (cr7r) {};
\draw (r8r) node[circle, draw=black!80, fill=red, inner sep=0pt, minimum size=12pt, line width=0.5mm] (cr8r) {};

\draw[thick, line width=0.8mm, minimum size=12pt, color=black] (s1r) to[out=20,in=117] (s2r); 

\end{tikzpicture}
	\end{center}
  \caption[\bel{Complexity vs. generalization.}]{The zig-zag line on the left panel is consistent over the blue and red training samples, but it is a complex separation surface that is not likely to generalize well to unseen data. In contrast, the decision surface on the right panel is simpler and might generalize better in spite of its misclassification of few points of the training samples. (Reference:  \cite{Mohri2012bookML})}
  \label{fig:fig0-0-appendix4}
\end{figure}
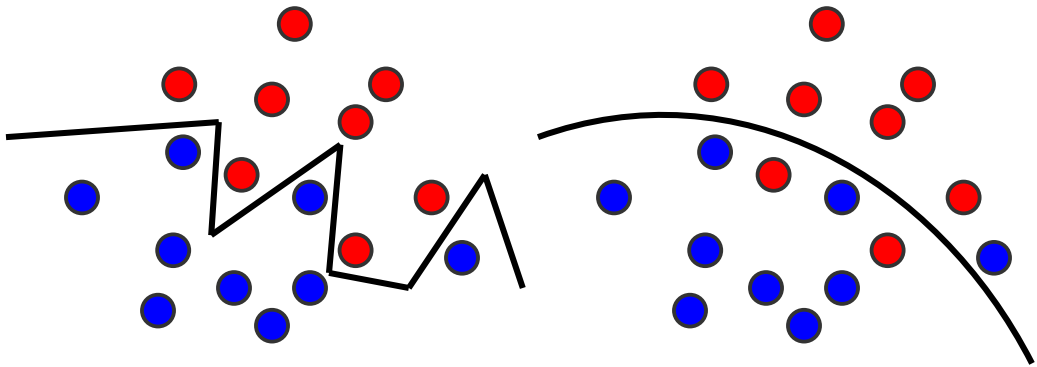

Thus, the performance of a model is of course evaluated based on its test error and not its error on the training samples. A model may be consistent, that is it may commit no error on the examples of the training data, and yet have a poor performance on the test data. This occurs for consistent models defined by very complex decision surfaces, as illustrated in Fig.\ref{fig:fig0-0-appendix4}, which tend to memorize a relatively small training samples instead of seeking to generalize well. This highlights the key distinction between memorization and generalization, which is the fundamental property sought for an accurate model.

In the following, we describe some common learning scenarios.

\subsection{Learning Scenarios}
\label{sub:appendix4learningscenarios}
We briefly describe common machine learning scenarios \cite{Mohri2012bookML}. These scenarios differ in the type of training data available to the learner, the order and method by which training data is received and the test data used to evaluate the model.

\begin{itemize}
    \item \emph{Supervised learning}: The learner receives a set of labeled examples as training data and makes predictions for unseen points. This is the most common scenario associated with classification, regression, and ranking problems.
    \item \emph{Unsupervised learning}: The learner exclusively receives unlabeled training data, and makes prediction for unseen points. Since in general no labeled example is available in that setting, it can be difficult to quantitatively evaluate the performance of a learner. Clustering and dimensionality reduction are example of unsupervised learning problems.
    \item \emph{Semi-supervised learning}: The learner receives some training samples which consist of both labeled and unlabeled data, and makes predictions for unseen points. Semi-supervised learning is common in setting where unlabeled data is easily accessible but labels are expensive to obtain. Various types of problems arising in applications, including classification, regression, or ranking tasks, can be framed as instances of semi-supervised learning. The hope is that the distribution of unlabeled data accessible to the learner can help to achieve a better performance than in the supervised setting.
    \item \emph{Transductive inference}: As in the semi-supervised scenario, the learner receives labeled training samples along with a set of unlabeled test points. However, the objective of transductive inference is to predict labels only for these particular test points. Transductive inference appears to be an easier task and matches the scenario encountered in a variety of modern applications. However, the assumptions under which a better performance can be achieved in this setting are research questions that have not been fully solved.
    \item \emph{Online learning}: In contrast with the previous scenarios, the online scenario involves multiple rounds and training and testing phases are intermixed. At each round, the learner receives an unlabeled training point, makes a prediction, receives the true label, and incurs a loss. The objective in the online setting is to minimize the cumulative loss over all rounds. Unlike the previous settings just discussed, no distributional assumption is made in online learning.
    \item \emph{Reinforcement learning}: The training and the testing phases are also intermixed in reinforcement learning. To collect information, the learner actively interacts with the environment and in some cases affects the environment, and receives an immediate reward for each action. The object of the learner is to maximize its reward over a course of actions and iterations with the environment. However, no long-term reward feedback is provided by the environment, and the learner is faced with the exploration versus exploitation dilemma, since it must choose between exploring unknown actions to gain more information versus exploiting the information already collected.
    \item \emph{Active learning}: The learner adaptively or interactively collects training examples, typically by querying an oracle to request labels for new points. The goal in active learning is to achieve a performance comparable to the standard supervised learning scenario, but with fewer labeled examples. Active learning is often used in applications where labels are expensive to obtain, for example computational biology applications.
\end{itemize}
In practice, many other intermediate and somewhat more complex learning scenarios may be encountered.

\section[Technical Details]{Technical Details}
\label{sec:appendix4technicaldetails}

\subsection{Bias-variance Tradeoff}
\label{sub:appendix4biasvariance}

The bias-variance tradeoff is the problem of minimizing two sources of errors that prevent a model from well generalizing beyond the train data. The first error source is named the bias error which comes from erroneous assumption about the complexity of the model which means this error depends on the performance of the model in average while considering an infinite number of training samples. In the other hand, the variance error which comes from the sensitivity of the model to small variations in the data. This depends on the model capacity to model the random noise (small perturbations) in the data.

The bias-variance decomposition \cite{Geman1992} provides a way to analyze the expected generalization error of a model. This decomposition gives the generalization error as a sum of two terms: the bias and the variance.

In order to formalize this decomposition, let us consider a regression problem over the distribution $p_{(\x, \y)}$. $\mathbb{D}_{\text{train}} \sim p_{(\x, \y)}$ is a sampled training data. For clearer exposition, let us take $\y = y$ to be a one-dimentional, although the results apply more generally \cite{Geman1992}. $\E[y |\xvec]$ denotes as a deterministic function that gives the value $y$ conditioned on a fixed $\xvec$. $\E[y |\xvec]$ can be seen as the best value $y$ for $\xvec$. For any function $f(\xvec)$, and any fixed $\xvec$, the regression error is \cite{Geman1992}
\begin{align}
  \E\left[(y - f(\xvec))^2|\xvec \right] & = \E\left[((y - \E[y|\xvec]) + (\E[y|\xvec] - f(\xvec)))^2 | \xvec \right] \label{eq:eqapp1-20-0}\\
  &= \E\left[(y - \E[y|\xvec])^2|\xvec \right] + (\E[y|\xvec] - f(\xvec))^2 \nonumber \\ 
    & + 2 \E \left[(y - \E[y|\xvec])|\xvec \right] \cdot (\E[y|\xvec] - f(\xvec)) \label{eq:eqapp1-20-1}\\
  &=\E\left[(y - \E[y|\xvec])^2|\xvec \right] + (\E[y|\xvec] - f(\xvec))^2 \nonumber \\ 
  &+ 2(\E[y|\xvec] - \E[y|\xvec]) \cdot (\E[y|\xvec] - f(\xvec)) \label{eq:eqapp1-20-2}\\
  & = \E\left[(y - \E[y|\xvec])^2|\xvec \right] + (\E[y|\xvec] - f(\xvec))^2 \label{eq:eqapp1-20-3} \; .
\end{align}
\noindent In other words, among all functions of $\xvec$, $f(\xvec) = \E[y|\xvec]$ is the best predictor of $y$ given $\xvec$, in the mean-squared-error.

Now, let us introduce the dependency of $f(\xvec)$ to its training sample $\mathbb{D}_{\text{train}}$ and let us note $f(\xvec; \mathbb{D}_{\text{train}})$. For clarity, we refer to $\mathbb{D}_{\text{train}}$ by $\mathbb{D}$. Now, the generalization error over a new fixed example $\xvec$ and fixed training sample $\mathbb{D}$ is computed as follows \cite{Geman1992}
\begin{align}
  \label{eq:eqapp1-20-4}
  \E\left[(y - f(\xvec; \mathbb{D}))^2 | \xvec, \mathbb{D} \right] & = \E\left[(y - \E[y|\xvec, \mathbb{D}])^2 | \xvec, \mathbb{D}\right]  + (f(\xvec; \mathbb{D}) - \E[y|\xvec])^2 \; .
\end{align}
\noindent The term ${\E\left[(y - \E[y|\xvec, \mathbb{D}])^2 | \xvec, \mathbb{D}\right]}$ does not depend on the data $\mathbb{D}$, nor on $f$. It is simply the variance of $y$ given $\xvec$. Therefore, only the term ${(f(\xvec; \mathbb{D}) - \E[y|\xvec])^2}$ measures the effectiveness of $f$ to predict $y$. The mean-squared error of $f$ as an estimator of the best prediction $\E[y|\xvec]$ is given by \cite{Geman1992}
\begin{equation}
  \label{eq:eqapp1-20-5}
  \E_{\mathbb{D}}\left[(f(\xvec; \mathbb{D}) - \E[y|\xvec])^2\right] \; ,
\end{equation}
\noindent where $\E\limits_{\mathbb{D}}$ is the expectation with respect to the training set, $\mathbb{D}$, that is the average over all possible sampled training samples $\mathbb{D}$.

The error measured in Eq.\ref{eq:eqapp1-20-5} can be further developed for any $\xvec$ as follows \cite{Geman1992}
\begin{align}
  \E_{\mathbb{D}}\left[(f(\xvec; \mathbb{D}) - \E[y|\xvec])^2\right] & = 
  \E_{\mathbb{D}}\left[( (f(\xvec; \mathbb{D}) + \E_{\mathbb{D}}[f(\xvec; \mathbb{D})]) + (\E_{\mathbb{D}}[f(\xvec; \mathbb{D})] -\E[y|\xvec]) )^2\right] \label{eq:eqapp1-20-5-1} \\
  & = \E_{\mathbb{D}}\left[(f(\xvec; \mathbb{D}) - \E_{\mathbb{D}}[f(\xvec; \mathbb{D})])^2\right] + \E_{\mathbb{D}}\left[(\E_{\mathbb{D}}[f(\xvec; \mathbb{D})] - \E[y|\xvec])^2\right] \nonumber \\ 
  &+ 2 \E_{\mathbb{D}}\left[(f(\xvec;\mathbb{D}) - \E_{\mathbb{D}}[f(\xvec; \mathbb{D})]) \cdot (\E_{\mathbb{D}}[f(\xvec;\mathbb{D})] - \E[y|\xvec]) \right]
  \label{eq:eqapp1-20-6} \\
  & = \E_{\mathbb{D}}\left[(f(\xvec; \mathbb{D}) - \E_{\mathbb{D}}[f(\xvec; \mathbb{D})])^2\right] +
  (\E_{\mathbb{D}}[f(\xvec; \mathbb{D})] - \E[y|\xvec])^2 \nonumber \\ 
  & + 2 \E_{\mathbb{D}}\left[f(\xvec;\mathbb{D}) - \E_{\mathbb{D}}[f(\xvec; \mathbb{D})]\right] \cdot (\E_{\mathbb{D}}[f(\xvec;\mathbb{D})] - \E[y|\xvec]) \label{eq:eqapp1-20-7} \\
  & = \underbrace{(\E_{\mathbb{D}}[f(\xvec; \mathbb{D})] - \E[y|\xvec])^2}_{\text{Bias}}  + \underbrace{\E_{\mathbb{D}}\left[(f(\xvec; \mathbb{D}) - \E_{\mathbb{D}}[f(\xvec; \mathbb{D})])^2\right]}_{\text{Variance}} \label{eq:eqapp1-20-8} \; .
\end{align}

From Eq.\ref{eq:eqapp1-20-8}, one can see that the bias is the mean error of the average models which are trained over infinite samples. Therefore, this error depends only on the model capability to model the data, i.e., model complexity. Similarly, the variance shows the capability of the model to model the variations of the data, again this is related to the complexity of the model.

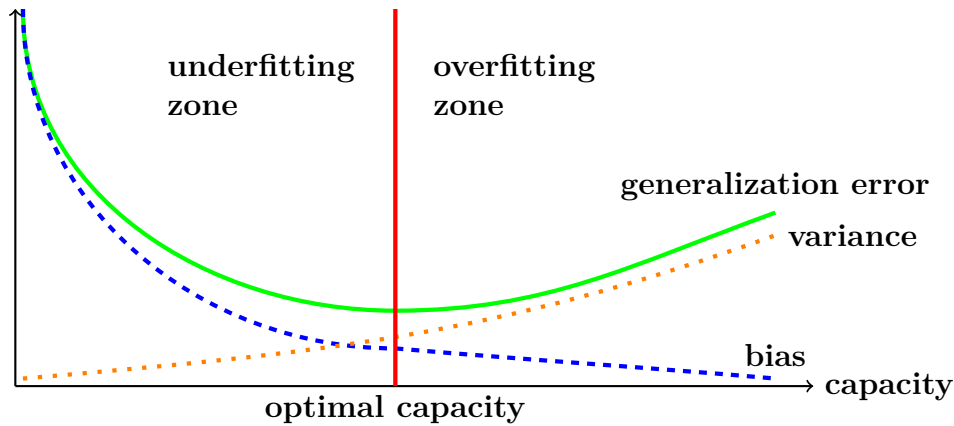
\begin{figure}[!htbp]
  \begin{center}
		\begin{tikzpicture}
\coordinate (center1) at (0, 0);
\coordinate (xneg) at (0, 0);
\coordinate (xpos) at (10.5, 0);
\coordinate (ypos) at (0, 5);
\coordinate (yneg) at (0, 0);

\coordinate (sge) at (0.1, 5);
\coordinate (sgem) at (5, 1);
\coordinate (ege) at (10, 2.3);

\coordinate (ste) at (0.1, 5);
\coordinate (stem) at (5, 0.5);
\coordinate (ete) at (10, 0.1);

\coordinate (sve) at (0.1, 0.1);
\coordinate (eve) at (10, 2);

\coordinate (optl) at (5, 0); 
\coordinate (optu) at (5, 5);

\draw[->, thick] (xneg) -- (xpos) node[right] (xaxis) {\textbf{capacity}};
\draw[->, thick] (yneg) -- (ypos) node[left] (yaxis) {};

\draw[color=green, ultra thick] (sge) to[out=270,in=180] (sgem) node[above, color=black]  {};
\draw[color=green, ultra thick] (sgem) to[out=0,in=200] (ege) node[above, color=black]  {\textbf{generalization error}};

\draw[color=blue, dashed, ultra thick] (ste) to[out=270,in=180] (stem) node[above, color=black]  {};
\draw[color=blue, dashed, ultra thick] (stem) to[out=-5,in=175] (ete) node[above, color=black]  {\textbf{bias}};

\draw[color=orange, loosely dotted, ultra thick] (sve) to[out=5,in=200] (eve) node[right, color=black]  {\textbf{variance}};

\draw[color=red, ultra thick] (optl) -- (optu);
\node at ($(optl) + (0mm, -3mm)$) {\textbf{optimal capacity}};
\node[text width=3cm] at ($(optu) + (20mm, -10mm)$) {\textbf{overfitting \\ zone}};
\node[text width=3cm] at ($(optu) + (-15mm, -10mm)$) {\textbf{underfitting \\ zone}};

\end{tikzpicture}
	\end{center}
  \caption[\bel{Relationship between model capacity, its bias and variance and the generalization error.}]{Typical relationship between model capacity, its bias and variance and the generalization error. (Reference: \cite{Goodfellowbook2016})}
  \label{fig:figapp1-4-appendix4}
\end{figure}

As a consequence, models with small capacity will tend to have high bias because they are enable to fit well the data and low variance because they do not consider the variation in the data. In the other hand, models with high capacity will have lower bias because they can fit well the data, but they have high variance because they are sensitive to changes in the data. Hence, a tradeoff is necessary to select a model with lower bias and variance. Fig.\ref{fig:figapp1-4-appendix4} shows a typical behavior of model bias and variance with respect to the model capacity. When the capacity of the model increases, the bias tends to decrease and the variance to increase yielding an U-shape of the generalization error. Above the optimum capacity, the model tends to have lower bias and higher variance. This relation is similar to the relation between the capacity, underfitting and overfitting.

When applying the regularization over a model that overfits (high capacity), we attempts to bring it from the overfitting regime toward the optimal regime by reducing its variance but without introduction significant bias.

\FloatBarrier

\subsection[Feedforward Neural Networks]{Feedforward Neural Networks}
\label{sub:appendix4fnn}

We cover in this section more technical details on neural networks including
\begin{enumerate*}
  \item backpropagation, derivatives computation, and issues, (Sec.\ref{subsub:appendix4backprop}),
  \item nonlinear activation functions (Sec.\ref{subsub:appendix4nonlinearfunctions}, \ref{subsub:appendix4perceptronforclassification}),
  \item universal approximation theorem (Sec.\ref{subsub:appendix4univapproxtheoremanddepth}),
  \item and other neural architectures (Sec.\ref{subsub:appendix4otherneuralarchitectures}).
\end{enumerate*}

\subsubsection{Backpropagation, Computing the Derivatives, and Issues}
\label{subsub:appendix4backprop}
The error minimization using gradient descent \cite{hadamard1908memoire} in the parameters space of differentiable systems has been discussed since the 60s \cite{kelley1960gradient, bryson1961, BRYSON-DENHAM-61A}. Steepest descent in the weights space of such systems can be performed \cite{bryson1961, kelley1960gradient, bryson1969applied} by iterating the chain rule \cite{leibniz1676, de1716analyse} to dynamic programming \cite{Bellman1957}. A simplified derivation of this backpropagation using the chain rule can be found in \cite{dreyfus1962}.

Explicit Backpropagation for arbitrary and discrete neural network-like systems was first described in 1970 in the master thesis of the Finnish mathematician and computer scientist Seppo Linnainmaa \cite{Linnainmaa1970, Linnainmaa1976} although without referencing neural networks. He did implement it in FORTRAN. Now, this method is mostly known as automatic differentiation \cite{Griewank2012}.

In 1974, Paul Werbos was the first to suggest the possibility to use the backpropagation described by Seppo Linnainmaa, after studying it in depth in his thesis \citep{Werbos74}, to train neural networks. However, Werbos did not publish his work at the time probably because of the AI winter until 1981 \cite{Werbos81sensitivity}. Related works were published later \cite{Parker85, LeCun85, LeCun89}. In 1986, a paper \cite{Rumelhart86} by David Rumelhart, Geoffrey Hinton, and Ronald Williams made the backpropagation a popular method for training neural networks with multi-layers where they showed useful internal representations learned at the hidden layers.

\begin{figure}[!htbp]
  \begin{center}
  \begin{turn}{-90}
		\begin{minipage}{20cm}
			\begin{tikzpicture}
  \coordinate (p1) at (0, 0);
  \coordinate (xn) at (0, 1.2);
  \coordinate (xdots) at (0, 2.4);
  \coordinate (x3) at (0, 3.6);
  \coordinate (x2) at (0, 4.8);
  \coordinate (x1) at (0, 6);
  
  \coordinate (sum1) at (3, 0.6);
  \coordinate (sumdots) at (3, 2.4);
  \coordinate (sum2) at (3, 3.6);
  \coordinate (sum3) at (3, 6);
  
  \coordinate (ym) at (5.7, 0.6);
  \coordinate (ydots) at (5.7, 2.4);
  \coordinate (y2) at (5.7, 3.6);
  \coordinate (y1) at (5.7, 6);
  
  \draw (p1) node[circle, draw] (cp1) {$+1$};
  \draw (xn) node[circle, draw, inner sep=0pt] (cxn) {$\frac{\partial \ell}{\partial x^{D_0}}$};
  \draw (xdots) node (cxdots) {$\vdots$};
  \draw (x3) node[circle, draw, inner sep=0pt] (cx3) {$\frac{\partial \ell}{\partial x^{3}}$};
  \draw (x2) node[circle, draw, inner sep=0pt] (cx2) {$\frac{\partial \ell}{\partial x^{2}}$};
  \draw (x1) node[circle, draw, inner sep=0pt] (cx1) {$\frac{\partial   \ell}{\partial x^{1}}$};
  
  \draw (sum1) node[circle, draw, inner sep=0pt] (csum1) {$\frac{\partial \ell}{\partial z^{D_1}_1}$};
  \draw (sumdots) node (csumdots) {$\vdots$};
  \draw (sum2) node[circle, draw, inner sep=0pt] (csum2) {$\frac{\partial \ell}{\partial z^{2}_1}$};
  \draw (sum3) node[circle, draw, inner sep=0pt] (csum3) {$\frac{\partial \ell}{\partial z^{1}_1}$};
  
  \draw (ym) node[circle, draw, inner sep=0pt] (cym) {$\frac{\partial \ell}{\partial \hat{y}^{D_1}_1}$};
  \draw (ydots) node (cydots) {$\vdots$};
  \draw (y2) node[circle, draw, inner sep=0pt] (cy2) {$\frac{\partial \ell}{\partial \hat{y}^2_1}$};
  \draw (y1) node[circle, draw, inner sep=0pt] (cy1) {$\frac{\partial \ell}{\partial \hat{y}^1_1}$};

  \draw[->, thick] (csum3) -- (cx1) node[pos=0.5, above] {$\bm{W}^{\top}_1$};
  \draw[->, thick] (csum3) -- (cx2) node {};
  \draw[->, thick] (csum3) -- (cx3) node {};
  \draw[->, thick] (csum3) -- (cxn) node {};
  \draw[->, thick] (csum3) -- (cp1) node {};
  
  \draw[->, thick] (csum2) -- (cx1) node {};
  \draw[->, thick] (csum2) -- (cx2) node {};
  \draw[->, thick] (csum2) -- (cx3) node {};
  \draw[->, thick] (csum2) -- (cxn) node {};
  \draw[->, thick] (csum2) -- (cp1) node {};
  
  \draw[->, thick] (csum1) -- (cx1) node {};
  \draw[->, thick] (csum1) -- (cx2) node {};
  \draw[->, thick] (csum1) -- (cx3) node {};
  \draw[->, thick] (csum1) -- (cxn) node {};
  \draw[->, thick] (csum1) -- (cp1) node {};

  \draw[<-, thick] (csum3) -- (cy1) node[pos=0.5, above] (nonlinear1) {$\cdot \nabla_{z^1_1}\phi_1(z^1_1)$};
  \draw[<-, thick] (csum2) -- (cy2) node[pos=0.5, above] (nonlinear2) {$\cdot \nabla_{z^2_1}\phi_1(z^2_1)$};
  \draw[<-, thick] (csum1) -- (cym) node[pos=0.5, above] (nonlinear3) {$\cdot \nabla_{z^{D_1}_1}\phi_1(z^{D_1}_1)$};
  
  \draw (ydots) node (y) [right=1mm] {$\frac{\partial \ell}{\partial \hat{\yvec}_1}$};
  \draw (xdots) node (x) [left=1mm] {$\frac{\partial \ell}{\partial \xvec}$};
  
  \draw (cy1) node (cdotsh1) [right=7mm] {$\cdots$};
  \draw (cy2) node (cdotsh2) [right=7mm] {$\cdots$};
  \draw (cym) node (cdotshm) [right=7mm] {$\cdots$};
  \coordinate (p12) at (8, 0);
  \coordinate (xn2) at (8, 1.2);
  \coordinate (xdots2) at (8, 2.4);
  \coordinate (x32) at (8, 3.6);
  \coordinate (x22) at (8, 4.8);
  \coordinate (x12) at (8, 6);
  
  \coordinate (sum12) at (11, 0.6);
  \coordinate (sumdots2) at (11, 2.4);
  \coordinate (sum22) at (11, 3.6);
  \coordinate (sum32) at (11, 6);
  
  \coordinate (ym2) at (13.7, 0.6);
  \coordinate (ydots2) at (13.7, 2.4);
  \coordinate (y22) at (13.7, 3.6);
  \coordinate (y12) at (13.7, 6);
  
  \draw (p12) node[circle, draw] (cp12) {$+1$};
  \draw (xn2) node[circle, draw, inner sep=0pt] (cxn2) {\small $\frac{\partial \ell}{\partial \hat{y}^{D_{K-1}}_{K-1}}$};
  \draw (xdots2) node (cxdots) {$\vdots$};
  \draw (x32) node[circle, draw, inner sep=0pt] (cx32) {\small $\frac{\partial \ell}{\partial \hat{y}^3_{K-1}}$};
  \draw (x22) node[circle, draw, inner sep=0pt] (cx22) {\small $\frac{\partial \ell}{\partial \hat{y}^2_{K-1}}$};
  \draw (x12) node[circle, draw, inner sep=0pt] (cx12) {\small $\frac{\partial \ell}{\partial \hat{y}^1_{K-1}}$};
  
  \draw (sum12) node[circle, draw, inner sep=0pt] (csum12) {\small $ \frac{\partial \ell}{\partial z^{D_K}_K}$};
  \draw (sumdots2) node (csumdots2) {$\vdots$};
  \draw (sum22) node[circle, draw, inner sep=0pt] (csum22) {\small $ \frac{\partial \ell}{\partial z^{2}_K}$};
  \draw (sum32) node[circle, draw, inner sep=0pt] (csum32) {\small $ \frac{\partial \ell}{\partial z^{1}_K}$};
  
  \draw (ym2) node[circle, draw, inner sep=0pt] (cym2) {\small $\frac{\partial \ell}{\partial \hat{y}^{D_K}_K}$};
  \draw (ydots2) node (cydots2) {$\vdots$};
  \draw (y22) node[circle, draw, inner sep=0pt] (cy22) {\small $\frac{\partial \ell}{\partial \hat{y}^2_K}$};
  \draw (y12) node[circle, draw, inner sep=0pt] (cy12) {\small $\frac{\partial \ell}{\partial \hat{y}^1_K}$};

  \draw[->, thick] (csum32) -- (cx12) node[pos=0.5, above] {$\bm{W}^{\top}_K$};
  \draw[->, thick] (csum32) -- (cx22) node {};
  \draw[->, thick] (csum32) -- (cx32) node {};
  \draw[->, thick] (csum32) -- (cxn2) node {};
  \draw[->, thick] (csum32) -- (cp12) node {};
  
  \draw[->, thick] (csum22) -- (cx12) node {};
  \draw[->, thick] (csum22) -- (cx22) node {};
  \draw[->, thick] (csum22) -- (cx32) node {};
  \draw[->, thick] (csum22) -- (cxn2) node {};
  \draw[->, thick] (csum22) -- (cp12) node {};
  
  \draw[->, thick] (csum12) -- (cx12) node {};
  \draw[->, thick] (csum12) -- (cx22) node {};
  \draw[->, thick] (csum12) -- (cx32) node {};
  \draw[->, thick] (csum12) -- (cxn2) node {};
  \draw[->, thick] (csum12) -- (cp12) node {};

  \draw[<-, thick] (csum32) -- (cy12) node[pos=0.5, above] (nonlinear12) {$\cdot \nabla_{z^1_K}\phi_K(z^1_K)$};
  \draw[<-, thick] (csum22) -- (cy22) node[pos=0.5, above] (nonlinear22) {$\cdot \nabla_{z^2_K}\phi_1(z^2_K)$};
  \draw[<-, thick] (csum12) -- (cym2) node[pos=0.5, above] (nonlinear32) {$\cdot \nabla_{z^{D_K}_K}\phi_1(z^{D_K}_K)$};

  \draw (ydots2) node (yy2) [right=1mm] {$\frac{\partial \ell}{\partial \hat{\yvec}_{K}}$};
  \draw (xdots2) node (xx2) [left=1mm] {$\frac{\partial \ell}{\partial \hat{\yvec}_{K-1}}$};
  
  \coordinate (L) at (15.5, 3.6);
  \draw (L) node[circle, draw] (cL) {${\ell}$};
  
  \draw[<-, thick] (cy12) -- (cL);
  \draw[<-, thick] (cy22) -- (cL);
  \draw[<-, thick] (cym2) -- (cL);

  \coordinate (fictif) at (5, 0);
  \draw [thick,decoration={brace,mirror,raise=0.55cm}, decorate, yshift=2ex, line width=2pt]
(fictif.west) -- (cp12.east) node [pos=0.5,anchor=north,yshift=-0.7cm] {\footnotesize
\textbf{Hidden layers}};

\end{tikzpicture}
		\end{minipage}
	\end{turn}
  \end{center}
  \caption[\bel{The backward pass of the backpropagation algorithm.}]{The backward pass of the backpropagation algorithm. (Notation: $D_k, k=1, \cdots, K$ is the dimension of the output of the layer $k$. $D_0=D$ is the dimension of the input $\bm{x}$ of the network. $D_K=M$ is the dimension of the output $\hat{\bm{y}}$ of the network, i.e. $M$. $x^i$ is the $i^{th}$ component of $\bm{x}$. $\hat{y}^j_k$ is the $j^{th}$ component of the output representation $\hat{\bm{y}}_k$ at the layer $k$)}
  \label{fig:figapp2-9}
\end{figure}
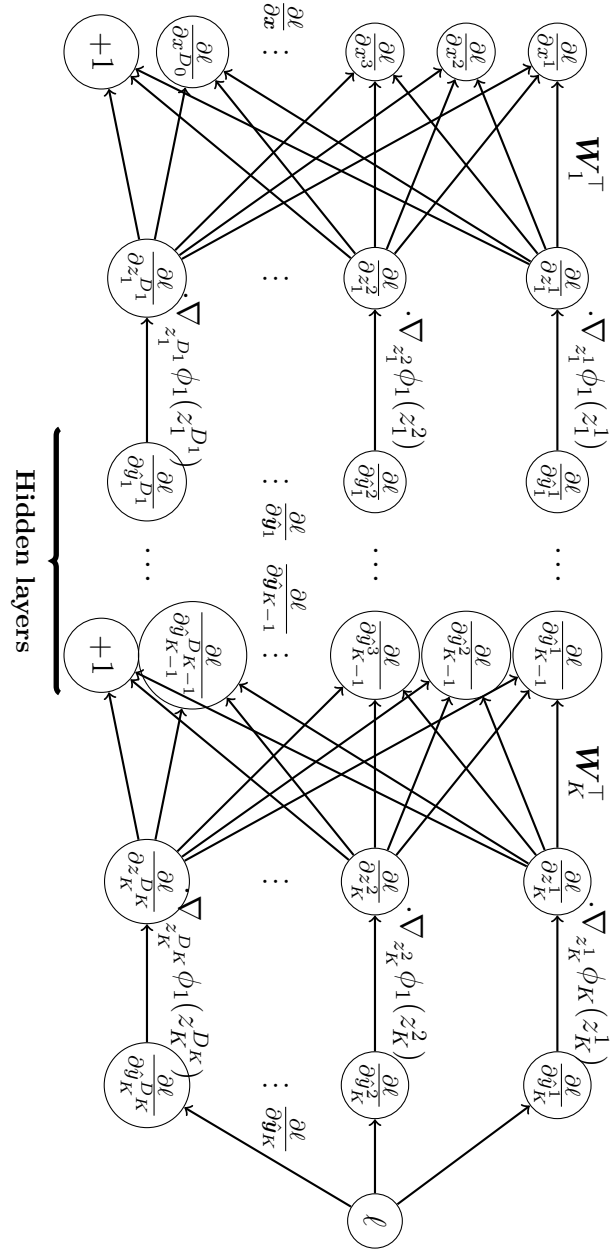

Up to the day of writing this thesis, backpropagation algorithm is the most dominant approach for training neural networks.
\bigskip
\\
\textbf{Technical Details}
\bigskip
\\
Let us consider a multi-layer perceptron with $K$ layers where $\hat{\yvec}_k, k=1, \cdots, K$ is the output of each layer. $\hat{\yvec}_0= \bm{x}$ is the input vector of size $1 \times D+1$. The following layers can be computed as
\begin{equation}
    \label{eq:eqapp2-40}
    \hat{\yvec}_k = \phi_k(\hat{\yvec}_{k-1} \cdot \bm{W}_k), \forall k= 1,\cdots, K \; ,
\end{equation}
\noindent with $\phi(\cdot)$ is an activation function, $\hat{\yvec}_K$ is the output vector of size $1 \times M$. $\ell(\cdot, \cdot)$ is the per-sample loss function (Eq.\ref{eq:eq0-32}). If $\phi(\cdot)$ is differentiable, one can compute the gradient for each layer $\hat{\yvec}_{k-1}$ using the gradient of the layer $\hat{\yvec}_k$ using the chain rule as follows \cite{Goodfellowbook2016, Demyanovtehsis2015}
\begin{align}
    \frac{\partial \ell}{\partial \hat{y}^i_{k-1}} &= \sum^M_{j=1} \frac{\partial \ell}{\partial \hat{y}^j_k} \cdot \frac{\partial \hat{y}^j_k}{\partial \hat{y}^i_{k-1}} \label{eq:eqapp2-41-0} \\
    &= \sum^M_{j=1} \frac{\partial \ell}{\partial \hat{y}^j_k} \cdot \frac{\partial \phi_k(z^j)}{\partial z^j}\Bigg|_{z^j=\hat{\yvec}_{k-1} \cdot \bm{W}_k^j} \cdot \frac{\partial z^j}{\partial \hat{y}^i_{k-1}} \label{eq:eqapp2-41-1} \\
    &= \sum^M_{j=1} \frac{\partial \ell}{\partial \hat{y}^j_k} \cdot \frac{\partial \phi_k(z^j)}{\partial z^j}\Bigg|_{z^j=\hat{\yvec}_{k-1} \cdot \bm{W}_k^j} \cdot W^{ij}_k \label{eq:eqapp2-41}\; ,
\end{align}
\noindent where $\bm{W}_k^{j}$ is the vector of weights connecting the layer $\hat{\yvec}_{k-1}$ with the neuron $j$ at the layer $\hat{\yvec}_k$. Using Eq.\ref{eq:eqapp2-41}, one can compute the gradient of the previous layer $\frac{\partial \ell}{\partial \hat{\yvec}_{k-1}}$ using: the gradient of the current layer $\frac{\partial \ell}{\partial \hat{\yvec}_{k}}$, 
the derivative of the activation function $\nabla_{\bm{z}}\phi_k(\bm{z})$, and the layer weights $W_k$. Therefore, it is possible to compute iteratively the gradient of all layers from $K-1$ to $0$ using $\frac{\partial \ell}{\partial \hat{y}_{K}}$ in Eq.\ref{eq:eq0-35} at the first iteration. Using the chain rule, one can compute the weights gradients $\frac{\partial \ell}{\partial \bm{W}_{k}}$ as follows \cite{Goodfellowbook2016, Demyanovtehsis2015}
\begin{align}
    \frac{\partial \ell}{\partial W_k^{ij}} & = \frac{\partial \ell}{\partial \hat{y}^j_k} \cdot \frac{\partial \hat{y}^j_k}{\partial W_k^{ij}} \label{eq:eqapp2-42-0} \\
    & = \frac{\partial \ell}{\partial \hat{y}^j_k} \cdot \frac{\partial \phi_k(z^j)}{\partial z^j} \Bigg|_{z^j = \hat{\yvec}_{k-1} \cdot \bm{W}^j_k} \cdot \frac{\partial z^j}{\partial W_k^{ij}} \label{eq:eqapp2-42-1} \\
    & = \frac{\partial \ell}{\partial \hat{y}^j_k} \cdot \frac{\partial \phi_k(z^j)}{\partial z^j} \Bigg|_{z^j = \hat{\yvec}_{k-1} \cdot \bm{W}^j_k} \cdot \hat{y}^i_{k-1}     \label{eq:eqapp2-42} \; .
\end{align}
\noindent Therefore, using Eq.\ref{eq:eqapp2-42}, one can compute the gradients of all weights at each layer from $K$ to $1$ once $\frac{\partial \ell}{\partial \hat{\yvec}_k}$ is available.
\bigskip
\\
The backpropagation algorithm can be divided into three steps: forward, backward and weight gradient computation.
\begin{enumerate}
    \item \emph{Forward pass}: Initialize the input vector $\hat{\yvec}_0$ to some input samples $\xvec$. Then, iteratively compute the following layers $\hat{\yvec}_k, \text{for } k=1, \cdots, K$. Fig.\ref{fig:fig0-8} illustrates the forward pass.
    \item \emph{Backward pass}: Initialize the gradients estimations $\frac{\partial \ell}{\partial \hat{\yvec}_K}$ using Eq.\ref{eq:eq0-35}. Then, propagate them back through the network layers from $K-1$ to $0$ to estimate $\frac{\partial \ell}{\partial \hat{\yvec}_k}$ using Eq.\ref{eq:eqapp2-41}. Fig.\ref{fig:figapp2-9} illustrates the backward pass.
    \item \emph{Weight updates}: The weights gradients can be obtained in parallel or after finishing the whole backward pass using $\hat{\yvec}_k$ and $\frac{\partial \ell}{\partial \hat{\yvec}_k}$ at each layer (Eq.\ref{eq:eqapp2-42}). The weights gradients are then used to update the weights using Eq.\ref{eq:eq0-33}. Besides stochastic gradient descent, different methods have been developed to search the minimum of a loss function, such as AdaGrad \cite{zeiler1} which adapts the learning rate for each weight, Nesterov accelerated gradient descent \cite{Nesterov1983wy} with a convergence rate of $\frac{1}{t^2}$, LBFGS algorithm \cite{Liu1989} which uses second order gradients information. However, SGD remains the most commonly used method for training neural networks.
\end{enumerate}
 
\FloatBarrier

\textbf{Matrix Representation of the Backpropagation}
\bigskip
\\
In practice, it is easier to use a matrix representation of the backpropagation for efficient computation. For this, we consider every single transformation linear or nonlinear as a single layer. However, for simplicity, we keep the notation $K$ as the total number of layers. In this case, a layer $k$ can be formalized as a function as follows 
\begin{equation}
    \label{eq:eqapp2-43}
    \hat{\yvec}_k = f_k(\hat{\yvec}_{k-1}, \bm{W}_k) \; ,
\end{equation}
\noindent which transforms the the vector $\hat{\yvec}_{k-1}$ into the vector $\hat{\yvec}_k$ using the weights $\bm{W}_k$. Moreover, the loss function $\ell(\cdot, \cdot)$ is considered as the last layer $\hat{\yvec}_{K+1}$ with dimension $1\times 1$. For simplicity, we refer to it as $\ell(\bm{x})$. The forward pass is the composition of all functions $\ell(\xvec) = f_{K+1}(f_K(\cdots f_1(\xvec) \cdots))$ applied to the input vector $\xvec$.

\begin{figure}[!htbp]
  \begin{center}
		\begin{tikzpicture}
\coordinate (dy0) at (0, 0);
\coordinate (dy1) at (3, 0);
\coordinate (dy2) at (6, 0);
\coordinate (dy3) at (9, 0);

\coordinate (l) at (12, 1);

\coordinate (y0) at (0, 2);
\coordinate (y1) at (3, 2);
\coordinate (y2) at (6, 2);
\coordinate (y3) at (9, 2);

\coordinate (w1) at (1.5, 1);
\coordinate (wK) at (7.5, 1);

\draw (dy0) node[circle, draw] (cdy0) {$\bm{dy}_0$};
\draw (dy1) node[circle, draw] (cdy1) {$\bm{dy}_1$};
\draw (dy2) node[circle, draw, inner sep=0pt] (cdy2) {$\bm{dy}_{K-1}$};
\draw (dy3) node[circle, draw] (cdy3) {$\bm{dy}_K$};

\draw (y0) node[circle, draw] (cy0) {$\hat{\bm{y}}_0$};
\draw (y1) node[circle, draw] (cy1) {$\hat{\bm{y}}_1$};
\draw (y2) node[circle, draw, inner sep=0pt] (cy2) {$\hat{\bm{y}}_{K-1}$};
\draw (y3) node[circle, draw] (cy3) {$\hat{\bm{y}}_K$};

\draw (l) node[circle, draw] (cl) {$\ell$};

\draw (w1) node[circle, draw, inner sep=0pt] (cw1) {$\bm{dW}_1$};
\draw (wK) node[circle, draw, inner sep=0pt] (cwK) {$\bm{dW}_K$};

\draw[->, thick] (cy0) -- (cy1) node[pos=0.5, above] {$f_1(\cdot)$};
\draw[->, thick, dotted] (cy1) -- (cy2) node {};
\draw[->, thick] (cy2) -- (cy3) node[pos=0.5, above] {$f_K(\cdot)$};
\draw[->, thick] (cy3) -- (cl) node {};
\draw[->, thick] (cl) -- (cdy3) node {};
\draw[->, thick] (cdy3) -- (cdy2) node[pos=0.5, below] {$\bar{f}_K(\cdot)$};
\draw[->, thick, dotted] (cdy2) -- (cdy1) node {};
\draw[->, thick] (cdy1) -- (cdy0) node[pos=0.5, below] {$\bar{f}_1(\cdot)$};

\draw[->, thick] (cdy1) -- (cw1) node {};
\draw[->, thick] (cy0) -- (cw1) node {};

\draw[->, thick] (cdy3) -- (cwK) node {};
\draw[->, thick] (cy2) -- (cwK) node {};

\end{tikzpicture}
	\end{center}
  \caption[\bel{The backpropagation algorithm in a matrix form.}]{The backpropagation algorithm in a matrix form. (Reference: \cite{Demyanovtehsis2015})}
  \label{fig:figapp2-10}
\end{figure}
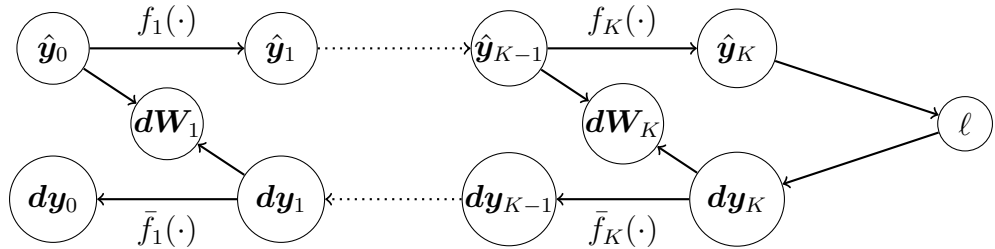

Let us consider the following notations:
\begin{enumerate}
    \item The vector of derivatives with respect to layer values
    \begin{equation}
        \label{eq:eqapp1-44}
        \bm{dy}_k = \frac{\partial \ell}{\partial \hat{\yvec}_k} \; .
    \end{equation}
    \item The backward backpropagation functions, referred to as reverse functions \\ ${\bar{f}(\bm{dy}_k, \bm{W}_k)}$ which computes the vector of derivatives of the previous layer $\hat{\yvec}_{k-1}$ using the derivatives of the next layer $\bm{dy}_k$ and its parameters $\bm{W}_k$. From Eq.\ref{eq:eqapp2-41}, one can write
    \begin{equation}
        \label{eq:eqapp2-45}
        \bm{dy}_{k-1} = \bar{f}(\bm{dy}_k, \bm{W}_k) = \bm{dy}_k \cdot \bm{J}_{\hat{\yvec}_k}(\hat{\yvec}_{k-1}) \; ,
    \end{equation}
    \noindent where $\bm{J}_{\hat{\yvec}_k}(\hat{\yvec}_{k-1})$ is the Jacobian matrix of the derivatives $\frac{\partial \hat{\yvec}_k}{\partial \hat{\yvec}_{k-1}}$. We note that the first Jacobian $\bm{J}_{\hat{\yvec}_{K+1}}(\hat{\yvec}_K) = \bm{dy}_K = \frac{\partial \ell}{\partial \hat{\yvec}_K}$. 
\end{enumerate}
Most of the functions $f_k$ are one of the two following types:
\begin{itemize}
    \item \emph{Linear with weights}: Therefore, $\hat{\yvec}_k = \hat{\yvec}_{k-1} \cdot \bm{W}_k \Rightarrow \bm{dy}_{k-1} = \bm{dy}_k \cdot \bm{W}^{\top}_k$.
    \item \emph{Nonlinear without weights}: Therefore, $\hat{\yvec}_k = \phi_k(\hat{\yvec}_{k-1}) \Rightarrow \bm{dy}_{k-1} = \bm{dy}_k \cdot \bm{J}_{\phi_k}(\hat{\yvec}_{k-1})$. Usually, the nonlinear function $\phi_k(\cdot)$ is element-wise, therefore, they have a squared and diagonal Jacobian $\bm{J}_{\phi_k}(\hat{\yvec}_{k-1})$.
\end{itemize}
Now, we introduce the notation of the vector weight gradients
\begin{equation}
    \label{eq:eqapp2-46}
    \bm{dW}_k = \frac{\partial \ell}{\partial \bm{W}_k} \; .
\end{equation}
\noindent Then, from Eq.\ref{eq:eq0-34}, we can write
\begin{equation}
    \label{eq:eqapp2-47}
    \bm{dW}_k = \bm{J}_{\hat{\yvec}_{k}}(\bm{W}_k) \cdot \bm{dy}_k \; ,
\end{equation}
\noindent where $\bm{J}_{\hat{\yvec}_{k}}(\bm{W}_k)$ is the Jacobian matrix of the derivatives with respect to the weights $\frac{\partial \hat{\yvec}_k}{\partial \bm{W}_k}$. In this case, only the linear functions with weights are considered. As a consequence, their Jacobian is equivalent to $\hat{\yvec}^{\top}_{k-1}$:
\begin{equation}
    \label{eq:eqapp2-48}
    \bm{dW}_k = \hat{\yvec}^{\top}_{k-1} \cdot \bm{dy}_k \; .
\end{equation}
The backpropagation diagram using a matrix representation is illustrated in Fig.\ref{fig:figapp2-10}.
\bigskip
\\
\textbf{Backpropagation Issues}
\bigskip
\\
By the end of $1980^{\prime}$s, it seems that the backpropagation by itself was not enough to train neural networks with many hidden layers, i.e., deep networks. Most applications focused on neural networks with few hidden layers, i.e., shallow networks. Adding more hidden layers did not often offer empirical benefits. This practical limitation of the backpropagation was accepted at the time. Moreover, the idea of using shallow networks was motivated furthermore by a theorem \cite{Kolmogorov65, hecht1989, hornik1989} (Sec.\ref{subsub:appendix4univapproxtheoremanddepth}) that states that a neural network with one hidden layer with enough units can approximate any multivariate continuous function with arbitrary accuracy.

The issue raised in training deep neural networks using backpropagation was fully understood by 1991 where Hochreiter presented in his diploma thesis \cite{Hochreiter91} a breakthrough work which clarified this issue. Hochreiter's work formally identified a major reason of the backpropagation failure to train deep networks. Typically, deep networks suffer from what is now known as vanishing gradients or exploding gradients. With standard activation functions such as the sigmoid function, cumulative backpropagated error signals can either shrink rapidly and vanish, or grow out of bounds and explode. The former is more likely to happen in feedforward networks while the later is more known as an issue in recurrent neural networks.

Over the years, several approaches to partially overcome this fundamental deep learning issue have been proposed and most of them are based on augmenting the backpropagation using unsupervised learning. First, the model is pre-trained using unsupervised learning, then fine-tuned using supervised learning. For example, this technique has been explored in recurrent neural networks \cite{schmidhuber1992, mydeep2013}. Deep feedforward networks can be also pre-trained in a layer-wise fashion by stacking auto-encoders \cite{ballard1987modular, bengio07, vincent2008}. Similarly, Deep Belief Networks (DBNs) can be pre-trained \cite{hinton06SCI, hinton06NC} by stacking many Restricted Boltzmann Machines (RBMs) \cite{smolensky86}. Recurrent neural networks can benefit from gradient clipping \cite{mikolov2012statistical, Pascanu2012arxiv, PascanuMB13} to avoid exploding the gradients. Activation functions that can saturate may lead quickly to vanishing the gradients such as the sigmoid or the hyperbolic tangent functions which saturate at either tail (0 or 1 for the sigmoid and -1 and 1 for the hyperbolic tangent). This saturation leads local gradients to be almost zero which causes the gradients to vanish. To avoid this, new activation functions with no saturation regime have been proposed such as the Rectified Linear Unit (ReLU) \cite{JarrettKRL09ICCV, NairHicml10, AISTATS2011GlorotBB11}, Leaky ReLU \cite{Maas13rectifiernonlinearities}, Parametric ReLU \cite{HeZRS15iccv} and maxout \cite{GoodfellowWMCB13ICML} which generalizes the Leaky and the Parametric ReLU (Sec.\ref{subsub:appendix4nonlinearfunctions}). Hessian-free optimization can help alleviate the problem for feedforward neural networks \cite{Moller93, Pearlmutter93, Martens10} and recurrent neural networks \cite{Martens2011hessfreeICML}. Moreover, today's GPUs provide a huge computational power that allows for propagating errors a few layers further down within reasonable time. This makes implementing deep neural network easier, accessible, and helps popularizing such models in different domains.

In the next section, we present some types of nonlinear activation functions.

\subsubsection{Nonlinear Activation Functions}
\label{subsub:appendix4nonlinearfunctions}
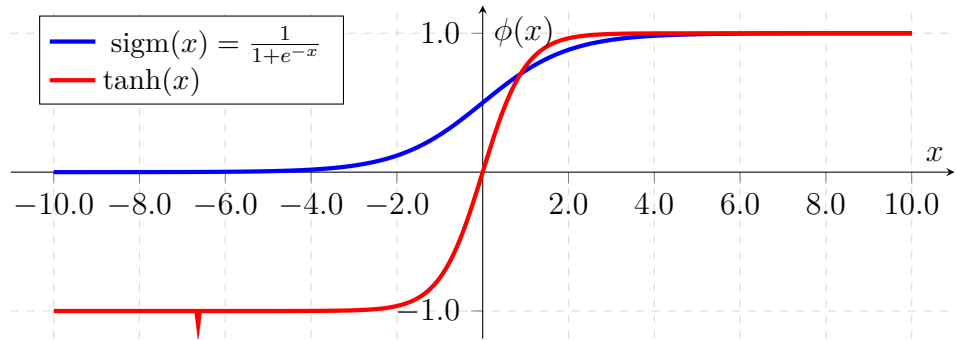
\begin{figure}[!htbp]
        \begin{center}
      		\begin{tikzpicture}
  \begin{axis}[
    	legend pos=north west,
        axis x line=middle,
        axis y line=middle,
        x tick label style={/pgf/number format/fixed,
                            /pgf/number format/fixed zerofill,
                            /pgf/number format/precision=1},
        y tick label style={/pgf/number format/fixed,
                            /pgf/number format/fixed zerofill,
                            /pgf/number format/precision=1},
        grid = major,
        width=14cm,
        height=6cm,
        grid style={dashed, gray!30},
        xmin=-11,     
        xmax= 11,    
        ymin= -1.2,     
        ymax= 1.2,   
        xlabel=$x$,
        ylabel=$\phi(x)$,
        tick align=outside,
        enlargelimits=false]
      \addplot[domain=-10:10, blue, ultra thick,samples=500] {1/(1+exp(-x))};
      \addplot[domain=-10:10, red, ultra thick,samples=500] {tanh(x)};
      \addlegendentry{\small $\sigm(x)=\frac{1}{1+e^{-x}}$}
      \addlegendentry{\small $\tanh(x) \phantom{=\tanh(x)}$}
    \end{axis}
\end{tikzpicture}
      	\end{center}
    \caption[\bel{Examples of nonlinear activation functions.}]{Examples of nonlinear activation functions. \emph{blue}: Logistic sigmoid function. \emph{red}: Hyperbolic tangent function}
    \label{fig:figapp2-13}
\end{figure}

Training neural networks using gradient based methods requires all the activation functions to be differentiable. For this reason, the Heaviside step function can no longer be used in such setup. Therefore, its approximation with another differentiable function which has similar shape is used instead. We mention two most commonly used approximations of the sigmoid functions which are the logistic function
\begin{equation}
    \label{eq:eqapp2-49}
    \phi(z) = \sigm(z) = \frac{1}{1 + \exp^{(-z)}} \Rightarrow \forall z,\; \sigm(z) \in [0, 1],
\end{equation}
\noindent and hyperbolic tangent function
\begin{equation}
    \label{eq:eqapp2-50}
    \phi(z) = \tanh(z) = \frac{1 - \exp^{(-2z)}}{1 + \exp^{(-2z)}} \Rightarrow \forall z,\; \tanh(z) \in [-1, 1] \; .
\end{equation}
Both sigmoid and hyperbolic tanget function are related to each other
\begin{equation}
    \label{eq:eqapp2-51}
    \tanh(z) = 2  \sigm(2z) - 1 \; .
\end{equation}
\noindent Fig.\ref{fig:figapp2-13} illustrates both functions.

Both activation functions have a convenient gradient $\nabla_z\phi(z)$ that can be computed using the function itself. For logistic function it is formulated as
\begin{equation}
    \label{eq:eqapp2-52}
    \frac{\partial \sigm(z)}{\partial z} = \frac{\exp^{(-z)}}{(1+\exp^{(-z)})^2} = \sigm(z) (1 - \sigm(z)) \; ,
\end{equation}
\noindent while for the hyperbolic tangent function, it is formulated as
\begin{equation}
    \label{eq:eqapp2-53}
    \frac{\partial \tanh(z)}{\partial z} = \frac{4\exp^{(-2z)}}{(1+\exp^{(-2z)})^2} = 1 - \tanh^2(z) \; .
\end{equation}

The activation functions described in this section are mostly used in the hidden units of a neural network. However, they can still be used at the output layer, depending on the output target. In Sec.\ref{subsub:appendix4perceptronforclassification}, we discuss a specific type of activation function for the output layer.

Sigmoid activation functions are not always the best choice. They raise an issue for gradient learning methods. Sigmoid functions are sensitive to $z$ only around $0$. They have an inconvenient property by falling into a saturation regime once the magnitude of $z$ is very high. This saturation drives the gradient to be very close to $0$ which either stops the learning process or makes it very slow.

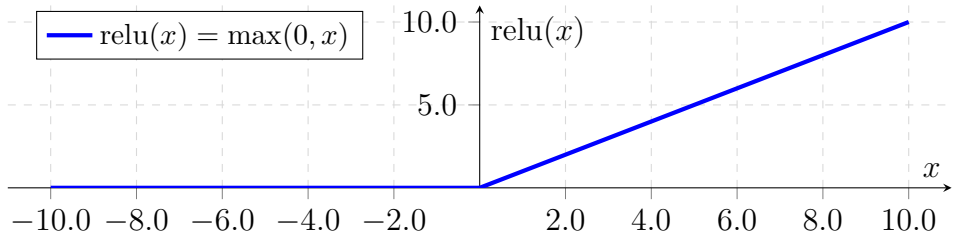
\begin{figure}[!htbp]
  \centering
  \begin{center}
    \begin{tikzpicture}
  \begin{axis}[
    	legend pos=north west,
        axis x line=middle,
        axis y line=middle,
        x tick label style={/pgf/number format/fixed,
                            /pgf/number format/fixed zerofill,
                            /pgf/number format/precision=1},
        y tick label style={/pgf/number format/fixed,
                            /pgf/number format/fixed zerofill,
                            /pgf/number format/precision=1},
        grid = major,
        width=14cm,
        height=4cm,
        grid style={dashed, gray!30},
        xmin=-11,     
        xmax= 11,    
        ymin= 0,     
        ymax= 11,   
        xlabel=$x$,
        ylabel=$\relu(x)$,
        tick align=outside,
        enlargelimits=false]
      \addplot[domain=-10:10, blue, ultra thick,samples=500] {max(0, x)};
      \addlegendentry{\small $\relu(x)=\max(0, x)$}
    \end{axis}
\end{tikzpicture}
  \end{center}
  \caption{Rectified linear unit function.}
  \label{fig:figapp2-14}
\end{figure}

Designing new activation functions is an active field. Recently, new functions have been proposed with much care about the saturation issue. For instance, REctified Linear Unit \cite{JarrettKRL09ICCV, NairHicml10, AISTATS2011GlorotBB11} (Fig.\ref{fig:figapp2-14})
\begin{equation}
    \label{eq:eqapp2-54}
    \phi(z) = \relu(z) = \max(z, 0) = z \cdot \ind_{(z>0)} \; ,
\end{equation}
\noindent is much faster in computation and does not have the saturation problem. Moreover, it allows a faster training \cite{krizhevsky12}. Its derivative is $1$ when $z$ is positive, and $0$ when it is negative
\begin{equation}
    \label{eq:eqapp2-55}
    \frac{\partial \relu(z)}{\partial z} = \ind_{(z>0)} \; .
\end{equation}
\noindent ReLU function is differentiable everywhere except in $z=0$. However, this does not seem to raise issues in practice \cite{krizhevsky12}. This issue is dealt with in implementation where usually the value $1$ is returned as the value of the derivative at $0$ \cite{Goodfellowbook2016}.

Later, more improvements were proposed such as Leaky ReLU \cite{Maas13rectifiernonlinearities}, parametric ReLU \cite{HeZRS15iccv}. Maxout activation function \cite{GoodfellowWMCB13ICML} generalizes ReLU function. Instead of applying an element-wise function, maxout function divides $z$ into $k$ groups of values. Each maxout unit then outputs the maximum element of the group
\begin{equation}
    \label{eq:eqapp2-56}
    \maxout_i(\bm{z}) = \max\limits_{j\in[1,k]} z_{ij} \; .
\end{equation}
\noindent For instance, in the case of convolutional layer, maxout takes the maximum over a group of feature maps.
\bigskip
\\
Many other types of activation functions are possible but are less commonly used. It is impractical to list them all within this thesis. More details on this matter can be found in \cite{Goodfellowbook2016}.

In the next section, we present an activation function for the output layer.

\subsubsection[Activation Function for Output Units: Classification Case]{Activation Function for Output Units: \\ Classification Case}
\label{subsub:appendix4perceptronforclassification}
The activation function of the output layer is considered separately because they must be set according to the target nature $\y$. For instance, in the case of regression with $\yvec \in [0, 1]^M$ one can consider using the logistic function. In the case $\yvec \in [-1, 1]^M$, the hyperbolic tangent function can be more suitable. In either cases, one can consider a linear output.

In classification case, things are slightly different. In the case of binary classification, one can consider one single output unit to model a Bernoulli distribution ${\hat{y} = \hat{y}_K = P(y = 1 | \xvec)}$. In this case, the logistic function, which has the range $[0, 1]$, is enough \cite{Goodfellowbook2016}
\begin{equation}
    \label{eq:eqapp2-57}
    \hat{y} = \hat{y}_K = \sigm_K(z) = P(y=1 | \xvec) \quad ,\; \text{such as } z = \hat{\yvec}_{K-1} \cdot \bm{W}_K  \; .
\end{equation}
\noindent Therefore,
\begin{equation}
    \label{eq:eq0-58}
    1 - \hat{y}_K = 1 - \sigm_K(z) = P(y=0 | \xvec) \quad ,\; \text{such as } z = \hat{\yvec}_{K-1} \cdot \bm{W}_K  \; .
\end{equation}
\noindent These results are valid under the use of the maximum log-likelihood as a cost function \cite{Goodfellowbook2016}.

In the case of $M$ classes, $M > 2$, we need $M$ output units to predict a vector of values to model a multinoulli distribution
\begin{equation}
    \label{eq:eq0-59}
    \hat{y}^j_K = P(y=j| \xvec) \; .
\end{equation}
In this case, the output vector $\hat{\yvec}$ must satisfy two conditions in order to form a probability distribution
\begin{equation}
    \label{eq:eq0-60}
    \forall j \in [1, M], \hat{y}^j_K \geq 0, \; \sum^M_{j=1} \hat{y}^j_K = 1 \; .
\end{equation}
The softmax function was designed to fill the two conditions
\begin{equation}
    \label{eq:eq0-61}
    \softmax(\bm{z})_j = \frac{\exp^{(z_j)}}{\sum^M_{t=1} \exp^{(z_t)}} \quad , \; \bm{z} = \hat{\yvec}_{K-1} \cdot \bm{W}_K \; .
\end{equation}
\noindent To compute the value of one single output unit, the softmax needs to use the value of all the output units.

When using the maximum log-likelihood, we want to maximize \cite{Goodfellowbook2016}
\begin{equation}
    \label{eq:eqapp2-62}
    \log P(y = j | \xvec) = \log \softmax(\bm{z})_j  = z_j - \log \sum^M_{t=1} z_t \; .
\end{equation}
\noindent When maximizing the log-likelihood, the first term encourages $z_j$ to be pushed up while the second term encourages all $z_t$ to be pushed down. If we consider that $\exp^{(z_t)}$ is insignificant for any $z_t$ that is noticeably less than $\max\limits_t z_t$, the second term can be approximated as: $\log\sum^M_{t=1}z_t \approx \max\limits_t z_t$. Therefore, the maximum log-likelihood always strongly penalizes the most active incorrect prediction. If the correct answer has the strongest activation, then the term $\log \sum^M_{t=1} \exp^{(z_t)} \approx \max\limits_t z_t = z_j$ will roughly cancel the first term $z_j$.

The softmax function has a convenient derivative
\begin{equation}
    \label{eq:eqapp2-63}
    \frac{\partial \softmax(\bm{z})_i}{\partial z_j} = \softmax(\bm{z})_i \times (\delta_{ij} - \softmax(\bm{z})_j) \; ,
\end{equation}
\noindent where $\delta_{ij}$ is the Kronecker delta function. We note that the Jacobian matrix (Eq.\ref{eq:eqapp2-45}) of the softmax is still squared but no longer diagonal, however it is symmetric.

The cross-entropy or the negative maximum likelihood are the common losses used for optimization combined with the softmax function \cite{Goodfellowbook2016}
\begin{equation}
    \label{eq:eqapp2-64}
    \ell(f(\xvec), \yvec) = - \sum^M_{j=1} y_j \times \log f(\xvec)_j, \quad  \; f(\xvec)_j = \softmax(\bm{z})_j = \softmax(\hat{y}^j_K \times \bm{W}^{:, j}_K) \; .
\end{equation}
\noindent To simplify the notation, let us set ${\ell(\bm{z}) = \ell(f(\xvec), \yvec)}$ and ${\phi(\bm{z}) = \softmax(\bm{z})}$. Therefore, ${l(\bm{z}) = \sum^M_{j=1} y_j \times \log \phi(\bm{z})_j}$. Now let us compute the derivative of the softmax function with respect to its input values $\bm{z}$:
\begin{align}
    \frac{\partial \ell(\bm{z})}{\partial z_j} & = \sum^M_{i=1} \frac{\partial \ell(\bm{z})}{\partial \phi(\bm{z})_i} \times \frac{\partial \phi(\bm{z})_i}{\partial z_j} \label{eq:eqapp2-65-0} \\
    & = \sum^M_{i=1} \frac{y_i}{\phi(\bm{z})_i} \times \phi(\bm{z})_i \times (\delta_{ij} - \phi(\bm{z})_j) \label{eq:eqapp2-65-1} \\
    & = \sum^M_{i=1} y_i \times \delta_{ij} - \phi(\bm{z})_j \times \sum^M_{i=1}y_i \label{eq:eqapp2-65-2} \\
    & = y_i - \phi(\bm{z})_i \Rightarrow \frac{\partial \ell(\bm{z})}{\partial \bm{z}} = \yvec - \phi(\bm{z}) \; . \label{eq:eqapp2-65}
\end{align}
\noindent Therefore, the derivative of the softmax function with respect to the input values $\bm{z}$ has a simple formulation which simplifies its implementation and improves numerical stability. In practice, other tricks are actually used even to compute the softmax function in order to avoid numerical issues.

In the next section, we present a well known theoretical property of neural networks. We discuss also their depth and its relation to generalization in practice.

\subsubsection{Universal Approximation Properties and Depth}
\label{subsub:appendix4univapproxtheoremanddepth}
Neural networks models has acquired the reputation that they lack theoretical foundations. The universal approximation theorem \cite{Cybenko1989, hornik1989, Funahashi1989} is one of the few theoretical justifications of the capability of neural networks. For instance in \cite{Cybenko1989}, Cybenko demonstrated that a feedforward network with a linear output and at least one hidden layer with any \quotes{squashing} activation function such as the logistic function can approximate any continuous function on a compact subset of $[0, 1]^m$ with any desired nonzero error, provided that the hidden layer has enough units. A geometric way to understand this result is as follows:
\begin{itemize}
    \item A continuous function on a compact set can be approximated by a piecewise constant function.
    \item A piecewise constant function can be represented as a neural network as follows: for each region where the function is constant, use a neural network as an indicator function for that region. Then, build a final layer with a single node, whose input linear combination is the sum of all the indicators multiplied by a weight equals to the constant value of the corresponding region in the original piecewise constant function.
\end{itemize}
Simply, one can take the continuous function over $[0, 1]^m$, and some target error $\epsilon > 0$, then grid the space $[0, 1]^m$ at a scale $\rho > 0$ to end up with $(1/\rho)^m$ subcubes so that the function which is constant over each subcube is within $\epsilon$ of the target function. A neural network can not precisely represent an indicator function, but it can get very close to it. For $m=1$, it can be shown that a neural network with one hidden layer containing two hidden units can build a \quotes{bump} function which is constant over an interval $[x_0, x_1]$ within the original compact set $[0, 1]$. The \quotes{squashing} function is important in Cybenko theorem. It must have a known limit when the input goes to $+\infty$ where the limit is $1$ and when the input goes to $-\infty$ the limit is $0$. One can handcraft the neural network parameters in order to fix $x_0, x_1$ and in order to build the indicator function. For instance, this can be done by setting the weights of the hidden layer to be extremely huge in order to push the input of the activation function to its limit ($0$ or $1$). Cybenko was aiming at providing a general theorem with minimal layers. He showed that one hidden layer with enough units is enough to construct an approximation with $\epsilon$ error.

Formally, let $\phi(\cdot)$ be a non-constant, bounded and monotonically increasing continuous function. Let $\mathbb{I}_m$ denote the $m$-dimensional unit hypercube $[0, 1]^m$. The space of continuous functions on $\mathbb{I}_m$ is denoted by $C(\mathbb{I}_m)$. Then, given any $\epsilon > 0$ and any function $f \in C(\mathbb{I}_m)$, there exists an integer $N$, real constants $v_i\;, b_i \in \R$ and real vector $\bm{w}_i \in \R^m$, where $i=1, \cdots, N$ such that we may define 
\begin{equation}
    \label{eq:eqapp2-66}
    F(\xvec) = \sum^N_{i=1} v_i \phi(\bm{w}^{\top}_i \xvec + b_i) \; ,
\end{equation}
\noindent as an approximate realization of the function $f$ where $f$ is independent of $\phi$; that is 
\begin{equation}
    \label{eq:eqapp2-67}
    \big|F(\xvec) - f(\xvec) \big| < \epsilon \;, \; \forall \xvec \in \mathbb{I}_m \;.
\end{equation}
Cybenko used in his proof the Hahn-Banach theorem \cite{Luenberger1969}. In the same year, Hornik found the same results \cite{hornik1989} using the Stone-Weiestrass theorem \cite{rudin1964principles}. Independently, Funahashi proved similar theorem \cite{Funahashi1989} using an integral formula presented in \cite{irie1988capabilities}.

A year later, Hornik showed \cite{Hornik1990} that feedforward networks with sigmoid units can approximate not only an unknown function, but also its derivative. Using a theorem in \cite{Sun1992}, Light extended \cite{light1992ridge} Cybenko's results to any continuous function on $\R^n$. Moreover, Light showed that the sigmoid function can be replaced by any continuous function that satisfies some conditions.
\bigskip

Earlier to this, in 1957, Kolmogorov provided a theorem \cite{kolmogorov1957} that states that one can express a continuous multivariate function on a compact set in terms of sums and compositions of a finite number of single variable functions. The main difference between Cybenko's theorem \cite{Cybenko1989} is that Kolmogorov \cite{kolmogorov1957} uses heterogeneous activation functions at the hidden layer whereas Cybenko \cite{Cybenko1989} uses the same activation function. Kolmogorov formulation can be seen as a composition of different neural networks in parallel where each one has a specific type of activation function. Due to this aspect of the theorem, part of the research community \cite{hechtnielsenkolmogorov1987, HechtNielsen1989, Lippmann1988} suggested that Kolmogorov \cite{kolmogorov1957} theorem provided theoretical support for the universality of neural networks, while others disagreed \cite{Girosi1989}.

More details on the universal approximation theorem can be found in \cite{Hassoun1995, csaji2001mscthesis, Anastassiou2016}. \cite{anastassiou2002approximation, steffens2007history, anastassiou2016intelligent} cover more studies on approximation theory.

As much as the universal approximation theorem \cite{Cybenko1989, hornik1989, Funahashi1989} provides a strong theoretical support for neural networks, it does not provide much insight on its application. None of the authors of the theorem provides a way to determine how many hidden neurons are required. Most importantly, they did not provide a learning algorithm to find the network parameters. In addition, the authors use sigmoid functions in the hidden units which we know that they are not the best choice as they cause saturation and prevent learning of the network when using gradient based methods. Later, the universal approximation theorem has been proved using a wider class of activation functions which include the rectified linear unit  \cite{LeshnoLPS93NN}.

The universal approximation theorem simply states that independently of what function we are seeking to learn, there exist a multilayer perceptron with one large hidden layer that is able to represent this function with certain precision. \cite{barron93universalapproximation} provides some bounds on the size of a single layer network needed to approximate a broad class of functions. In the worst case, an exponential number of hidden units may be required. Even if we know the exact number of hidden units which is high-likely to be large, there is no guaranty that the learning algorithm will succeed to learn the right parameters. One of the reasons of this failure is that most likely the network will overfit the training set due to the large number of parameters. In addition, the no free lunch theorem \cite{Wolpert1996} shows that there is no universally superior learning machine. Therefore, the universal approximation theorem proves that feedforward networks are an universal system for representing any function in a sense that given a function, there exists a feedforward network with a one large hidden layer that approximates this function with certain precision. Nor the size of the hidden layer, nor the training procedure to find the network parameters are known. Moreover, there is no guaranty that the network will generalize well to unseen inputs.
\bigskip
\\
\textbf{Depth of the Network and Generalization}
\bigskip
\\
In practice and in many circumstances, it was found that using deeper models can reduce the need of large number of hidden units to represent the desired function and can reduce the generalization error. \cite{Montufar2014NIPS} showed that functions representable with a deep rectifier network can require an exponential number of hidden units with a network with one hidden layer. More precisely, they showed that piecewise networks which can be obtained using rectifier nonlinearities or maxout unit can represent functions with a number of regions that is exponential in the depth of the network.

\begin{figure}[!htbp]
  \centering
  \includegraphics[scale=0.3]{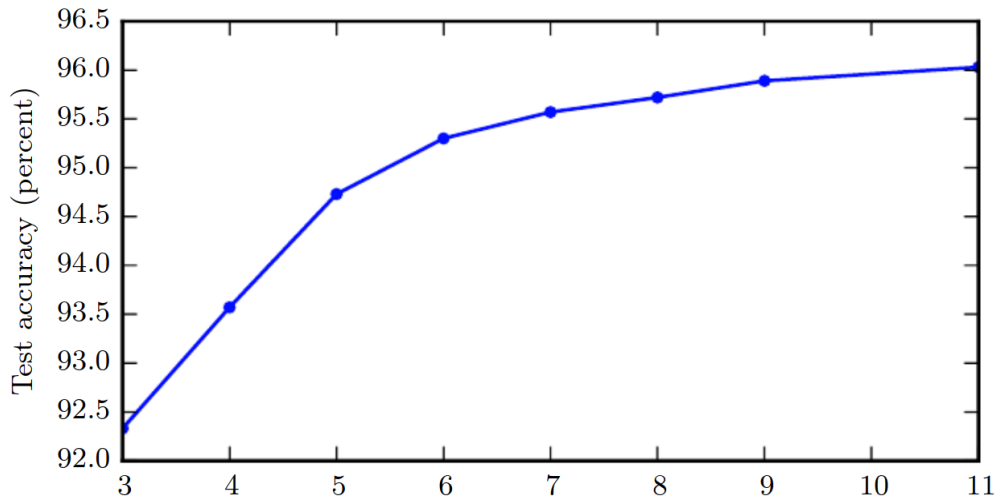}
  \caption[\bel{Effect of depth on the generalization of neural networks.}]{Effect of depth. Empirical results showing that deeper networks generalize better when used to transcribe multidigit numbers from photographs of addresses. Image from \cite{goodfellow2014iclr}. (Credit: \cite{Goodfellowbook2016})}
  \label{fig:figapp2-15}
\end{figure}

\begin{figure}[!htbp]
  \centering
  \includegraphics[scale=0.3]{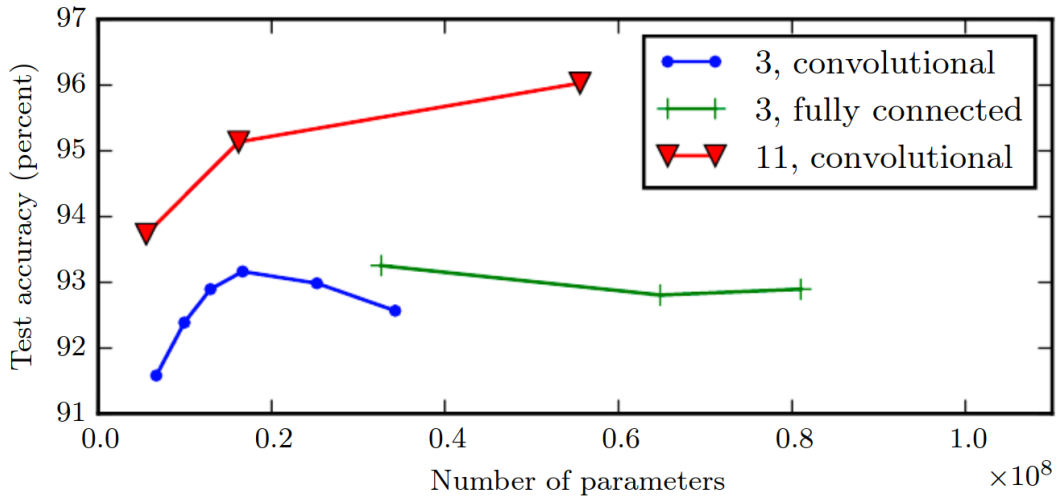}
  \caption[\bel{Effect of the number of parameters on the generalization of neural networks.}]{Effect of the number of parameters. Deeper models tend to perform better compared to less deep models with the same number of parameters. Image from \cite{goodfellow2014iclr}. (Credit: \cite{Goodfellowbook2016})}
  \label{fig:figapp2-16}
\end{figure}

Deep models can also be motivated from statistical view \cite{Goodfellowbook2016}. Choosing a deep model encodes a general belief that the function we want to learn should involve the composition of several simple functions. This can be interpreted from a representation point of view as saying that the learning problem consists of discovering a set of underlying factors of variation that can in turn be described in terms of other simpler underlying factors of variation. Alternatively, one may interpret using deep models as computer program that process the data step by step in order to reach the output decision. The intermediate steps are not necessarily factors of variation but may be seen as pointers that the network uses to organize its internal process. Empirically, greater depth seems  to result in better generalization for a wide variety of tasks \cite{krizhevsky12, SimonyanZ14aCORR} (Fig.\ref{fig:figapp2-15}, \ref{fig:figapp2-16}). However, the need to deep model is merely an empirical fact which is not supported by any theoretical foundation. Recently, it was shown that shallow networks can perform as well as deep networks \cite{Ba2014NIPS}.

Neural network domain has a wide range of architectures that differ mostly depending on the task in hand. In the next section, we describe some of the common architectures.

\subsubsection{Other Neural Architectures}
\label{subsub:appendix4otherneuralarchitectures}
Up to now, we have discussed only feedforward networks which are composed of sequential layers one after the other such as multilayer perceptron and convolutional neural networks. These models are directed acyclic graphs where the information streams from a layer to the next one from the input layer toward the output layer without loops. Sometimes, the information at a layer $i$ can \quotes{skip} the layer $i+1$ to reach the layer $i+2$ or higher \cite{heZRS16, HeZRS16ECCV, Srivastava2016arxiv}. All our work presented in this thesis is validated on feedforward neural networks type.
\bigskip
\\
\textbf{Boltzmann Machines}
\bigskip
\\
One can build a neural model by considering an undirected graph, without direct loop from one neuron to itself such as Boltzmann Machines (BMs) \cite{FahlmanHS83AAAI, AckleyCognScie1985, hintontechrepo1984}. BMs are a type of neural models built using stochastic neurons where each neuron is directly connected to every neuron in the model. They are used to learn arbitrary probability distributions over binary vectors. The set of neurons (stochastic variables) are usually divided into two sets to indicate that the input signal is partial observable. A set of variables indicates the observable part of the signal referred to as visible units while the hidden part is referred to as hidden units. The joint probability distribution of all the variables is described using an energy function. The hidden units state can be inferred given the visible states. Moreover, the visible states can also be inferred given the hidden states which makes this type of architecture a generative model. BMs can be simplified by removing the connection between units at the same layer which leads to what is known as Restricted Boltzmann Machines (RBMs). RBMs can be stacked to form a deep network which can be used for feature extraction or for discrimination such as in Deep Belief Networks (DBNs) \cite{hinton06, Hinton2007Trends}. Unlike RBMs which contain only one layer of hidden units,  Deep Boltzmann Machines (DBMs) have several layers of hidden units \cite{SalHinton07DBM} which have been applied in several tasks including document modeling \cite{SrivastavaDropout2013}. Training Boltzmann Machines-like models is known to be difficult \cite{Hinton12CD} which includes using contrastive divergence \cite{Hinton2002CD} or stochastic maximum likelihood algorithms \cite{Goodfellowbook2016}.
\bigskip
\\
\textbf{Recurrent Neural Networks}
\bigskip
\\
Most feedforward networks such as MLPs can process only fixed length input. In many applications, the input is a sequence and the same goes for the output. Instead of using one MLP per element of the sequence, the MLP is shared across all the inputs which is similar to the idea of weight sharing in convolutional networks where the same filter is applied across the whole 2D image. Moreover, when dealing with sequence data, it is useful to remember what happened in the past. Therefore, the MLP needs its previous states as input, hence, a recurrence is made. Recurrent Neural Networks (RNNs) \cite{Rumelhart86} are a simple MLP duplicated over the input sequence where they take as input their previous output and the current input in the sequence. In theory an RNN with sufficient number of hidden units can approximate any measurable sequence-to-sequence mapping to arbitrary accuracy \cite{Hammer98onthe}. For many sequence labeling tasks, it is beneficial to have access to future as well past context. Bidirectional Recurrent Neural Networks (BRNNs) \cite{Schuster1997BRNN, schuster1999supervisedtehsis, baldi1999exploiting} offer the possibility to do so and showed interesting improvements such as in protein secondary structure prediction \cite{baldi1999exploiting, ChenC04aISNN} and speech processing \cite{schuster1999supervisedtehsis, fukada99boundary}. Training RNNs is done using a variant of backpropagation algorithm by taking in consideration the dependencies to compute the gradient. It is referred to as BackPropagation Through Time (BPTT). When increasing the size of the input sequence, RNNs trained with BPTT seem to have difficulties learning long-term dependencies \cite{Pascanu2013}. This problem is mostly due the vanishing/exploding  gradients issue \cite{Hochreiter91, Hochreiter2001, Bengio1994RNNLTD}. Many attempts were proposed to deal with this issue. Among them, we cite the Long Short-Term Memory (LSTM) \cite{Hochreiter1997LSM} architecture which is similar to vanilla RNN but with a crucial addition to help the network memorizes long-term dependencies. The LSTM models have been found very successful in many applications, such as unconstrained handwriting recognition \cite{GravesNIPS2008}, speech recognition \cite{GravesMH13ICASSP, Graves2014ICML}, handwriting generation \cite{Graves13ARXIVCORR}, machine translation \cite{Sutskever2014SSL}, image captioning \cite{KirosSZ14CORRARXIV, VinyalsTBE15CVPR, XuBKCCSZB15ICML}, and parsing \cite{VinyalsKKPSH15NIPS}. Variant architectures based on LSTMs were proposed known as Gated Recurrent Units (GRUs) \cite{ChoMBB14SSST, Chung2014DLWNIPS, ChungGCB15ICML} where the main difference is how to control the network memory. More work has been done in order to improve the memory aspect of RNNs to deal with sequence-data by providing an entire working memory that allows the networks to hold and manipulate the most relevant information. \cite{weston2014memory} introduced memory networks that contain memory cells that can be addressed via an addressing system using a supervised way. \cite{graves2014neuralturingmachine, GravesNature2016} introduced the neural Turing machine which enables learning to read/write arbitrary content in the memory cells without explicit supervision based on an attention mechanism \cite{BahdanauCB14CORR}. This addressing mechanism has become a standard \cite{SukhbaatarSWF15CORR, JoulinM15NIPS, kumar16pmlr}. The attention mechanism is probably the next big step in recurrent neural networks. Instead of processing all the input sequence and map it into a fixed length vector, the attention mechanism allows the use of all the internal states (memory cells) of the network by taking their weighted combination in order to produce a single output. This can be interpreted as a memory access in computer. However, in the attention mechanism, the access is performed to all the memory instead of selected cells. More details on recurrent neural networks and the last advances can be found in  \cite{Goodfellowbook2016, Graves2012CSI}.
\bigskip
\\
\textbf{Generative Models}
\bigskip
\\
Neural networks can also be used to generate new samples. For instance, variational auto-encoders 
\cite{Doersch16CORR, KingmaW13, SalimansKW15} make a strong assumption about the distribution of the latent variables, for example, assuming a Gaussian distribution. One can sample new hidden codes from the learned hidden distribution, then, use these new hidden codes to generate new samples. Generative Adversarial Networks (GANs) \cite{GoodfellowNIPS2014} are probably one of the hottest topics in neural networks domain these years. GANs are a class of algorithms composed of two networks. A generative network that takes random noise as input and outputs new samples. A discriminative network that attempts to discriminate between true (real) samples and the samples generated by the generative network which are considered fake. Using this technique, one can learn to generate new samples that look real in an unsupervised way. This topic is an ongoing research subject.
\\
\bigskip
\\
More details on the last advances in neural networks can be found in \cite{Goodfellowbook2016}.

\subsubsection{Experimenting in Deep Learning}
\label{subsub:appendix4experimentingindl}
Experimenting in science is encouraged which may lead to more understating of the matter and probably new discoveries. However, heavy experimenting may lead to waist of effort, resources and more importantly, it may open the door to personal interpretations which may bias the results and miss-guide the upcoming research. Unfortunately, deep learning domain is a heavily experimental field where most the founded results are empirical which makes it weak compared to concurrent methods in machine learning, despite its practical high performance. There are many unanswered questions about neural networks and its performance. This makes it difficult to understand its results and makes it seem as a \quotes{magical black box}. Hopefully, neural network field will get more theoretical support in order to set it on the right and solid direction.

In the following, we provide some technical details on regularization.

\FloatBarrier

\subsection[Regularization]{Regularization}
\label{sub:appendix4regularization}
We cover in this section, from a theoretical perspective, the impact of using $L_p$ norm or early stopping, as a regularization, on the parameters of the obtained solution of a learning algorithm.

\subsubsection[$L_p$ Parameters Norm]{$L_p$ Parameters Norms}
\label{subsub:appendix4lpnorm}

We provide more details on the $L_p$ parameters norm regularization for $p \in \{1, 2\}$ (Sec.\ref{para:appendix4l1norm}, Sec.\ref{para:appendix4l2norm}). Let us consider the regularized training objective function
\begin{align}
    \label{eq:eqapp3-post-68}
    \widetilde{J}(\bm{\theta}; \bm{X}, \yvec) = J(\bm{\theta}; \bm{X}, \yvec) + \alpha \Omega(\bm{\theta}) ,\quad \Omega(\bm{\theta}) = \frac{1}{p} \lVert \bm{\theta} \rVert^{p}_p \; .
\end{align}

\paragraph[$L_2$ Parameters Norm]{$L_2$ Parameters Norm}
\label{para:appendix4l2norm}
\mbox{}
\bigskip
\\
\indent For $L_2$ parameters norm regularization, we have
\begin{equation}
    \label{eq:eqapp3-post-69}
    \Omega(\bm{\theta}) = \frac{1}{2} \lVert \bm{w} \rVert^2_2 \; .
\end{equation}

For the next analysis, no bias parameters are assumed, so $\bm{\theta}$ represents only $\bm{w}$. Let us consider a general form of the objective function
\begin{equation}
    \label{eq:eqapp3-post-70}
    \widetilde{J}(\bm{w}; \bm{X}, \yvec) = \frac{\alpha}{2} \bm{w}^{\top} \bm{w} + J(\bm{w}; \bm{X}, \yvec) \; ,
\end{equation}
\noindent and its corresponding gradient
\begin{equation}
    \label{eq:eqapp3-post-71}
    \nabla_{\bm{w}}\widetilde{J}(\bm{w}; \bm{X}, \yvec) = \alpha \bm{w} + \nabla_{\bm{w}} J(\bm{w}; \bm{X}, \yvec) \; .
\end{equation}
\noindent Considering $\epsilon$ a learning rate, the update rule of the gradient descent at each step is performed as follows \cite{Goodfellowbook2016}

\begin{align}
    \bm{w} &\leftarrow \bm{w} - \epsilon (\alpha \bm{w} + \nabla_{\bm{w}} J(\bm{w}; \bm{X}, \yvec)) \label{eq:eqapp3-post-72-0}  \\
    \bm{w} &\leftarrow (1 - \epsilon \alpha) \bm{w} - \epsilon \nabla_{\bm{w}} J(\bm{w}; \bm{X}, \yvec) \label{eq:eqapp3-post-72} \; .
\end{align}
\noindent The last equation (Eq.\ref{eq:eqapp3-post-72}) shows that the $L_2$ regularization modifies the learning rule by multiplicatively shrinking the weight by a constant factor on each step just before performing the updates. In order to get more insights on what happens over the entire training process,  further simplification of the analysis can be made by considering a quadratic approximation of the objective function around the value of the weights that obtains minimal unregularized training cost, ${\bm{w}^* = \argmin_{\bm{w}} J(\bm{w})}$ as follows \cite{Goodfellowbook2016}
\begin{align}
    \hat{J}(\bm{w}) &= J(\bm{w}^*) + \nabla_{\bm{w}} J(\bm{w}^*)^{\top}(\bm{w} - \bm{w}^*) + \frac{1}{2} (\bm{w} - \bm{w}^*)^{\top} \bm{H} (\bm{w} - \bm{w}^*)  \label{eq:eq0-73-0} \\
    \hat{J}(\bm{w}) &= J(\bm{w}^*) + \frac{1}{2} (\bm{w} - \bm{w}^*)^{\top} \bm{H} (\bm{w} - \bm{w}^*) \label{eq:eqapp3-post-73} \; ,
\end{align}
\noindent where $\bm{H}$ is the Hessian matrix of $J$ with respect to $\bm{w}$ evaluated at $\bm{w}^*$. By definition, $\nabla_{\bm{w}}J(\bm{w}^*)=0$ at the minimum $\bm{w}^*$. Given that $\bm{w}^*$ is a local minimum, $\bm{H}$ is a positive semidefinite matrix. In order to find the analytic form of the minimum, the gradient of $\hat{J}$ is computed as follows \cite{Goodfellowbook2016}
\begin{equation}
    \label{eq:eqapp3-post-74}
    \nabla_{\bm{w}} \hat{J}(\bm{w}) = \bm{H} (\bm{w} - \bm{w}^*) \; ,
\end{equation}
\noindent and solve $\nabla_{\bm{w}} \hat{J}(\bm{w}) = 0$ for $\bm{w}$. To study the effect of the weight decay, its gradient is added to Eq.\ref{eq:eqapp3-post-74}. Now, the regularized version of $\hat{J}$ can be solved. Let $\widetilde{\bm{w}}$ denotes the location of the minimum, therefore \cite{Goodfellowbook2016}
\begin{align}
    \alpha \widetilde{\bm{w}} + \bm{H}(\widetilde{\bm{w}} - \bm{w}^*) & = 0 \label{eq:eqapp3-post-75}  \\
    (\bm{H} + \alpha \bm{I}) \widetilde{\bm{w}} & = \bm{H} \bm{w}^* \label{eq:eqapp3-post-76} \\
    \widetilde{\bm{w}} & = (\bm{H} + \alpha \bm{I})^{-1} \bm{H} \bm{w}^* \label{eq:eqapp3-post-77} \; .
\end{align}

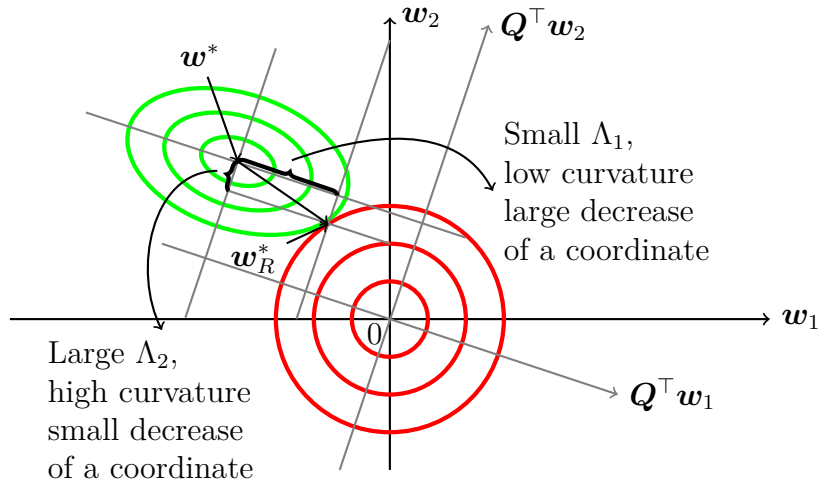
\begin{figure}[!htbp]
  \begin{center}
		\begin{tikzpicture}
\coordinate (center1) at (0, 0);
\coordinate (xneg) at (-5, 0);
\coordinate (xpos) at (5, 0);
\coordinate (ypos) at (0, 4);
\coordinate (yneg) at (0, -2);

\coordinate (qw2low) at (-0.66, -2);
\coordinate (qw2up) at (1.3, 3.9);

\coordinate (qw1low) at (-3, 1);
\coordinate (qw1up) at (3, -1);

\coordinate (center2) at (-2, 2.09);
\coordinate (wstar) at (-2.5, 3.5);
\coordinate (wrstar) at (-1.8, 0.8);

\coordinate (qw2lowg) at (-4, 2.74);
\coordinate (qw2upg) at (1, 1.08);

\coordinate (qw1lowg) at (-2.69, 0.02);
\coordinate (qw1upg) at (-1.5, 3.59);

\coordinate (decomp1) at (0, 3.7);
\coordinate (decomp2) at (-1.23, 0);

\coordinate (decomp3) at (0, 1);
\coordinate (decomp4) at (-2.15, 1.71);

\coordinate (intersection) at (-0.81, 1.26);
\coordinate (p1) at (-0.69, 1.64);
\coordinate (p2) at (-2.12, 1.7);

\coordinate (text1) at (3, 1.7);
\coordinate (text2) at (-3, -1.2);
 
\node at ($(center1) + (-2mm, -2mm)$) {$0$};
\draw[->, thick] (xneg) -- (xpos) node[right] (xaxis) {$\bm{w}_1$};
\draw[->, thick] (yneg) -- (ypos) node[right] (yaxis) {$\bm{w}_2$};

\draw[red, ultra thick] (center1) circle [radius=0.5] node (cr1) {};
\draw[red, ultra thick] (center1) circle [radius=1] node (cr2) {};
\draw[red, ultra thick] (center1) circle [radius=1.5] node (cr3) {};

\draw[->, thick, gray] (qw2low) -- (qw2up) node[right] {\color{black}$\bm{Q}^{\top}\bm{w}_2$};

\draw[->, thick, gray] (qw1low) -- (qw1up) node[right] {\color{black}$\bm{Q}^{\top}\bm{w}_1$};

\draw[green, ultra thick] (center2) circle [x radius=0.5, y radius=0.3, rotate=-18] node (cg1) {};

\draw[green, ultra thick] (center2) circle [x radius=1, y radius=0.6, rotate=-18] node (cg2) {};

\draw[green, ultra thick] (center2) circle [x radius=1.5, y radius=0.9, rotate=-18] node (cg3) {};

\draw[-, thick, gray] (qw2lowg) -- (qw2upg) ;
\draw[-, thick, gray] (qw1lowg) -- (qw1upg);

\node at (wstar) (wstarn) {$\bm{w}^*$};
\draw[->, thick] (wstarn) -- (center2);

\draw[-, thick, gray] (decomp1) -- (decomp2);
\draw[-, thick, gray] (decomp3) -- (decomp4);

\draw[->, thick] (center2) -- (intersection);
\node at (wrstar) (wrstarn) {$\bm{w}^*_R$};
\draw[->, thick] (wrstarn) -- (intersection);

\draw [ultra thick,decoration={brace}, decorate]
(center2) -- (p1) node (brace1) {};
\draw [ultra thick,decoration={brace, mirror}, decorate]
(center2) -- (p2) node (brace2) {};

\node[text width=3cm] at (text1) (ntext1) {Small $\Lambda_1$, \\ low curvature large decrease \\ of a coordinate};

\draw[->, thick, minimum size=12pt, color=black] ($(brace1.north) + (-6mm, 3mm)$) to [out=20,in=117] (ntext1.west); 

\node[text width=3cm] at (text2) (ntext2) {Large $\Lambda_2$, \\ high curvature small decrease \\ of a coordinate};

\draw[->, thick, minimum size=12pt, color=black] ($(brace2.north) + (-1.5mm, 1mm)$) to [out=180,in=117] (ntext2.north); 
\end{tikzpicture}
	\end{center}
  \caption[\bel{Effect of $L_2$ norm regularization.}]{Effect of $L_2$ norm regularization: it scales the weights coordinates depending on the corresponding eigenvalues.  (\emph{red contours}): indicates the $L_2$ cost ($L_2$ norm of $\bm{w}$). (\emph{green contours}): indicate the unregularized cost $J$. One contour indicates a set of parameters $\bm{w}$ that have the same cost. For instance, all the coordinates $(w_1, w_2)$ that belong to the central red circle have the same $L_2$ norm. $\bm{w}=\bm{0}$ is the optimum solution for the $L_2$ cost. $\bm{w}=\bm{w}^*$ is the minimal solution of $J$. $\bm{w}=\bm{w}^*_R$
  represents the optimum solution of the regularized cost $\widetilde{J}$ (Eq.\ref{eq:eqapp3-post-70}).(Reference: \cite{Demyanovtehsis2015})}
  \label{fig:figapp3-post-15-1}
\end{figure}

As $\alpha$ approaches $0$, the regularized solution $\widetilde{\bm{w}}$ approaches $\bm{w}^*$. In order to have an idea on what happens when $\alpha$ grows, one can proceed using matrix decomposition. $\bm{H}$ has an eigen-decomposition using a diagonal matrix $\bm{\Lambda}$ and an orthogonal basis of eigenvectors $\bm{Q}$ such that
\begin{equation}
    \label{eq:eqapp3-post-78}
    \bm{H} = \bm{Q} \bm{\Lambda} \bm{Q}^{\top} \; .
\end{equation}
\noindent Applying this decomposition to Eq.\ref{eq:eqapp3-post-77}, the following is obtained \cite{Goodfellowbook2016}
\begin{align}
    \widetilde{\bm{w}} & = (\bm{Q} \bm{\Lambda} \bm{Q}^{\top} + \alpha \bm{I})^{-1} \bm{Q} \bm{\Lambda} \bm{Q}^{\top} \bm{w}^* \label{eq:eqapp3-post-79} \\
   \widetilde{\bm{w}} & = \left[ \bm{Q} (\bm{\Lambda} + \alpha \bm{I}) \bm{Q}^{\top}\right]^{-1} \bm{Q} \bm{\Lambda} \bm{Q}^{\top} \bm{w}^* \label{eq:eqapp3-post-80} \\
   \widetilde{\bm{w}} & = \bm{Q} (\bm{\Lambda} + \alpha \bm{I})^{-1} \bm{\Lambda} \bm{Q}^{\top} \bm{w}^* \label{eq:eqapp3-post-81-0} \\
   \underbrace{\bm{Q}^{\top} \widetilde{\bm{w}}}_{\text{Projection of } \widetilde{\bm{w}} \text{ in } \bm{Q}^{\top}} & = \underbrace{(\bm{\Lambda} + \alpha \bm{I})^{-1} \bm{\Lambda}}_{\text{Scaling factor= } \frac{\Lambda_i}{\Lambda_i + \alpha}} \underbrace{\bm{Q}^{\top} \bm{w}^*}_{\text{Projection of } \bm{w}^* \text{ in } \bm{Q}^{\top}} \label{eq:eq0-81} \; .
\end{align}
\noindent Therefore, one can see that the effect of weight decay is to rescale $\bm{w}^*$ along the axes defined by the eigenvectors $\bm{Q}^{\top}$. The components of $\bm{w}^*$ that are aligned with the $i$-th eigenvector of $\bm{H}$ are rescaled by a factor of $\frac{\Lambda_i}{\Lambda_i + \alpha}$. Therefore, in the case where $\Lambda_i \gg \alpha$, the effect of the regularization is relatively small. While, in the case where $\Lambda_i \ll \alpha$, the components of the parameters will be shrunk to have nearly zero magnitude. As a result, only directions along which the parameters contribute significantly to reducing the objective function, determined by eigenvalues with high values, are preserved relatively intact. In directions that do not contribute much in reducing the objective function, a small eigenvalue of the Hessian indicates that the movement in this direction will not significantly increase the gradient. Components of the weight vector corresponding to such unimportant directions are pushed toward zero. Fig.\ref{fig:figapp3-post-15-1} illustrates the effect of $L_2$ norm regularization on the parameters search.

So far we have seen the effect of the weight decay over a general quadratic cost function. Using the same analysis, one can see its impact on a true quadratic function such as the linear regression
\begin{equation}
    \label{eq:eeqapp3-post-82}
    \yvec = \bm{X} \bm{w} \; .
\end{equation}
\noindent Its unregularized objective function is defined as 
\begin{equation}
    \label{eq:eqapp3-post-83}
    J(\bm{w}; \bm{X}, \yvec) = \lVert \bm{X} \bm{w} - \yvec \rVert^2_2 \; ,
\end{equation}
\noindent and its regularized objective function is defined as
\begin{equation}
    \label{eq:eqapp3-post-84}
    \widetilde{J}(\bm{w}; \bm{X}, \yvec) = \lVert \bm{X} \bm{w} - \yvec \rVert^2_2 + \alpha \lVert \bm{w} \rVert^2_2 \; .
\end{equation}
The solution for the normal equation Eq.\ref{eq:eqapp3-post-83} is given by
\begin{equation}
    \label{eq:eqapp3-post-85}
    \widetilde{\bm{w}} = (\bm{X}^{\top} \bm{X})^{-1} \bm{X}^{\top} \yvec \; ,
\end{equation}
\noindent while the solution for the regularized form is given as
\begin{equation}
    \label{eq:eqapp3-post-86}
    \widetilde{\bm{w}} = (\bm{X}^{\top} \bm{X} + \alpha \bm{I})^{-1} \bm{X}^{\top} \yvec \; .
\end{equation}

The matrix $(\bm{X}^{\top} \bm{X})$ in Eq.\ref{eq:eqapp3-post-85} is proportional to the convariance matrix $\frac{1}{m}(\bm{X}^{\top} \bm{X})$. $L_2$ norm regularization replaces this matrix by $(\bm{X}^{\top} \bm{X} + \alpha \bm{I})$ in Eq.\ref{eq:eqapp3-post-86}. The new matrix is similar to the old one but with the addition of a positive constant $\alpha$ to the diagonal. The diagonal entries in $(\bm{X}^{\top} \bm{X})$ correspond to the variance of each input feature. Therefore, $L_2$ norm regularization makes the input look like it has a high variance. This shrinks the weights on features whose covariance with the output target is low compared to the added variance. From computation perspective, $L_2$ norm regularization reduces the numerical instability of inverting $(\bm{X}^{\top} \bm{X})$ by making it non-singular. More interpretations of the $L_2$ norm regularization can be found in \cite{Boyd2004}.

\paragraph[$L_1$ Parameters Norm]{$L_1$ Parameters Norm}
\label{para:appendix4l1norm}
\mbox{}
\bigskip
\\
\indent $L_1$ parameters norm regularization is formulated as follows
\begin{equation}
    \label{eq:eqapp3-post-87}
    \Omega(\theta) = \lVert \bm{w} \rVert_1 = \sum_i |w_i| \; .
\end{equation}

While we present the $L_1$ norm regularization, we highlight the differences between $L_1$ and $L_2$ forms of regularization by considering a linear regression problem. $L_1$ weight decay controls the strength of the norm penalty by scaling $\Omega$ using a positive hyperparameter $\alpha$ as in $L_2$ norm regularization. Thus, the regularized objective function is given by 
\begin{equation}
    \label{eq:eqapp3-post-88}
    \widetilde{J}(\bm{w}; \bm{X}, \yvec) = \alpha \lVert \bm{w} \rVert_1 + J(\bm{w}; \bm{X}, \yvec)   \; ,
\end{equation}
\noindent with the corresponding gradient
\begin{equation}
    \label{eq:eqapp3-post-90}
    \nabla_{\bm{w}}\widetilde{J}(\bm{w}; \bm{X}, \yvec) = \alpha \sign(\bm{w}) + \nabla_{\bm{w}} J(\bm{w}; \bm{X}, \yvec)  \; ,
\end{equation}
\noindent where $\sign(\bm{w})$ is the sign of $\bm{w}$ applied element-wise.  Comparing $L_1$ (Eq.\ref{eq:eqapp3-post-90}) and $L_2$ (Eq.\ref{eq:eqapp3-post-71}), one can see that the $L_1$ regularization contribution is no longer scales linearly with each weight $w_i$ as in $L_2$; instead it is a constant factor with a sign equal to the sign of the parameter $w_i$.

Now, let us consider approximating the objective function around the minimum ${\bm{w}^* = \argmin_{\bm{w}} J(\bm{w})}$ by ${\hat{J}}$ using Taylor expansion. In this case, the gradient of $\hat{J}$ is computed as \cite{Goodfellowbook2016}
\begin{equation}
    \label{eq:eqapp3-post-91}
    \nabla_{\bm{w}}\hat{J}(\bm{w}) = \bm{H} (\bm{w} - \bm{w}^*) \; ,
\end{equation}
\noindent where $\bm{H}$ is the Hessian matrix of $J$ with respect to $\bm{w}$ evaluated at $\bm{w}^*$. However, this time further simplification are made by assuming that the Hessian is diagonal
\begin{equation}
    \label{eq:eqapp3-post-92}
    \bm{H} = \diag([H_{1,1}, \cdots, H_{n,n}]), \quad \forall i: H_{i,i} > 0 \; .
\end{equation}
\noindent This assumption holds if the data for the linear regression problem has been preprocessed to remove all correlation between the input features. Now, the quadratic approximation of the $L_1$ regularized objective function can be written as \cite{Goodfellowbook2016}
\begin{equation}
    \label{eq:eqapp3-post-93}
    \hat{J}(\bm{w}; \bm{X}, \yvec) = \J(\bm{w}^*; \bm{X}, \yvec) + \sum_i \left[\frac{1}{2} H_{i,i} (w_i - w^*_i)^2 + \alpha |w_i| \right] \; .
\end{equation}

Eq.\ref{eq:eqapp3-post-93} has an analytic solution for each dimension $w_i$ under the following form \cite{Goodfellowbook2016}
\begin{equation}
    \label{eq:eqapp3-post-94}
    w_i = \sign(w^*_i) \max\left\{|w^*_i| - \frac{\alpha}{H_{i,i}}, 0\right\} \; .
\end{equation}
\noindent When $w^*_i > 0$, there are two possible outcomes:
\begin{enumerate}
    \item $w^*_i \leq \frac{\alpha}{H_{i,i}}$: The optimal value of $w_i$ is simply $w_i = 0$. This occurs when the contribution of the $L_1$ regularization penalty takes over the objective function $J(\bm{w}; \bm{X}, \yvec)$. 
    \item $w^*_i > \frac{\alpha}{H_{i,i}}$: In this case, the regularization does not set the optimal value of $w_i$ to zero but instead shift it in the direction of zero by $\frac{\alpha}{H_{i,i}}$.
\end{enumerate}
\noindent In the case where $w^*_i < 0$, the $L_1$ penalty makes $w_i$ less negative by a distance $\frac{\alpha}{H_{i,i}}$ or set it to zero.

\bigskip

Comparing to $L_2$, $L_1$ regularization results in a solution where most of the parameters are zero, i.e., a sparse solution. The sparsity behavior in $L_1$ is different than the one in $L_2$ where the parameters are pushed toward zero in some cases. Eq.\ref{eq:eqapp3-post-81-0} gives the solution $\widetilde{\bm{w}}$ for $L_2$ regularization. If the same assumption, used in the case of $L_1$, is considered about the Hessian matrix, one can find that $\widetilde{w}_i = \frac{H_{i,i}}{H_{i,i} + \alpha} w^*_i$. Therefore, if $w^*_i$ is nonzero, $\widetilde{w}_i$ 
remains nonzero. This shows that $L_2$ does not promote sparse solutions, while $L_1$ regularization may set a subset of the parameters to zero for large enough $\alpha$. Fig.\ref{fig:figapp3-post-16-1} illustrates the geometric effect introduced by the $L_1$ regularization on the parameters search. This sparsity aspect plays an important role in machine learning particularly as a feature selection mechanism. Feature selection simplifies machine learning by choosing which subset of input features are relevant to predict the output target. This has a key role in application where the interpretation is highly important. For instance, when building a model to predict a disease based on set of input features, it is important to know which factors are implicated in the cause of the disease. The sparsity has been used for a long time, for instance, the well known LASSO \cite{Tibshirani94regressionshrinkage} (Least Absolute Shrinkage and Selection Operator) model integrates an $L_1$ penalty with a linear model. More details on the $L_1$ regularization and sparsity can be found in \cite{Hastie2015book}.

\begin{figure}[!htbp]
  \begin{center}
		\begin{tikzpicture}
\coordinate (center1) at (0, 0);
\coordinate (xneg) at (-5, 0);
\coordinate (xpos) at (5, 0);
\coordinate (ypos) at (0, 4);
\coordinate (yneg) at (0, -2);

\coordinate (qw2low) at (-0.66, -2);
\coordinate (qw2up) at (1.3, 3.9);

\coordinate (qw1low) at (-3, 1);
\coordinate (qw1up) at (3, -1);

\coordinate (center2) at (-1, 1.98);
\coordinate (wstar) at (-1., 3.5);
\coordinate (wrstar) at (-1.8, 0.8);

\coordinate (qw2lowg) at (-4, 2.74);
\coordinate (qw2upg) at (1, 1.08);

\coordinate (qw1lowg) at (-2.69, 0.02);
\coordinate (qw1upg) at (-1.5, 3.59);

\coordinate (decomp1) at (0, 3.7);
\coordinate (decomp2) at (-1.23, 0);

\coordinate (decomp3) at (0, 1);
\coordinate (decomp4) at (-2.15, 1.71);

\coordinate (intersection1) at (0, 1.05);
\coordinate (intersection2) at (-0.13, 1.5);
\coordinate (p1) at (-0.69, 1.64);
\coordinate (p2) at (-2.12, 1.7);

\coordinate (text1) at (4, 1.7);
\coordinate (text2) at (-2, -1.2);
 
\node at ($(center1) + (-2mm, -2mm)$) {$0$};
\draw[->, thick] (xneg) -- (xpos) node[right] (xaxis) {$\bm{w}_1$};
\draw[->, thick] (yneg) -- (ypos) node[right] (yaxis) {$\bm{w}_2$};

\node [draw,scale=2,diamond, ultra thick, red] at (center1) (cr4) {};
\node [draw,scale=4,diamond, ultra thick, red] at (center1) (cr5) {};
\node [draw,scale=6,diamond, ultra thick, red] at (center1) (cr6) {};

\draw[green, ultra thick] (center2) circle [x radius=0.5, y radius=0.3, rotate=-18] node (cg1) {};

\draw[green, ultra thick] (center2) circle [x radius=1, y radius=0.6, rotate=-18] node (cg2) {};

\draw[green, ultra thick] (center2) circle [x radius=1.5, y radius=0.9, rotate=-18] node (cg3) {};

\node at (wstar) (wstarn) {$\bm{w}^*$};
\draw[->, thick] (wstarn) -- (center2);

\draw[->, thick] (center2) -- (intersection1);
\draw[->, thick] (center2) -- (intersection2);
\node at (wrstar) (wrstarn) {$\bm{w}^*_R$};
\draw[->, thick] (wrstarn) -- (intersection1);
\draw[->, thick] (wrstarn) -- (intersection2);

\node[text width=5cm] at (text1) (ntext1) {Smaller $\alpha$, \\ further from the origin \\ no coordinates are $0$};

\draw[->, thick,  minimum size=12pt, color=black] ($(intersection2) + (-3mm, 3mm)$) to [out=20,in=117] (ntext1.west); 

\node[text width=5cm] at (text2) (ntext2) {Larger $\alpha$, \\ closer to the origin \\ some coordinates are $0$};

\draw[->, thick,  minimum size=12pt, color=black] ($(intersection1) + (-7mm, 5mm)$) to [out=180,in=117] ($(ntext2) + (-20mm, 9mm)$); 
\end{tikzpicture}
	\end{center}
  \caption[\bel{Effect of $L_1$ norm regularization.}]{Effect of $L_1$ norm regularization: Large $\alpha$ makes some parameters equal to 0. (\emph{red contours}): indicate the $L_1$ cost ($L_1$ norm of $\bm{w}$). (\emph{green contours}): indicate the unregularized cost $J$. $\bm{w}=\bm{w}^*$ is the minimal solution of $J$. $\bm{w}=\bm{w}^*_R$
  represent the optimum solutions of the regularized cost $\widetilde{J}$ (Eq.\ref{eq:eqapp3-post-88}). In this example, $L_1$ regularization allows two solutions $\bm{w}^*_R$ depending on the value of $\alpha$. Small $\alpha$ results in a solution a little far from the origin. However, large $\alpha$ provides a sparse solution where $w_1= 0$. (Reference: \cite{Demyanovtehsis2015})}
  \label{fig:figapp3-post-16-1}
\end{figure}

\subsubsection[Early Stopping as a Regularization]{Early Stopping as a Regularization}
\label{subsub:appendix4earlystopping}
We have mentioned in Sec.\ref{subsub:earlystopping0} that early stopping can play a role of a regularizer. Formal demonstration is provided in this section.

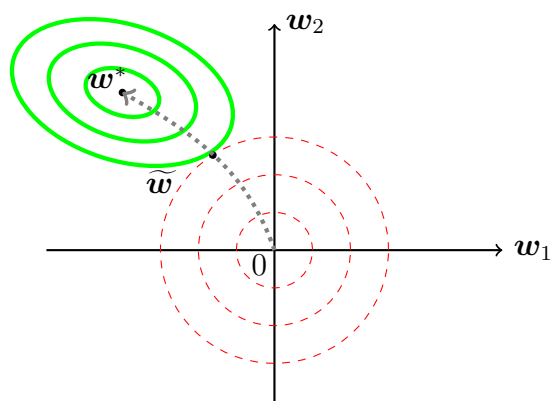
\begin{figure}[!htbp]
  \begin{center}
		\begin{tikzpicture}
\coordinate (center1) at (0, 0);
\coordinate (xneg) at (-3, 0);
\coordinate (xpos) at (3, 0);
\coordinate (ypos) at (0, 3);
\coordinate (yneg) at (0, -2);

\coordinate (qw2low) at (-0.66, -2);
\coordinate (qw2up) at (1.3, 3.9);

\coordinate (qw1low) at (-3, 1);
\coordinate (qw1up) at (3, -1);

\coordinate (center2) at (-2, 2.09);
\coordinate (wstar) at (-2.5, 3.5);
\coordinate (wrstar) at (-1.8, 0.8);

\coordinate (qw2lowg) at (-4, 2.74);
\coordinate (qw2upg) at (1, 1.08);

\coordinate (qw1lowg) at (-2.69, 0.02);
\coordinate (qw1upg) at (-1.5, 3.59);

\coordinate (decomp1) at (0, 3.7);
\coordinate (decomp2) at (-1.23, 0);

\coordinate (decomp3) at (0, 1);
\coordinate (decomp4) at (-2.15, 1.71);

\coordinate (intersection) at (-0.81, 1.26);
\coordinate (p1) at (-0.69, 1.64);
\coordinate (p2) at (-2.12, 1.7);

\coordinate (text1) at (3, 1.7);
\coordinate (text2) at (-3, -1.2);
 
\node at ($(center1) + (-2mm, -2mm)$) {$0$};
\draw[->, thick] (xneg) -- (xpos) node[right] (xaxis) {$\bm{w}_1$};
\draw[->, thick] (yneg) -- (ypos) node[right] (yaxis) {$\bm{w}_2$};

\draw[red, dashed] (center1) circle [radius=0.5] node (cr1) {};
\draw[red, dashed] (center1) circle [radius=1] node (cr2) {};
\draw[red, dashed] (center1) circle [radius=1.5] node (cr3) {};

\draw[green, ultra thick] (center2) circle [x radius=0.5, y radius=0.3, rotate=-18] node (cg1) {};

\draw[green, ultra thick] (center2) circle [x radius=1, y radius=0.6, rotate=-18] node (cg2) {};

\draw[green, ultra thick] (center2) circle [x radius=1.5, y radius=0.9, rotate=-18] node (cg3) {};

\node at ($(center2) + (-2mm, 2mm)$)  {$\bm{w}^*$};
\draw[black] (center2) node[circle, fill, inner sep=1pt] {};
\draw[black] (intersection) node[circle, fill, inner sep=1pt] {};

\node at ($(intersection) + (-7mm, -3.5mm)$) {$\widetilde{\bm{w}}$};

\draw[->, color=gray, dotted, ultra thick] (center1) to[out=109,in=330] (center2) node[above, color=black]  {};

\end{tikzpicture}
	\end{center}
  \caption[\bel{An illustration of the effect of early stopping.}]{An illustration of the effect of early stopping. (\emph{green contours}): indicate the contours the unregularized cost $J$ (no early stopping). (\emph{red dashed contours}): indicate the contours of the $L_2$ cost, which cause the minimum of the total cost to lie nearer the origin rather the minimum of the unregularized cost. (\emph{gray dotted path}): indicates the trajectory taken by the SGD starting from the origin. Rather than stopping at the optimum point $\bm{w}^*$ that minimizes the cost, early stopping results in the trajectory stopping at an earlier point $\widetilde{\bm{w}}$.  (Credit: \cite{bishop1995ICANN, Goodfellowbook2016})}
  \label{fig:figapp3-post-18}
\end{figure}

Let us consider a network with weights initialized from a distribution with zero mean. We will see formally how early stopping can be a regularizer. Many authors \cite{bishop1995ICANN, sjoberg1995} argued that early stopping has the effect of restricting the optimization procedure to a relatively small volume of parameter space in the neighborhood of the initial parameter $\bm{\theta}_0$ (Fig.\ref{fig:figapp3-post-18}). More specifically, consider taking $\tau$ optimization steps and with a learning rate $\epsilon$. One can view $\epsilon \tau$ as the effective capacity. Restricting the number of iteration and the learning rate limits the volume of the parameter space reachable from $\bm{\theta}_0$. In this case, $\epsilon \tau$ behaves as if it was the reciprocal of the coefficient used for weight decay.

To compare early stopping with the classical $L_2$ regularization, let us consider a setting where parameters are linear weights $\bm{\theta} = \bm{w}$. One can approximate the objective function $J$ with a quadratic form in the neighborhood of the empirically optimal value of the weights $\bm{w}^*$ \cite{Goodfellowbook2016}
\begin{equation}
    \label{eq:eqapp3-post-113}
    \hat{J}(\bm{w}) = J(\bm{w}^*) + \frac{1}{2} (\bm{w} - \bm{w}^*)^{\top} \bm{H} (\bm{w} - \bm{w}^*) \; ,
\end{equation}
\noindent where $\bm{H}$ is the Hessian matrix of $J$ with respect to $\bm{w}$ evaluated at $\bm{w}^*$. This makes $\bm{H}$ positive semidefinite. The gradient of the $\hat{J}$ is
\begin{equation}
    \label{eq:eqapp3-post-114}
    \nabla_{\bm{w}} \hat{J}(\bm{w}) = \bm{H} (\bm{w} - \bm{w}^*) \; .
\end{equation}
Now, let us study the trajectory followed by the parameter vector during training. For simplicity, the initial parameters are set to the origin, $\bm{w}^{(0)} = \bm{0}$. Then, the gradient descent updates are performed as follows \cite{Goodfellowbook2016}
\begin{align}
    \bm{w}^{(\tau)} & = \bm{w}^{(\tau -1)} - \epsilon \nabla_{\bm{w}} \hat{J}(\bm{w}^{(\tau -1)} \label{eq:eqapp3-post-115} \\
    &= \bm{w}^{(\tau-1)} - \epsilon \bm{H} (\bm{w}^{(\tau -1)} - \bm{w}^*) \label{eq:eqapp3-post-116} \\
    \bm{w}^{(\tau)}- \bm{w}^* &= (\bm{I} - \epsilon \bm{H}) (\bm{w}^{(\tau -1)} - \bm{w}^*) \label{eq:eqapp3-post-117} \; .
\end{align}
\noindent Now, using the eigendecomposition of ${\bm{H}}$: ${\bm{H} = \bm{Q} \bm{\Lambda} \bm{Q}^{\top}}$, where $\bm{\Lambda}$ is a diagonal matrix and $\bm{Q}$ is an orthonormal basis of eigenvectors, it results \cite{Goodfellowbook2016}
\begin{align}
    \bm{w}^{(\tau)}- \bm{w}^* &= (\bm{I} - \epsilon \bm{Q} \bm{\Lambda} \bm{Q}^{\top}) (\bm{w}^{(\tau -1)} - \bm{w}^*) \label{eq:eqapp3-post-118} \\
    \bm{Q}(\bm{w}^{(\tau)}- \bm{w}^*) &= (\bm{I} - \epsilon \bm{\Lambda})\bm{Q}^{\top} (\bm{w}^{(\tau -1)} - \bm{w}^*) \; . \label{eq:eqapp3-post-119}
\end{align}
\noindent Assuming that $\epsilon$ is chosen to be small enough to grantee ${|1 - \epsilon \lambda_i|< 1}$, it can be shown \cite{bishop1995ICANN, Goodfellowbook2016} that the parameter trajectory during training after $\tau$ parameter updates has the following form
\begin{equation}
    \label{eq:eqapp3-post-120}
    \bm{Q}^{\top} \bm{w}^{(\tau)} = \left[\bm{I} - (\bm{I} - \epsilon\Lambda)^{\tau}\right] \bm{Q}^{\top} \bm{w}^* \; .
\end{equation}
\noindent Eq.\ref{eq:eqapp3-post-81-0} can be rearranged as
\begin{equation}
    \label{eq:eqapp3-post-121}
    \bm{Q}^{\top} \widetilde{\bm{w}} = \left[ \bm{I} - (\Lambda + \alpha \bm{I}) ^{-1} \alpha \right] \bm{Q}^{\top} \bm{w}^* \; .
\end{equation}
\noindent Comparing Eq.\ref{eq:eqapp3-post-120} and Eq.\ref{eq:eqapp3-post-121}, one can see that if $\epsilon$, $\alpha$ and $\tau$ are chosen such that \cite{bishop1995ICANN, Goodfellowbook2016}
\begin{equation}
    \label{eq:eqapp3-post-122}
    (\bm{I} - \epsilon\Lambda)^{\tau} = (\Lambda + \alpha \bm{I}) ^{-1} \alpha \; ,
\end{equation}
\noindent then $L_2$ regularization and early stopping can be seen as equivalent (under the quadratic approximation and the previous stated assumptions). Going further, by approximating both sides of Eq.\ref{eq:eqapp3-post-122}, one can conclude that if all $\lambda_i$  are small, then \cite{bishop1995ICANN, Goodfellowbook2016}
\begin{align}
    \tau \approx \frac{1}{\epsilon \alpha} \label{eq:eqapp3-post-123} \; , \\
    \alpha \approx \frac{1}{\tau \epsilon} \label{eq:eqapp3-post-124} \; .
\end{align}
\noindent Thus, under these assumptions, the number of training iterations $\tau$ plays a role inversely proportional to the $L_2$ regularization parameter, and the inverse of $\tau \epsilon$ plays the role of the weight decay. Therefore, the parameter corresponding to the directions of significant curvature of the objective function are regularized less than directions of less curvature. In the context of early stopping, this means that parameters that correspond to directions of significant curvature tend to learn early relatively to parameters corresponding to directions of less curvature. This is actually very intuitive. Parameters that correspond to high curvature tend to learn faster, which means, that in a short time, they have already learned something. While given the same amount of time, parameters that correspond to low curvature will tend to learn slowly. Therefore, early stopping mimics $L_2$ regularization by repressing parameters corresponding to low curvature and preventing parameter corresponding to high curvature to significantly grow. One can note that early stopping has the advantage to be determined through one run of the training process while $L_2$ regularization requires many runs with different values.

\end{appendices}

\printthesisindex 

\end{document}